\documentclass[journal]{IEEEtran}
\usepackage{graphicx}
\usepackage{booktabs}
\usepackage{tikz}
\usepackage{comment}
\usepackage{amsmath,amssymb} 
\usepackage{color}
\usepackage{xcolor}
\usepackage{graphicx}
\usepackage{tikz}
\usepackage{comment}
\usepackage{amsmath,amssymb} 
\usepackage{stfloats}
\usepackage{graphicx}
\usepackage{booktabs}
\usepackage{array}
\usepackage{multirow}
\usepackage{color}
\usepackage{colortbl}
\usepackage{framed}
\usepackage{bm}
\usepackage{bbm}
\usepackage{xspace}
\usepackage{enumitem}
\usepackage{pifont}
\usepackage{xcolor}
\usepackage{scalerel,graphicx,xparse}
\usepackage{mwe}
\usepackage{caption}
\usepackage{subcaption}
\usepackage{cite}
\usepackage[accsupp]{axessibility}  

\newcommand{\tit}[1]{\smallbreak\noindent\textbf{#1.}}
\newcommand{\tinytit}[1]{\noindent\textbf{#1.}}
\begin{document}

\title{Better Fit: Accommodate Variations in Clothing Types for Virtual Try-on} 

\author{Dan Song, Xuanpu Zhang, Jianhao Zeng, Pengxin Zhan, Qingguo Chen, Weihua Luo\\and An-An Liu*~\IEEEmembership{Senior Member,~IEEE}
\IEEEcompsocitemizethanks{\IEEEcompsocthanksitem  Dan Song, Xuanpu Zhang, Jianhao Zeng, and An-An Liu are with the School of Electrical and Information Engineering, Tianjin University, Tianjin 300072, China. 
\IEEEcompsocthanksitem Pengxin Zhan, Qingguo Chen and Weihua Luo are with Alibaba Group, Hangzhou 311121, China.
\IEEEcompsocthanksitem	Corresponding author: An-An Liu, anan0422@gmail.com.}
}
\markboth{IEEE Transactions on Circuits and Systems for Video Technology}%
{Song \MakeLowercase{\textit{et al.}}: Better Fit}

\maketitle

\begin{abstract}
Image-based virtual try-on aims to transfer target in-shop clothing to a dressed model image, the objectives of which are totally taking off original clothing while  preserving the contents outside of the try-on area, naturally wearing target clothing and correctly inpainting  the gap between target clothing and  original clothing. 
Tremendous efforts have been made to facilitate this popular research area, but cannot keep the type of target clothing  with the try-on area affected by original clothing. In this paper, we focus on the unpaired virtual try-on situation where target clothing and original clothing on the model are different, i.e., the practical scenario.
To break the correlation between the try-on area and the original clothing and make the model learn the correct information to inpaint, we propose an adaptive mask training paradigm that dynamically adjusts training masks. It not only improves the alignment and fit of clothing but also significantly enhances the fidelity of virtual try-on experience. Furthermore, we for the first time propose two metrics for unpaired try-on evaluation, the Semantic-Densepose-Ratio (SDR) and Skeleton-LPIPS (S-LPIPS), to evaluate the correctness of clothing type and the accuracy of clothing texture. For unpaired try-on validation, we construct a comprehensive cross-try-on benchmark (Cross-27) with distinctive clothing items and model physiques, covering a broad try-on scenarios. Experiments demonstrate the effectiveness of the proposed methods, contributing to the advancement of virtual try-on technology and offering new insights and tools for future research in the field. The code, model and benchmark will be publicly released.

\end{abstract}


\section{Introduction}

Recently, fashion images attract attentions \cite{fashion_customization,fashion_retrieval} and image-based virtual try-on is a hot topic in the field of conditional image generation, aiming to create an image of a person wearing a target clothing item given the clothing image and the person image. It offers practical advantages for online shoppers and opens up new avenues for creativity in the fashion domain.

\begin{figure}[t]
\centering
\includegraphics[width=\linewidth]{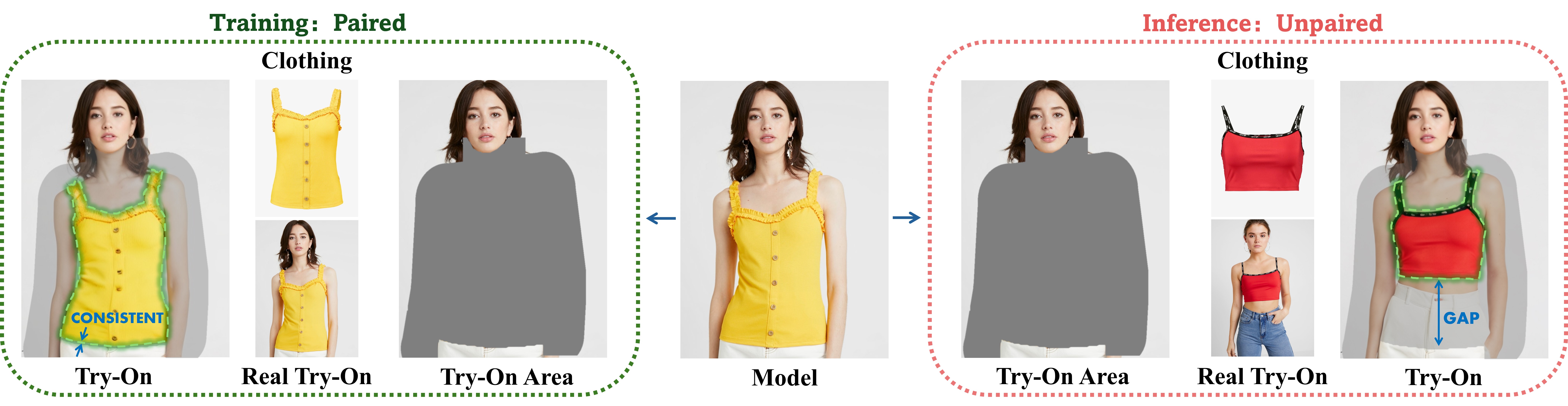}
\caption{There is a discrepancy between training and inference in virtual try-on tasks. During the paired training process, the try-on area is consistent with the target clothing, whereas during unpaired inference process, there is a gap between the try-on area and the target clothing.}
\label{fig:paired_unpaired}
\end{figure}

\begin{figure*}[t!]
\centering
\scriptsize
\setlength{\tabcolsep}{.2em}
\setlength{\arrayrulewidth}{1.25pt}
\resizebox{\linewidth}{!}{
\begin{tabular}{c  cccccc c c }
& \textbf{VITON-HD} & \textbf{HR-VTON} & \textbf{LaDI-VTON} & \textbf{DCI-VTON} & \textbf{GP-VTON} & \textbf{StableVITON} && \textbf{Ours} \\
\includegraphics[width=0.099\linewidth]{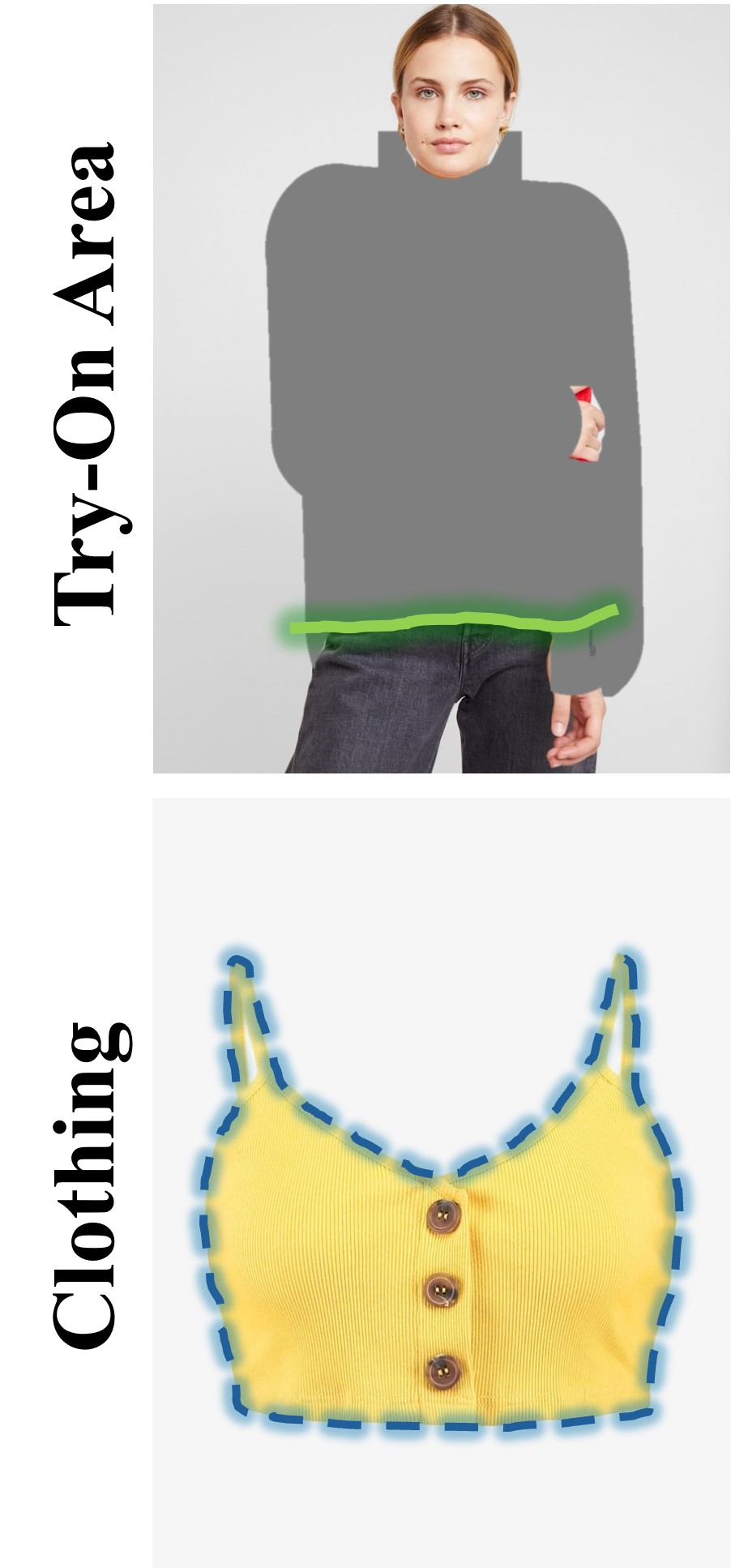} & 
\includegraphics[width=0.16\linewidth]{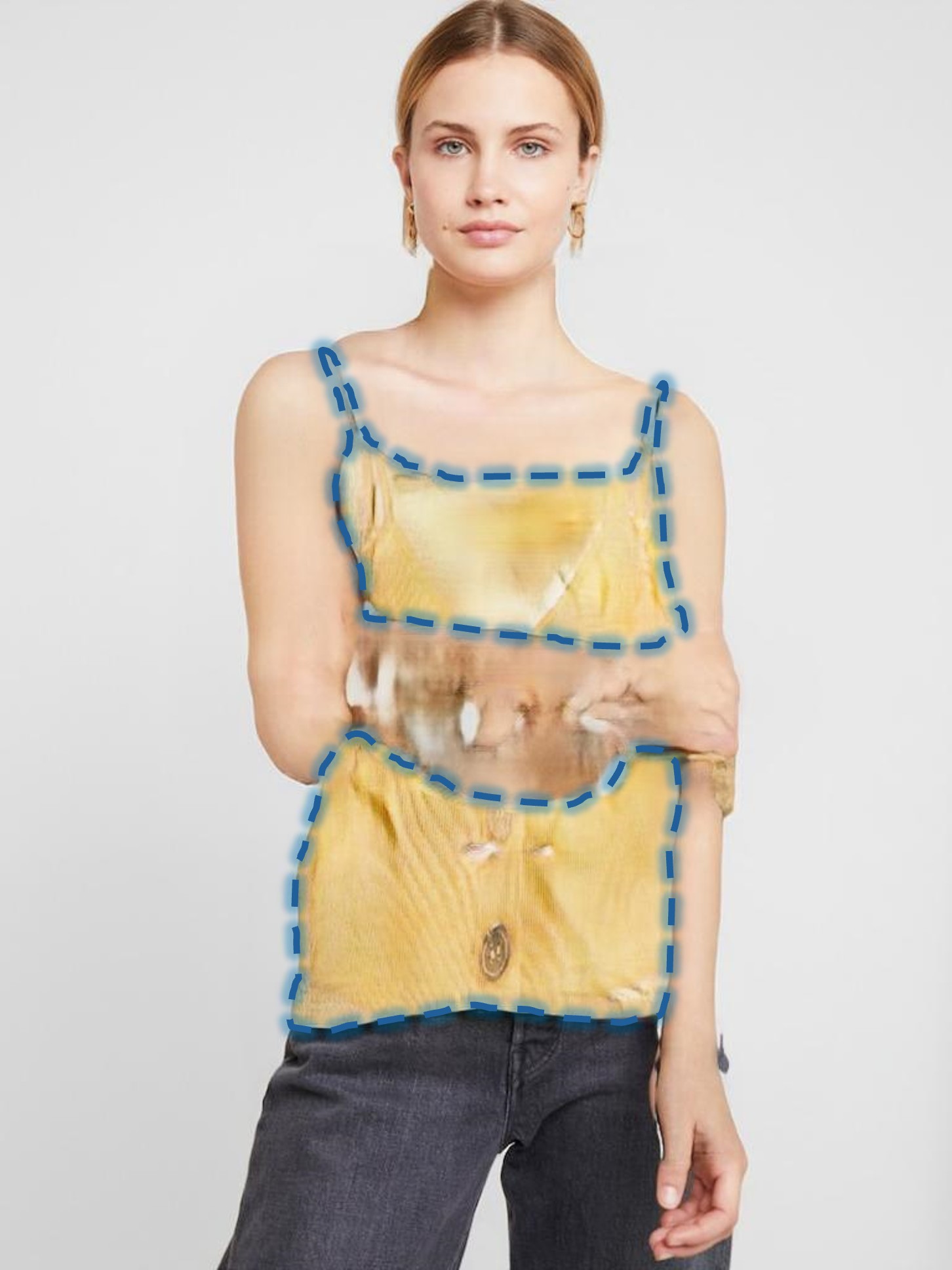} &
\includegraphics[width=0.16\linewidth]{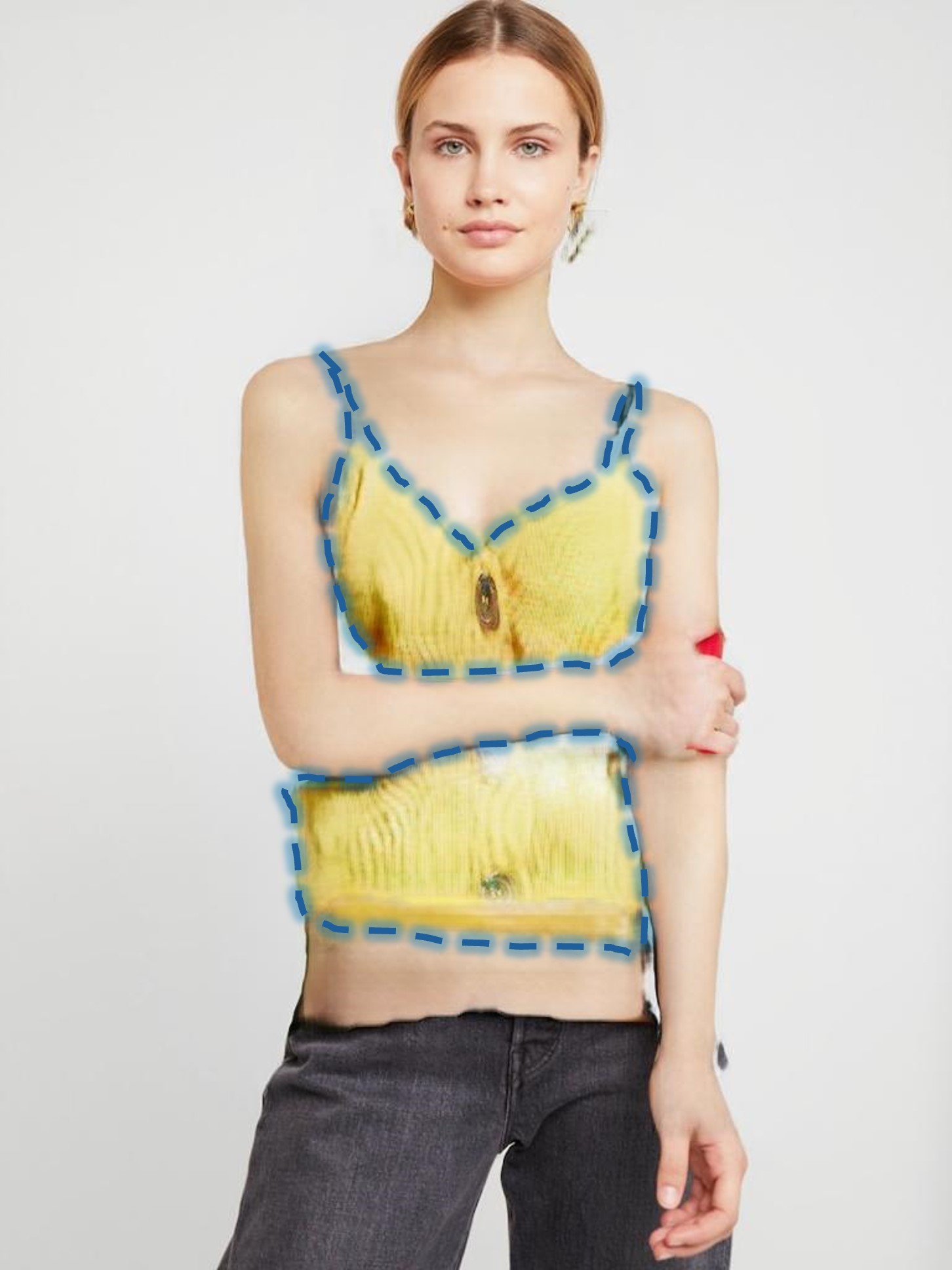} & 
\includegraphics[width=0.16\linewidth]{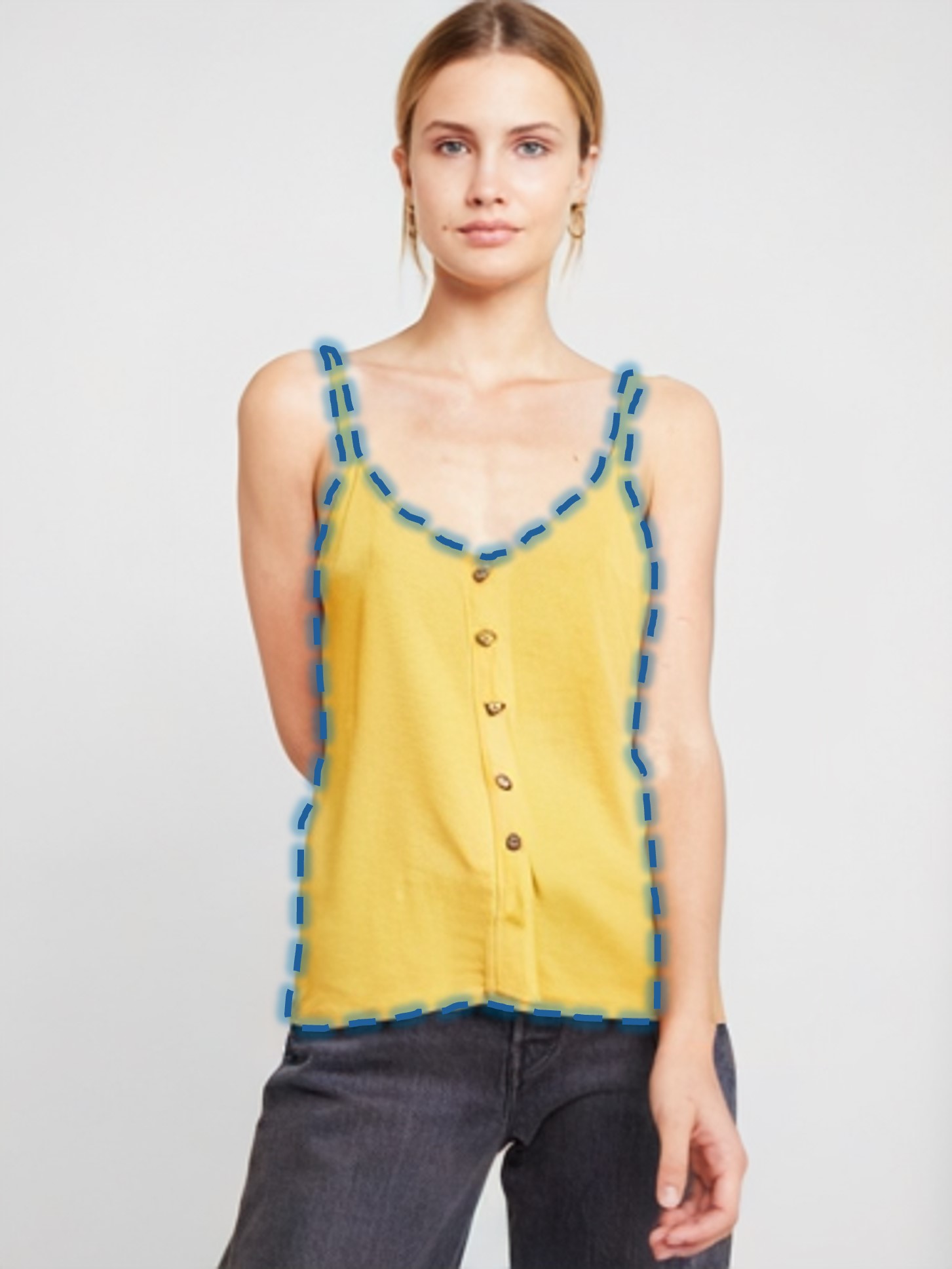} &
\includegraphics[width=0.16\linewidth]{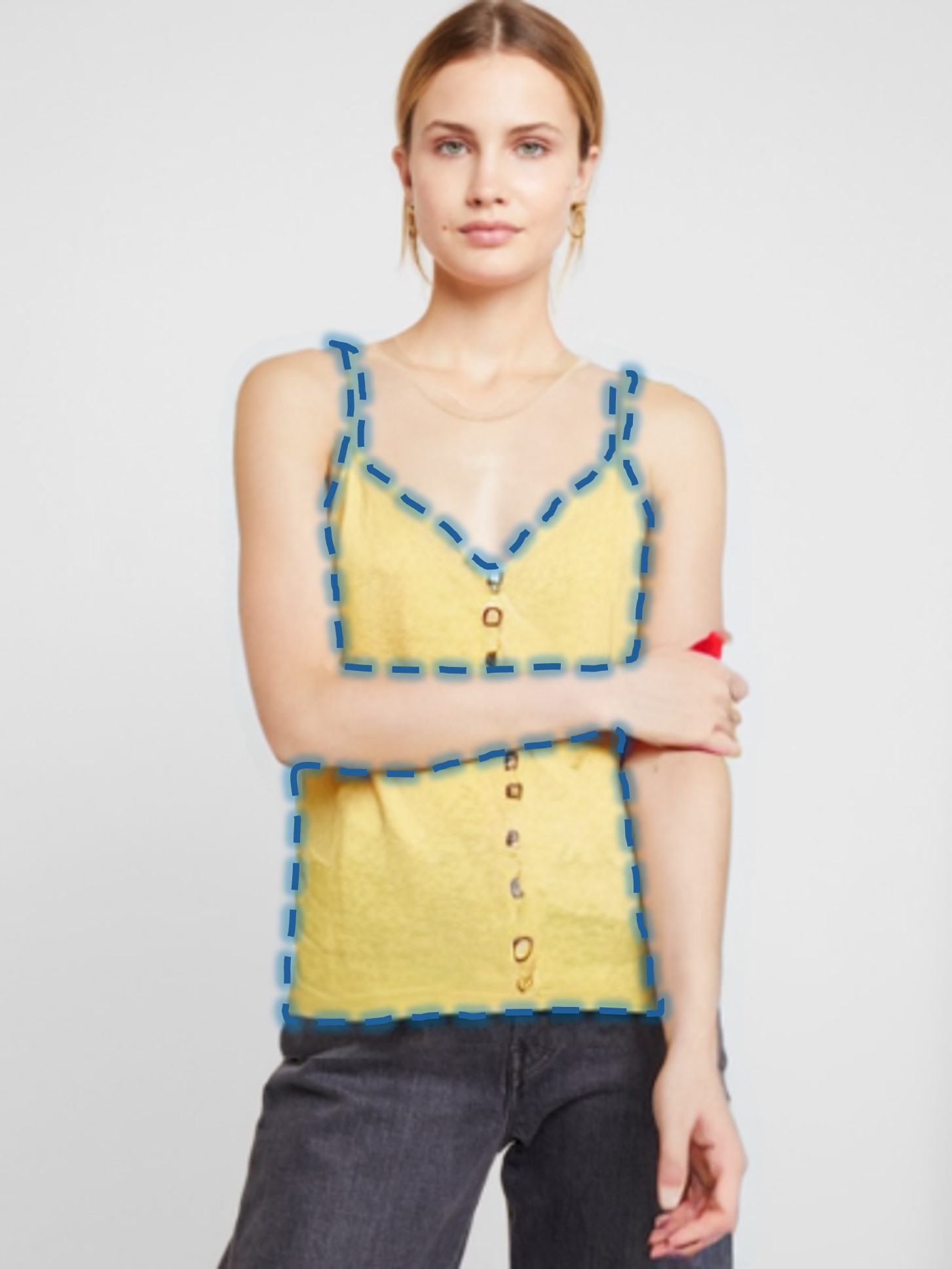} &
\includegraphics[width=0.16\linewidth]{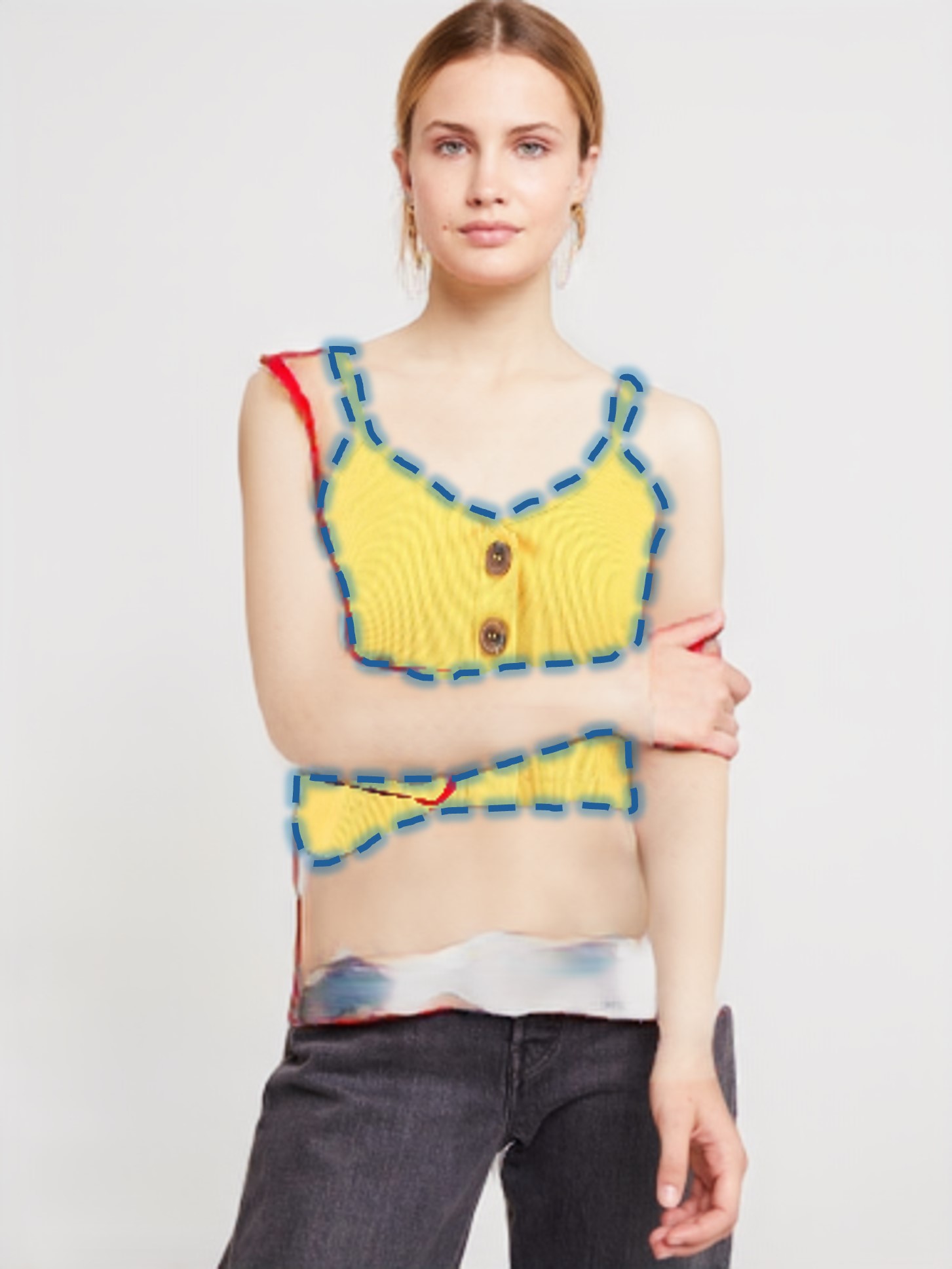} &
\includegraphics[width=0.16\linewidth]{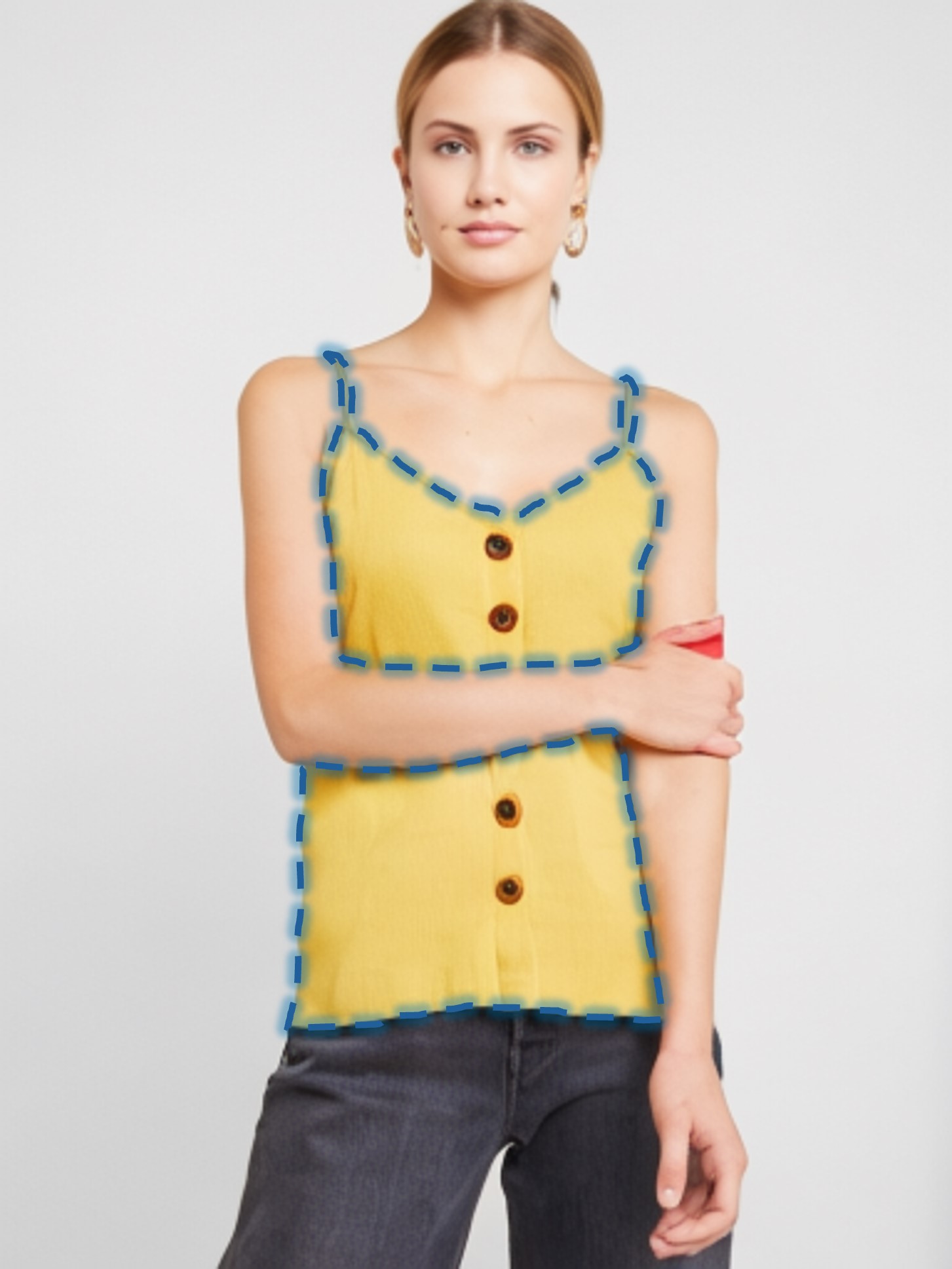} &&
\includegraphics[width=0.16\linewidth]{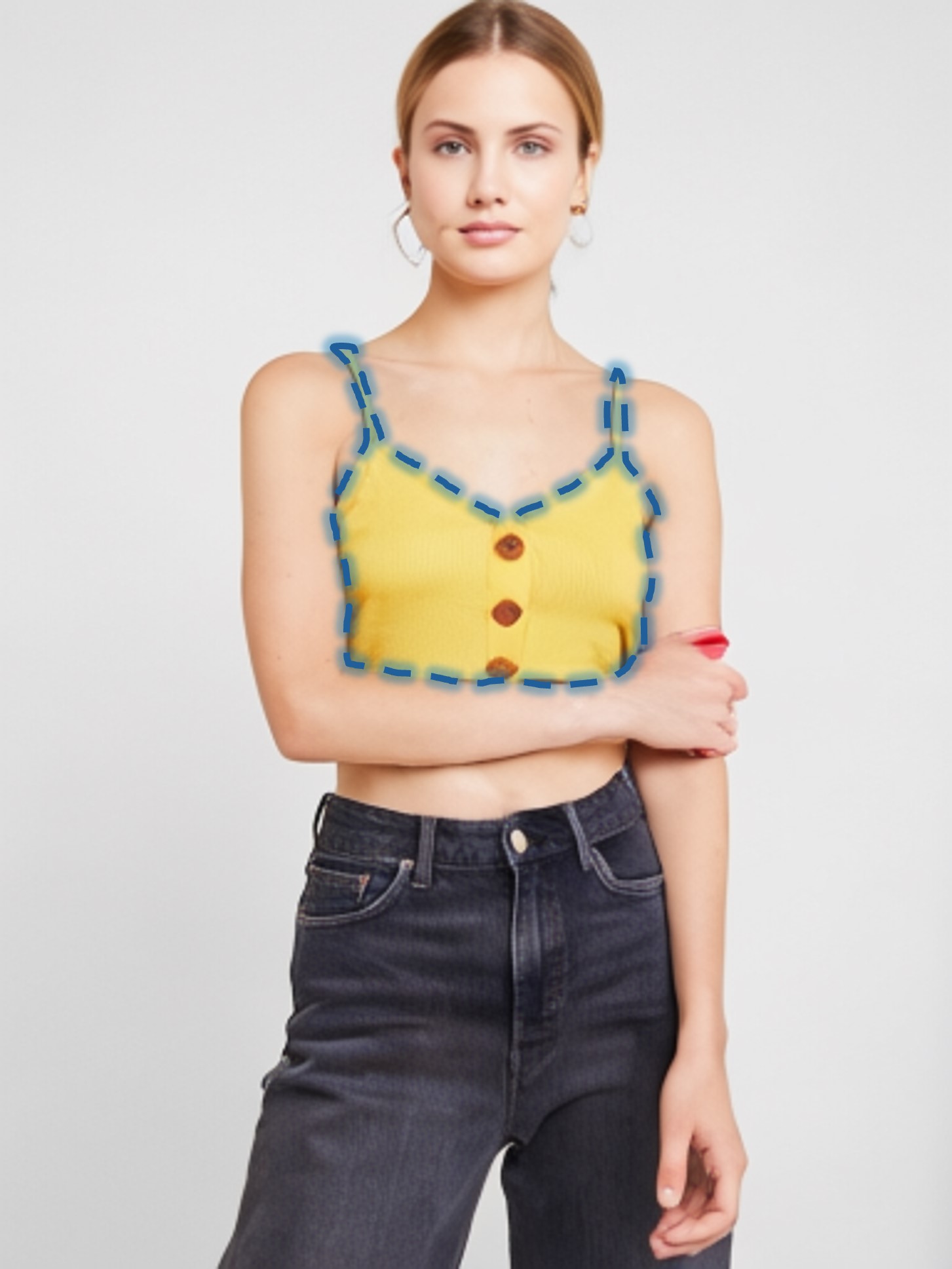}\\
\includegraphics[width=0.099\linewidth]{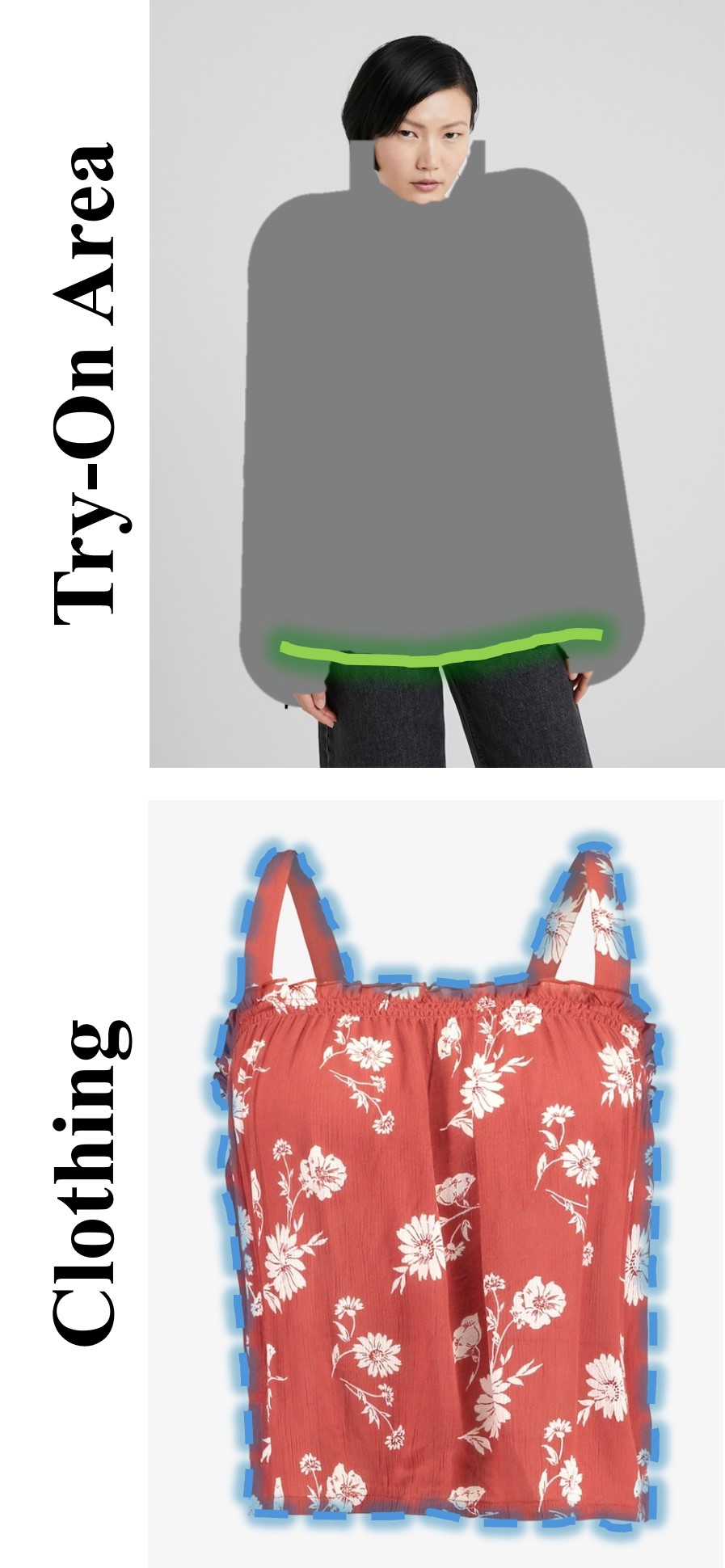} & 
\includegraphics[width=0.16\linewidth]{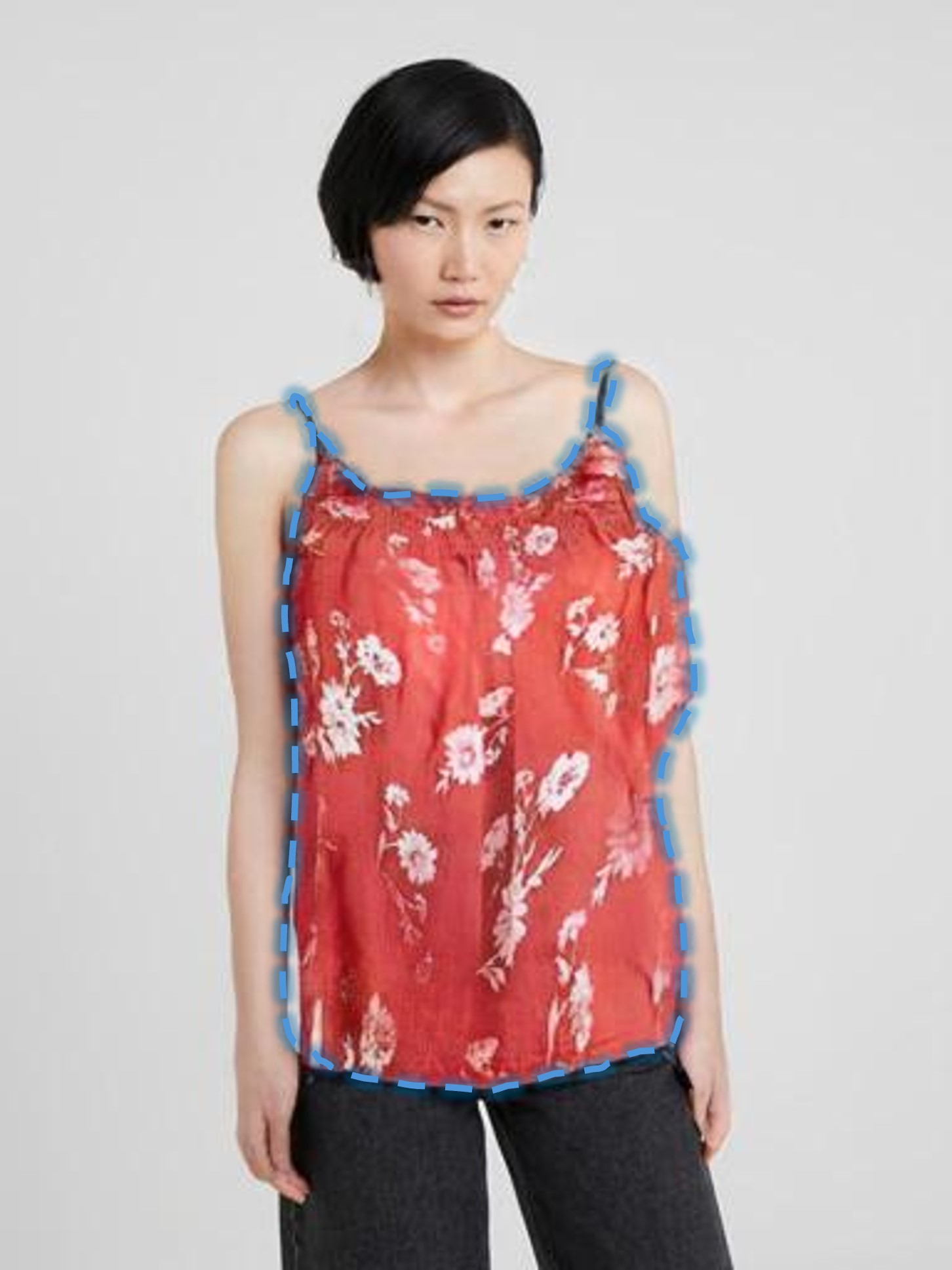} &
\includegraphics[width=0.16\linewidth]{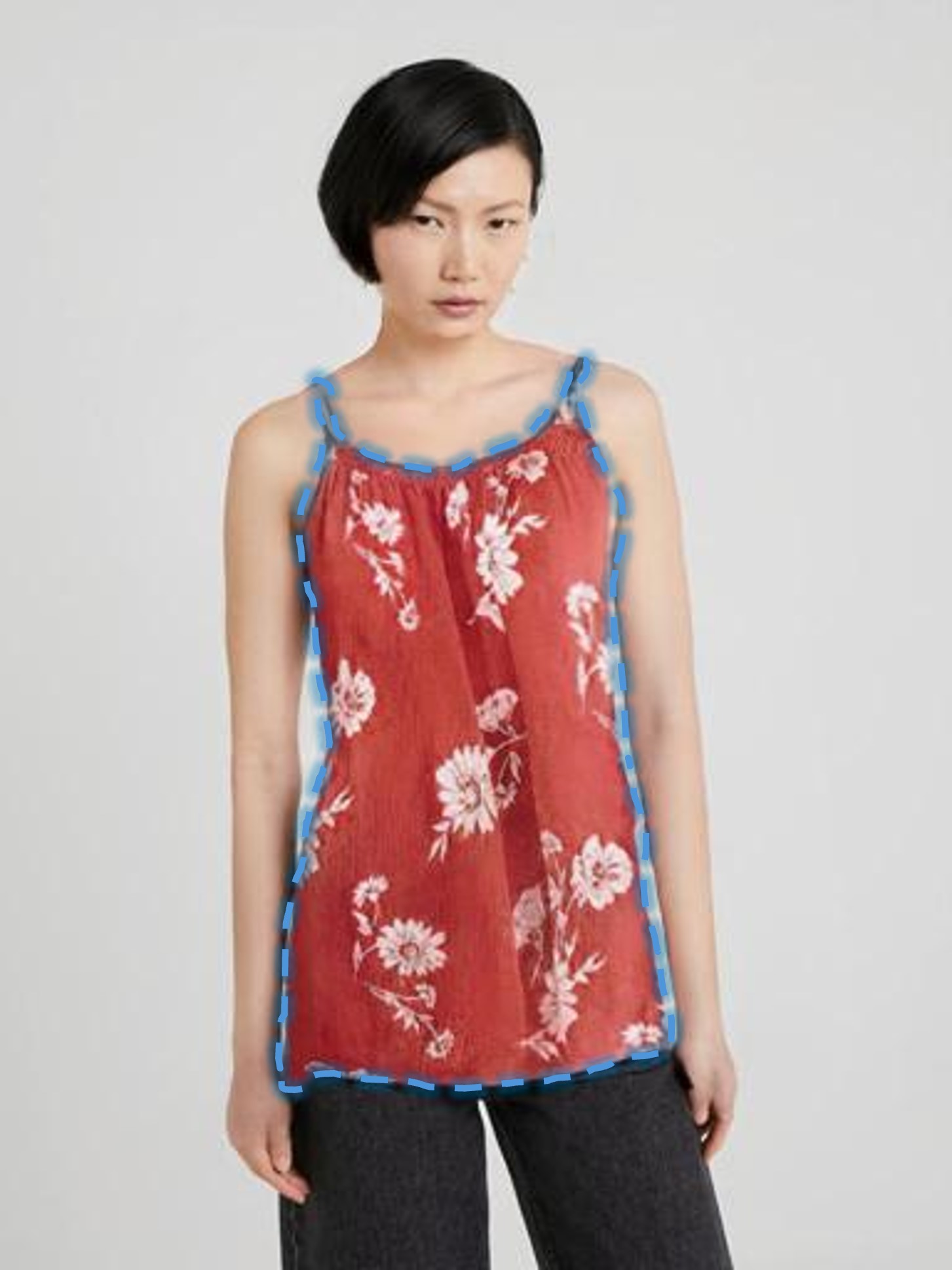} & 
\includegraphics[width=0.16\linewidth]{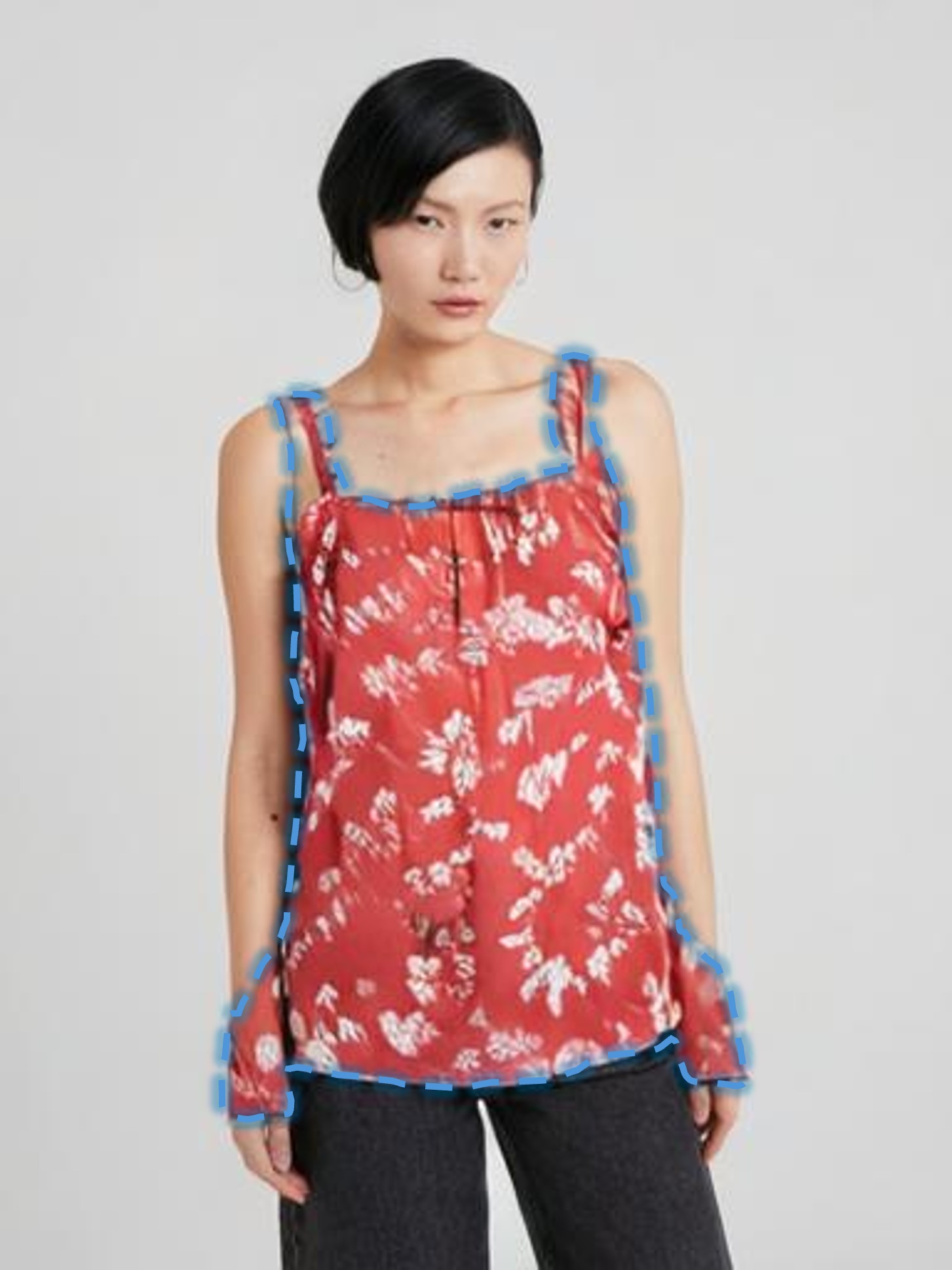} &
\includegraphics[width=0.16\linewidth]{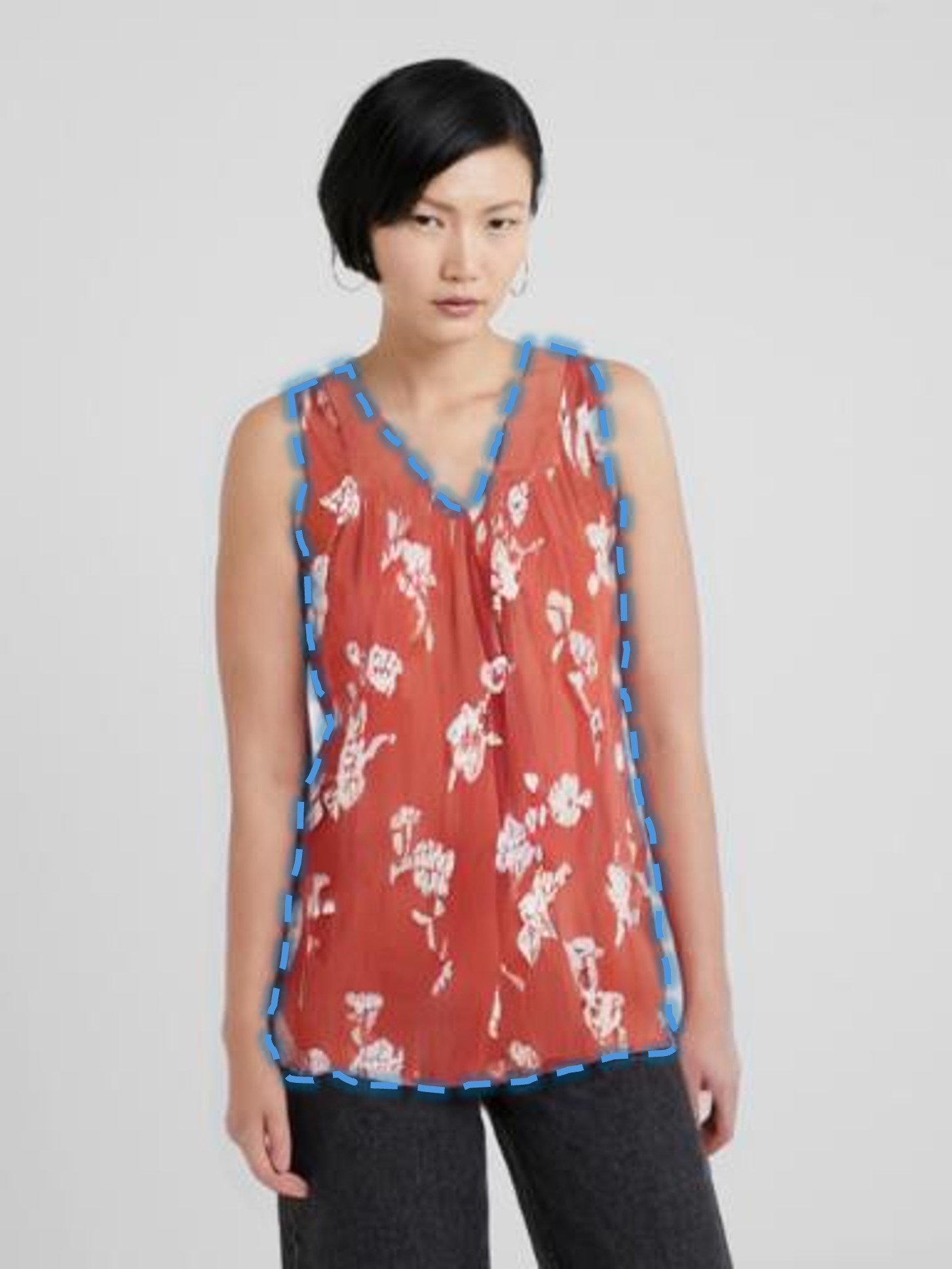} &
\includegraphics[width=0.16\linewidth]{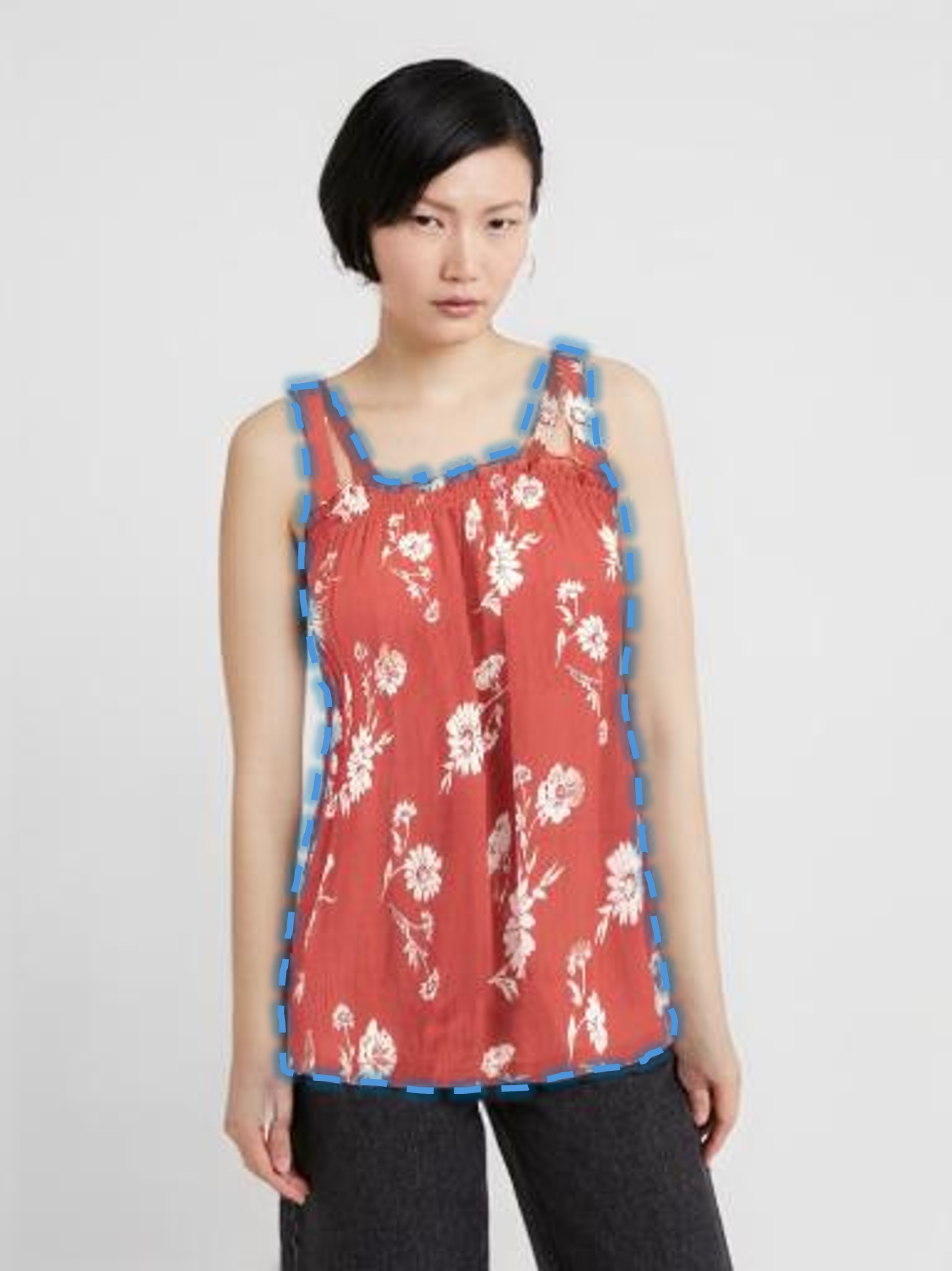} &
\includegraphics[width=0.16\linewidth]{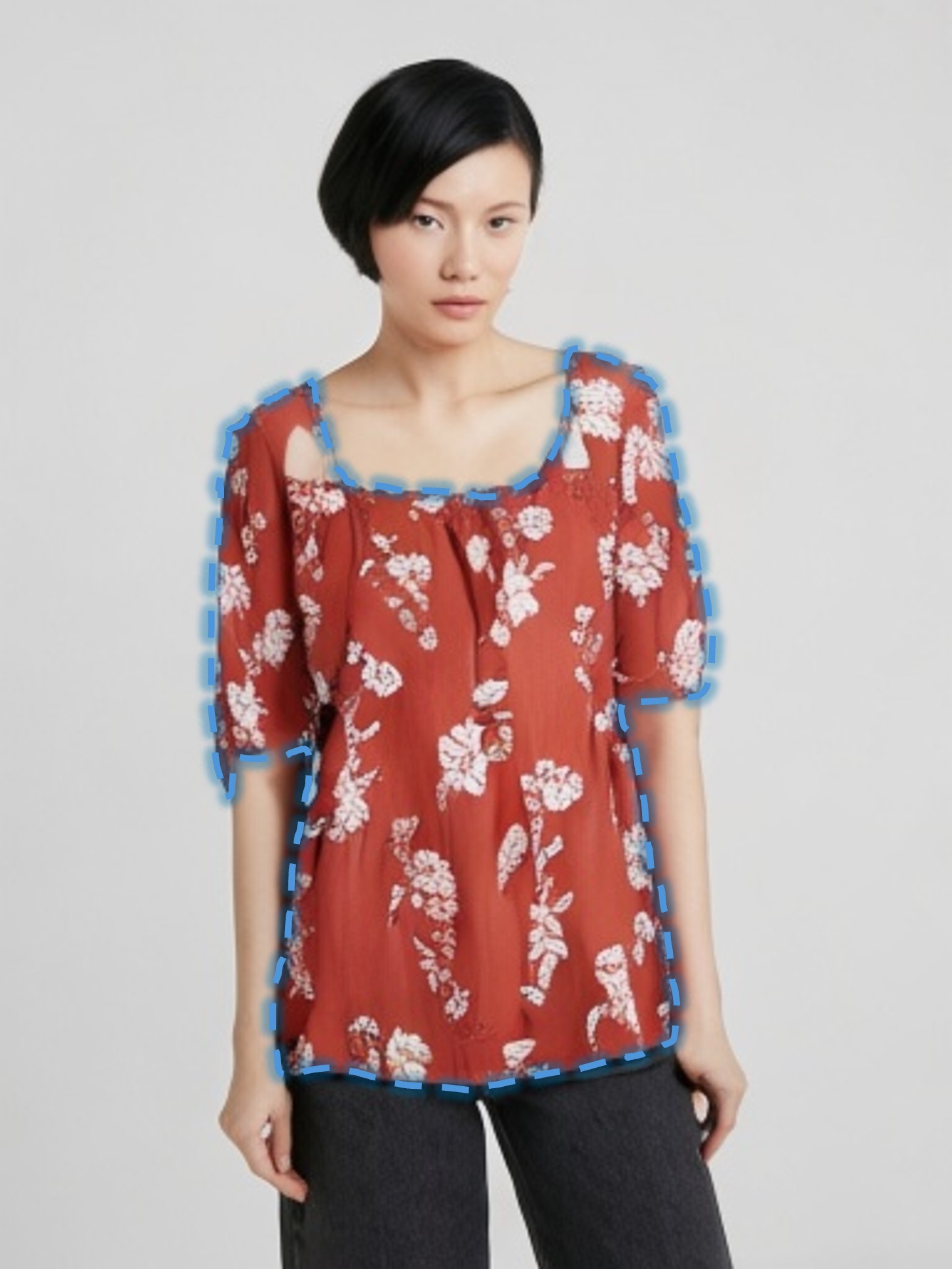} &&
\includegraphics[width=0.16\linewidth]{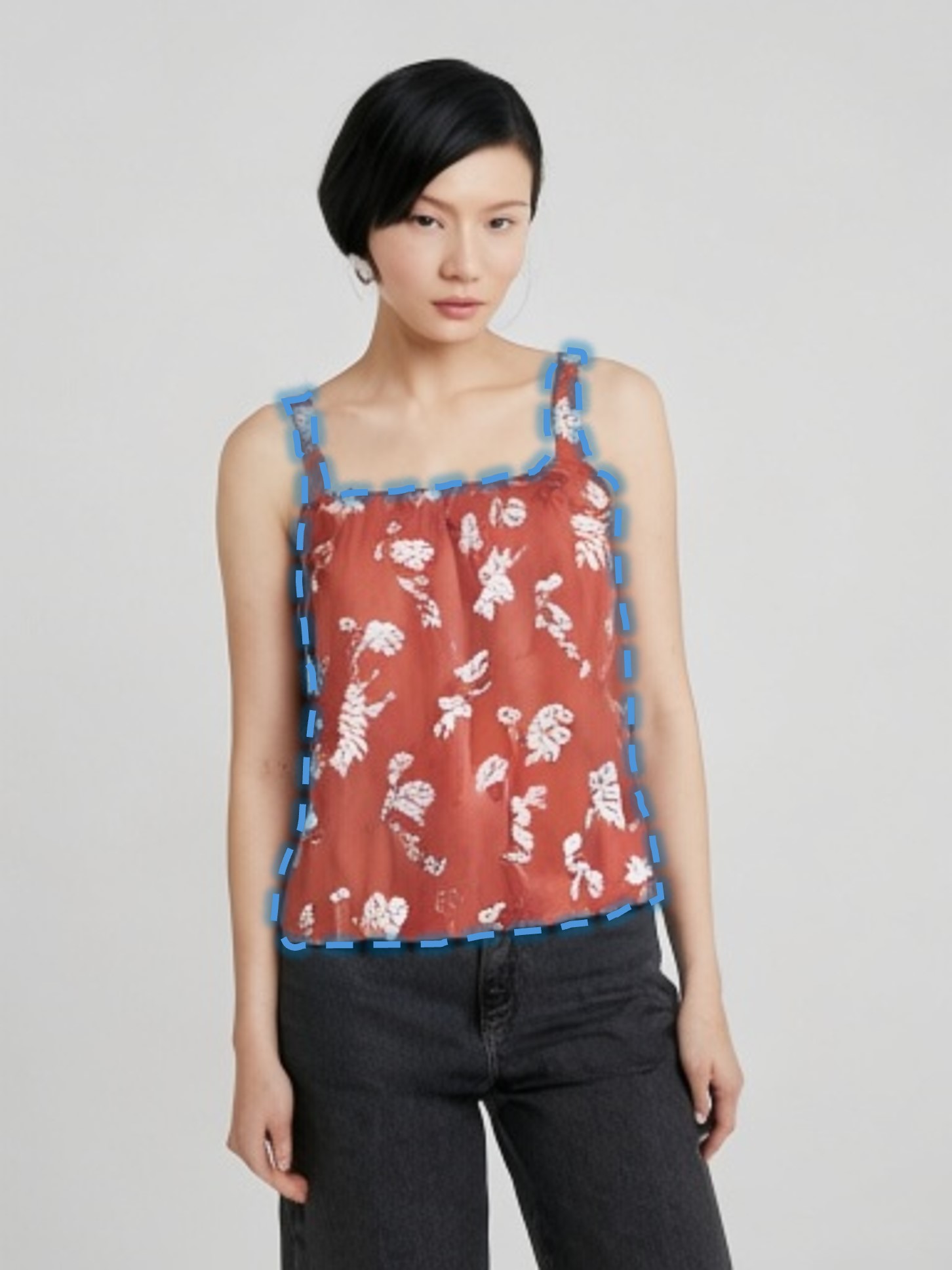}\\
\includegraphics[width=0.099\linewidth]{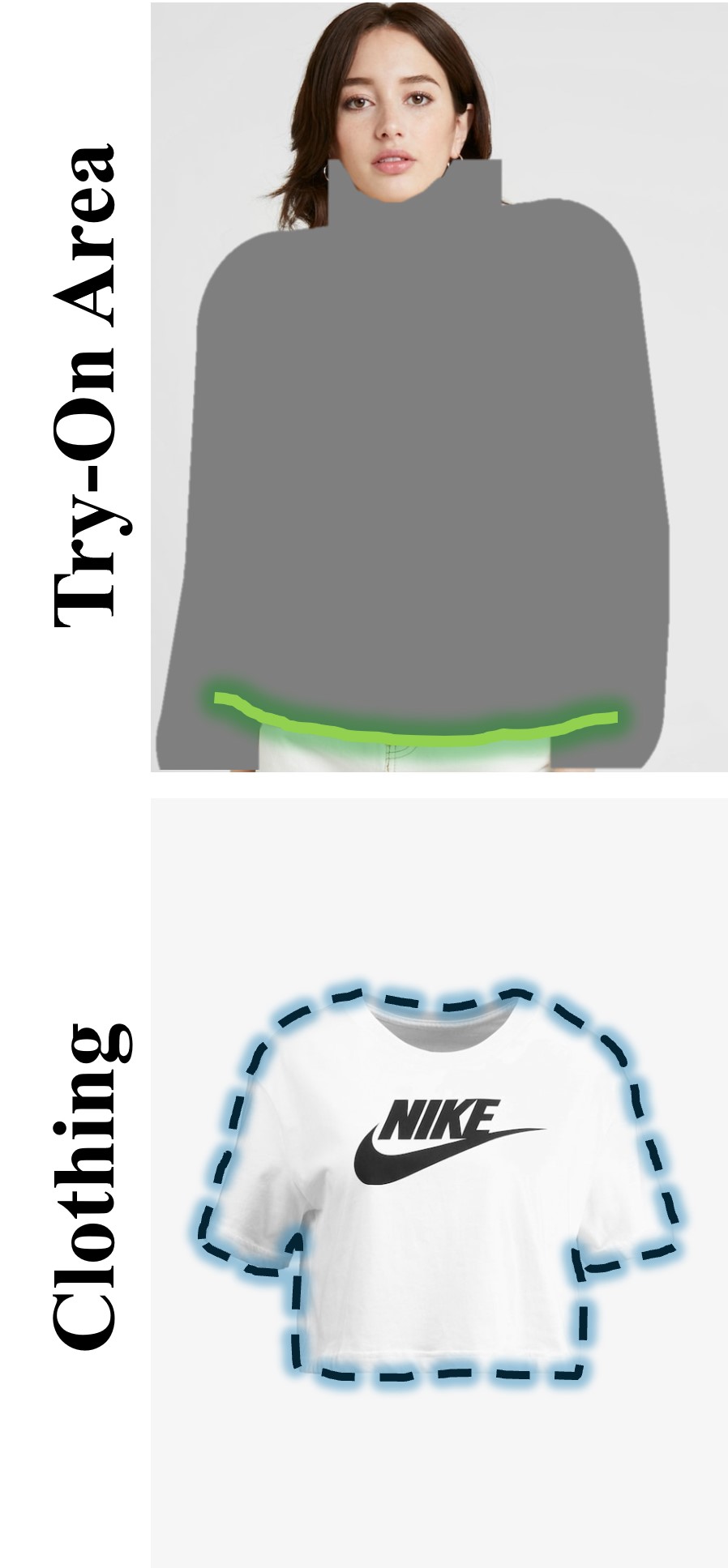} & 
\includegraphics[width=0.16\linewidth]{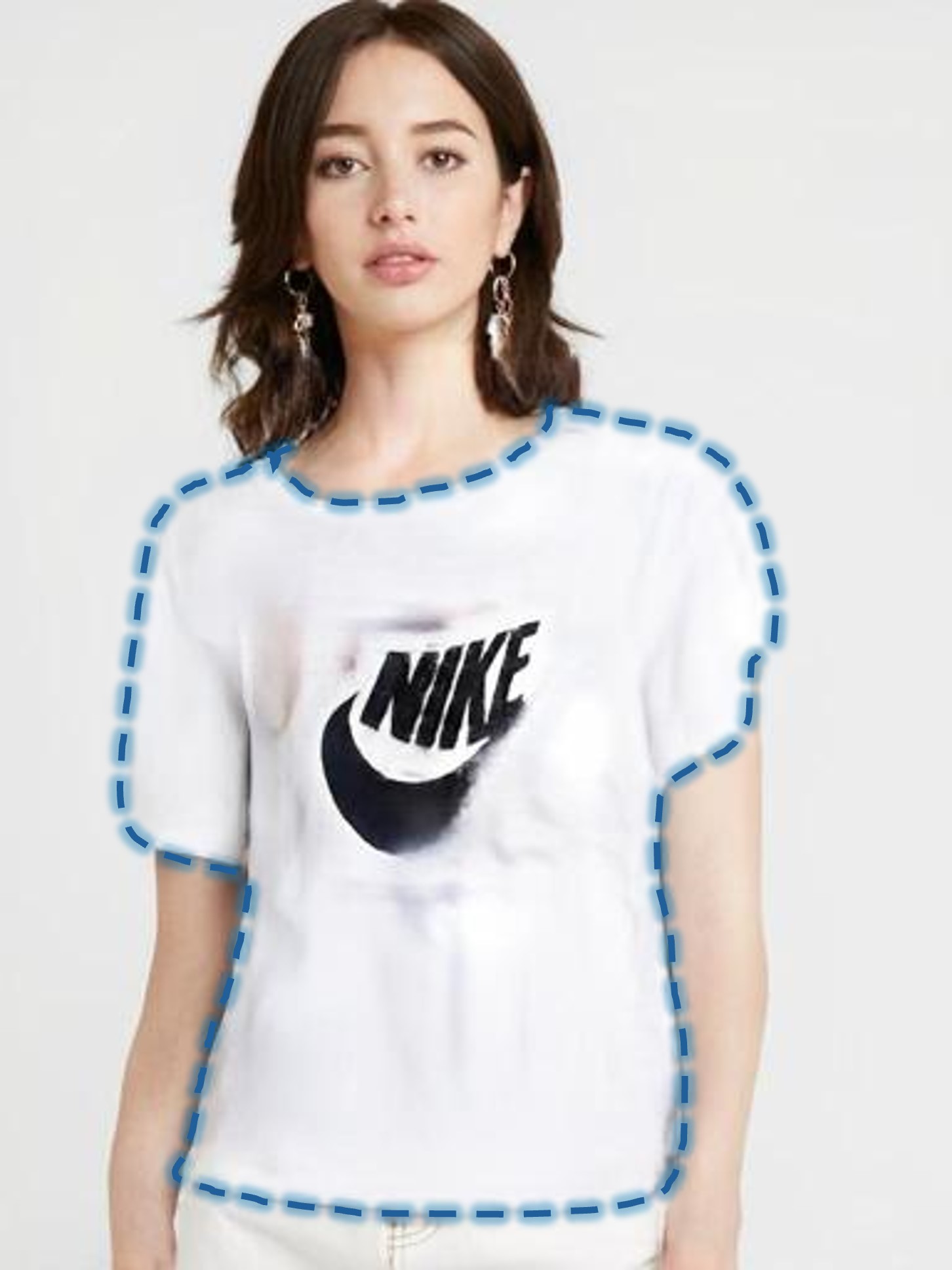} &
\includegraphics[width=0.16\linewidth]{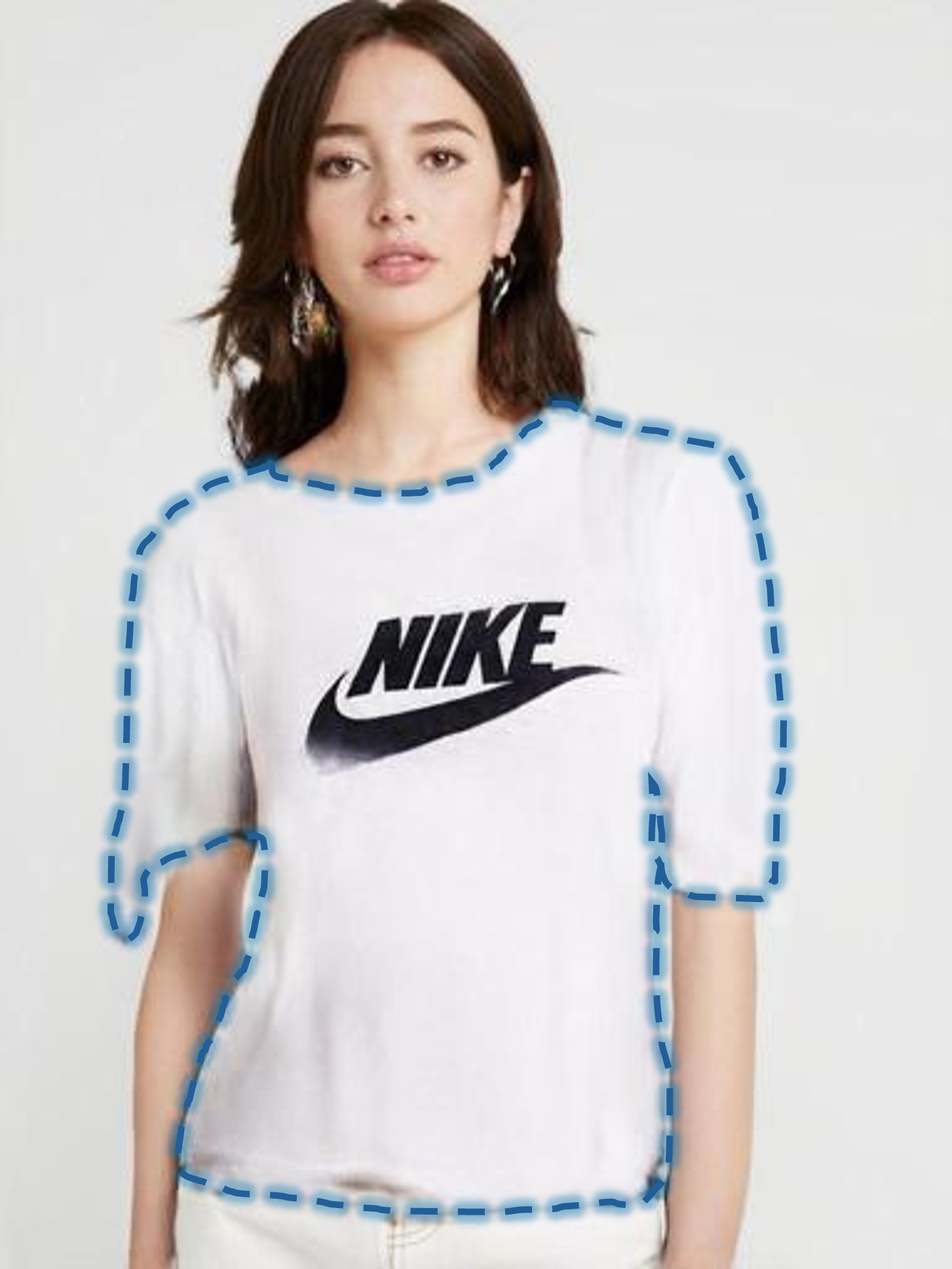} & 
\includegraphics[width=0.16\linewidth]{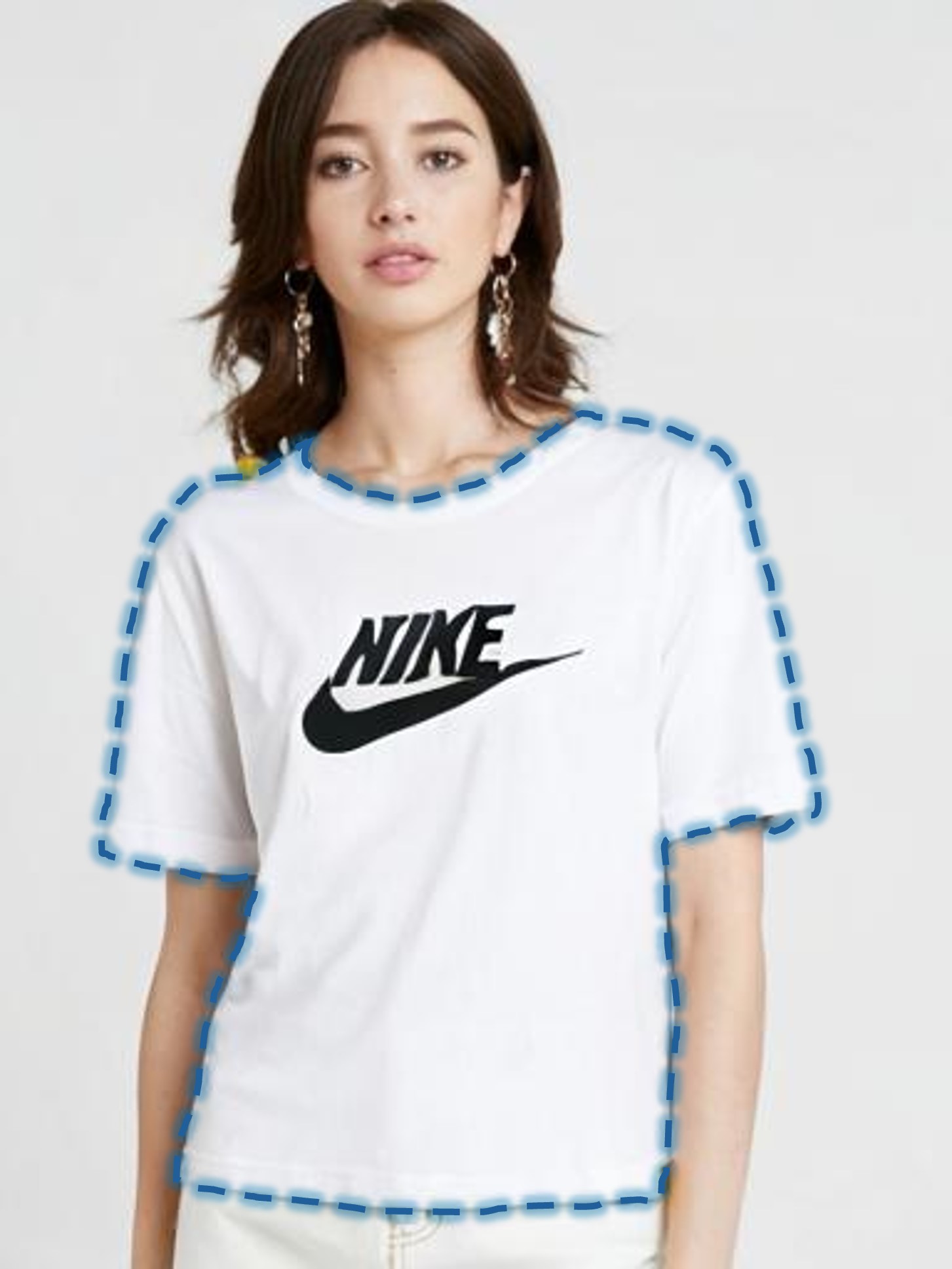} &
\includegraphics[width=0.16\linewidth]{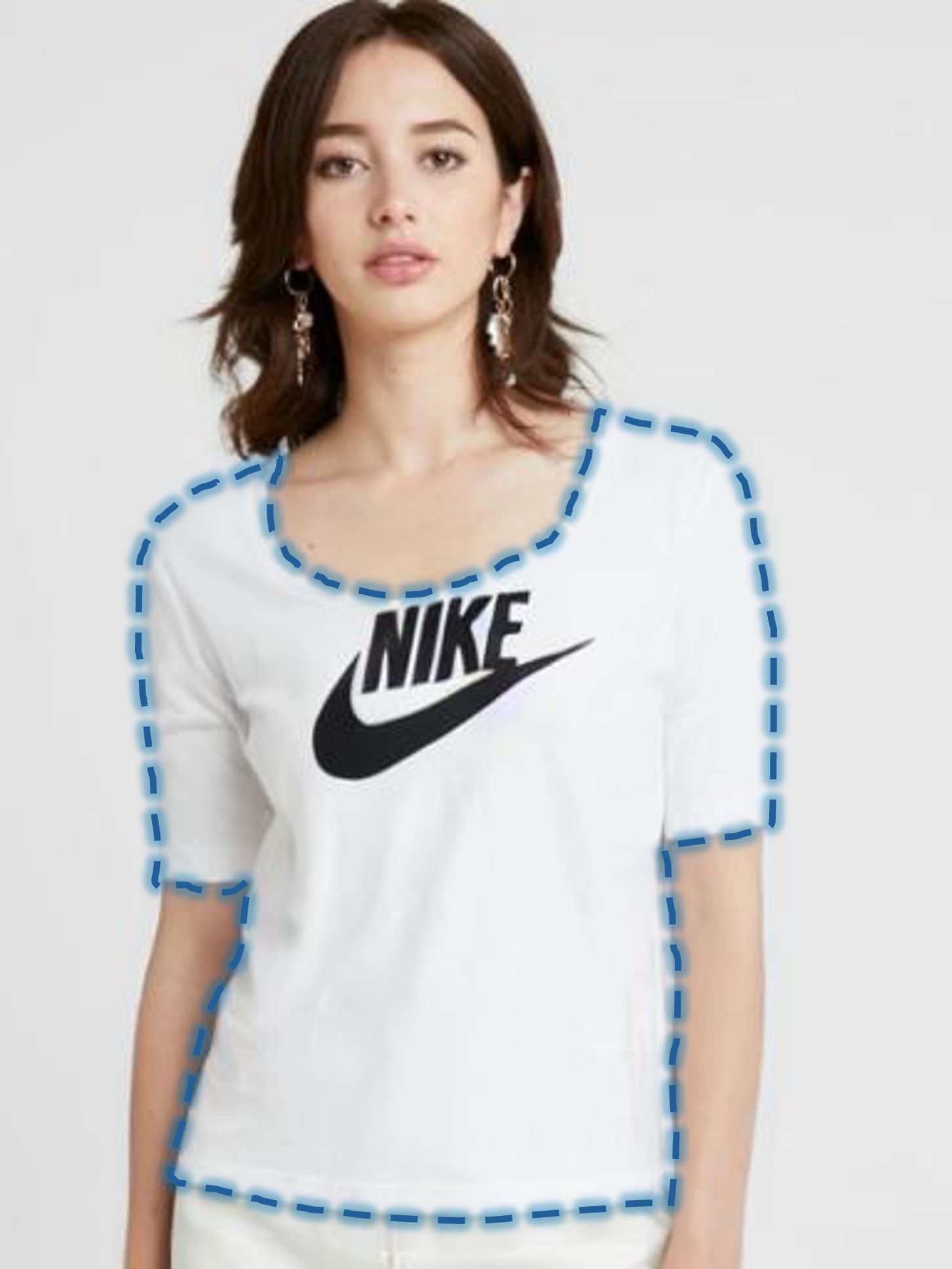} &
\includegraphics[width=0.16\linewidth]{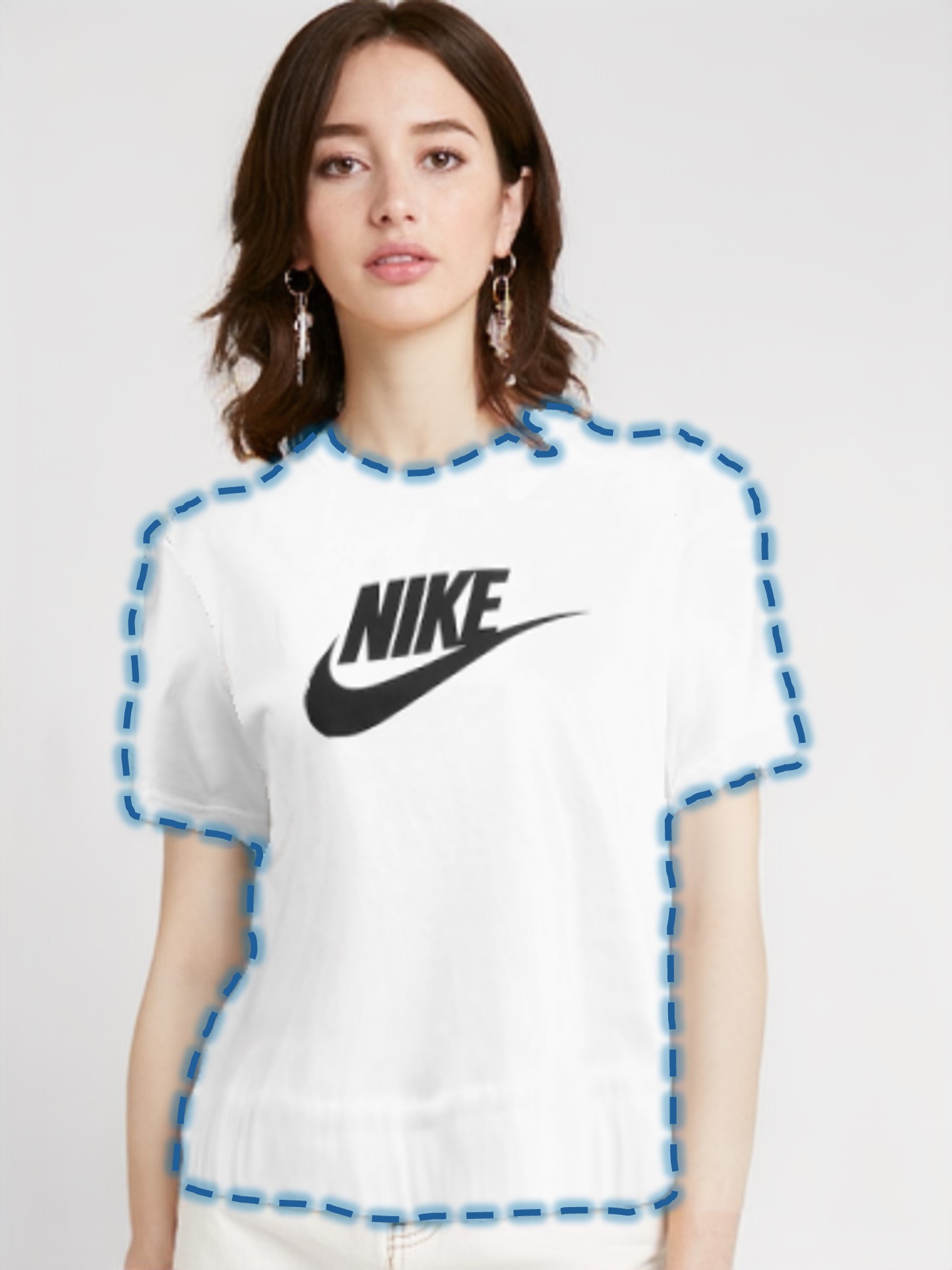} &
\includegraphics[width=0.16\linewidth]{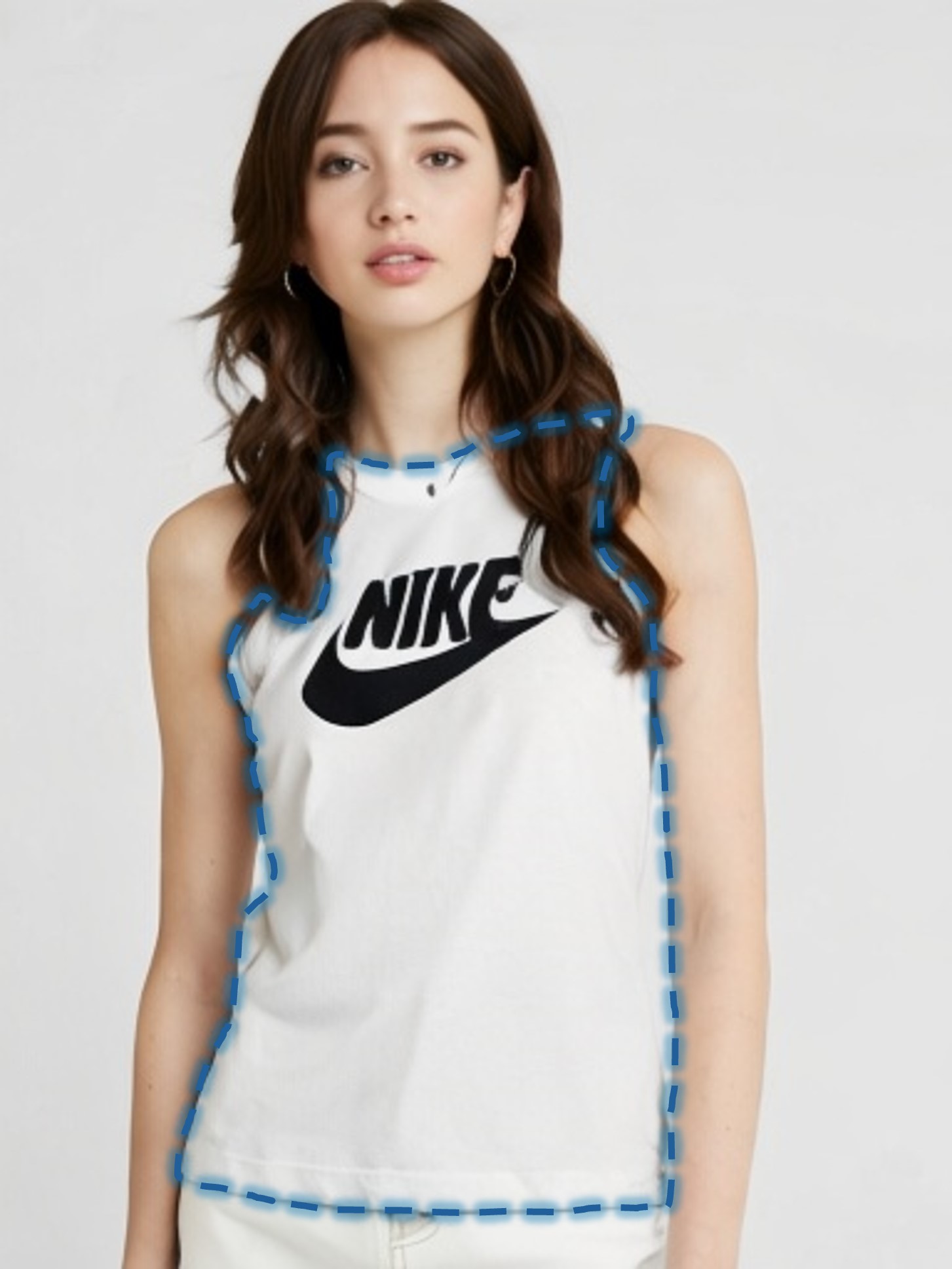} &&
\includegraphics[width=0.16\linewidth]{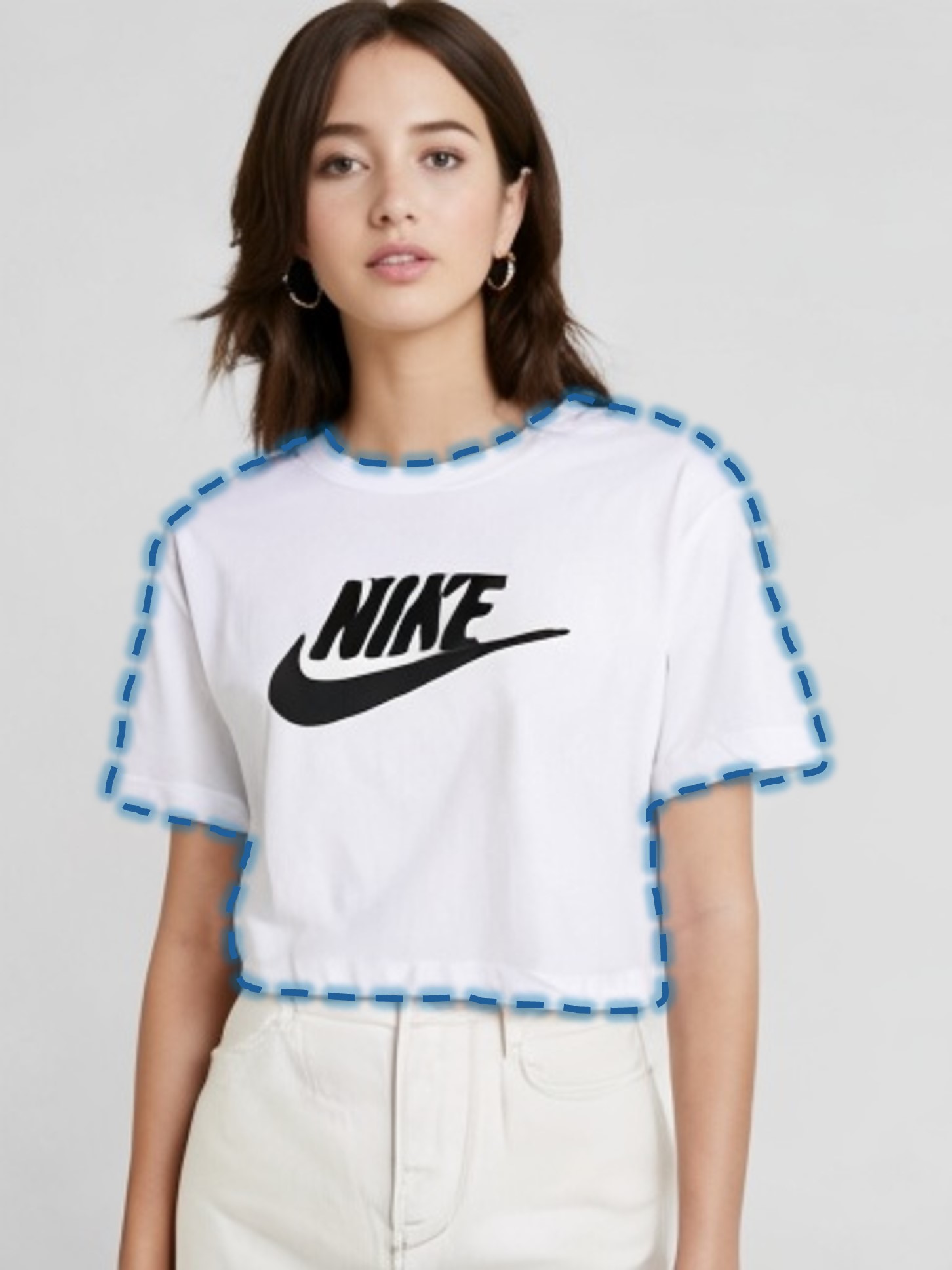}\\
\end{tabular}
}
\caption{
Existing methods change the type of target clothing to match the try-on area which is masked according to the original clothing, whereas our method breaks the correlation between the try-on area and original clothing during training and accurately inpaint the try-on area with clothing type preserved.
}
\label{fig:main}
\end{figure*}

Existing methods are trained with paired data (Fig. \ref{fig:paired_unpaired} Left) consisting of a clothing image and a model image wearing the same clothes. To facilitate the unpaired situation of trying different clothes (Fig. \ref{fig:paired_unpaired} Right), i.e., the practical scenario, VITON \cite{Viton} first proposes the clothing-agnostic person representation that discards the texture and obscures the contour of original clothes in the model image. Subsequent GAN-based methods \cite{CP-VTON,CP-VTON+,ACGPN,PFAFN,VITON-HD,HR-VITON,GP-VTON,StableVITON} adopt similar representation, and improve final generation quality in aspects of clothing warping and try-on sythesis. With the recent advancement of diffusion models \cite{DDIM,DDPM,IDDPM,LDM} in the image generation field, some studies \cite{LaDI-VTON,DCI-VTON,MGD} design diffusion-based models for virtual try-on, furthermore lifting the realism of generated images. In spite of the satisfying image realism, as shown in Fig. \ref{fig:main}, previous methods change the type of target clothing to match the try-on area, which goes against the objective of virtual try-on. 

The key reason of changing clothing types attributes to that the try-on area is affected by the original clothing that is desired to take off. Forcing constraints in clothing warping \cite{ACGPN} could keep target clothing away from excessive distortion, but also hinders flexible deformation particularly when facing challenging postures. Paying attention to the width-height ratio of clothing and perform truncation \cite{GP-VTON} could avoid squeezing clothing, but cannot deal well with various clothing types.
We propose a novel adaptive mask training paradigm to break the correlation between the try-on area and the original clothing, which  corrects the bias caused by paired training data, thus training the model to learn more accurate semantic correspondences for inpainting. 
As shown in the right-most column of Fig. \ref{fig:main}, the proposed strategy significantly enhances the performance of unpaired try-on.

Apart from the lack of effective approach to preserve clothing types in unpaired try-on, current benchmarks \cite{VITON-HD,Viton,DressCode} and evaluation metrics \cite{FID,CleanFID,KID,SSIM,LPIPS} also have shortages. Existing benchmarks do not cover sufficient kinds of unpaired try-on situations. Previous metrics \cite{FID,CleanFID,KID} are designed to evaluate general image generation tasks and are not fully capable of penalizing incorrect unpaired try-on results. 
Faced with different appearances between the generated result and the real try-on state in unpaired situation (Fig. \ref{fig:paired_unpaired} Right), how to objectively evaluate the type and texture of clothing during virtual try-on is challenging.

 
To evaluate the correctness of clothing type for unpaired try-on, we propose the Semantic-Densepose-Ratio (SDR) metric, which compares the area of target clothing relative to the model body. To assess the accuracy of clothing texture for unpaired try-on, we propose the Skeleton-LPIPS (S-LPIPS) metric, which makes visual comparisons at key semantic positions. Additionally, to comprehensively evaluate unpaired performance with various try-on situations, a cross-try-on benchmark (named as Cross-27) is constructed. We select 27 samples with different clothing types and model physiques, and obtain 729 try-on combinations with distinctive features.

Our contributions can be summarized as follows:
\begin{itemize}
    \item We propose a novel adaptive mask training paradigm to make the model capable of more accurate correspondence between target clothing and model images, leading to the superior performance of preserving clothing types.
    \item We propose two novel metrics to objectively compare the type and texture of target clothing in virtual try-on, overcoming the challenges of unpaired evaluation such as different poses and visual appearances. 
    \item We construct Cross-27, a new evaluation benchmark, with complex try-on situations and conduct comprehensive experiments to validate the effectiveness of the proposed methods.
\end{itemize}
\section{Related Work}

\subsection{Image Based Virtual Try-On}
Human-centered conditional image generation can be classified according to conditions, such as the text modality \cite{text_to_image_syntheis} and the image modality (e.g., pose \cite{pose_guided_image_generation}). Image-based virtual try-on takes the target clothing image as a condition and also the model image as input, which aims to synthesize an image where the model naturally wears target clothes.
\textcolor{black}{Early virtual try-on methods\cite{Viton,CP-VTON,CP-VTON+,ACGPN,VITON-HD,HR-VITON,PFAFN,style-flow} are based on GANs\cite{GANs}. These methods divide virtual try-on into two parts: clothing warping\cite{FlowNet,TPS,STN} and the fusion of warped clothing with the model. After Diffusion\cite{beats_gan} model demonstrated stronger image generation capabilities than GANs, works\cite{LaDI-VTON,DCI-VTON} use pretrained Diffusion models\cite{LDM,PBE} as the fusion module to generate final try-on images. However, limited by the performance of the clothing warping module, when model pose are complex, errors in clothing warping can propagate to the final try-on results. Recent methods\cite{TryonDiffusion,StableVITON} seek to accomplish virtual try-on solely using Mask Inpainting Diffusion models, directly generating the clothing and body in the try-on state within the mask area while preserving the content outside the mask. Existing methods do not consider the differences in try-on between training and inference, leading to the model erroneously interpreting the lower boundary of the mask area as the lower boundary of the target clothing, as illustrated in Fig. \ref{fig:paired_unpaired}. We successfully addressed this flaw by applying adaptive masks for the training samples during the training process.}

\subsection{Metrics For Virtual Ty-On}
\textcolor{black}{The metrics for virtual try-on can be divided into paired and unpaired metrics, corresponding to paired and unpaired try-on, respectively. Paired metrics include many mature and effective evaluation indicators: LPIPS\cite{LPIPS} and SSIM\cite{SSIM} use the pixel similarity between the generated image and the real image to reflect the quality of the try-on results, while Semantic IoU\cite{CP-VTON+,Dior,GP-VTON} uses the accuracy of clothing semantic regions to reflect the quality of clothing warping. In the case of unpaired try-on, where ground truths are lacking, it is not possible to simply evaluate the quality of each generated result. Earlier methods\cite{Viton,CP-VTON+} used the Inception Score\cite{IS} (IS) as a metric to evaluate the quality of unpaired try-on results. IS can reflect the diversity of the try-on images but cannot reflect the accuracy of the try-on. Subsequent work\cite{StableVITON,DCI-VTON,LaDI-VTON} has increasingly adopted FID\cite{FID} and KID\cite{KID} to evaluate unpaired results. FID and KID assess the similarity between the unpaired try-on images and paired real try-on images, indirectly reflecting the authenticity of the try-on images. To fill the lacking in the field of unpaired metrics, we introduce the SDR and S-LPIPS metrics, which directly evaluate the try-on image by referencing the real try-on images of the target clothing.}
\begin{figure*}[t]
\centering
\includegraphics[width=0.8\linewidth]{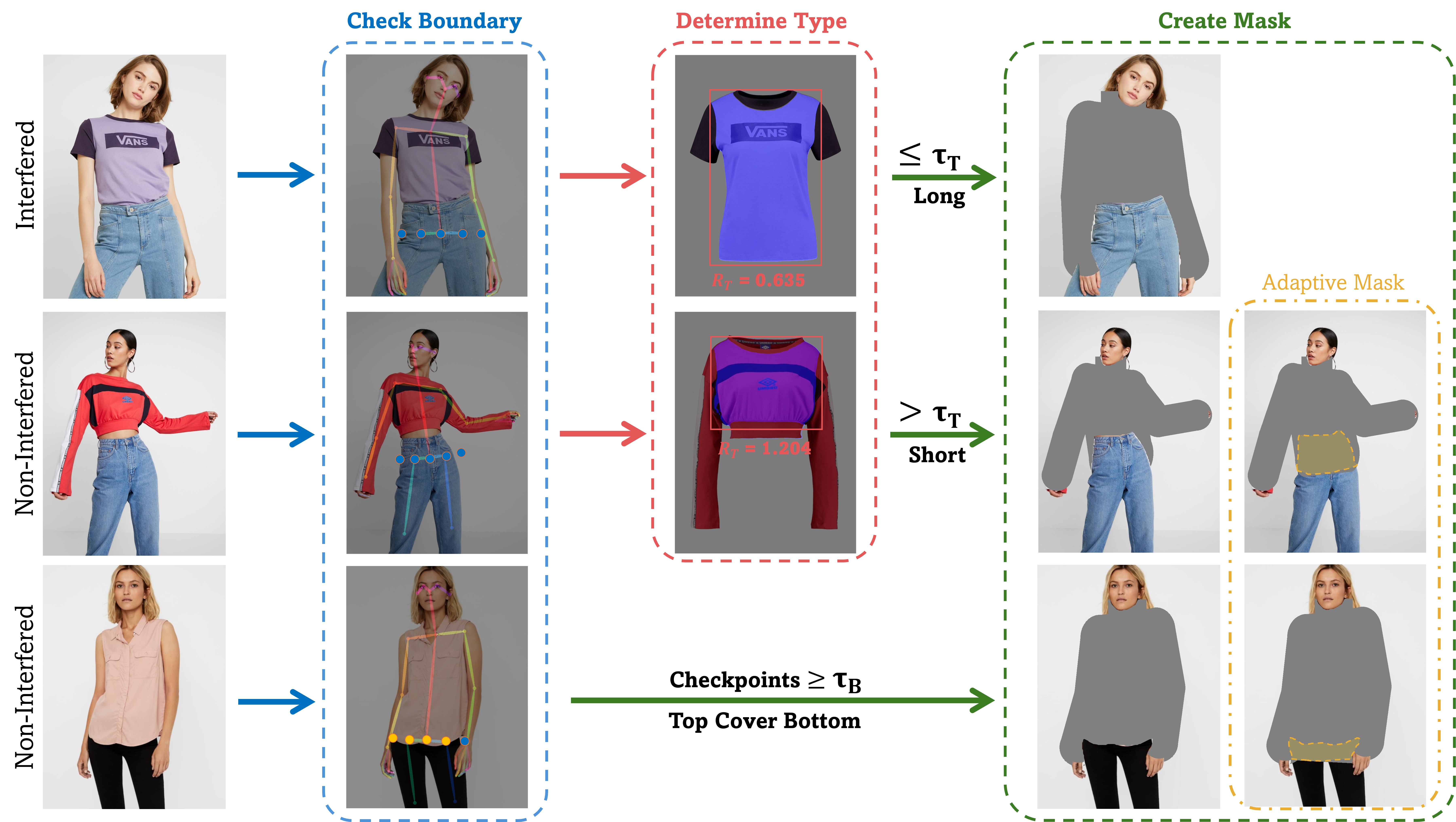}
\caption{The Adaptive Mask Maker consists of three steps. \textcolor[rgb]{0,0.4,0.8}{Check Boundary} step uses five checkpoints around the waist to identify samples where the top clothing covers the bottom clothing. \textcolor[rgb]{0.9,0.35,0.35}{Determine Types} step calculates the aspect ratio of the clothing's torso part to assess if the clothing is of a long type, thereby determining if the top is interfered. \textcolor[rgb]{0.25,0.6,0.2}{Create Mask} step define the final area of the mask.}
\label{fig:metric_maker}
\end{figure*}
\section{Adaptive Mask Inpainting for Virtual Try-On}

\textcolor{black}{During training, we adaptively adjust the mask area based on the wearing style of the samples to simulate the unpaired try-on scenario as closely as possible. We classify the training samples into two different wearing styles based on whether the top is interfered by the bottom, termed as \textit{Interfered} and \textit{Non-Interfered}. The $1^{st}$ row in Fig. \ref{fig:metric_maker} shows examples of the \textit{Interfered} style, where part of the top is tucked into the bottom, and thus the lower boundary of the top is determined by the bottom. The $2^{nd}$ and $3^{rd}$ rows in Fig. \ref{fig:metric_maker} display the \textit{Non-Interfered} wearing style. In addition to applying the same mask to all samples as in previous works\cite{StableVITON}, we also apply an adaptive mask to \textit{Non-Interfered} samples to simulate the scenario in unpaired try-on where there is a gap between the lower boundary of the target clothing and the lower boundary of the mask area.}

\subsection{Adaptive Mask Maker}

\textcolor{black}{We designed a pipeline that automatically determines the wearing style of the top clothing to create adaptive mask. We divide the mask creation process into three steps: (1) Check Boundary, Determine Types, and Create Mask, as shown in Fig. \ref{fig:metric_maker}. (1) Check Boundary involves using five checkpoints around the waist to determine whether the top clothing covers the bottom clothing. If the top covers the bottom, it can be directly classified as \textit{Non-Interfered}. (2) For cases where the top does not cover the bottom, we further determine the wearing style based on the long/short type of the clothing. (3) We create different masks based on the identified wearing styles.}

\textcolor{black}{For checking boundary, We first utilize the openpose\cite{OpenPose} parser to obtain three key points around the waist, and then interpolate an additional key point on each side, resulting in a total of five points used as checkpoints. The preliminary judgment on whether the bottom interfere with the top is made by checking if these five points fall within the semantic area\cite{CIHP} of the top. If more than $\tau_B$ checkpoints ($cp$) are in the area of the top, the wearing style is classified as \textit{Interfered}. Otherwise, we need to proceed to the next step for further analysis.}

\textcolor{black}{For determine type, we further assess whether this is due to the clothing's type being relatively short, which results in it not covering the checkpoints. We use the aspect ratio of clothing's torso part as a parameter $R_{T}$ for its type. If the ratio is greater than $\tau_T$, the clothing is considered to be a short type; otherwise, it is considered as a long type. If the clothing is short, we consider that the bottom non-interfere with the top. The value $\tau_T$ is the average aspect ratio of the torso part of the models in the dataset.}
\begin{equation}
\text{Wearing Style} = 
\begin{cases} 
  \text{~Non-Interfered}, &  \text{cp} > \tau_B \\
  \text{~Non-Interfered}, &  \text{cp} \leq \tau_B, ~R_{\text{T}} \geq \tau_T\\
  \text{~Interfered}  ,   &  \text{cp} \leq \tau_B, ~R_{\text{T}} < \tau_T
\end{cases}
\end{equation}

\textcolor{black}{For creating mask, building on the previous mask method\cite{VITON-HD}, we mask the entire upper body of the model. For the \textit{Interfered} wearing style, we completely preserve the bottom clothing area according to its semantic segmentation. For \textit{Non-Interfered}, we extended the lower boundary of the mask area downwards, thereby simulating the scenario where a gap between the unpaired clothing area and the mask area.}
\begin{figure*}[t]
\centering
\includegraphics[width=0.8\linewidth]{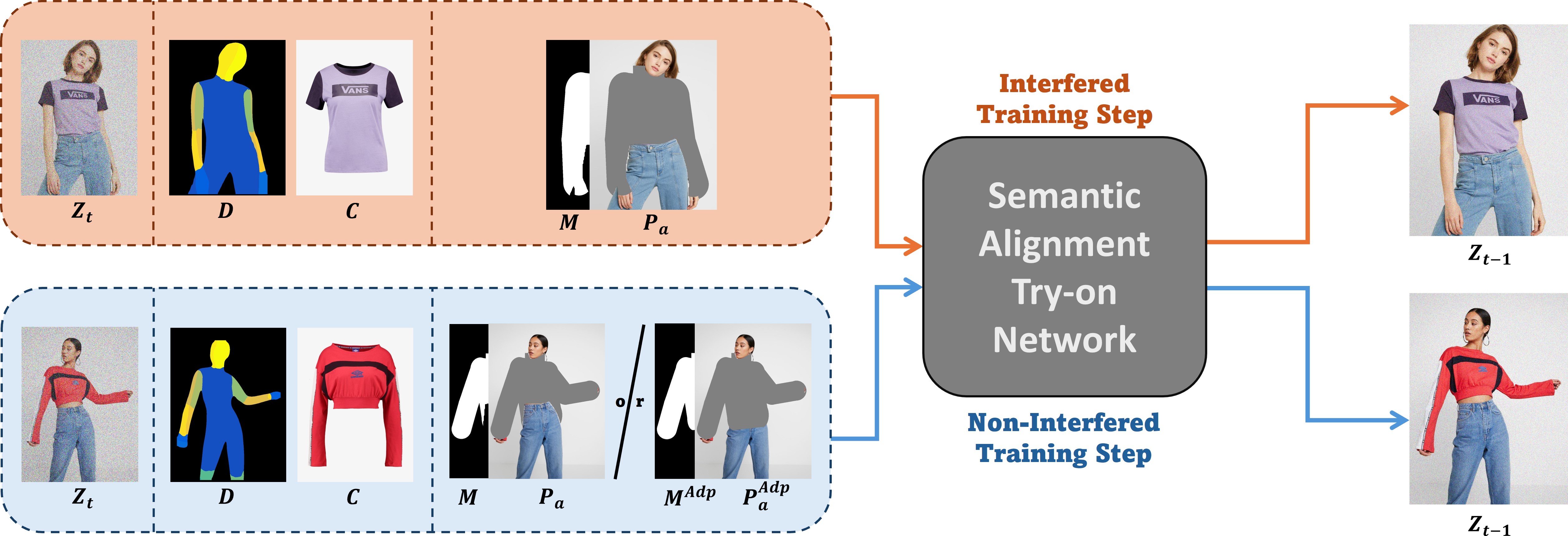}
\caption{For the \textcolor{orange}{Interfered} and \textcolor{blue}{Non-Interfered} wearing style, we input different Masked Person $\{M, P_a\}$ during training. In the Interfered training step, we use $\{M, P_a\}$ with the bottom clothing fully retained; in the Non-Interfered training step, we randomly use $\{M^{Adp},  P_a^{Adp}\}$ with parts of the bottom clothing eliminated.}
\label{fig:model}
\end{figure*}
\subsection{Adaptive Mask Training Paradigm}
\textcolor{black}{
To validate the effectiveness of the Adaptive Mask Training Paradigm, we selected StableVITON\cite{StableVITON} as our base model and trained it using our paradigm. StableVITON is a semantic alignment try-on network, which leverages the knowledge in pre-trained image generation model \cite{PBE} based on semantic alignment via cross-attention between target clothing and the model body. Different from StableVITON that inpaints the try-on area with clues only from clothing image, our paradigm enabling the network to learn more accurate semantic correspondences and automatically repair the gap between target clothing and mask area. }

\textcolor{black}{The input to the network consists of a latent noise $Z_t$ combined with try-on conditions, which include a densepose $D$, a clothing image $C$ and an masked Person $\{M, P_a\}$. The mask $M$ indicates the area where the mask operation is applied, while agnostic Person $P_a$ is used to retain pixel information in non-try-on areas. This approach ensures that important features of the person are maintained while allowing for editing in designated areas.} 
\begin{figure}[t]
\centering
\includegraphics[width=1.0\linewidth]{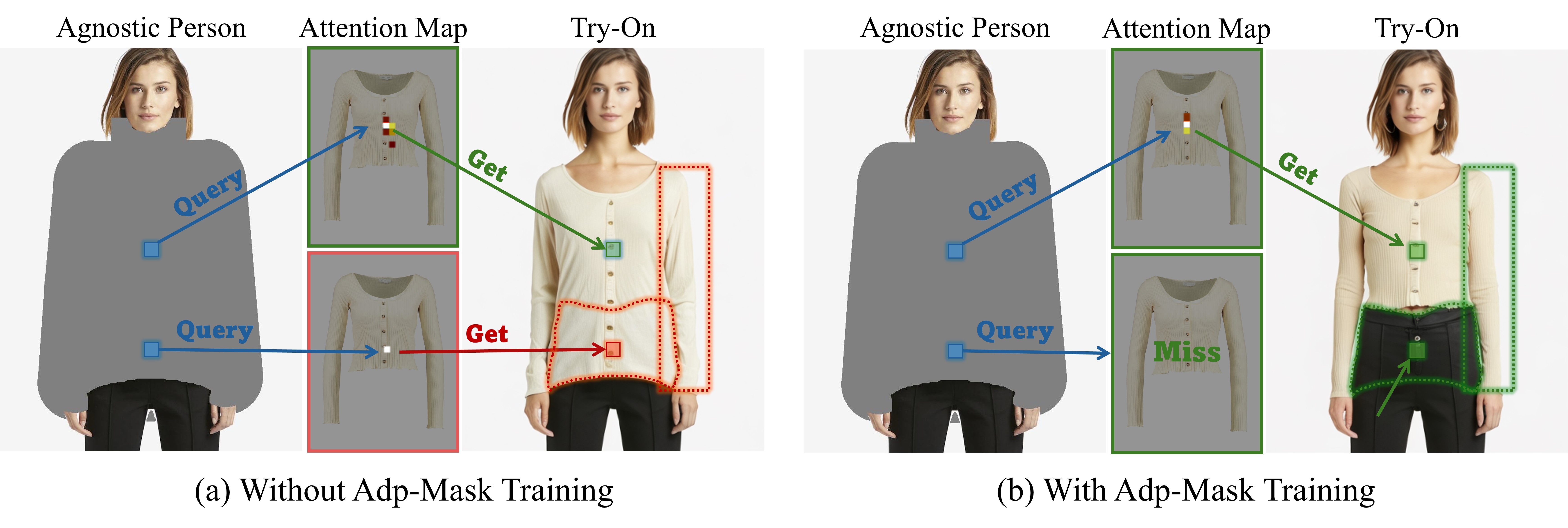}
\caption{The correspondence between model body semantics and clothing semantics. After Adp-Mask Training, the network faithfully retains clothing details within body areas aligned with clothing semantics and naturally repairs areas that are not aligned.}
\label{fig:miss_condition}
\end{figure}

As shown in Fig. \ref{fig:model}, we categorize the training steps based on the wearing style of the training samples into \textit{Interfered} and \textit{Non-Interfered} training steps. During the \textit{Interfered} training step, we use $\{M, P_a\}$ that preserves the information of upper boundary of the bottom clothing. In the \textit{Non-Interfered} training step, we replace the $\{M, P_a\}$ with an enhanced Mask Person $\{M^{Adp}, P_a^{Adp}\}$ that made by the adaptive mask maker with a probability $p$. In the context of adp-mask training, the network is required to determine the lower boundary of the clothing based on its type and inpaint the content at the junction of the top and bottom clothing. When there is a gap between the top and bottom clothing, network should repair the gap area by utilizing the information from the retained part of the bottom clothing. 

As shown in Fig. \ref{fig:miss_condition}, after adptive mask training, the model no longer samples from clothing features across the entire mask area. Instead, it initially discerns the boundaries between the top and bottom clothing areas, and subsequently, only the body semantics of the top clothing area exhibit high correlation with the clothing semantics. In addition to generating more accurate shapes of clothing, but the network also achieves greater accuracy and naturalness in generating the details of clothing textures. The folds in the clothing on the inner side of the model's arm are more in line with the original type of clothing.

\section{Evaluation Metrics for Unpaired Try-On}

\textcolor{black}{To quantitatively analyze the virtual try-on results for unpaired samples, we introduced two metrics: Semantic Densepose Area Ratio (SDR) and Skeleton-Learned Perceptual Image Patch Similarity (S-LPIPS). Due to the absence of ground truth in unpaired try-on, it is not feasible to evaluate the quality of try-on result by calculating the similarity on a pixel-by-pixel basis between the generated image and the ground truth, such as SSIM\cite{SSIM} and LPIPS\cite{LPIPS}. However, we can obtain the semantic relationship between clothing and the model body from real try-on images, and then assess the accuracy of the virtual try-on by comparing the similarity of the relationship between clothing and the model body in both virtual try-on images and real try-on images. We divide the accuracy of the semantic relationship into two aspects: the accuracy of the overall clothing types and the accuracy of the clothing content at key positions. Correspondingly, we use SDR to calculate the accuracy of the clothing types and S-LPIPS to measure the accuracy of the clothing content. This approach allows for a more nuanced assessment of virtual try-on quality in the absence of ground truth.}

\begin{figure}[t]
\centering
\includegraphics[width=\linewidth]{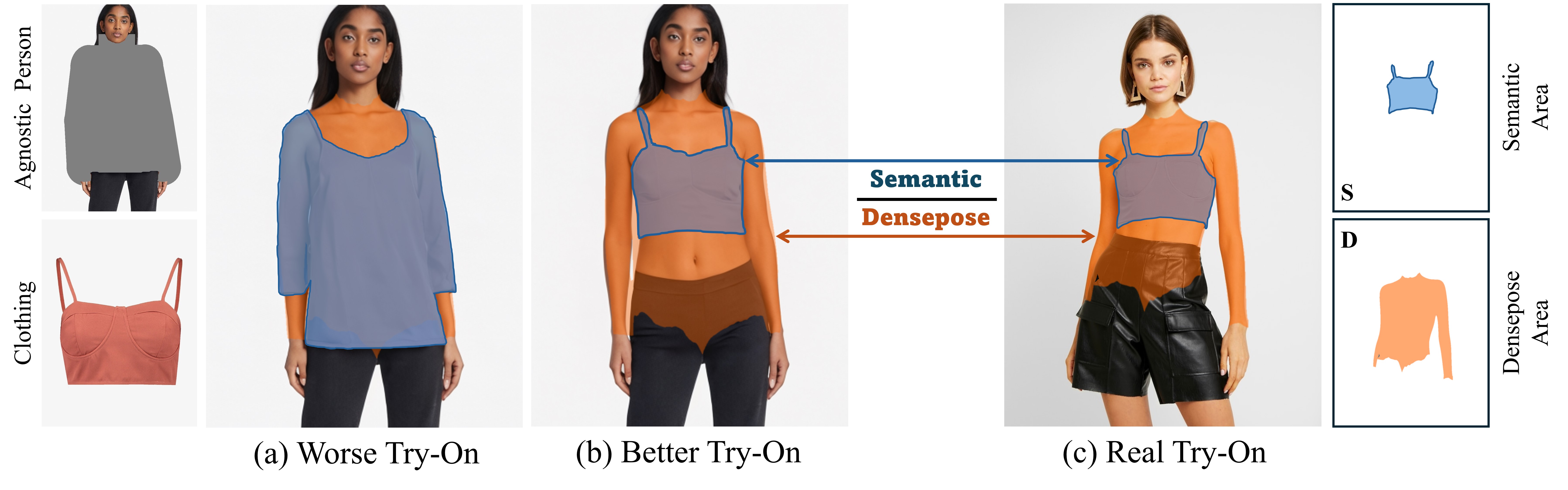}
\caption{The area ratio of the \textcolor{blue}{clothing semantic region} (Semantic) to the \textcolor{orange}{body region} (Densepose) can reflect the types of the clothing. During the try-on process, the type of the clothing should remain unchanged, thereby the SDR should remain similar to that of real try-on.}
\label{fig:matric_sdr}
\end{figure}
\subsection{Semantic Densepose Ratio}
\textcolor{black}{The accuracy of the clothing types is the most intuitive reflection of try-on, which can be roughly defined by the area of the clothing covering the region of the body (represented by the Densepose\cite{DensePose}) and the area of the semantic\cite{CIHP} region of the clothing. As shown in Fig. \ref{fig:matric_sdr}, taking the try-on of a top as an example, for a human body image in a fixed pose, the upper body area (the orange area) that can be covered is fixed. At this time, the semantic area (the blue area) of the top directly reflects the types of the clothing. As shown in the comparison between Fig. \ref{fig:matric_sdr}a and Fig. \ref{fig:matric_sdr}b, there is a clear difference in the semantic area between a long-sleeved top and a tank top, leading to different ratios of the clothing area to the upper body area. For a specific piece of clothing, it should maintain the same type on different bodies, that is, a similar ratio of clothing area to body area. Based on this causal relationship, we use the following formula to define the clothing types:}
\begin{equation}
    SDR = \frac{S}{D} 
\end{equation}
\noindent\textcolor{black}{Given that $S$ represents the area of the clothing semantic region, and $D$ represents the area of the upper body region of the model. When using the SDR to calculate the distance of the same clothing tried on two different model bodies, we can use the following formula:}
\begin{equation}
    SDR\ Distance = \alpha \cdot \beta \cdot \left| \frac{S_1}{D_1} - \frac{S_2}{D_2} \right|
\end{equation}
\noindent\textcolor{black}{We introduce correction factors utilizing the overlapping area between $S$ and $D$, denoted as $S \cap D$. \( \alpha = \frac{D}{S \cap D} \) represents the fabric area factor, which can amplify the distance between the SDR values of two results when the clothing has a type that uses less fabric, thereby capturing more subtle differences. \( \beta = \frac{S \cap D}{S} \) represents the clothing fit factor, which can reduce the distance of SDR values when the clothing has a looser fit, thus providing a more lenient evaluation for looser type. In the testing of unpair try-on, we calculate the $\alpha$ and $\beta$ factors using $S_R$ and $D_R$ from the real try-on. The similarity metric between the Virtual try-on result and the Real try-on result, based on the SDR, can be represented by the equation \ref{sdr_equation}. Our experiments indicate that the SDR distance can effectively measure the accuracy of clothing type in unpaired try-on, with more results referenced in Section \ref{sec:Experiment}.}
\begin{equation}
    \begin{split}
    SDR\ Distance &= \frac{D_R}{S_R \cap D_R} \cdot \frac{S_R \cap D_R}{S_R} \cdot \left| \frac{S_R}{D_R} - \frac{S_V}{D_V} \right|\\&= \left|1 - \frac{D_R S_V}{S_R D_V}\right|
    \end{split}
\label{sdr_equation}
\end{equation}

\subsection{Skeleton Based LPIPS}
\label{sec:metric_LPIPS}
\textcolor{black}{The Learned Perceptual Image Patch Similarity (LPIPS\cite{LPIPS}) is an effective metric for measuring the similarity of image content, but it is sensitive to the spatial location of pixels within images. In unpaired try-on, where there are changes in the shape and posture of model, it is not feasible to directly calculate the LPIPS distance between unpaired results and real try-on results. However, we found that despite differences in posture and shape, the semantic correspondence between clothing and the human body remains unchanged (for example, the shoulder seam of a top should correspond to the position of the model's shoulder). By leveraging this invariant semantic correspondence, we can measure the accuracy of clothing content generation by calculating the LPIPS distance between clothing pixels at the same semantic locations of the body in both virtual try-on and real try-on images.}
\begin{figure}[t]
\centering
\includegraphics[width=\linewidth]{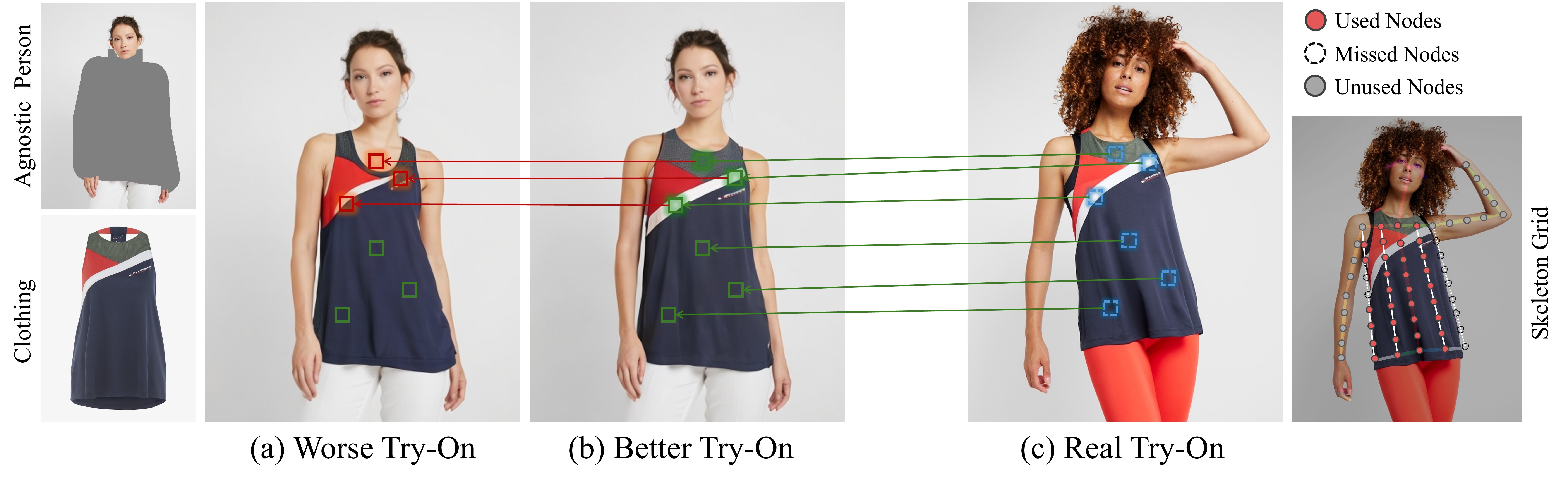}
\caption{The accuracy of clothing content can be assessed by measuring the similarity of pixels in areas near the same skeleton nodes. \textcolor[rgb]{0.2,0.7,0.2}{Green patches} indicate correct content, while \textcolor{red}{red patches} signify incorrect content.}
\label{fig:matric_slpips}
\end{figure}
\textcolor{black}{Specifically, taking the try-on of tops as an example, we first interpolate between the keypoints of the openpose\cite{OpenPose} skeleton to obtain a skeleton grid (as shown in Fig. \ref{fig:matric_slpips}c), which is composed of three parts: the left arm, the torso, and the right arm, including a total of 56 nodes (8+40+8=56). Then, we perform two rounds of node filtering: (1) Missed Nodes, filtering based on the foreground and background relationships of the human body, and removing nodes in positions that are obscured and not visible; (2) Unused Nodes, removing nodes outside the area of the clothing being tried on. For calculating the S-LPIPS distance, we segment out patch $p_i$ centered on the effective grid nodes $n_i$. The patches can be represented as $P\in\mathbb{R}^{H \times W \times N}$, where $W$ and $H$ represent the width and height of the patch, and $N$ denotes the number of effective nodes. As shown in Equation \ref{equation:slpips}, we calculate the LPIPS distance between corresponding semantic position patches and take the average as the S-LPIPS distance.}
\begin{equation}
S-LPIPS\ Distance=\frac{1}{5} \frac{1}{N} \sum_{j=1}^5 \sum_{i=1}^N \phi_j\left(p_R^i, p_V^i\right)
\label{equation:slpips}
\end{equation}
\noindent\textcolor{black}{Where $\phi_j$ represents the output of the $j^{th}$ layer of the VGG\cite{VGG} network, $p_R$ and $p_V$ respectively denote the same semantic patch in the Real try-on image and the Virtual try-on image. As shown in Fig. \ref{fig:matric_slpips}a and Fig. \ref{fig:matric_slpips}b, better try-on results' semantic patches that are closer to the real try-on results. In areas of the clothing's upper left corner where the information content is richer, the three green patches in the better try-on are generated correctly, while in the worse try-on, these three red patches have missing or incorrect information.}

\section{Experiment}
\label{sec:Experiment}

\subsection{Experimental Setup}

\begin{figure}[t]
\centering
\scriptsize
\setlength{\tabcolsep}{.2em}
\resizebox{\linewidth}{!}{
\begin{tabular}{ccc c ccc c ccc}
\multicolumn{7}{c}{\textbf{\textit{Non-Interfered}}} && \multicolumn{3}{c}{\textbf{\textit{Interfered}}} \\
\addlinespace[0.2cm]
\multicolumn{3}{c}{\textbf{Long}} && \multicolumn{3}{c}{\textbf{Short}} & & & &\\
\cmidrule{1-3} \cmidrule{5-7} \cmidrule{9-11}
\textbf{Sleeveless} & \textbf{Short Sleeves} & \textbf{Long Sleeves} && \textbf{Sleeveless} & \textbf{Short Sleeves} & \textbf{Long Sleeves} && \textbf{Sleeveless} & \textbf{Short Sleeves} & \textbf{Long Sleeves} \\

\includegraphics[width=0.16\linewidth]{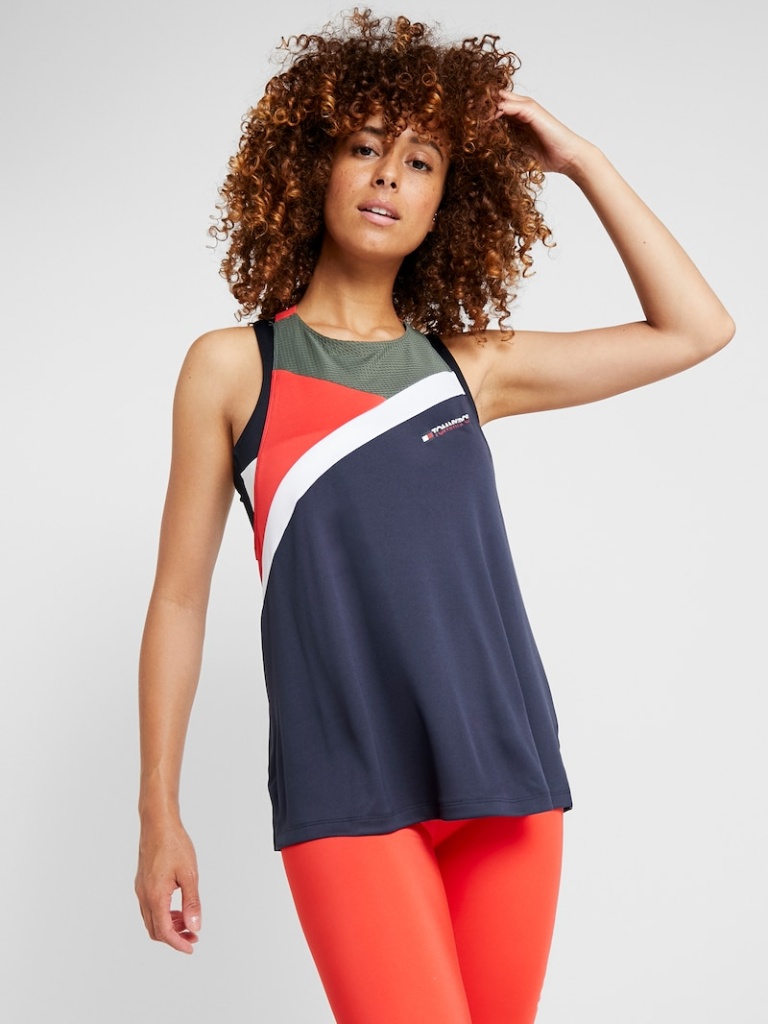} &
\includegraphics[width=0.16\linewidth]{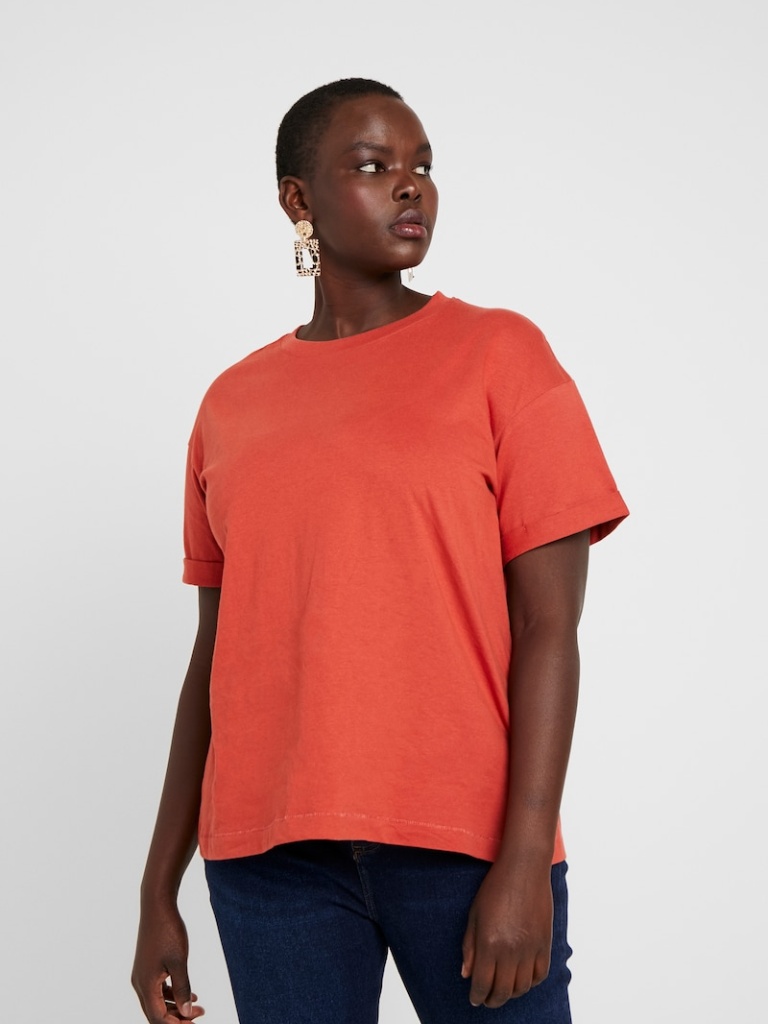} &
\includegraphics[width=0.16\linewidth]{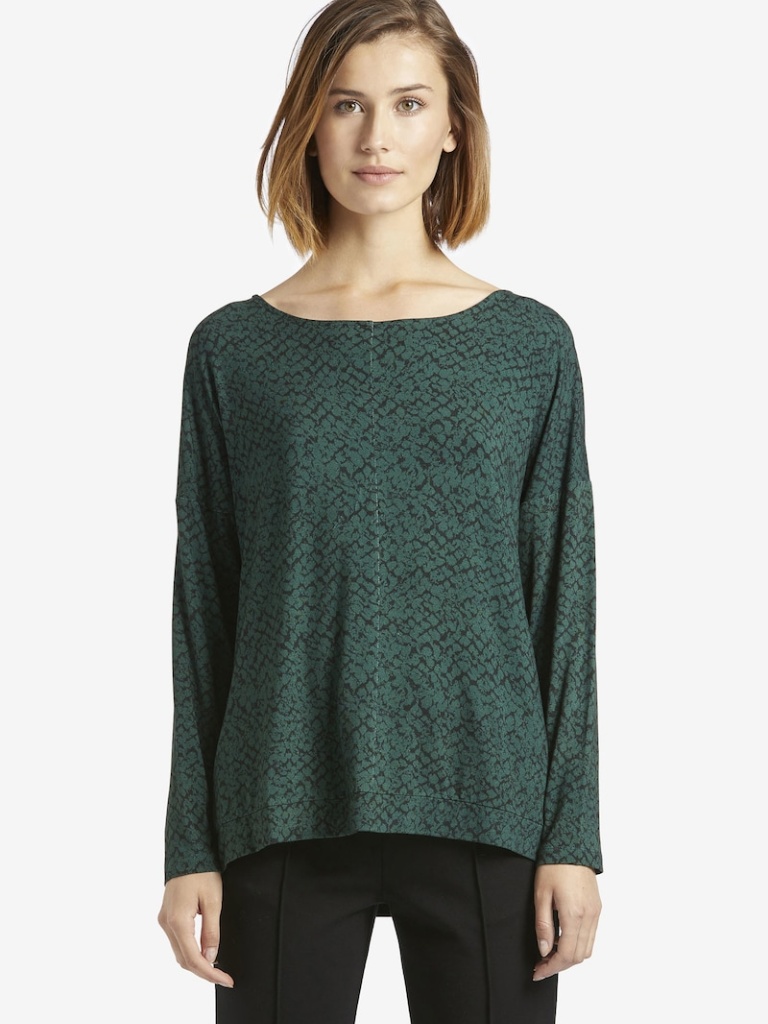} &&
\includegraphics[width=0.16\linewidth]{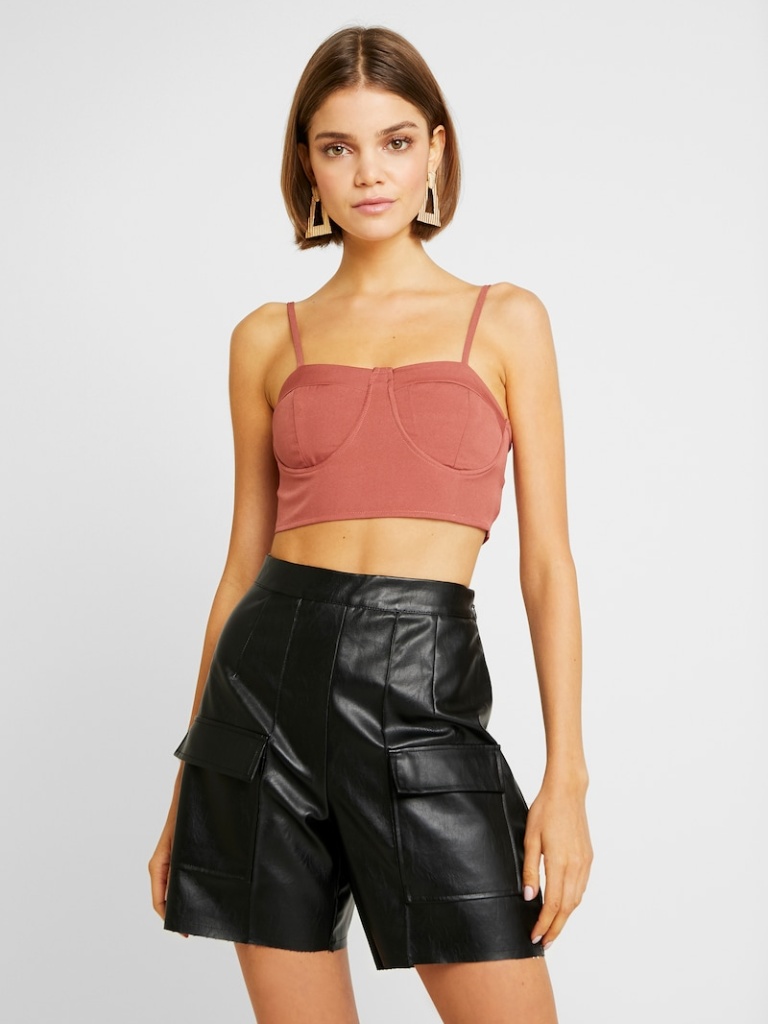} &
\includegraphics[width=0.16\linewidth]{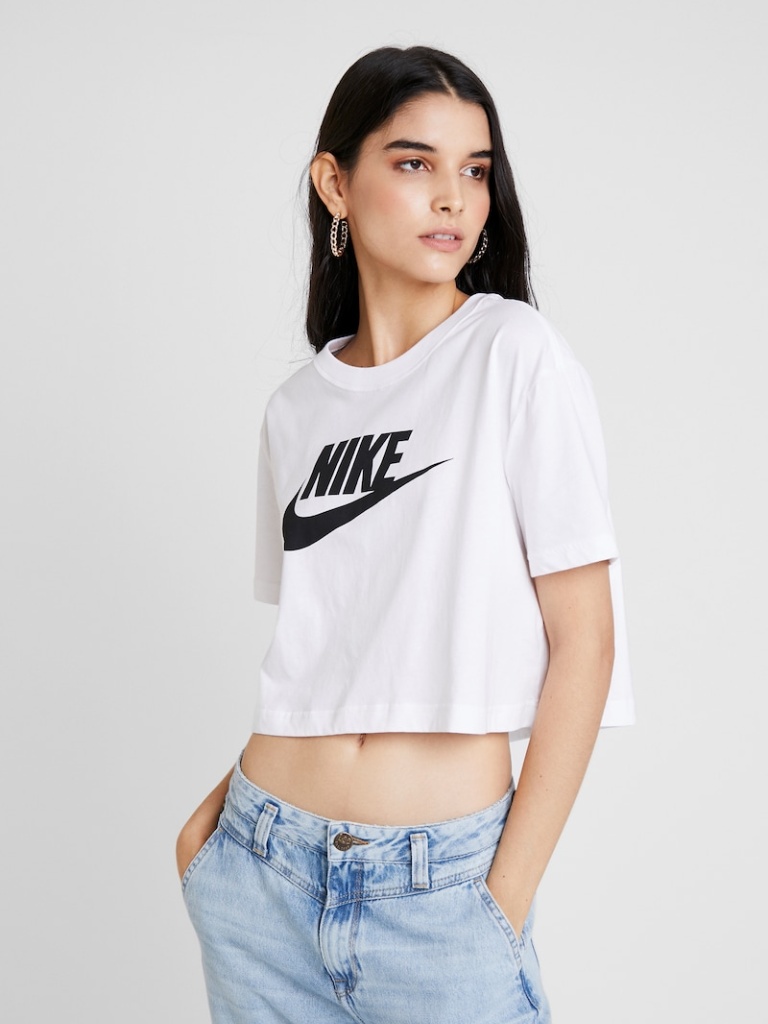} &
\includegraphics[width=0.16\linewidth]{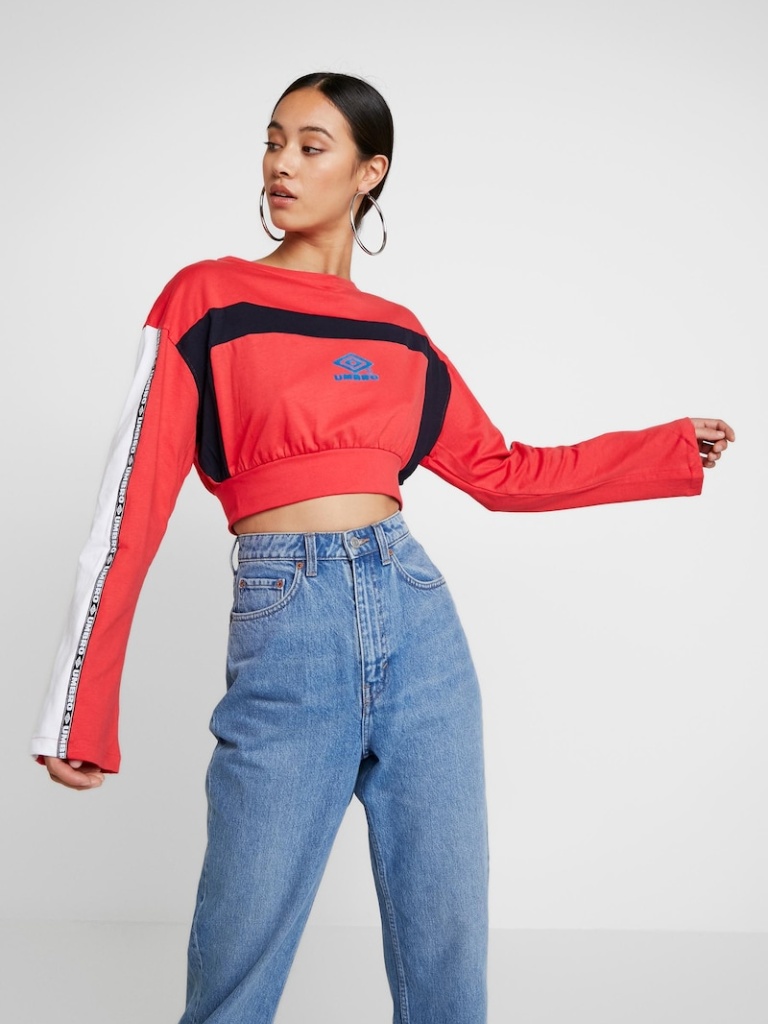} &&
\includegraphics[width=0.16\linewidth]{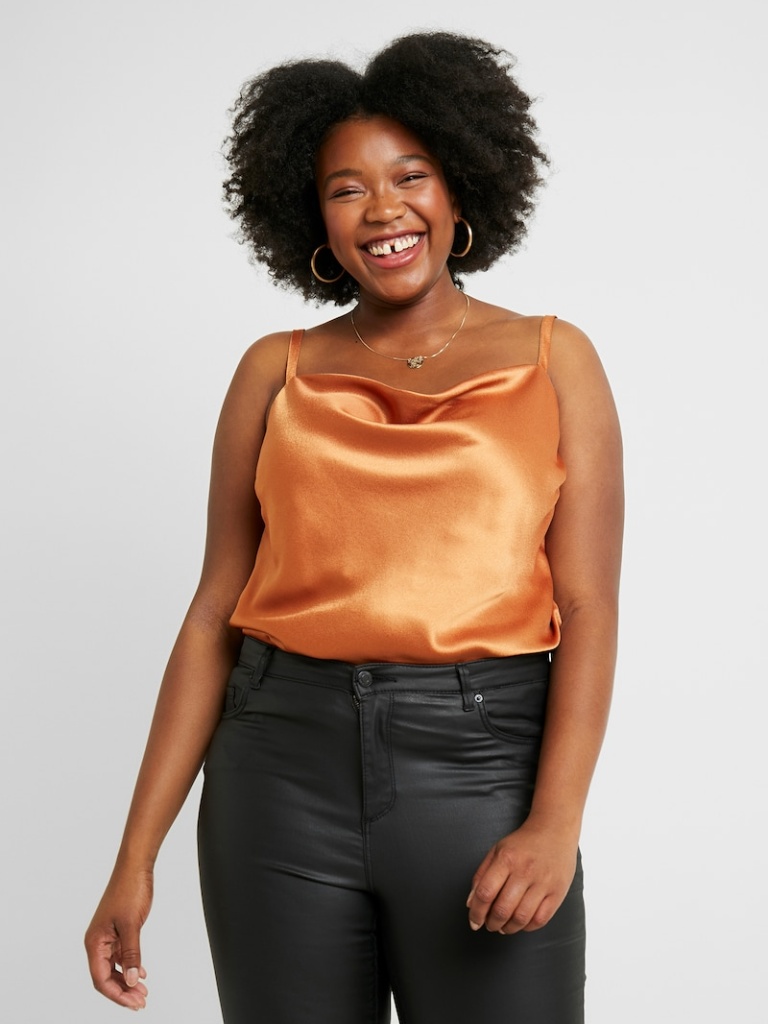} &
\includegraphics[width=0.16\linewidth]{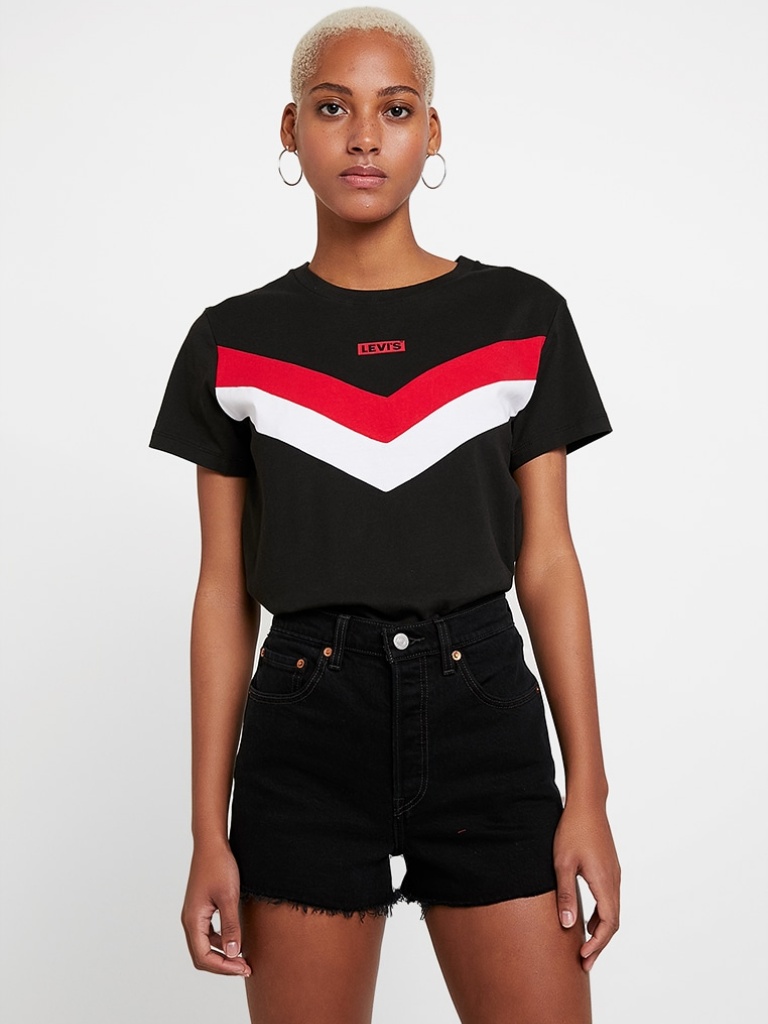} &
\includegraphics[width=0.16\linewidth]{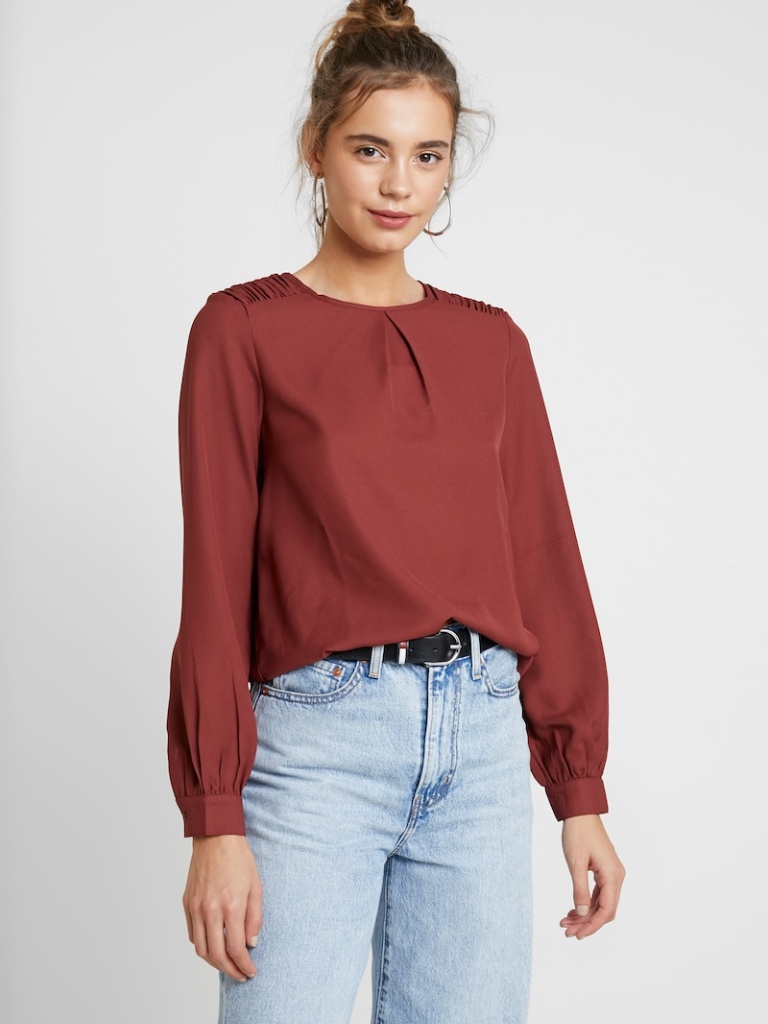} \\

\includegraphics[width=0.16\linewidth]{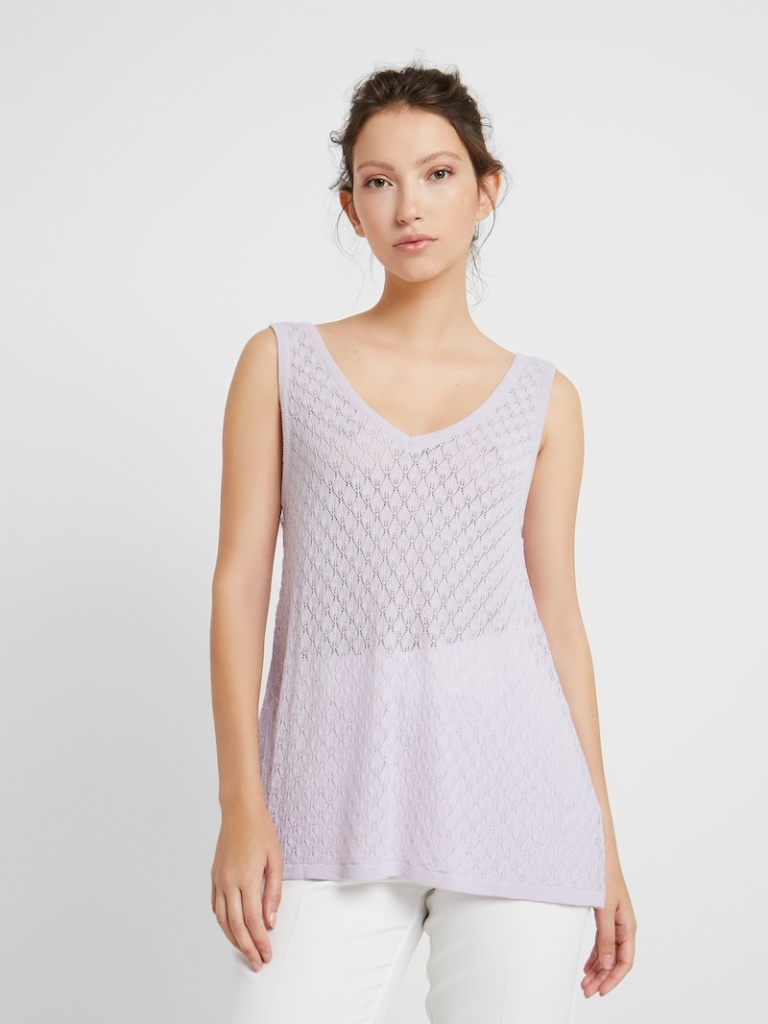} &
\includegraphics[width=0.16\linewidth]{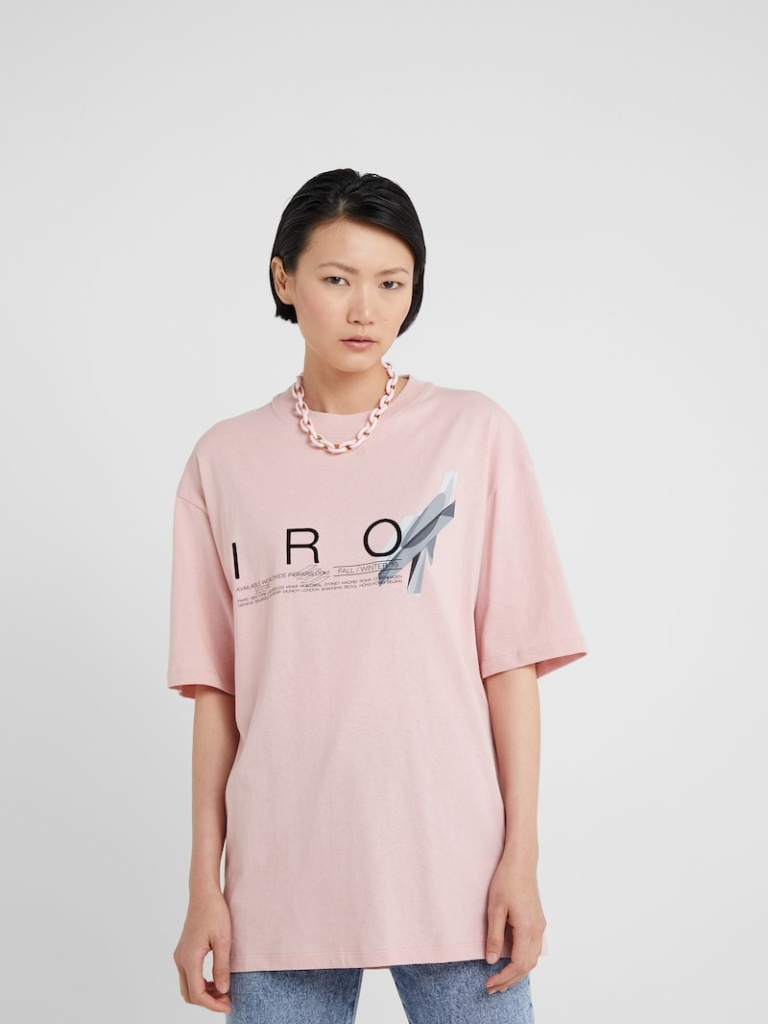} &
\includegraphics[width=0.16\linewidth]{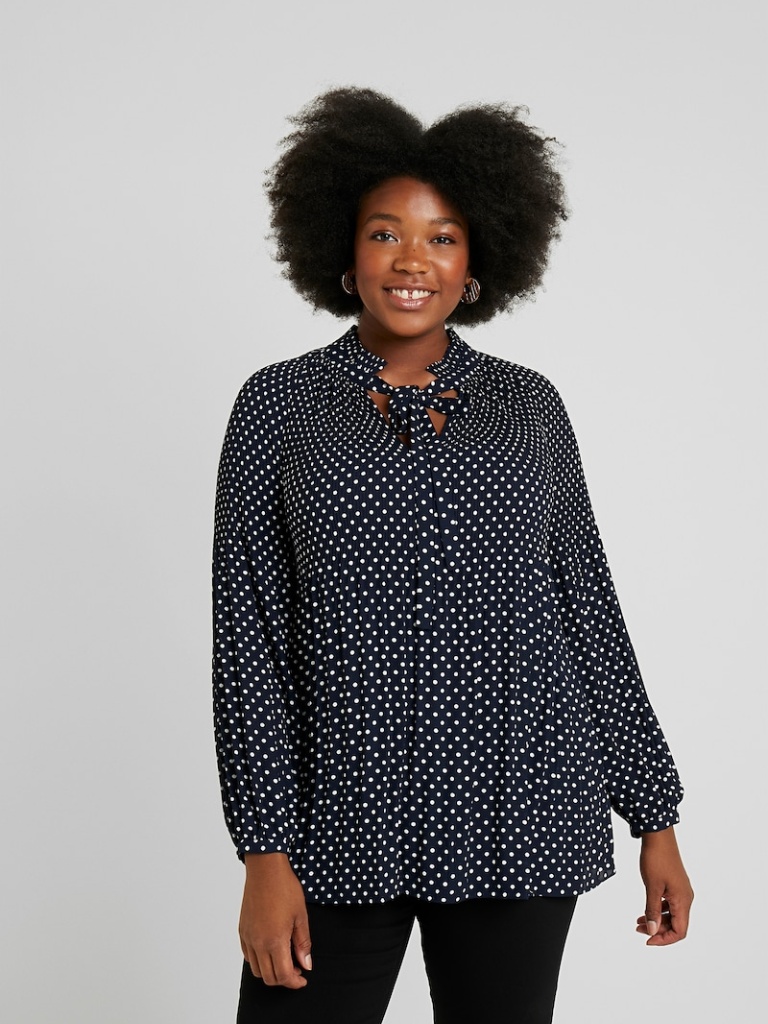} &&
\includegraphics[width=0.16\linewidth]{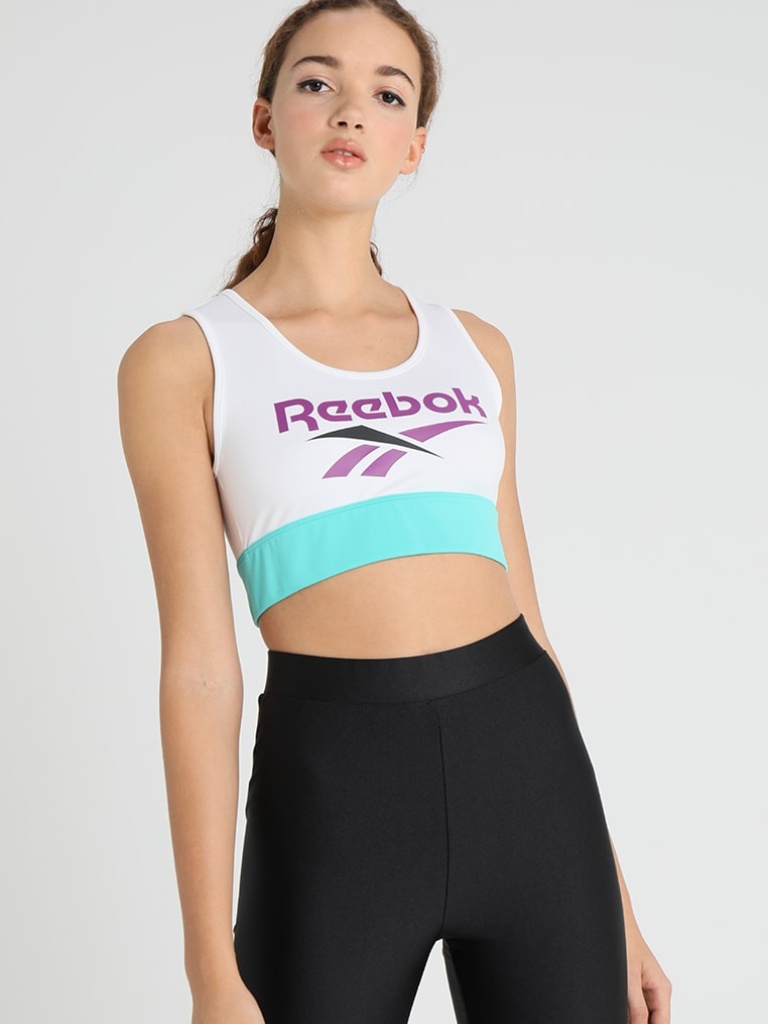} &
\includegraphics[width=0.16\linewidth]{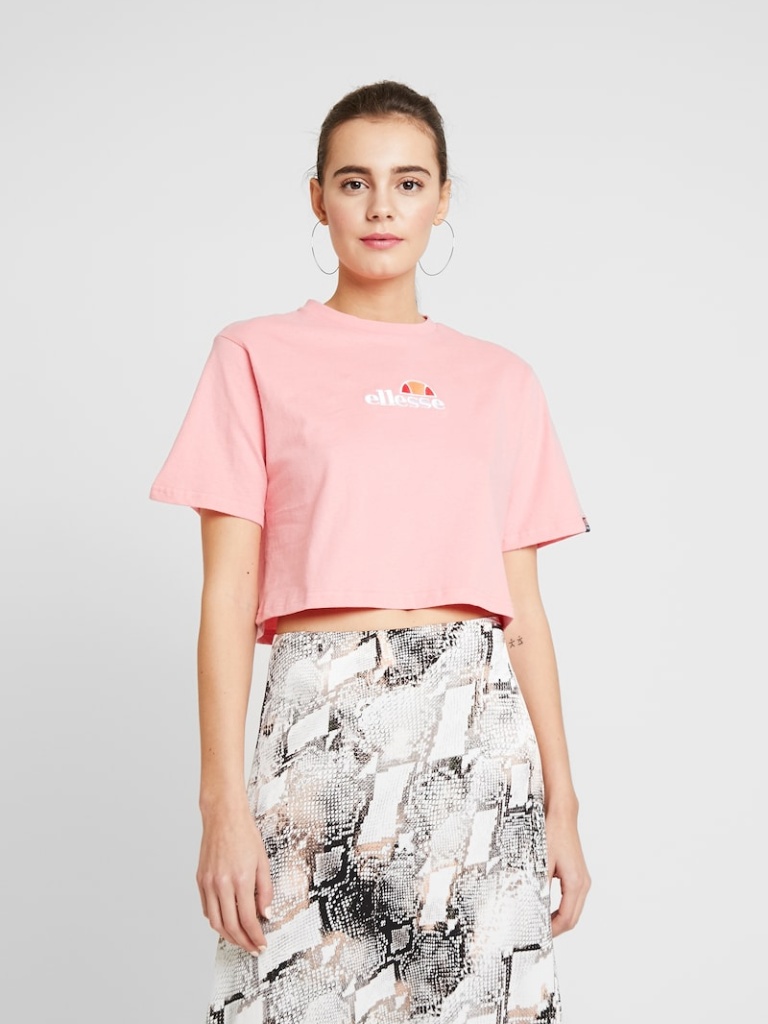} &
\includegraphics[width=0.16\linewidth]{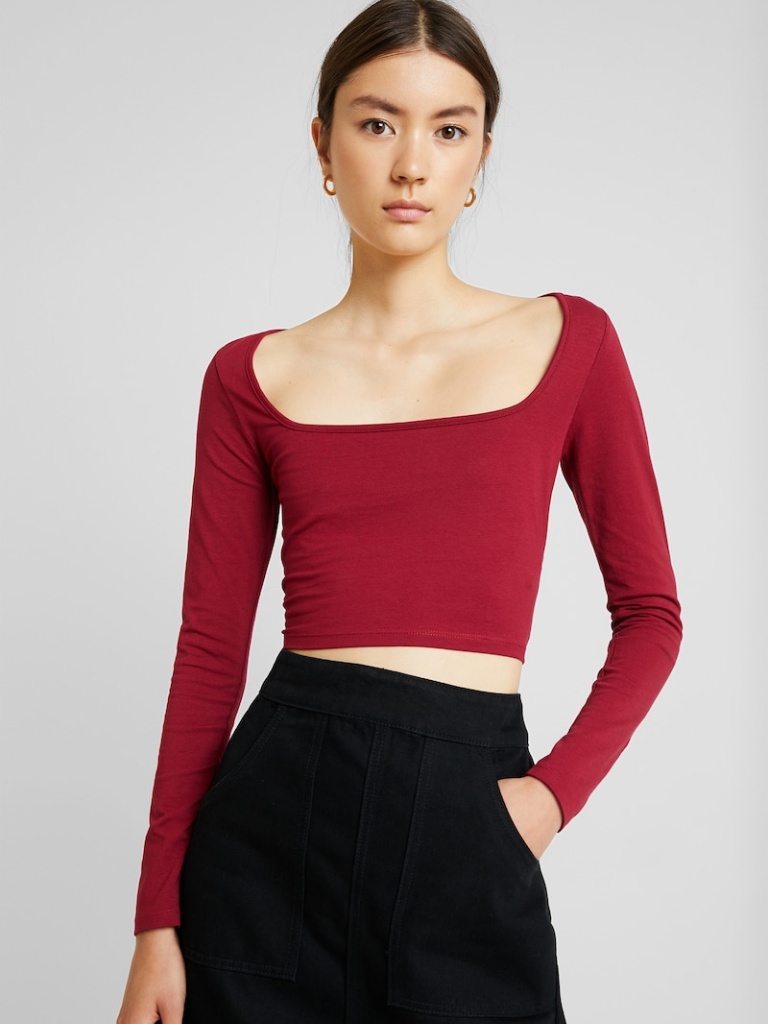} &&
\includegraphics[width=0.16\linewidth]{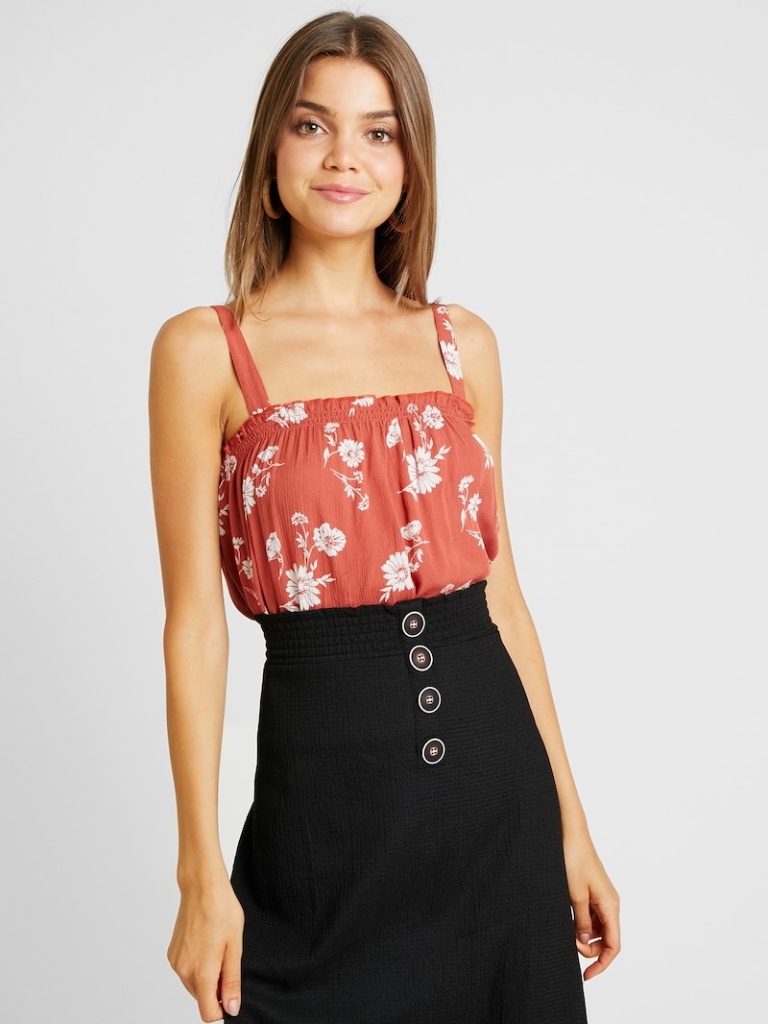} &
\includegraphics[width=0.16\linewidth]{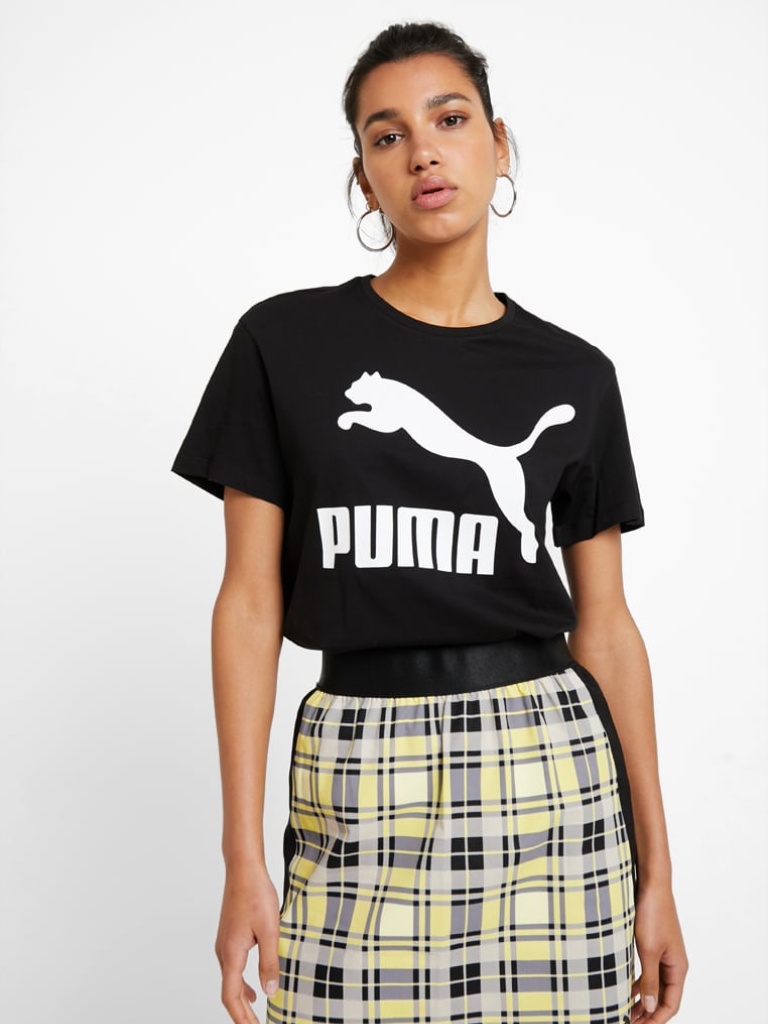} &
\includegraphics[width=0.16\linewidth]{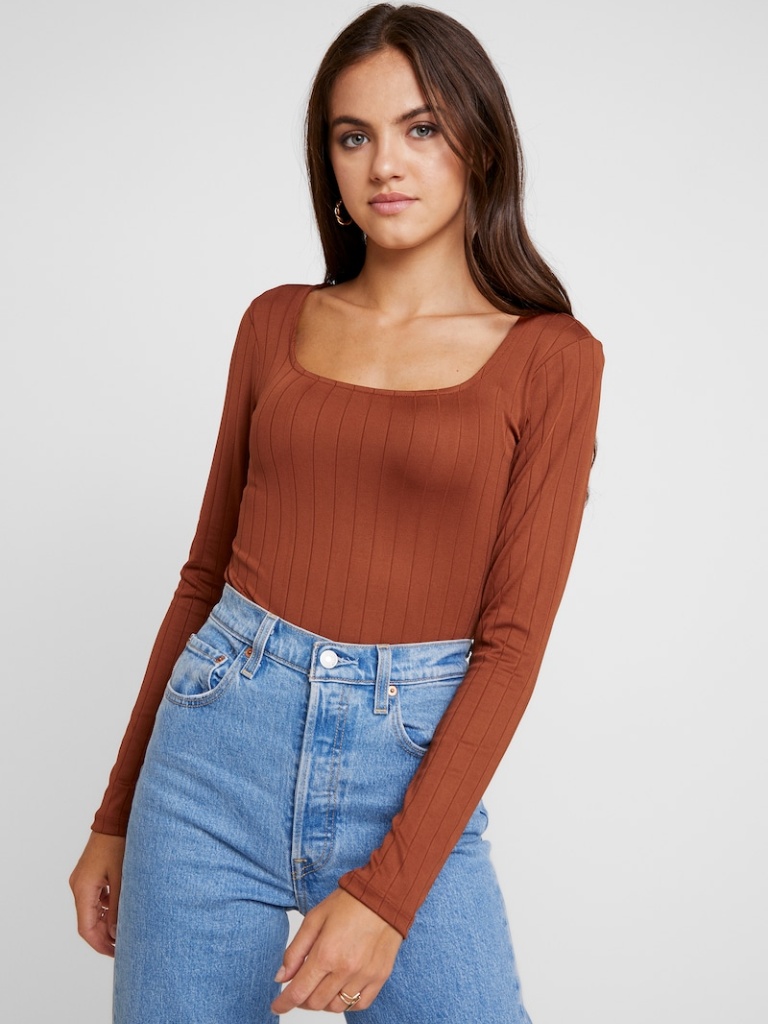} \\

\includegraphics[width=0.16\linewidth]{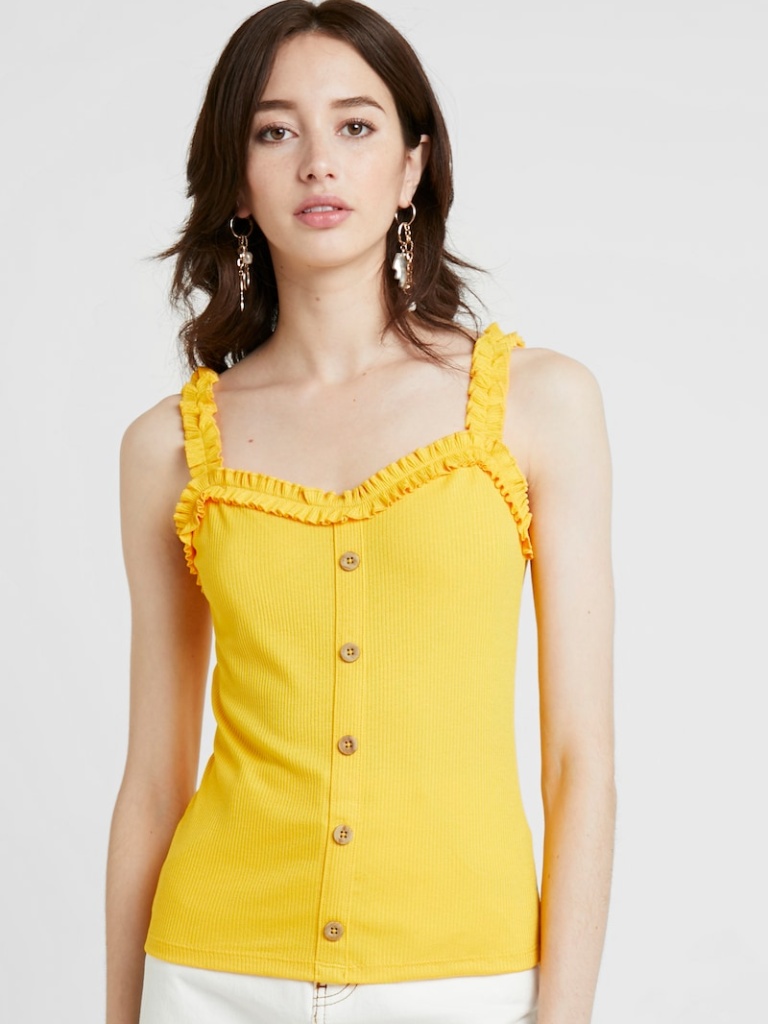} &
\includegraphics[width=0.16\linewidth]{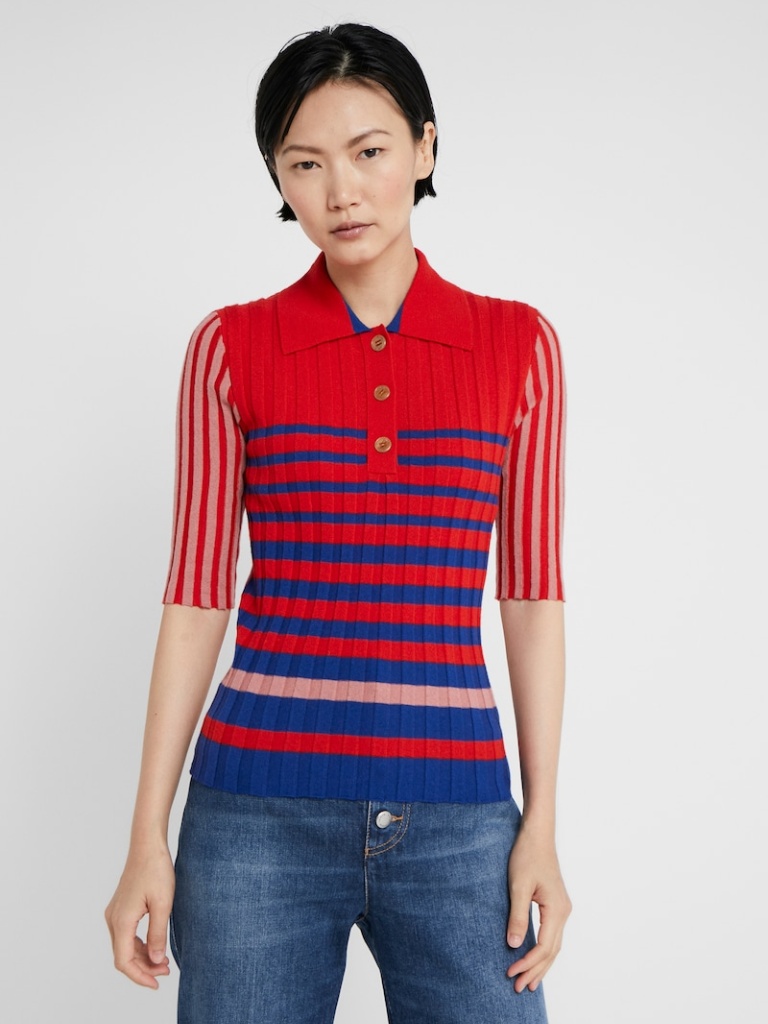} &
\includegraphics[width=0.16\linewidth]{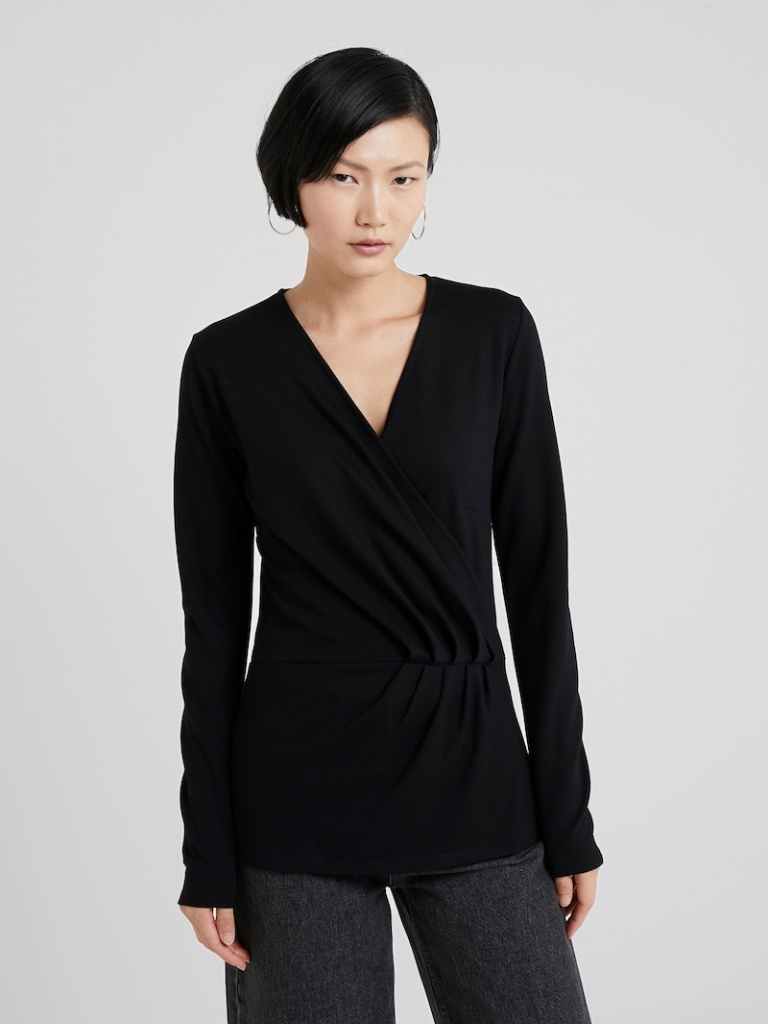} &&
\includegraphics[width=0.16\linewidth]{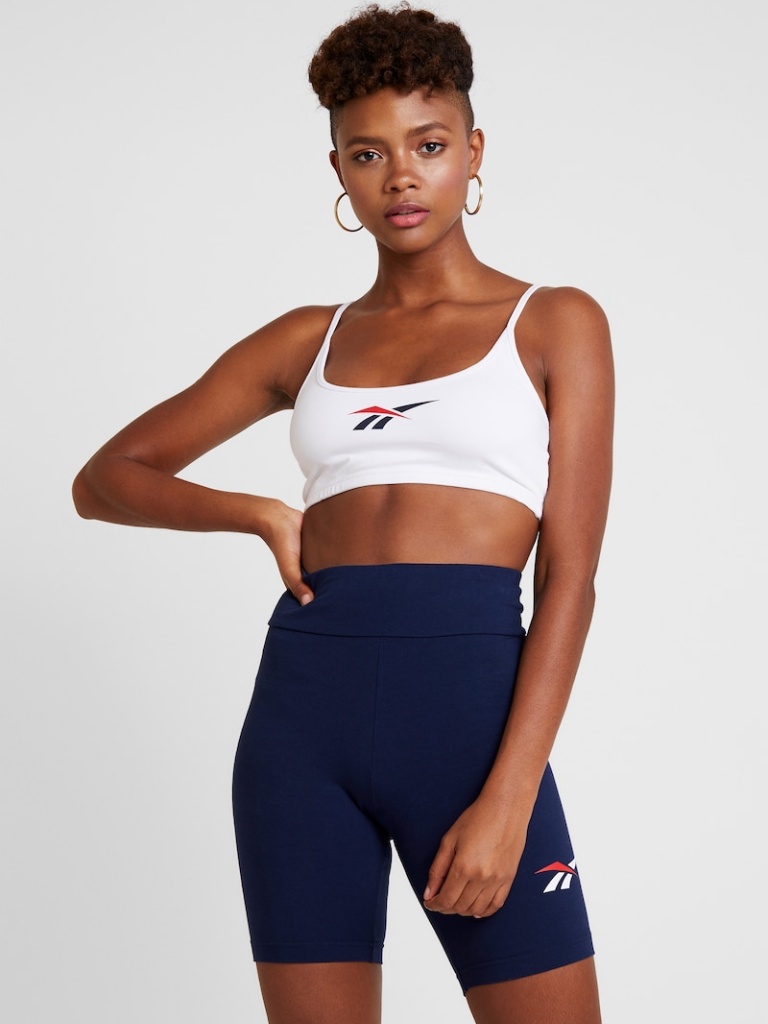} &
\includegraphics[width=0.16\linewidth]{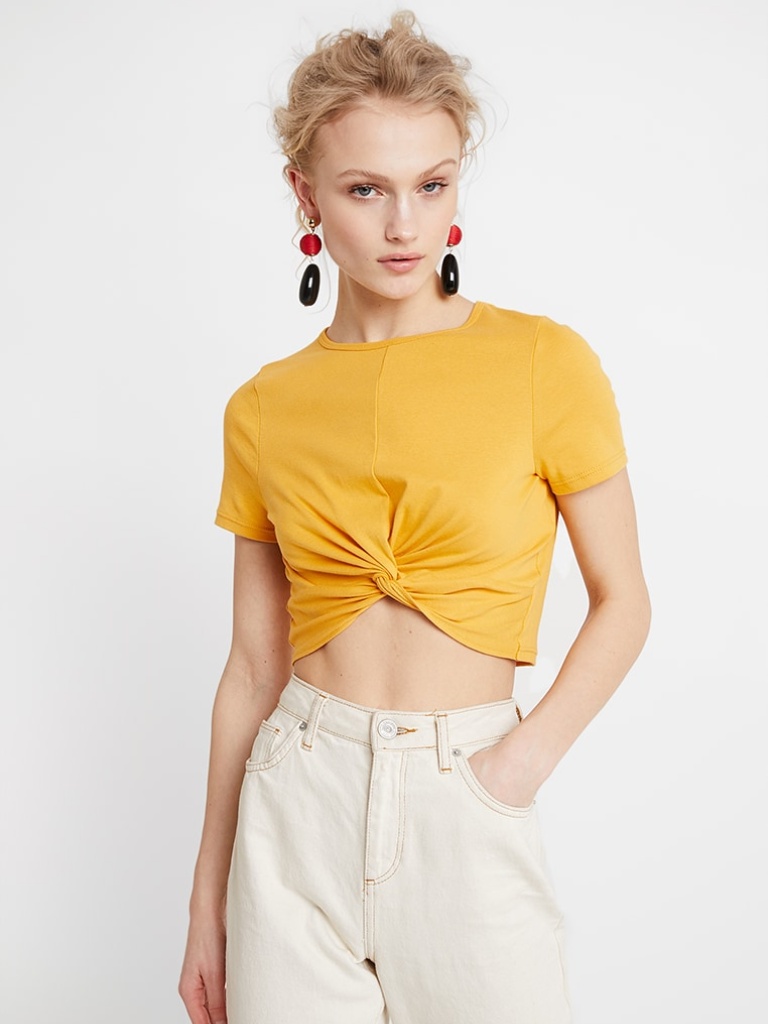} &
\includegraphics[width=0.16\linewidth]{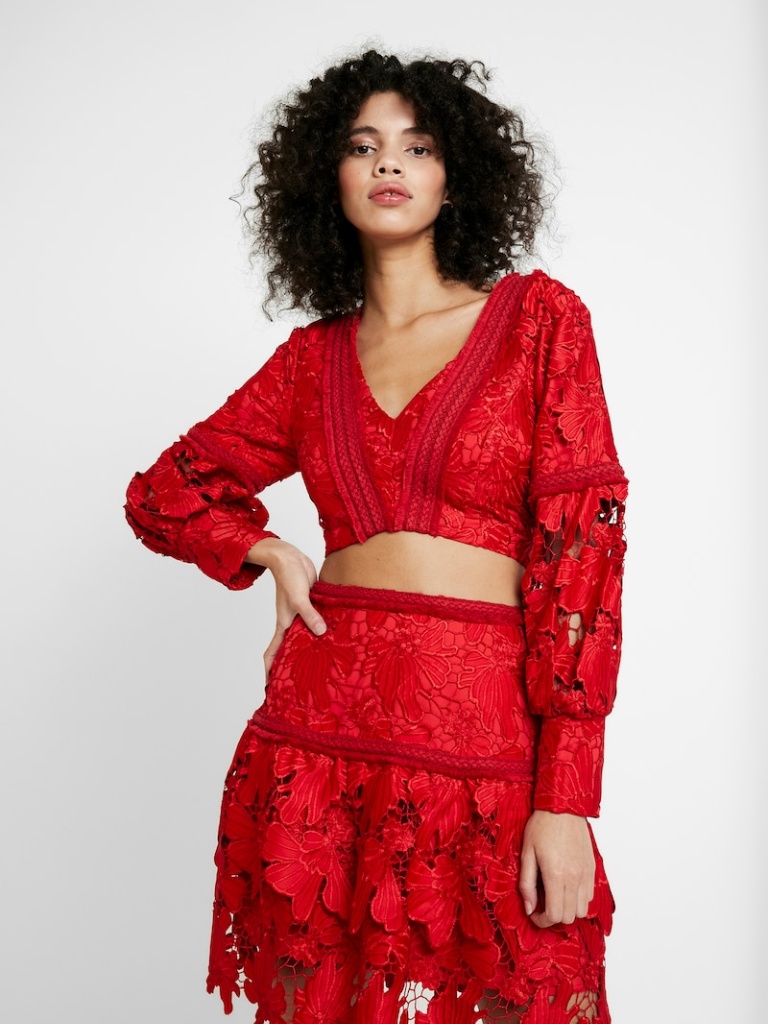} &&
\includegraphics[width=0.16\linewidth]{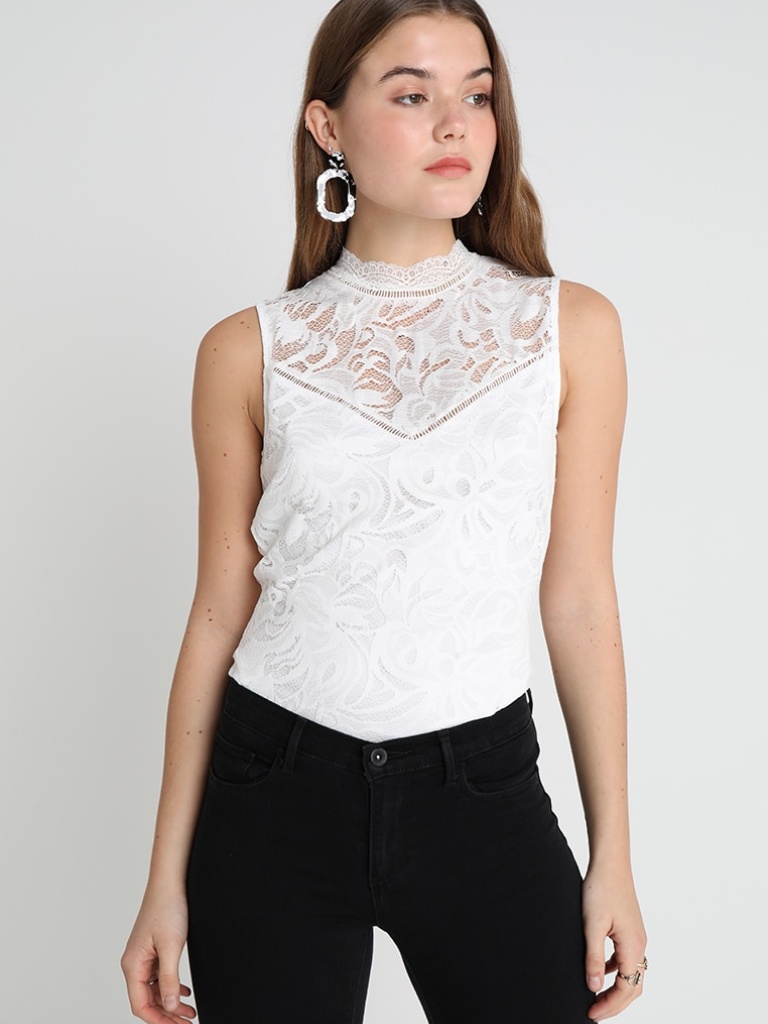} &
\includegraphics[width=0.16\linewidth]{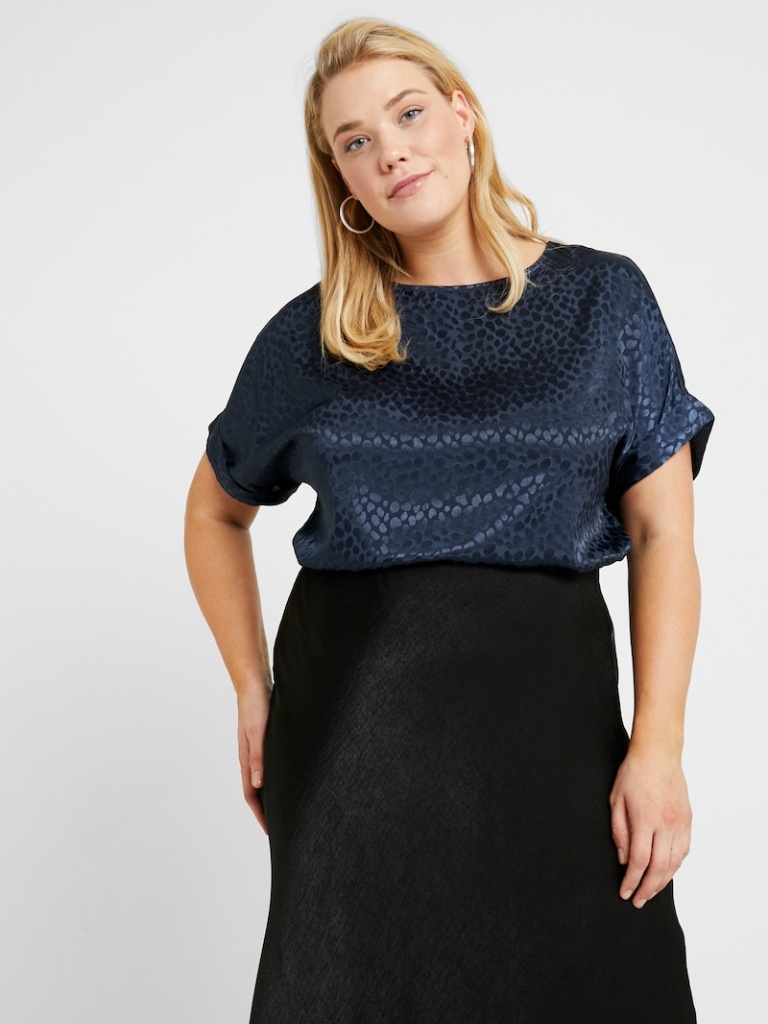} &
\includegraphics[width=0.16\linewidth]{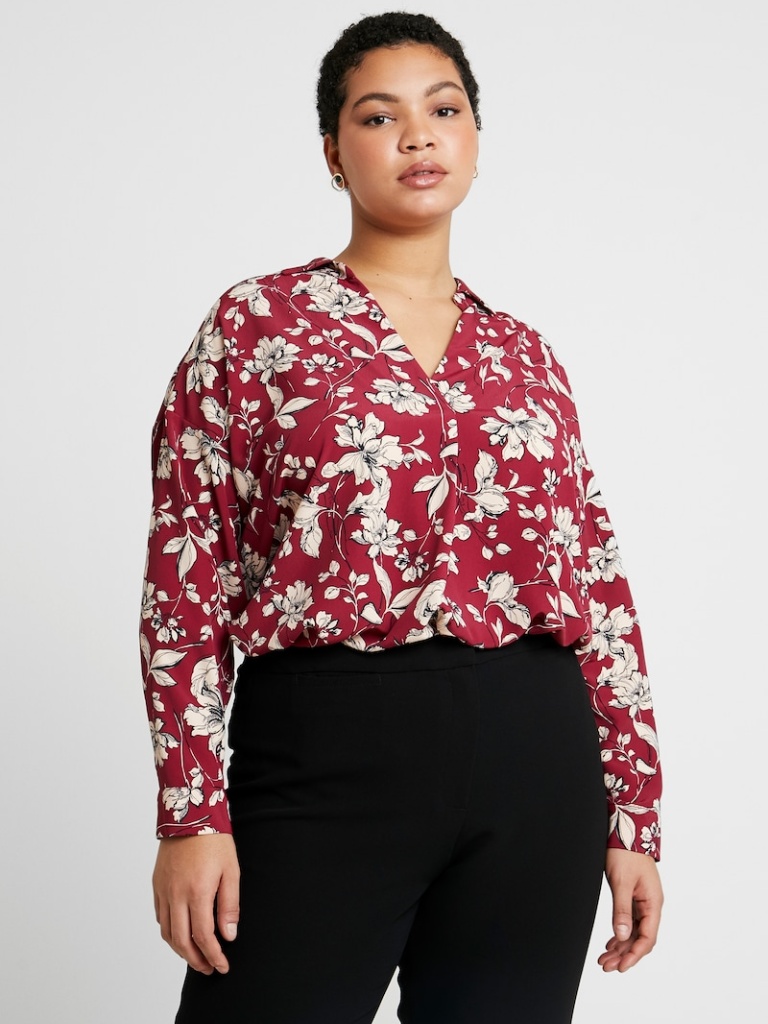} \\

\end{tabular}
}
\caption{All samples in Cross-27.}
\label{fig:benchmark_1}
\end{figure}

\tinytit{Datasets}
\textcolor{black}{We tested our training paradigm and evaluation metrics on VITON-HD \cite{VITON-HD}. The VITON-HD dataset contains 14,221 training samples and 2,032 testing samples. 
We conduct unpaired try-on experiments on two test sets: Unpair-2032 and Cross-27. Unpair-2032 is the original test set of VITON-HD, composed of random try-on combinations of 2032 test samples. Cross-27 is a manually curated cross-try-on benchmark. We selected 27 samples from the VITON-HD test set by considering three dimensions: long sleeves / short sleeves / sleeveless, long/short lengths, and \textit{Interfered} / \textit{Non-Interfered}. Specifically, for each condition, we select three samples by considering different body shapes and poses. The selected samples in Cross-27 are shown in Fig. \ref{fig:benchmark_1}, and each person image corresponds to a clothing image. Cross-27 Benchmark is used for evaluating the cross-try-on performance with various and  challenging try-on conditions, and 729 (i.e., $27\times27$) try-on images in total are obtained. Our training and testing were both conducted at a resolution of $512\times384$. }

\begin{figure}[t]
  \centering
  \includegraphics[width=1.0\linewidth]{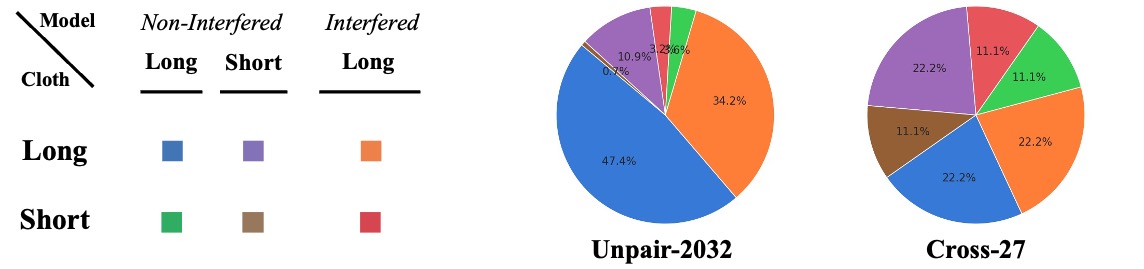}
   \caption{Distribution of try-on situations in test set.}
   \label{fig:benchmark_reb}
\end{figure}

Fig. \ref{fig:benchmark_reb} compares the distribution of try-on situations covered by Unpair-2032 and Cross-27, where Cross-27 exhibits a more balanced distribution. 
Try-on situations in blue and orange are less challenging, resulting in a bias on easy samples for VITON-HD. 

\textcolor{black}{\tit{Evaluation} The FID\cite{FID}, KID\cite{KID}, SDR, and S-LPIPS metrics are used for calculating results. The FID and KID metrics use the official implementation from Torch Metric\cite{CleanFID}. SDR, S-LPIPS metrics require the compared images to have the same resolution, so we resize the ground truth images from $1024\times768$ resolution to $512\times384$ resolution for calculation.}

\textcolor{black}{\tit{Training}
We train the model using an AdamW optimizer with a fixed learning rate of 1e-4 for 80k iterations, employing a batch size of 128. Then, we finetune the model with the attention total variation weight hyper-parameter $\lambda_{ATV} = 0.001$, using the same learning rate and batch size for 10K iterations. We train for about 50 hours using 16 NVIDIA A100 GPUs. We employ the same data augmentation methods as those used in StableVITON\cite{StableVITON}. For the Adp-Mask Maker, we set the parameter $\tau_B = 3$, and $\tau_T = 0.65$.}

\subsection{Mask Effectiveness Analysis}
\begin{figure}[t!]
\centering
\scriptsize
\setlength{\tabcolsep}{.2em}
\resizebox{\linewidth}{!}{
\begin{tabular}{c ccccccc}
\includegraphics[width=0.0785\linewidth]{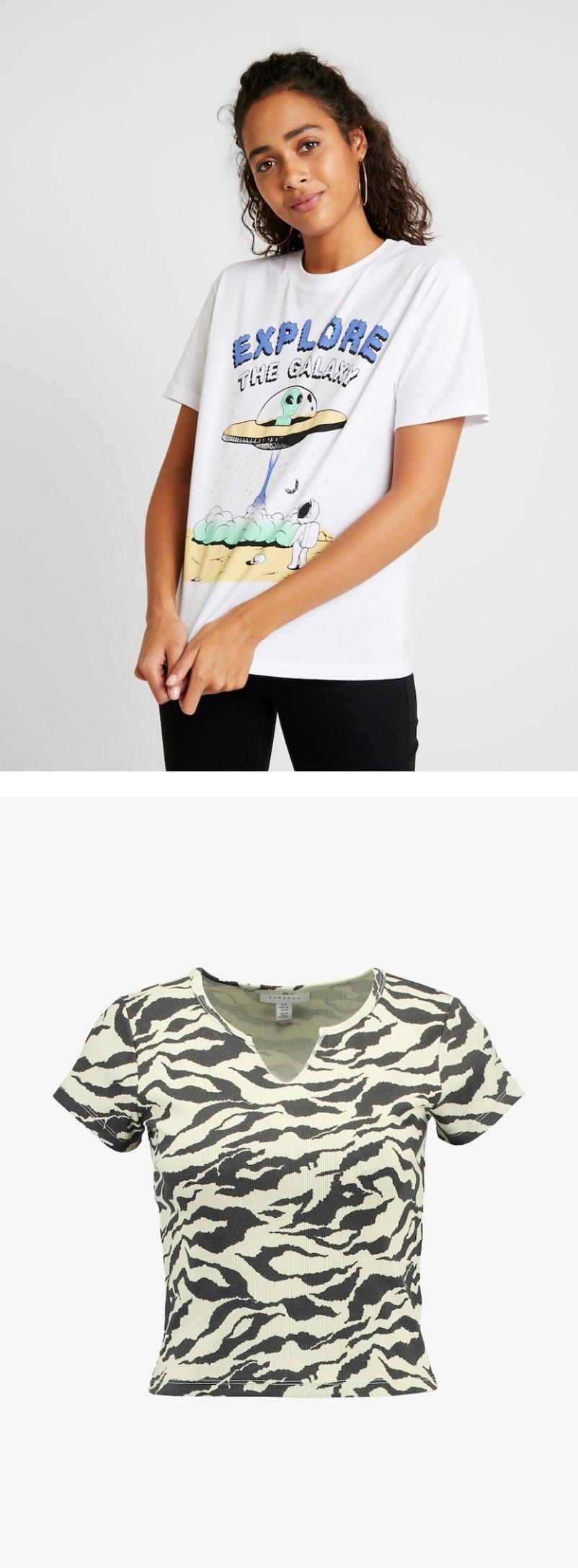} &
\includegraphics[width=0.16\linewidth]{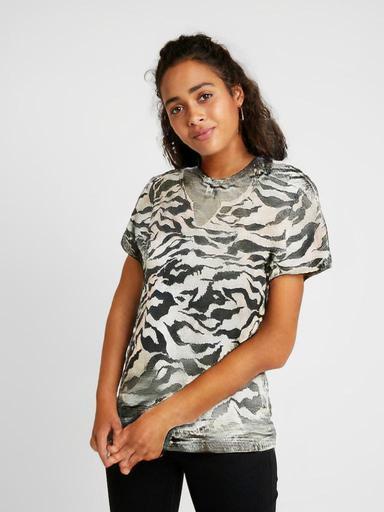} &
\includegraphics[width=0.16\linewidth]{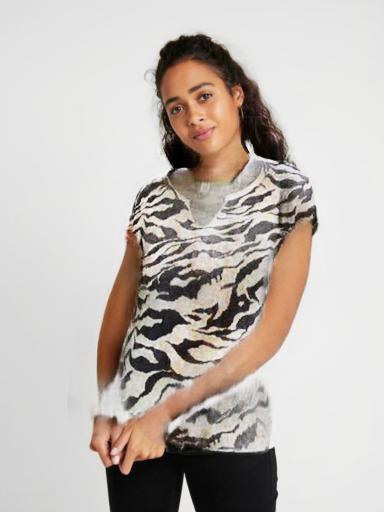} & 
\includegraphics[width=0.16\linewidth]{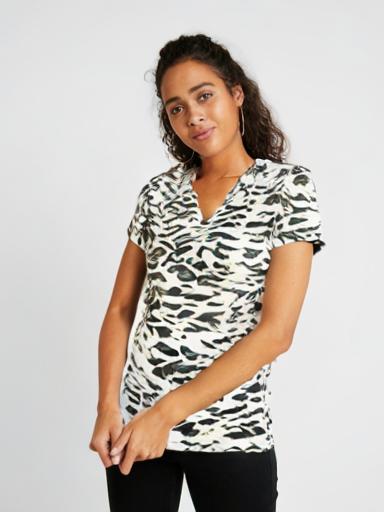} &
\includegraphics[width=0.16\linewidth]{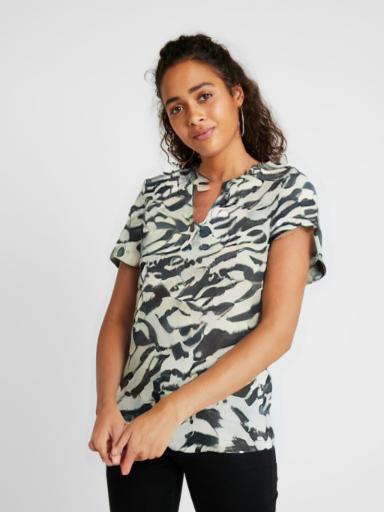} &
\includegraphics[width=0.16\linewidth]{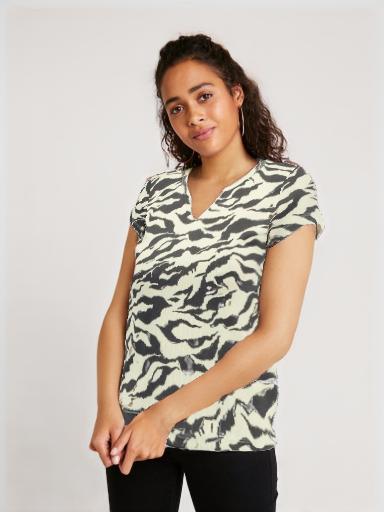} & 
\includegraphics[width=0.16\linewidth]{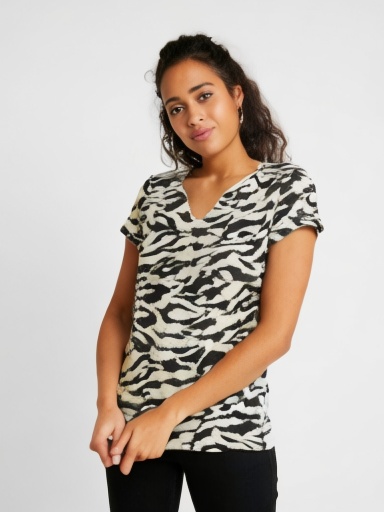} &
\includegraphics[width=0.16\linewidth]{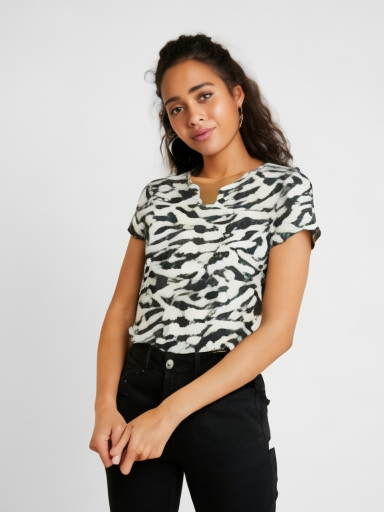} \\
\includegraphics[width=0.0785\linewidth]{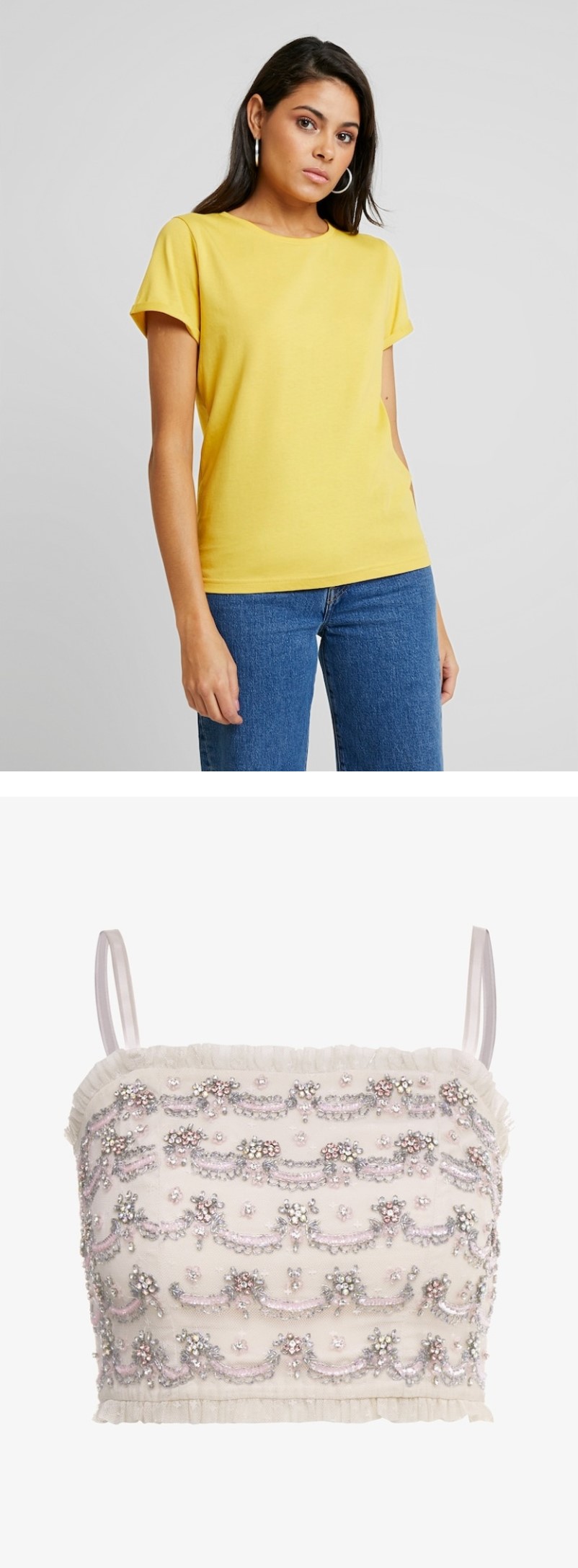} &
\includegraphics[width=0.16\linewidth]{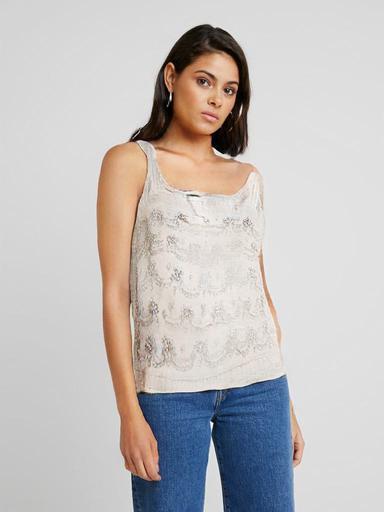} &
\includegraphics[width=0.16\linewidth]{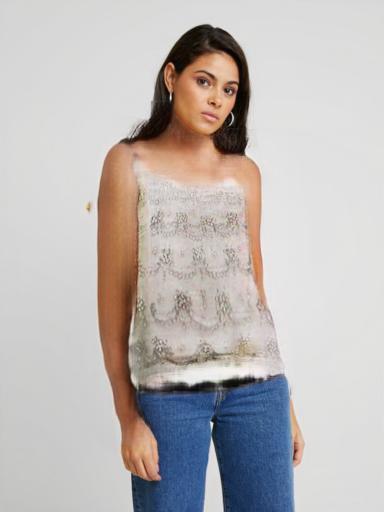} & 
\includegraphics[width=0.16\linewidth]{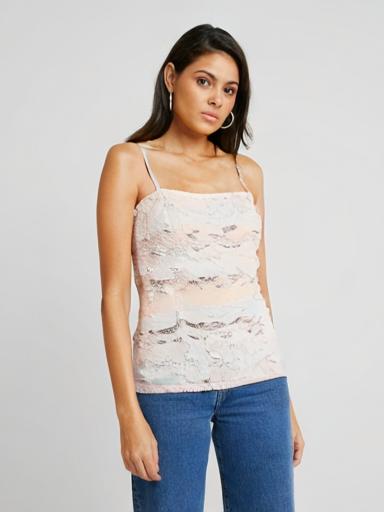} &
\includegraphics[width=0.16\linewidth]{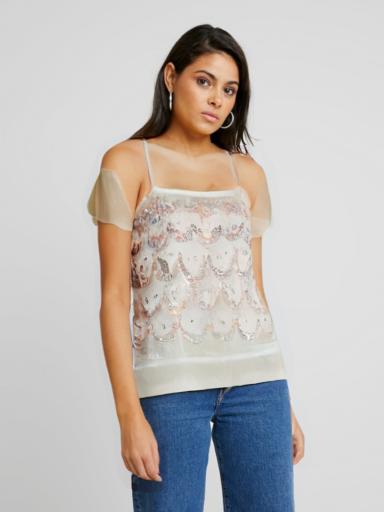} &
\includegraphics[width=0.16\linewidth]{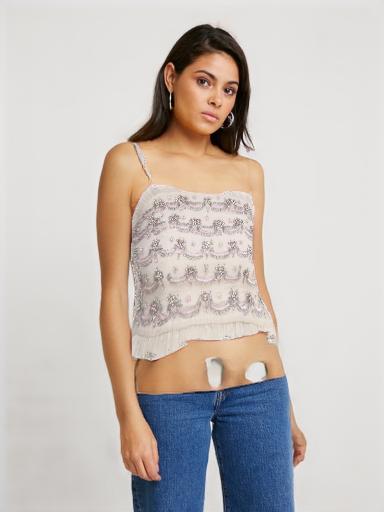} & 
\includegraphics[width=0.16\linewidth]{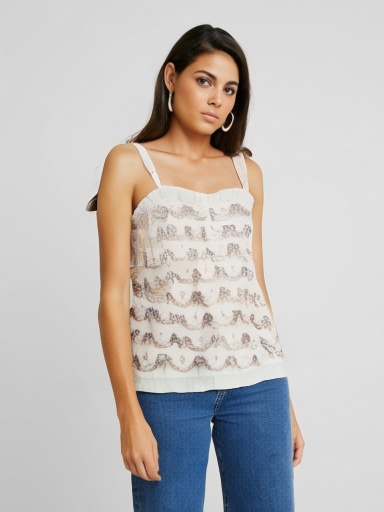} &
\includegraphics[width=0.16\linewidth]{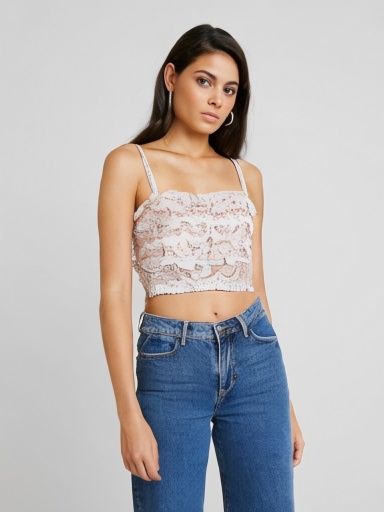} \\
\includegraphics[width=0.0785\linewidth]{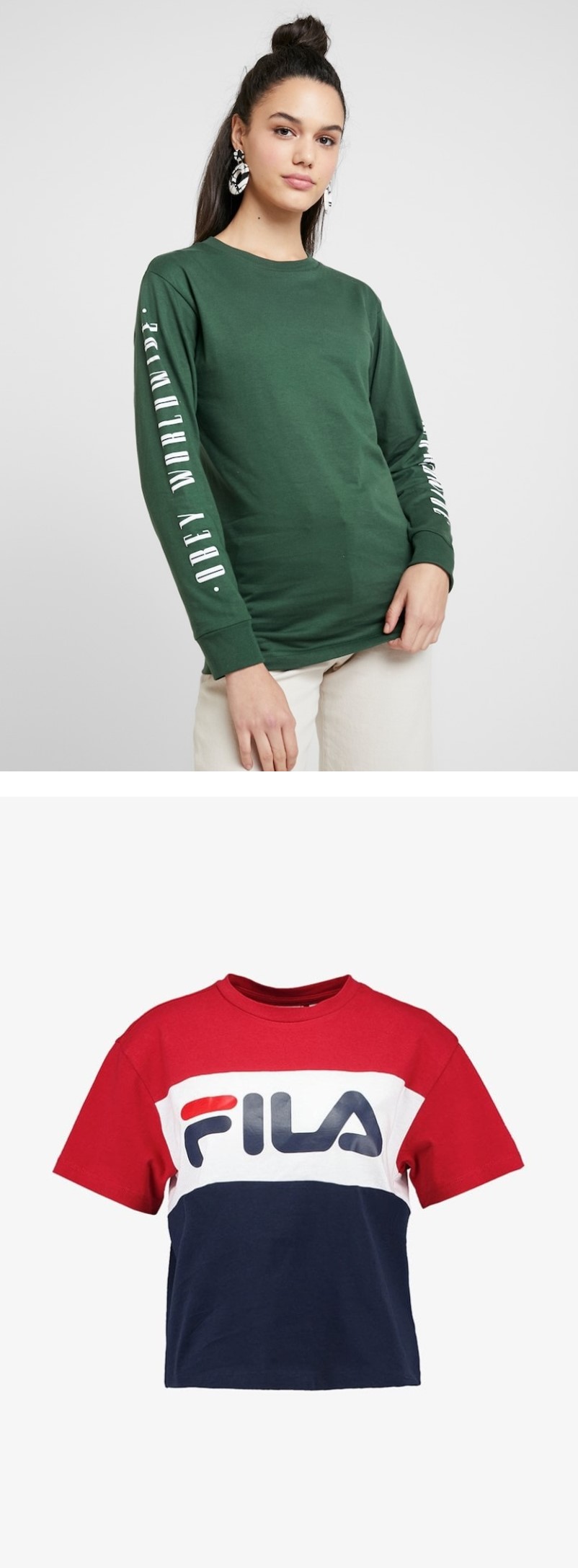} &
\includegraphics[width=0.16\linewidth]{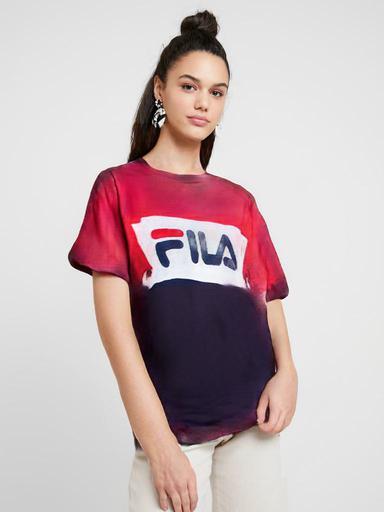} &
\includegraphics[width=0.16\linewidth]{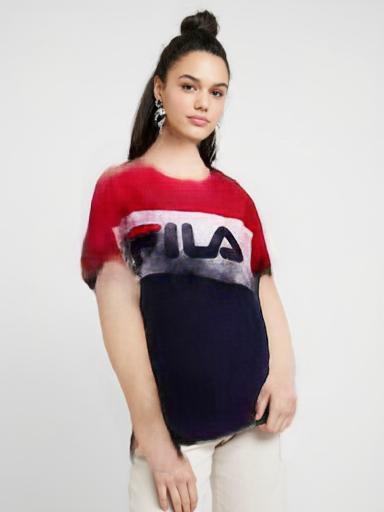} & 
\includegraphics[width=0.16\linewidth]{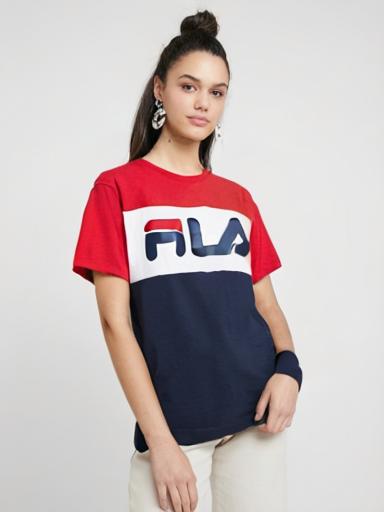} &
\includegraphics[width=0.16\linewidth]{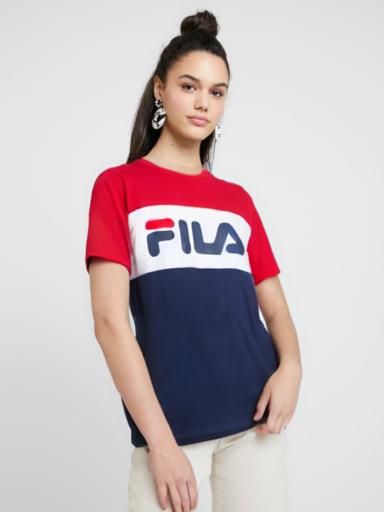} &
\includegraphics[width=0.16\linewidth]{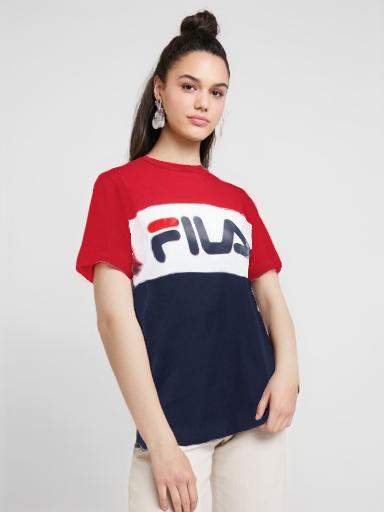} & 
\includegraphics[width=0.16\linewidth]{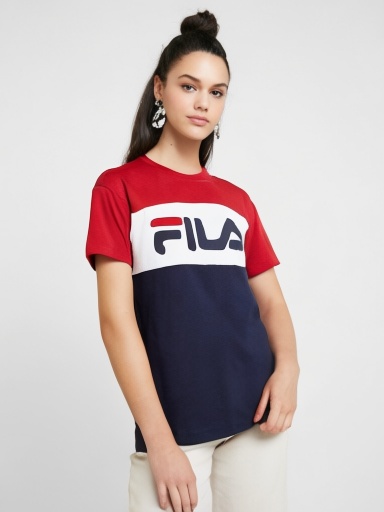} &
\includegraphics[width=0.16\linewidth]{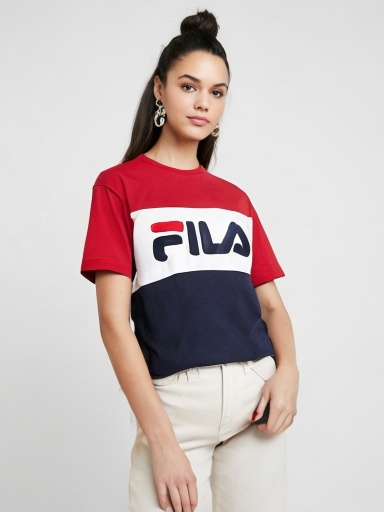} \\
\addlinespace[0.08cm]
 & \textbf{VITON-HD} & \textbf{HR-VTON} & \textbf{LaDI-VTON} & \textbf{DCI-VTON} & \textbf{GP-VTON} & \textbf{StableVITON} & \textbf{Ours} \\
\end{tabular}
}
\caption{Comparison results on Unpair-2032.}
\label{fig:comparison_vitonhd}
\vspace{-0.2cm}
\end{figure}

\begin{figure}[t!]
\centering
\scriptsize
\setlength{\tabcolsep}{.2em}
\resizebox{\linewidth}{!}{
\begin{tabular}{cccccccc}
\addlinespace[0.08cm]
\includegraphics[width=0.157\linewidth]{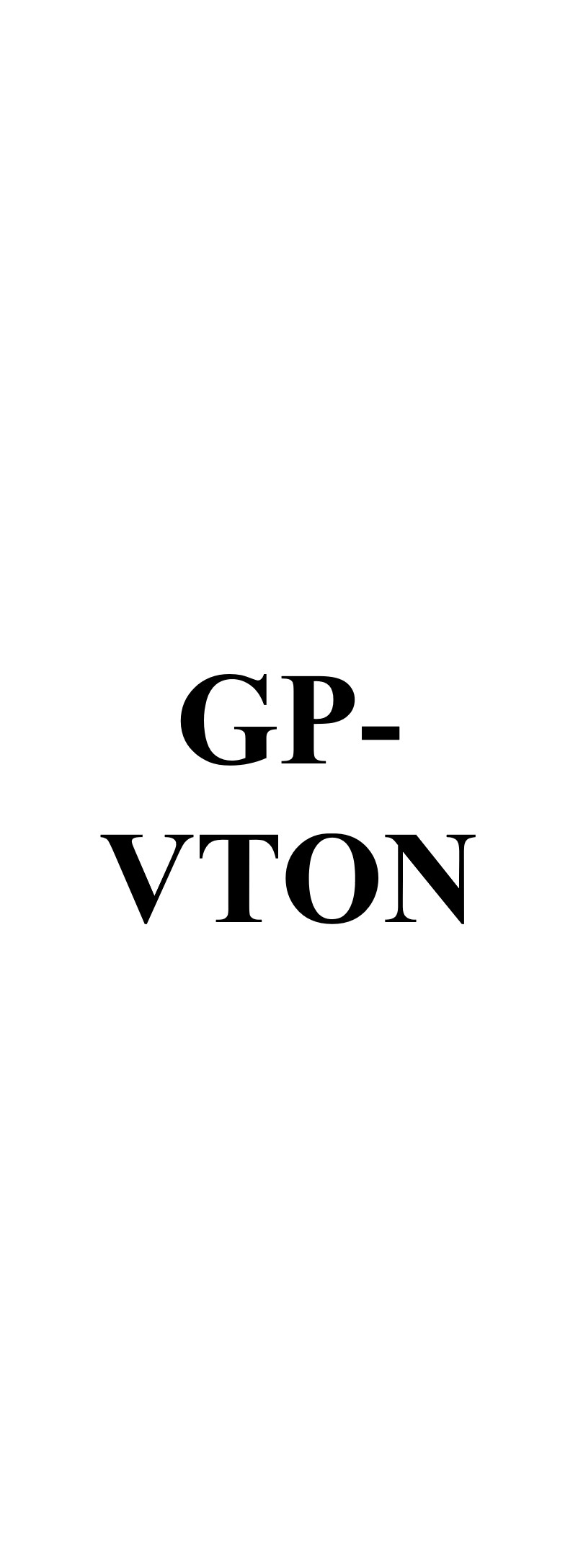}&
\includegraphics[width=0.32\linewidth]{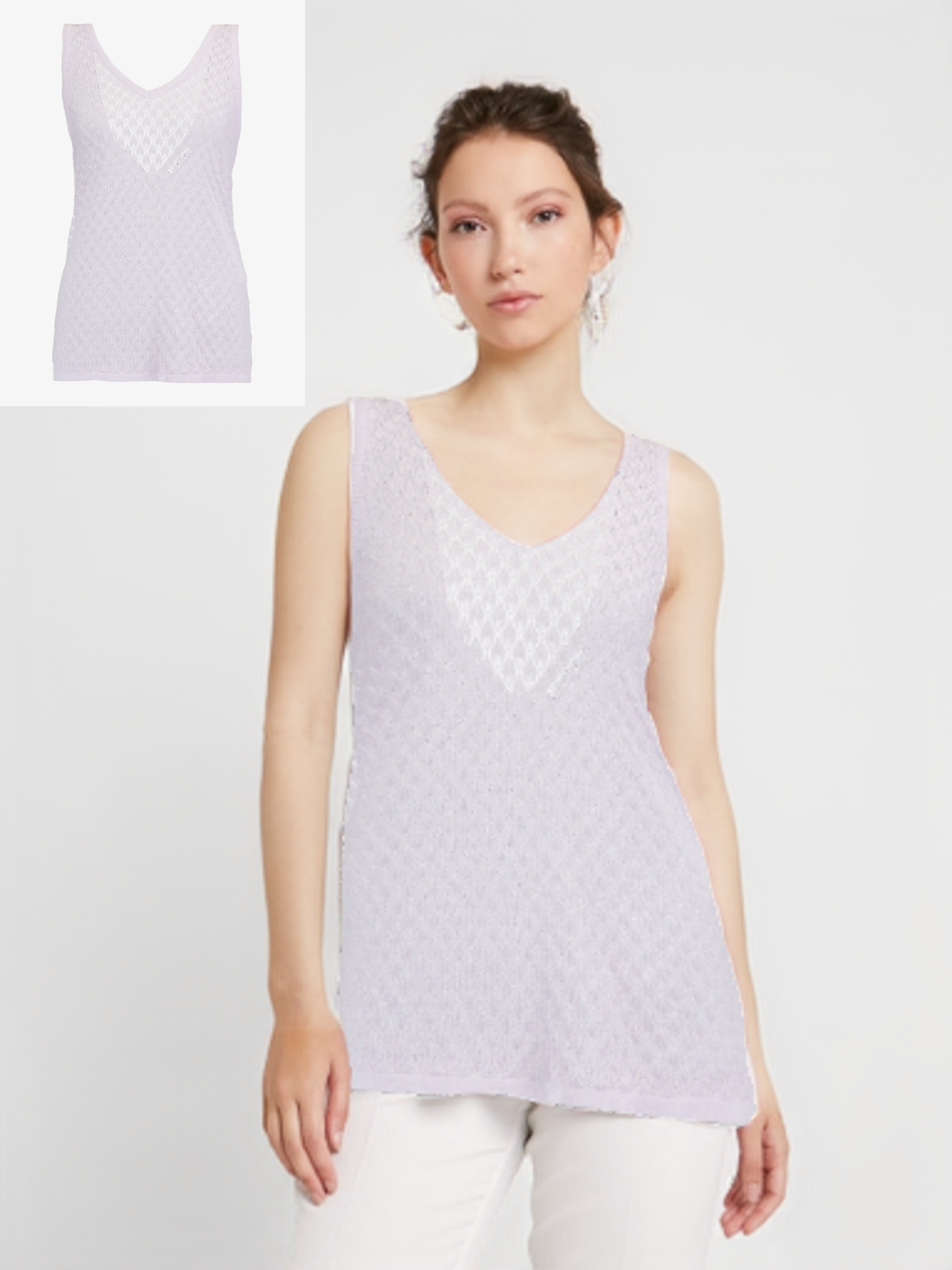} &
\includegraphics[width=0.32\linewidth]{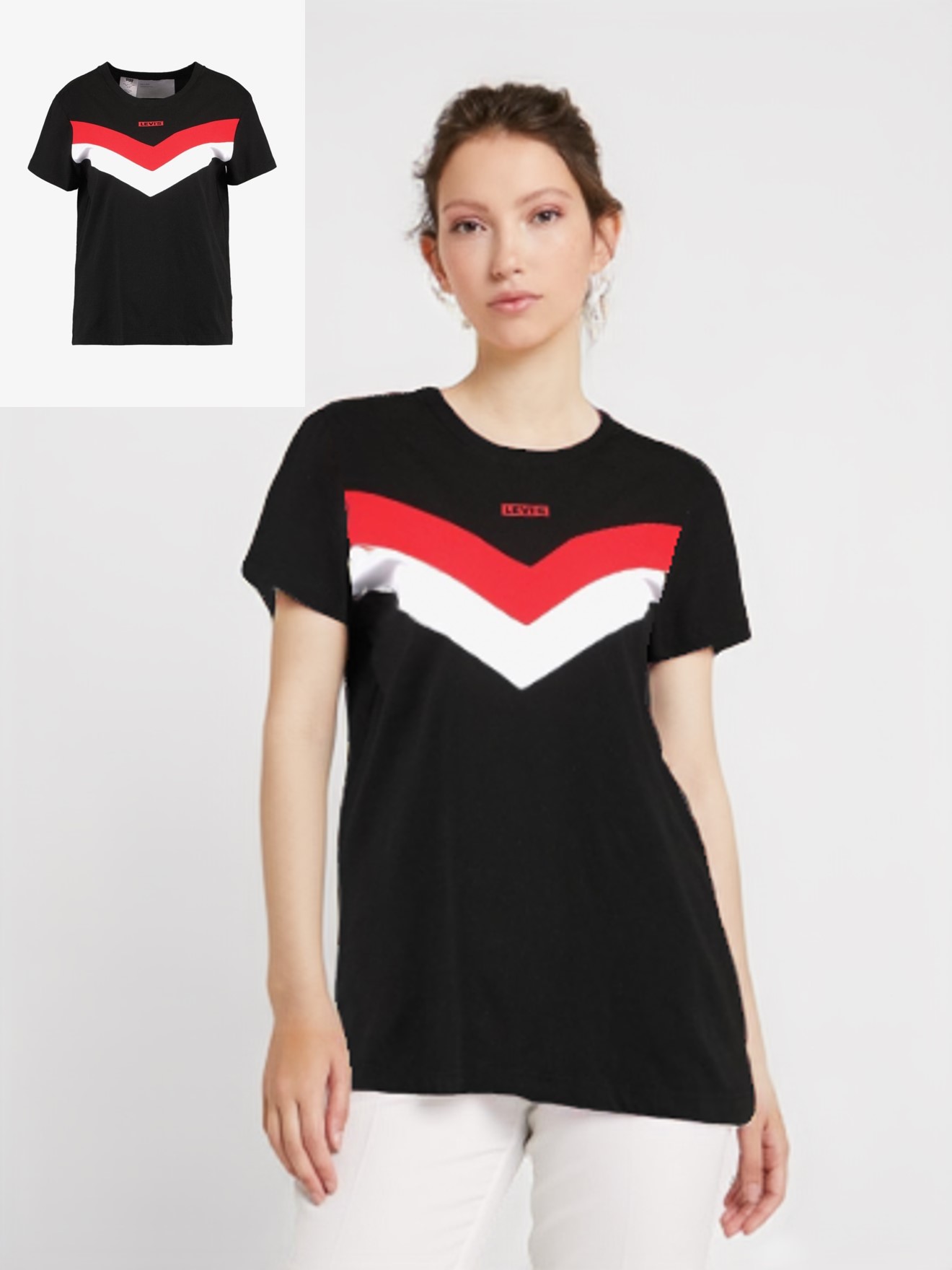} & 
\includegraphics[width=0.32\linewidth]{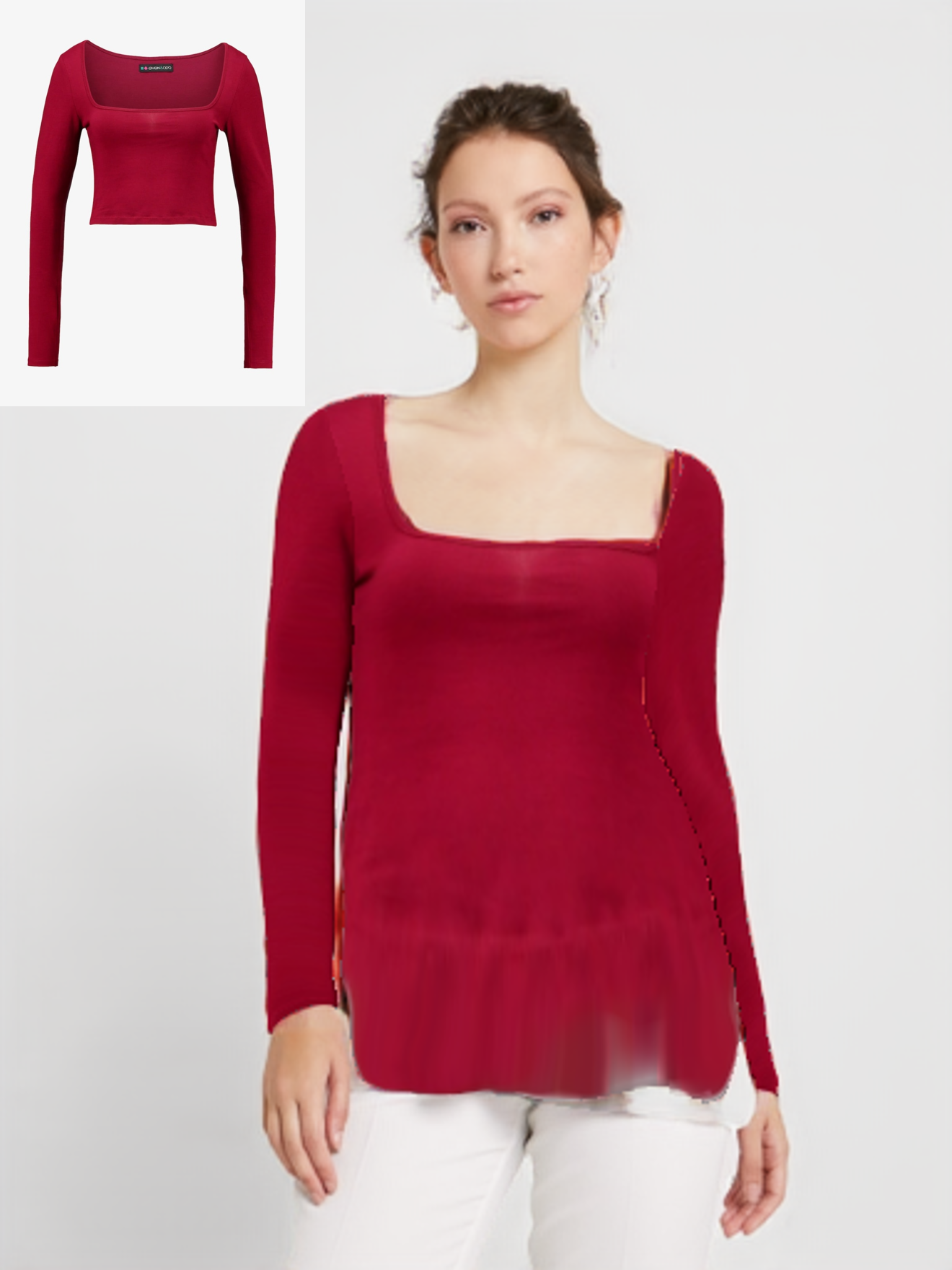} &
\includegraphics[width=0.32\linewidth]{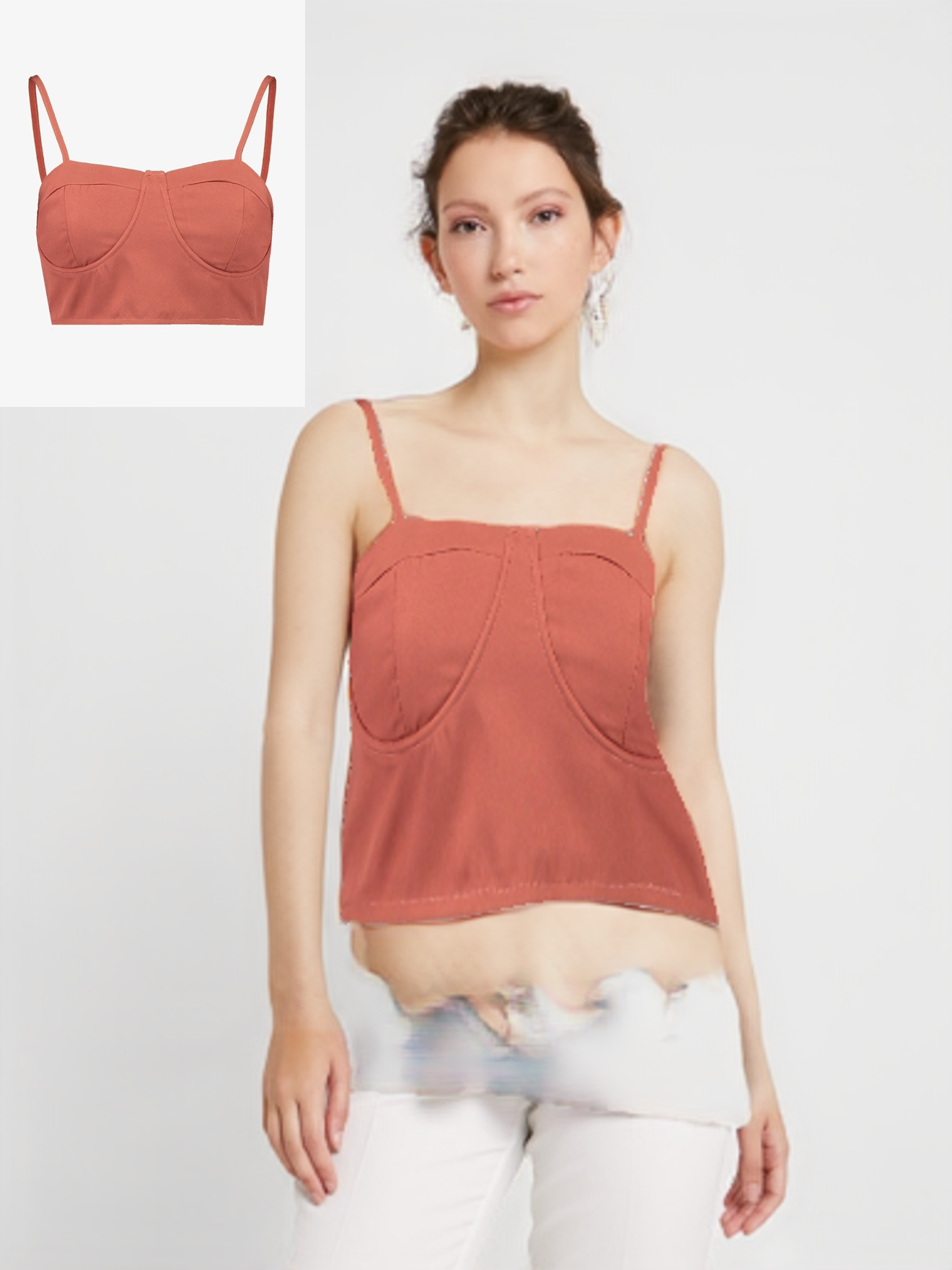} &
\includegraphics[width=0.32\linewidth]{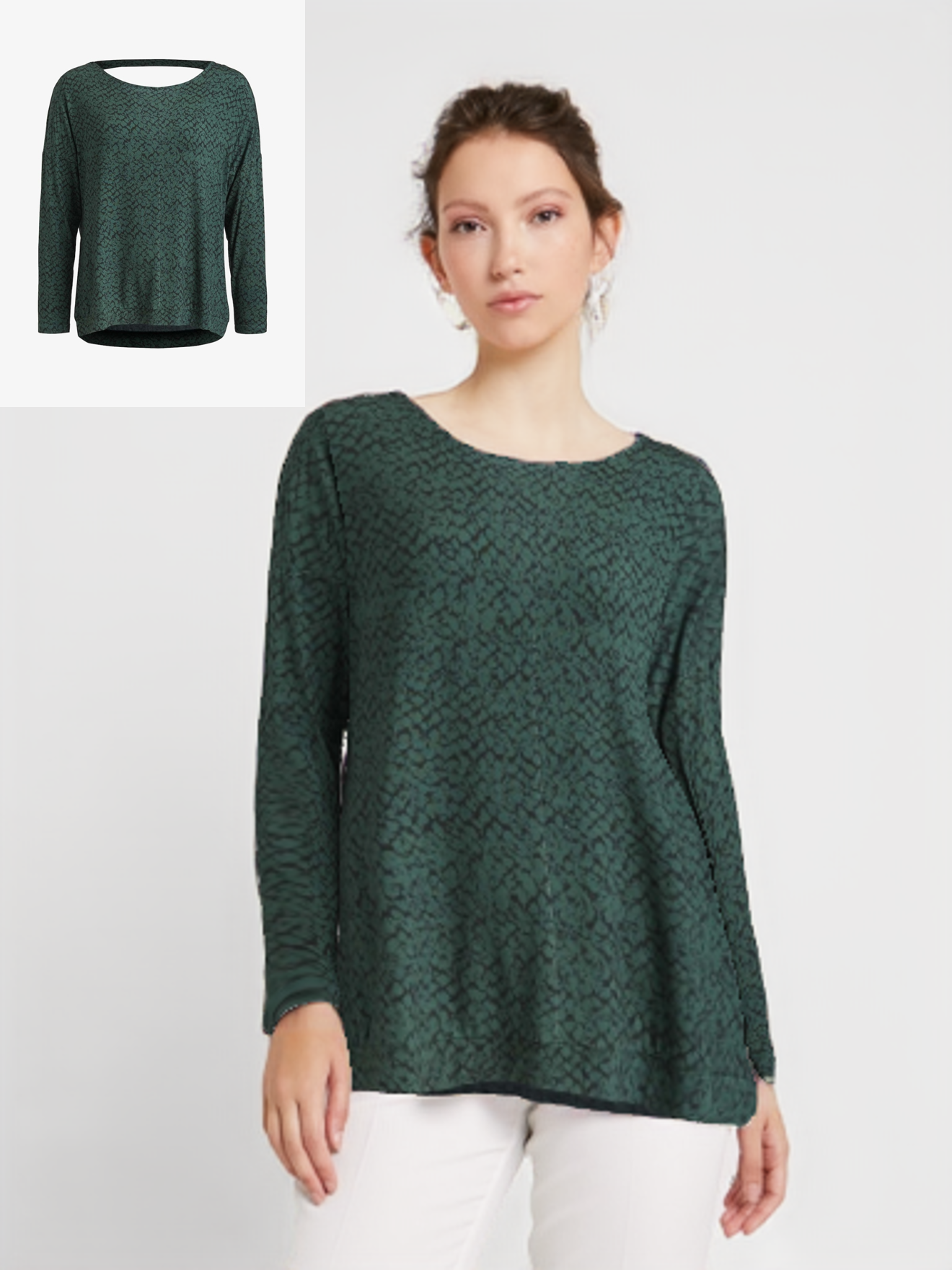} &
\includegraphics[width=0.32\linewidth]{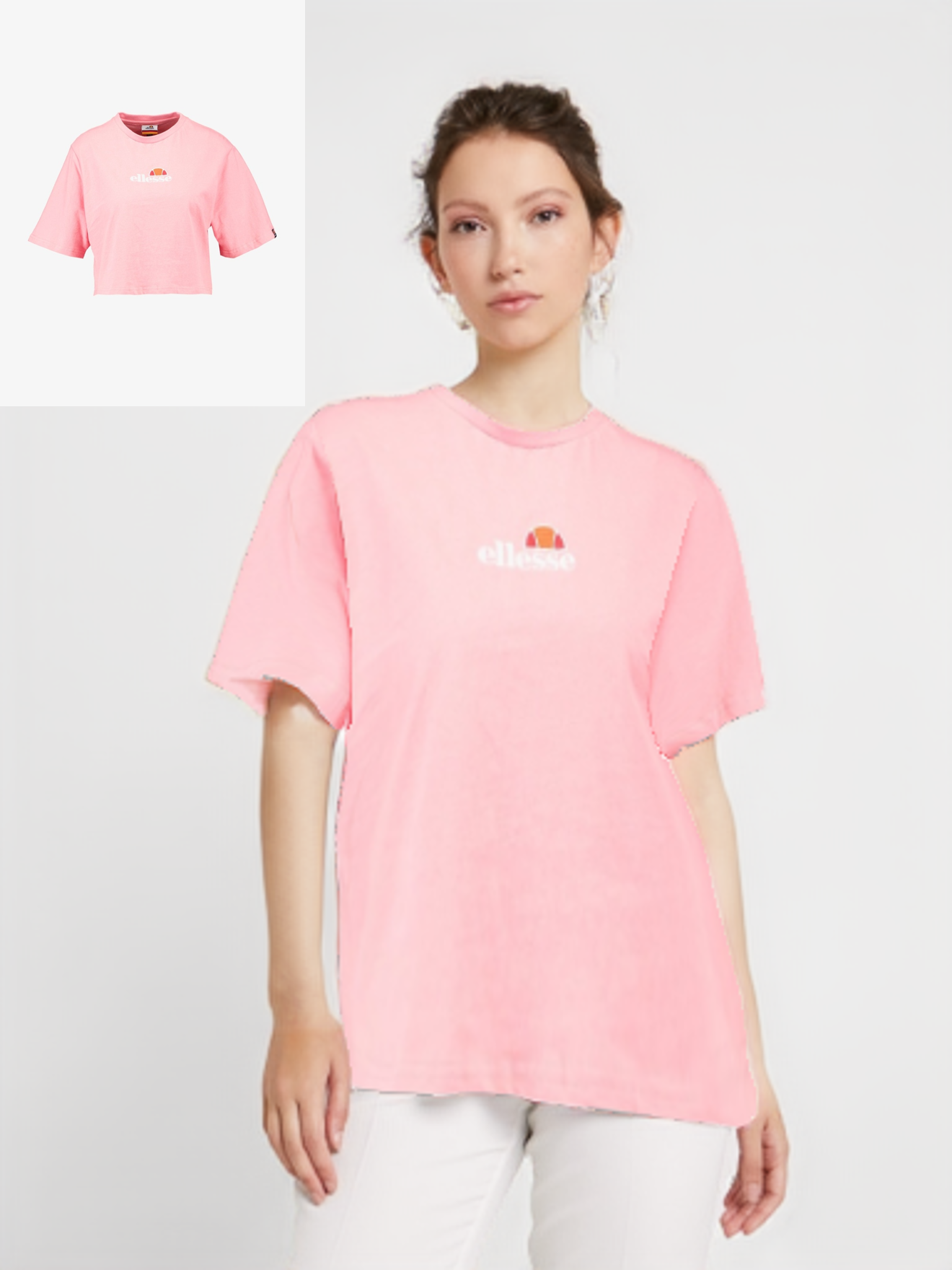} &
\includegraphics[width=0.32\linewidth]{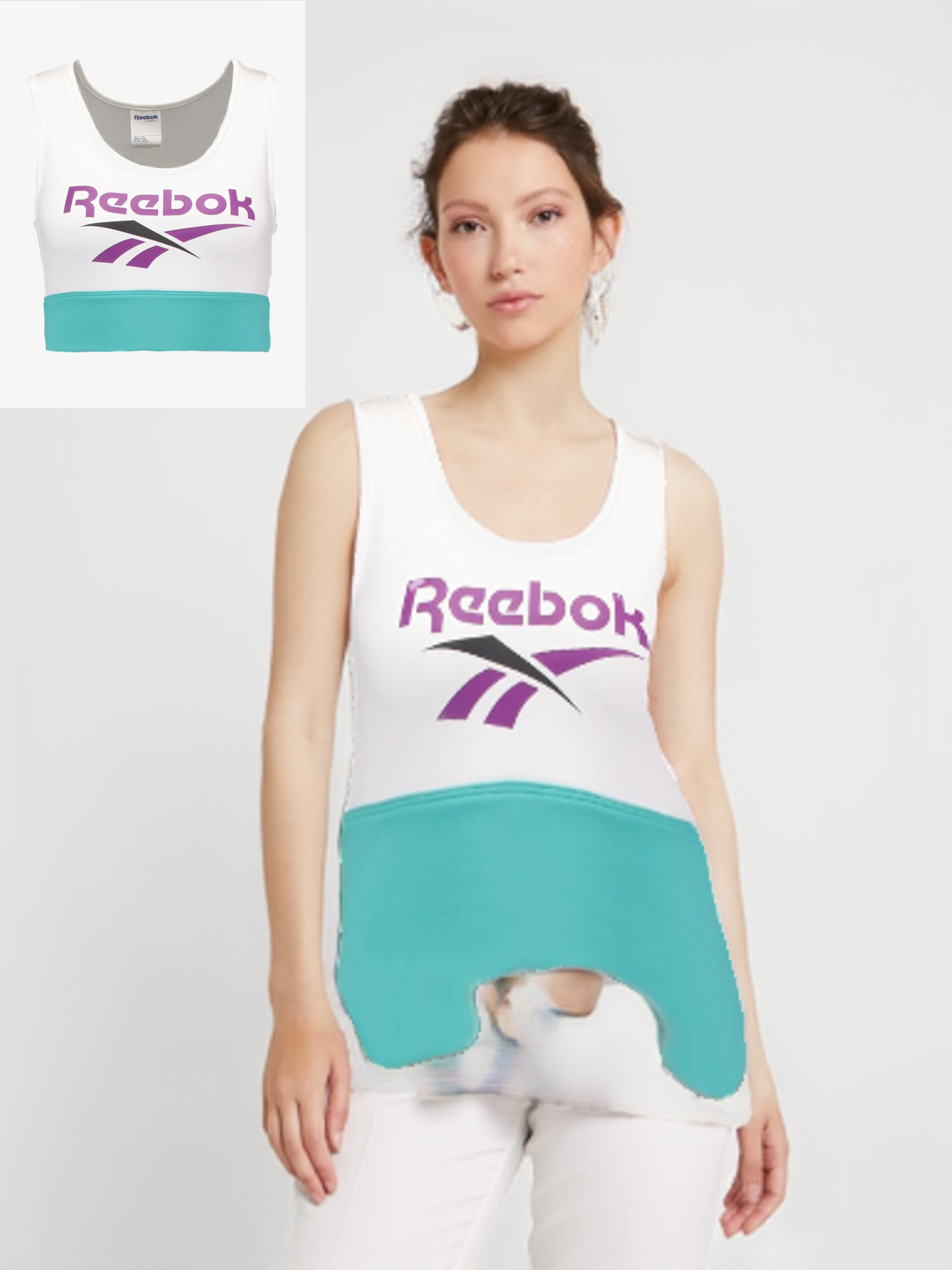} \\
\includegraphics[width=0.157\linewidth]{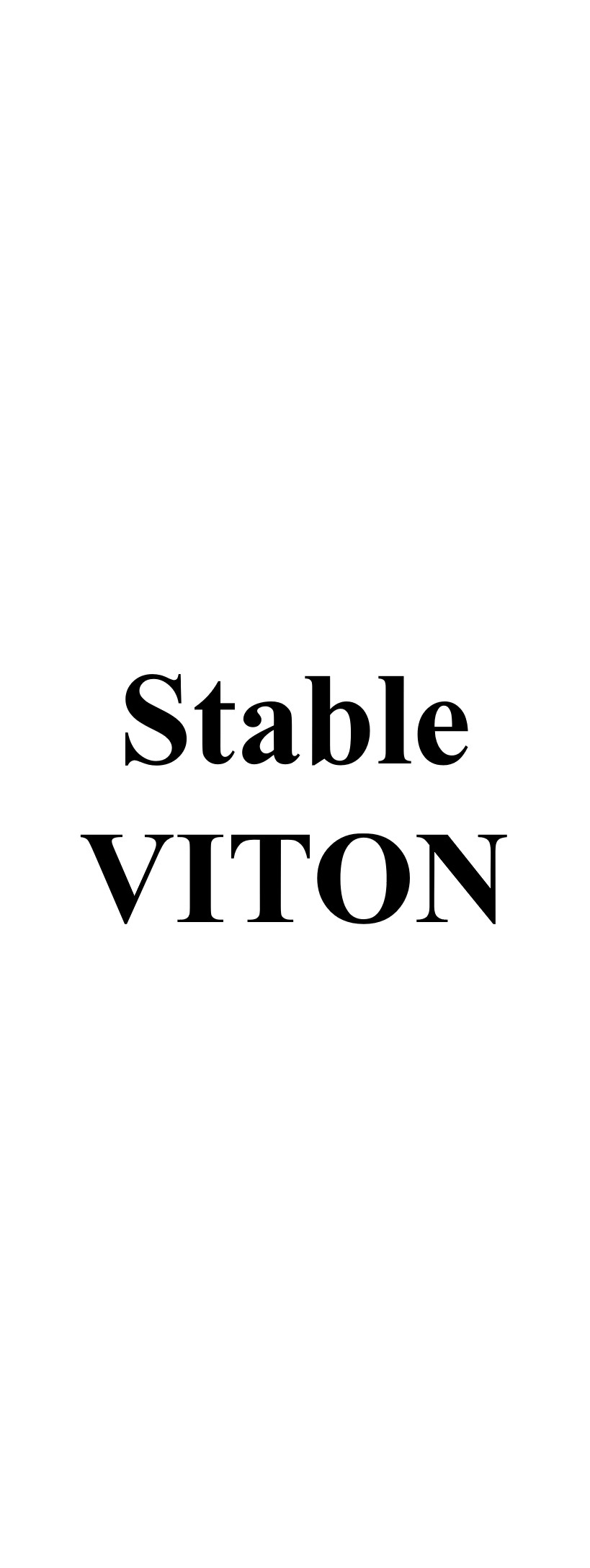} &
\includegraphics[width=0.32\linewidth]{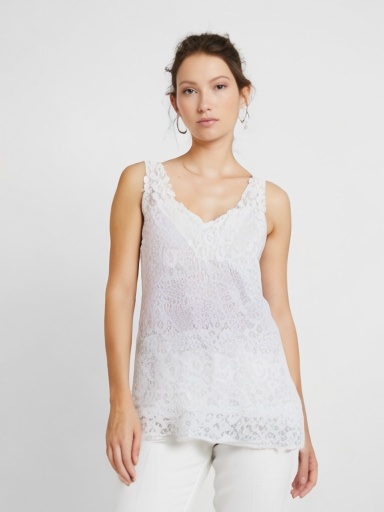} &
\includegraphics[width=0.32\linewidth]{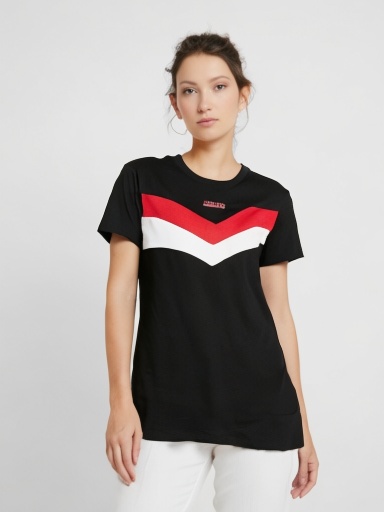} &
\includegraphics[width=0.32\linewidth]{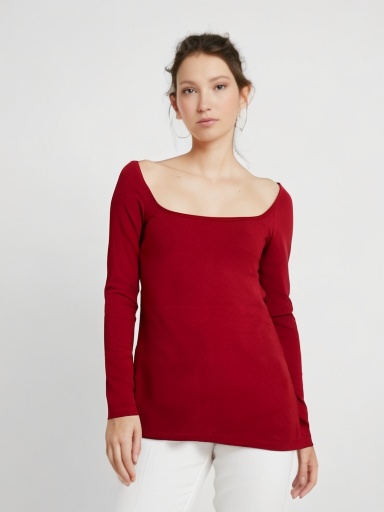} & 
\includegraphics[width=0.32\linewidth]{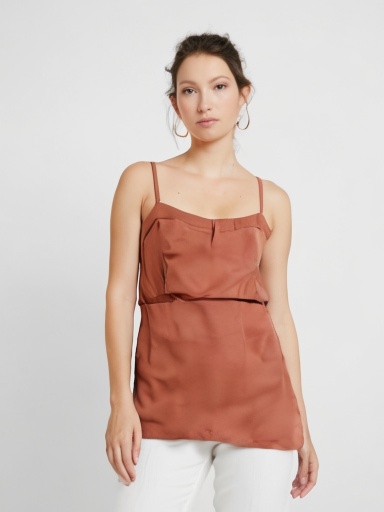} &
\includegraphics[width=0.32\linewidth]{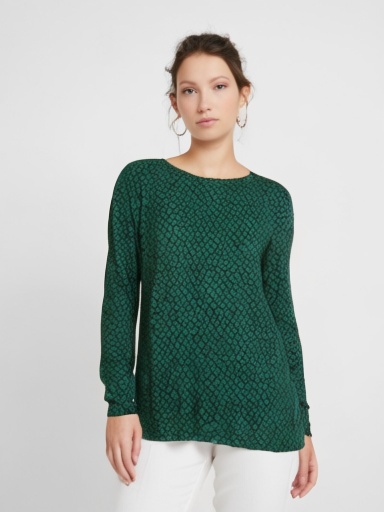} &
\includegraphics[width=0.32\linewidth]{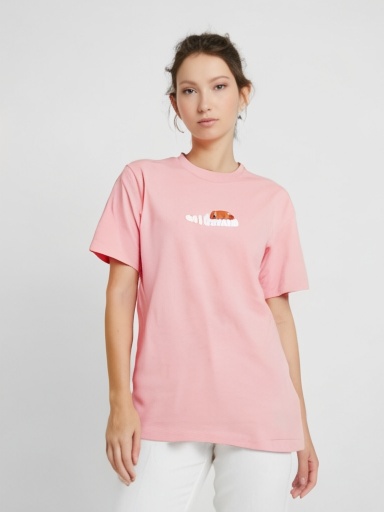} &
\includegraphics[width=0.32\linewidth]{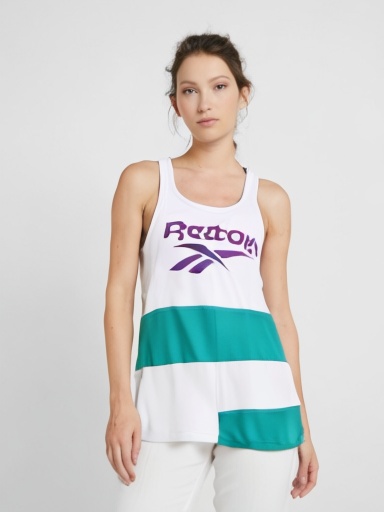} \\
\includegraphics[width=0.157\linewidth]{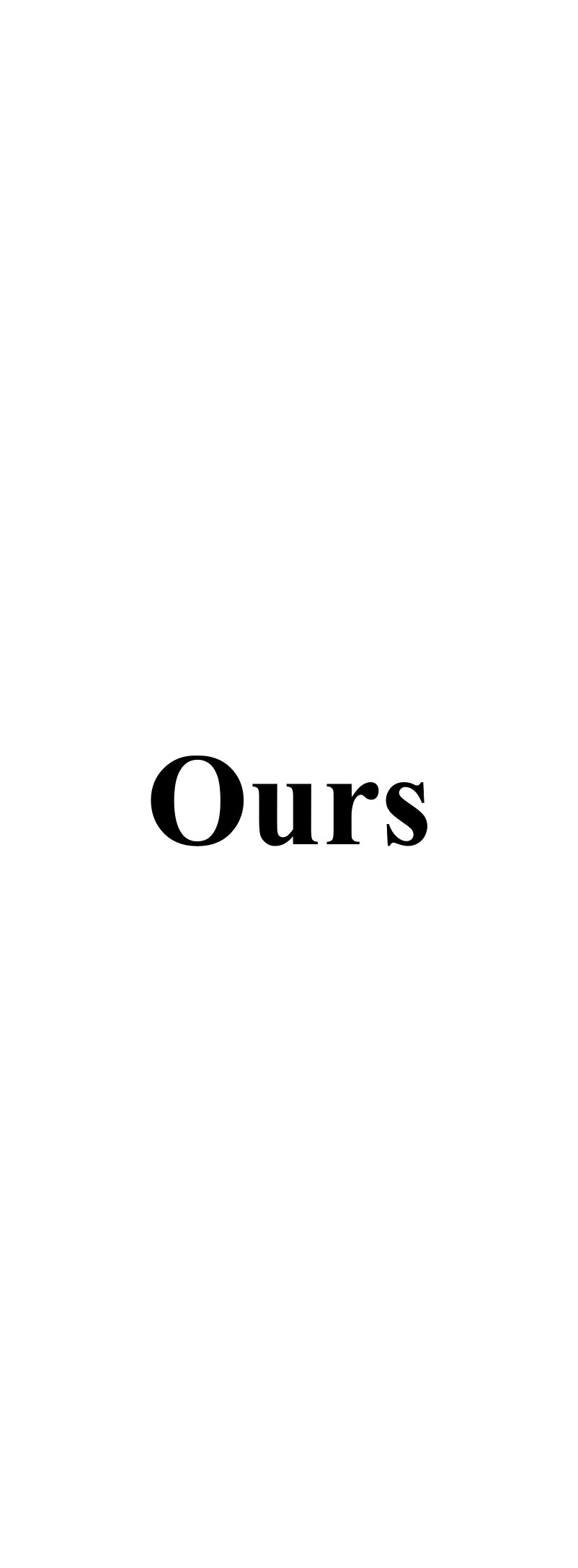} &
\includegraphics[width=0.32\linewidth]{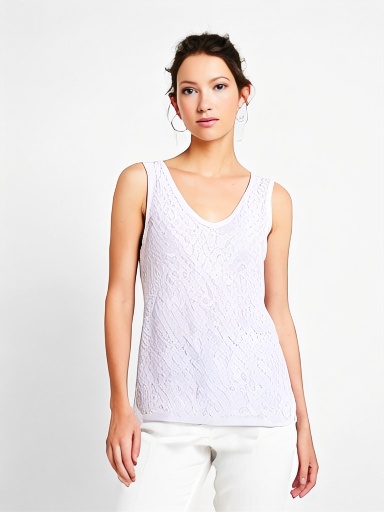} &
\includegraphics[width=0.32\linewidth]{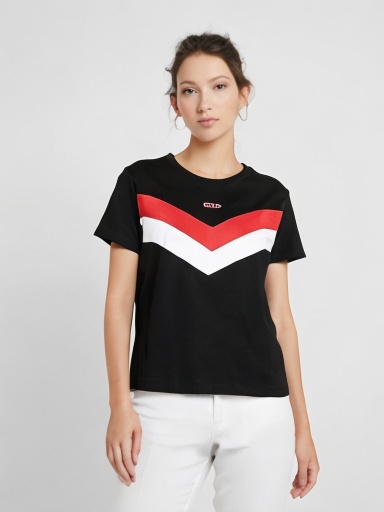} &
\includegraphics[width=0.32\linewidth]{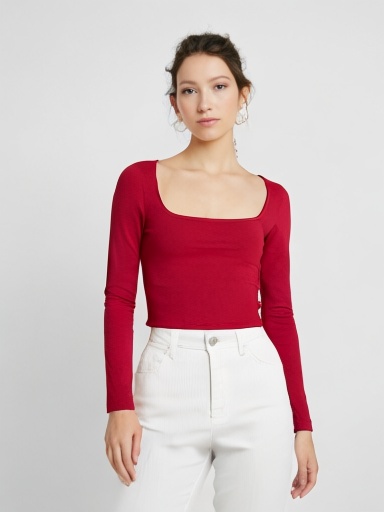} & 
\includegraphics[width=0.32\linewidth]{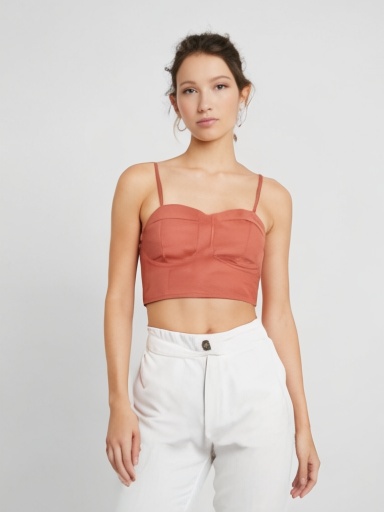} &
\includegraphics[width=0.32\linewidth]{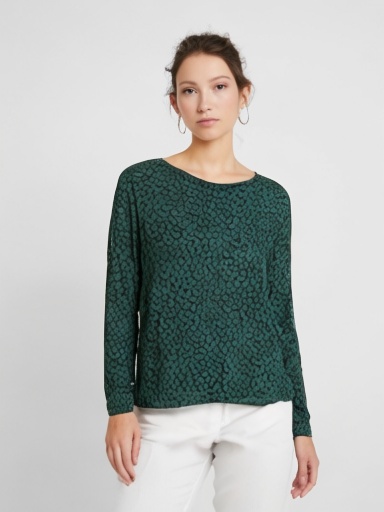} &
\includegraphics[width=0.32\linewidth]{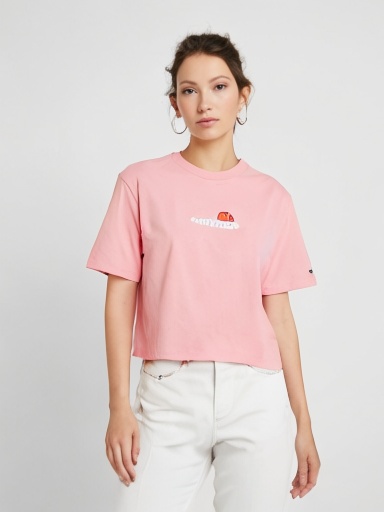} &
\includegraphics[width=0.32\linewidth]{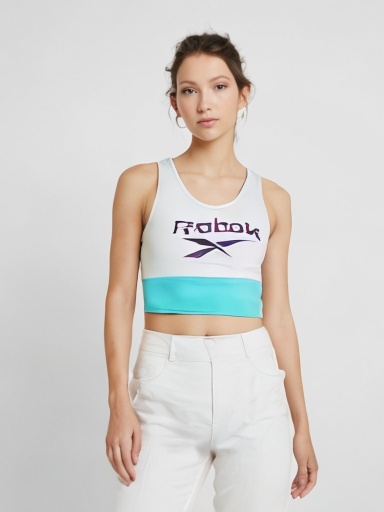}
\end{tabular}
}
\caption{Comparison results on Cross-27.}
\label{fig:comparsion_benchmark}
\end{figure}

\tit{Qualitative analysis} \textcolor{black}{As shown in Fig. \ref{fig:comparison_vitonhd}, compared to previous works, our method faithfully preserves the type of the clothing and correctly warps the clothing. Previous methods stretched the clothing forcefully, leading to very unnatural deformations while also distorting the types of the clothing. Our training paradigm enables the model to accurately predict the lower boundary of clothing and to realistically and reasonably inpaint the area between the top and bottom clothing, achieving a significant visual enhancement during unpaired try-on. Compared to the StableVITON\cite{StableVITON}, while retaining its capability to generate realistic texture details and patterns, we have enhanced the accuracy in semantic alignment between clothing and the model body, as shown in Fig. \ref{fig:comparison_vitonhd} Row 3. The lower boundary of the top is no longer bound to the upper boundary of the bottom clothing; instead, clothing is first aligned at the correct semantic location, and then the missing body or bottom clothing is reconstructed in the remaining area. Our method demonstrates a more pronounced ability to preserve clothing types on Cross-27, as shown in the Fig. \ref{fig:comparsion_benchmark}.}


\begin{table*}[t]
\centering
\caption{Quantitative comparison results on Unpair-2032 and Cross-27.}
\label{tab:try-on_res}
\begin{tabular}{lc cccc c cccc}
\toprule
\textbf{Testsets}& & \multicolumn{4}{c}{\textbf{Unpair-2032}}&&\multicolumn{4}{c}{\textbf{Cross-27}}\\
\cmidrule{3 -6} \cmidrule{8-11}
\textbf{Model} 
& &  \textbf{FID} $\downarrow$ & \textbf{KID} $\downarrow$ & \textbf{SDR} $\downarrow$ & \textbf{S-LPIPS} $\downarrow$ & & \textbf{FID} $\downarrow$ & \textbf{KID} $\downarrow$ & \textbf{SDR} $\downarrow$ & \textbf{S-LPIPS} $\downarrow$\\
\midrule
VITON-HD~\cite{VITON-HD}          & &  16.718          &  9.12             & 0.2156             & 0.1140 & & 39.651 & 10.60 & 0.3664 & 0.1271 \\
HR-VITON~\cite{HR-VITON}          & &  24.601          & 18.03             & 0.2179             & 0.1206 & & 41.862 & 11.73 & 0.3183 & 0.1247 \\
Ladi-VITON~\cite{LaDI-VTON}       & &   9.324          &  1.46             & 0.2264             & 0.1247 & & 38.920 & \textbf{9.49} & 0.3635 & 0.1037 \\
MGD~\cite{MGD}                    & &  12.949          &  3.76             & 0.2266             & 0.1247 & & -      & -     & -      & -      \\
DCI-VTON~\cite{DCI-VTON}          & &\textbf{8.750}    &  \textbf{0.94}             & 0.2347             & 0.1094 & & 39.775 & 10.75 & 0.3916 & 0.1044 \\
GP-VTON~\cite{GP-VTON}            & &  14.660          &  1.50             & 0.2076             & \underline{0.1010} & & 41.406 & 11.48 & 0.3149 & \underline{0.0915} \\
StableVITON~\cite{StableVITON}    & &\underline{9.005} &  \underline{1.05}    & 0.2127             & 0.1031 & & \textbf{37.687} & \underline{10.59} & 0.3549 & 0.0977 \\ \hline
Baseline                          & &   9.222          &  1.30 & \underline{0.2047} & 0.1014 & & \underline{38.353} & 10.69 & \underline{0.3047} & 0.0927 \\
Ours                              & &   9.517          &  1.38             & \textbf{0.1946}    & \textbf{0.0991} & & 38.783 & 10.79 & \textbf{0.2240} & \textbf{0.0904} \\
\bottomrule
\end{tabular}
\label{tab:comparison_vitonhd}
\end{table*}

\begin{table*}[t]
\centering
\caption{Quantitative comparison results on different masks.}
\begin{tabular}{lc cccc c cccc}
\toprule
\textbf{Testsets}& & \multicolumn{4}{c}{\textbf{Unpair-2032}}&&\multicolumn{4}{c}{\textbf{Cross-27}}\\
\cmidrule{3-6} \cmidrule{8-11}
\textbf{Model} 
& & \textbf{FID} $\downarrow$ & \textbf{KID} $\downarrow$ & \textbf{SDR} $\downarrow$ & \textbf{S-LPIPS} $\downarrow$ & & \textbf{FID} $\downarrow$ & \textbf{KID} $\downarrow$ & \textbf{SDR} $\downarrow$ & \textbf{S-LPIPS} $\downarrow$\\
\midrule
Big Mask        & & 10.584 & 2.08 & 0.2801 & 0.1169 & & \textbf{34.291} &  \textbf{7.49} & 0.5009 & 0.1158 \\
Small Mask      & &  9.748 & 1.76 & 0.2112 & 0.1014 & & 39.743 & 11.93 & 0.3050 & \underline{0.0923} \\
Normal Mask     & &  \textbf{9.222} & \textbf{1.30} & \underline{0.2047} & \underline{0.1014} & & \underline{38.353} &   \underline{10.69} & \underline{0.3047} & 0.0927 \\
Adaptive Mask        & &  \underline{9.517} & \underline{1.38} & \textbf{0.1946} & \textbf{0.0991} & & 38.783 & 10.79 & \textbf{0.2240} & \textbf{0.0904} \\
\bottomrule
\end{tabular}
\label{tab:mask_ablation}
\end{table*}

\tit{Quantitative analysis} We conducted quantitative analysis using the Unpair-2032 and our Cross-27. Since MGD\cite{MGD} has not open-sourced its sketch parser, we are unable to obtain the test results of MGD on the Cross-27. As of the writing of this manuscript, StableVITON\cite{StableVITON} has not provided training code, so we replicated the code based on the information provided by the paper, it is denoted as Baseline in Table \ref{tab:comparison_vitonhd}. On the Unpair-2032, our training paradigm inherits the advantages of StableVITON and achieves better SDR and S-LPIPS scores compared to Baseline. Although our FID and KID scores are worse than those of some previous works, this does not imply that our try-on quality is similarly inferior; we will explain the reasons for this occurrence in the section \ref{sec:metric_eff}. The effectiveness of our method is further demonstrated on Cross-27, where, compared to Baseline, we lead significantly in SDR and S-LPIPS scores.

\begin{figure*}[h]
\centering
\Large
\resizebox{\linewidth}{!}{
\begin{tabular}{cccc c cccc}
\textbf{Small Mask} & \textbf{Big Mask} & \textbf{Normal Mask} & \textbf{Adp-Mask} & \textbf{Clothing} &  \textbf{Small Mask} & \textbf{Big Mask} & \textbf{Normal Mask} & \textbf{Adp-Mask}\\
\addlinespace[0.08cm]
\includegraphics[width=0.32\linewidth]{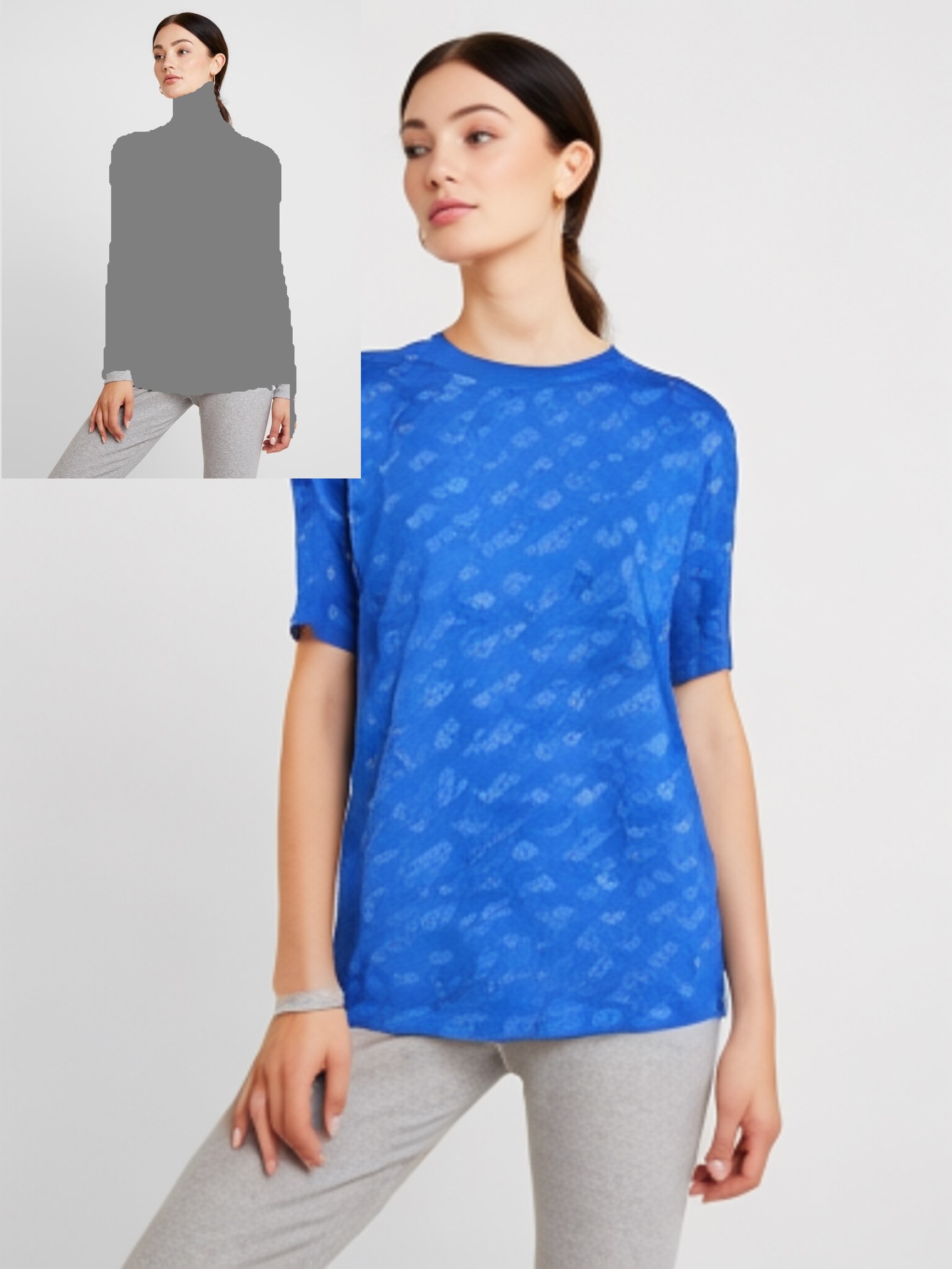} &
\includegraphics[width=0.32\linewidth]{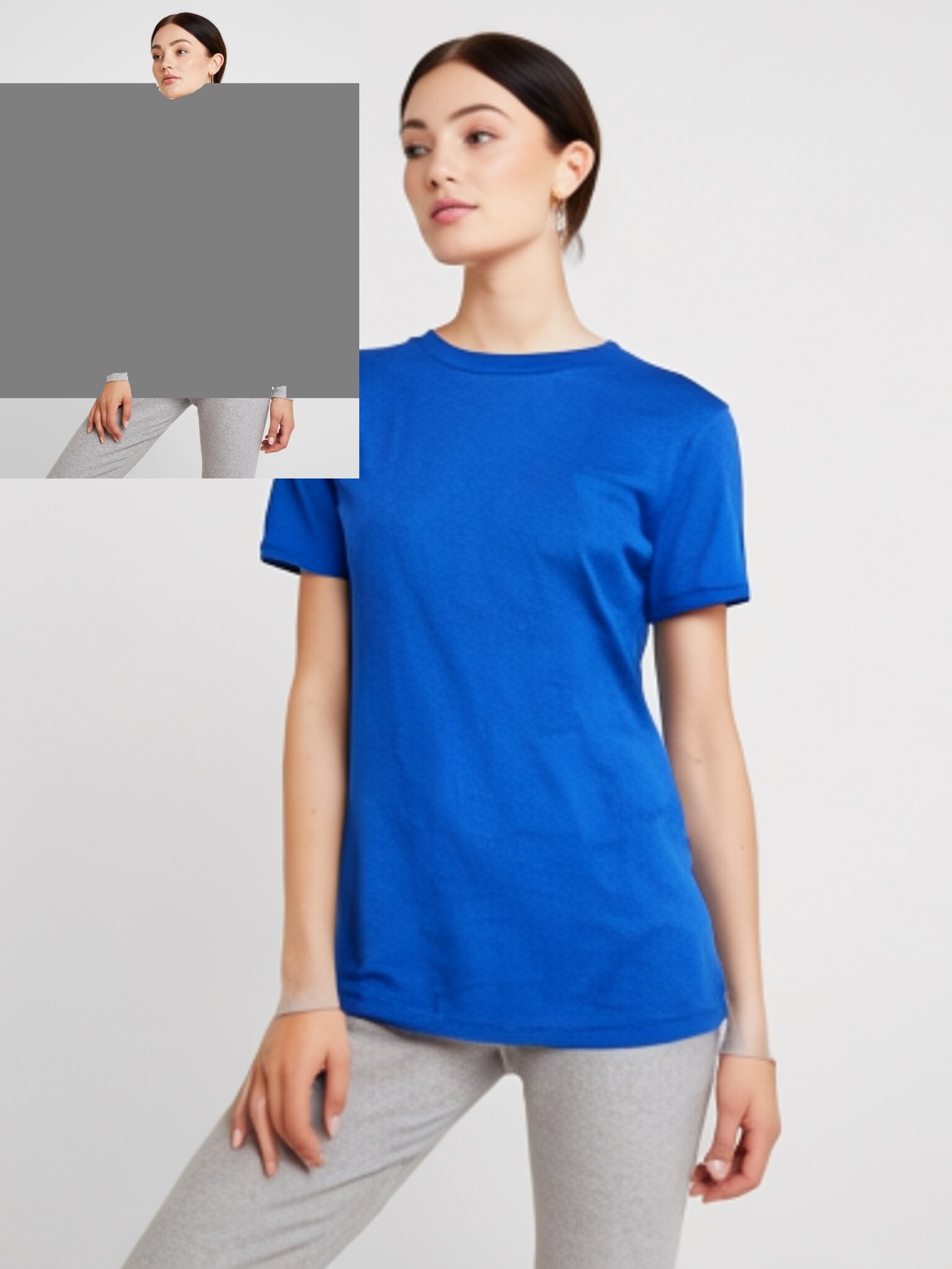} & 
\includegraphics[width=0.32\linewidth]{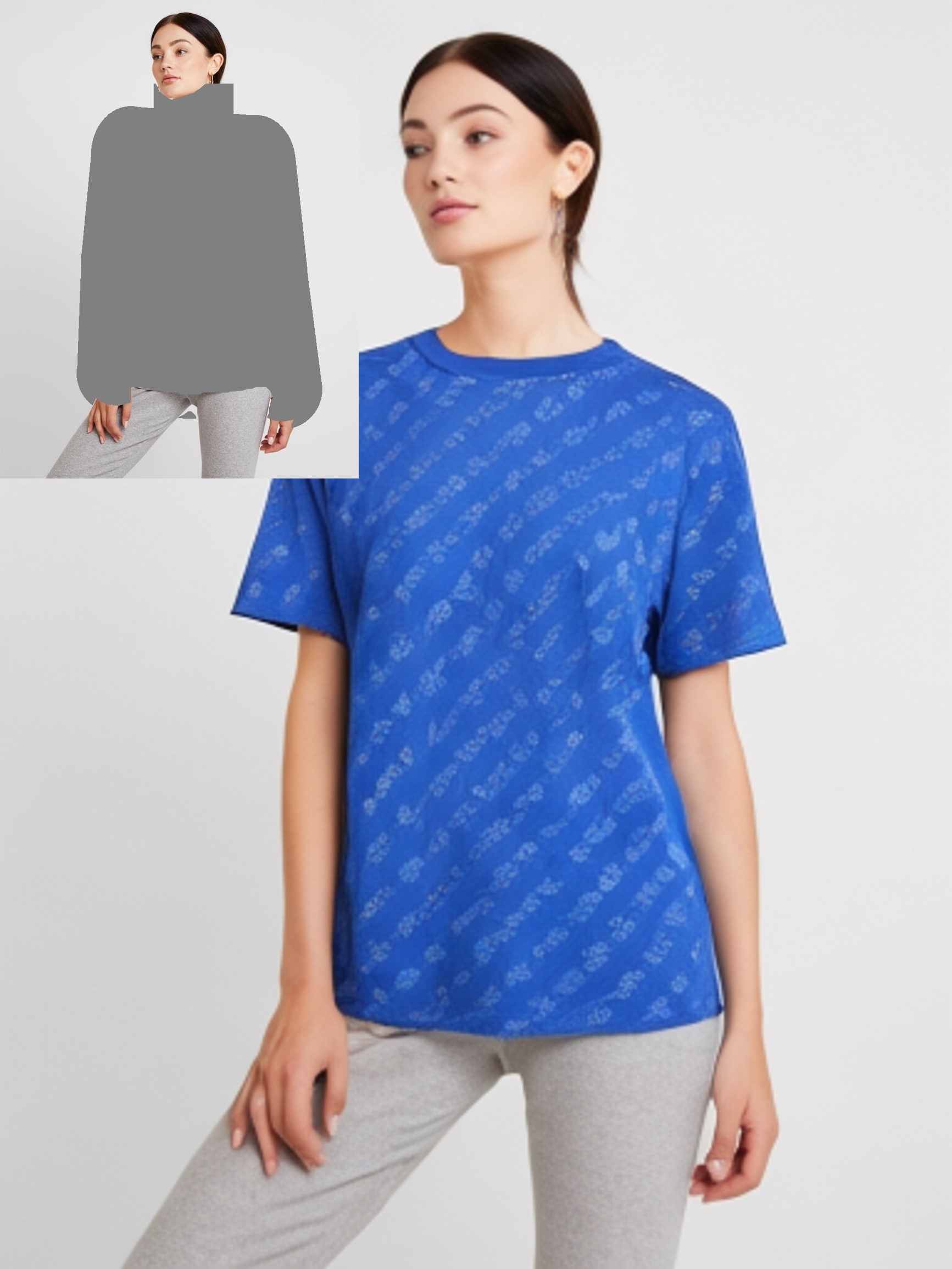} &
\includegraphics[width=0.32\linewidth]{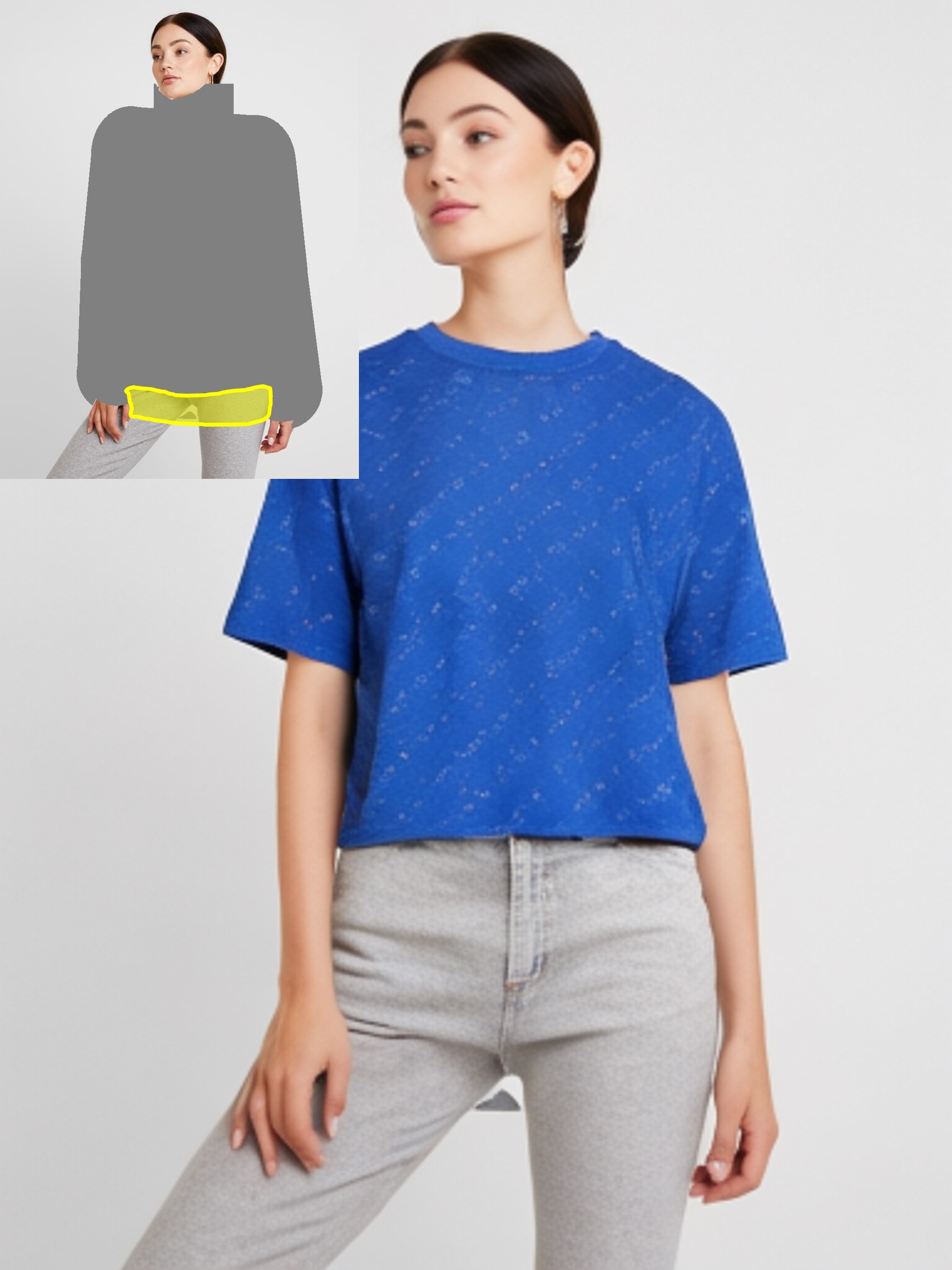} &
\includegraphics[width=0.157\linewidth]{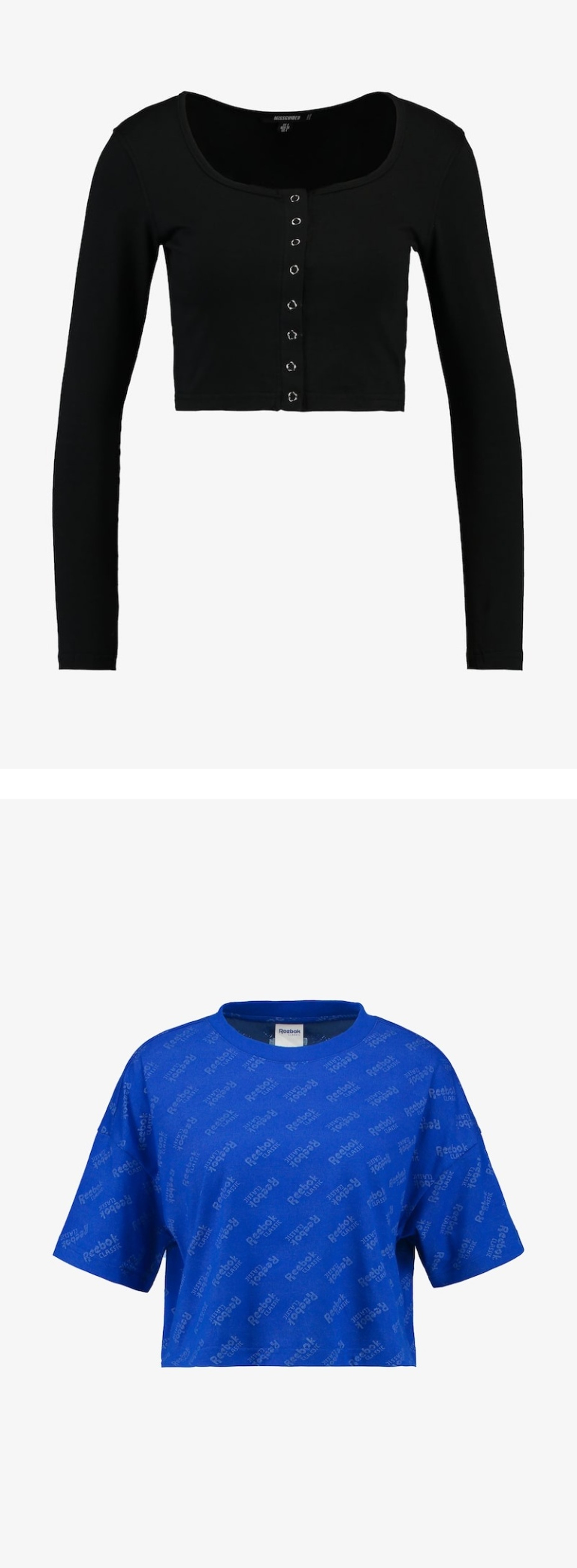} &
\includegraphics[width=0.32\linewidth]{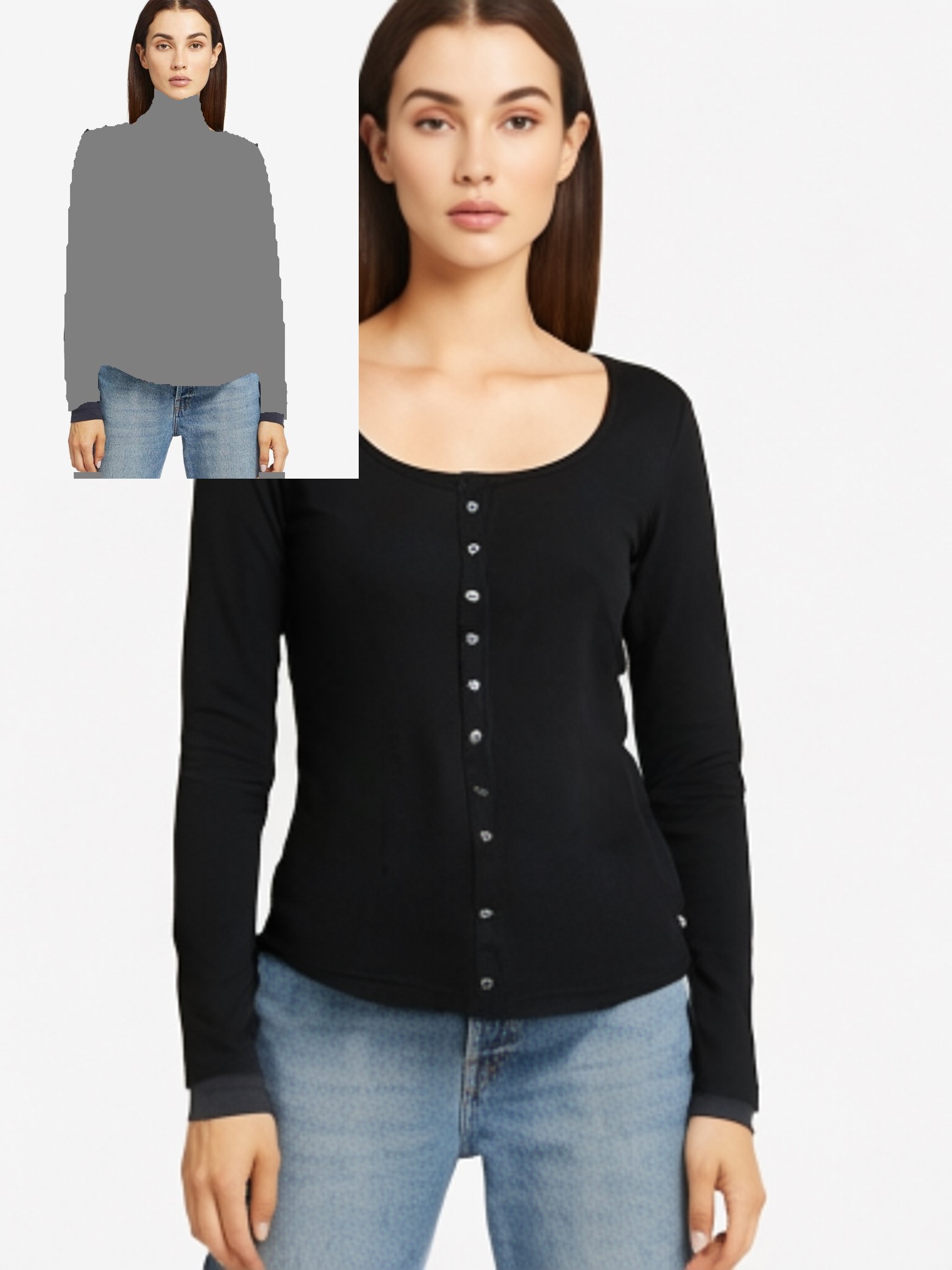} &
\includegraphics[width=0.32\linewidth]{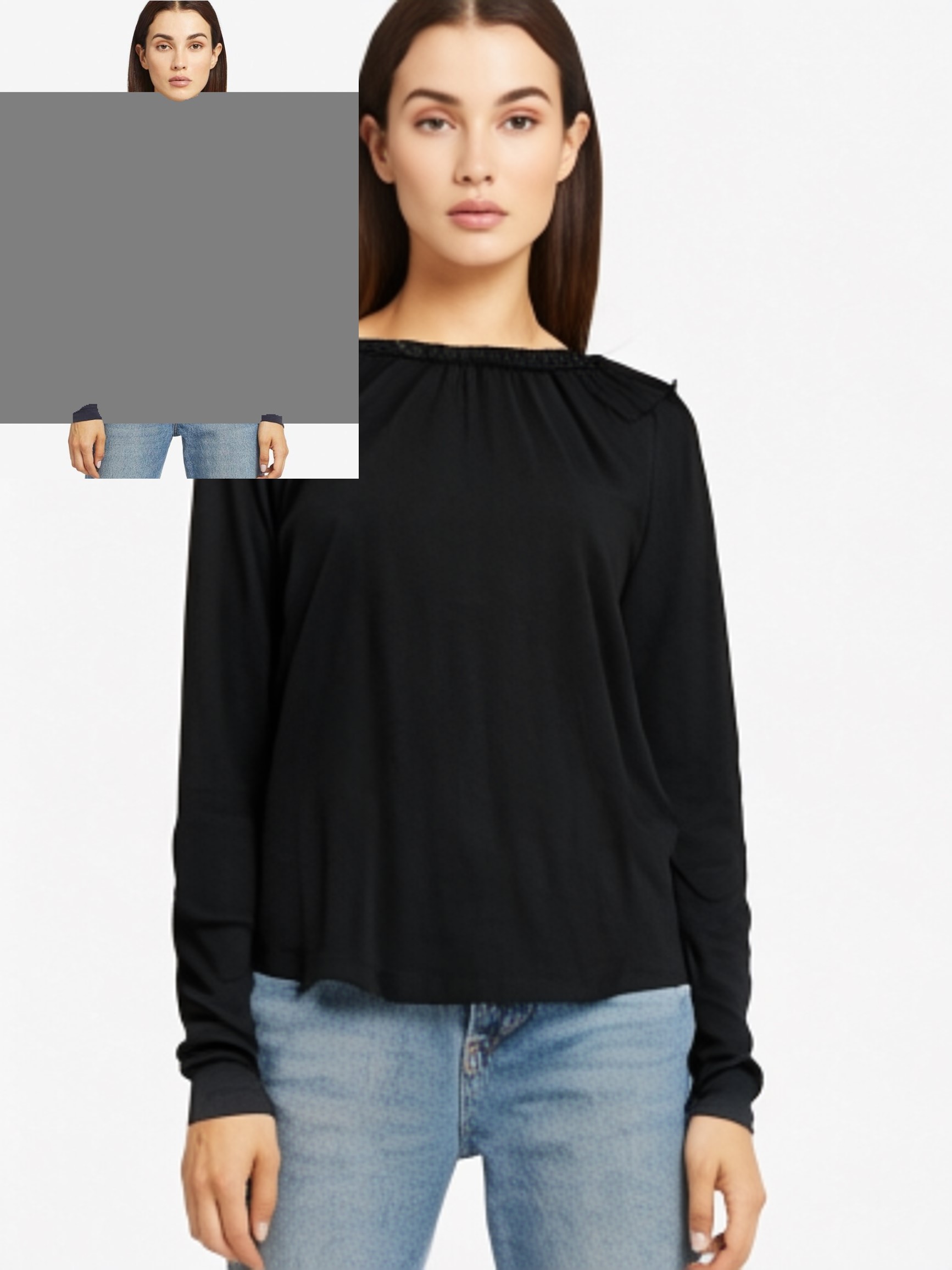} &
\includegraphics[width=0.32\linewidth]{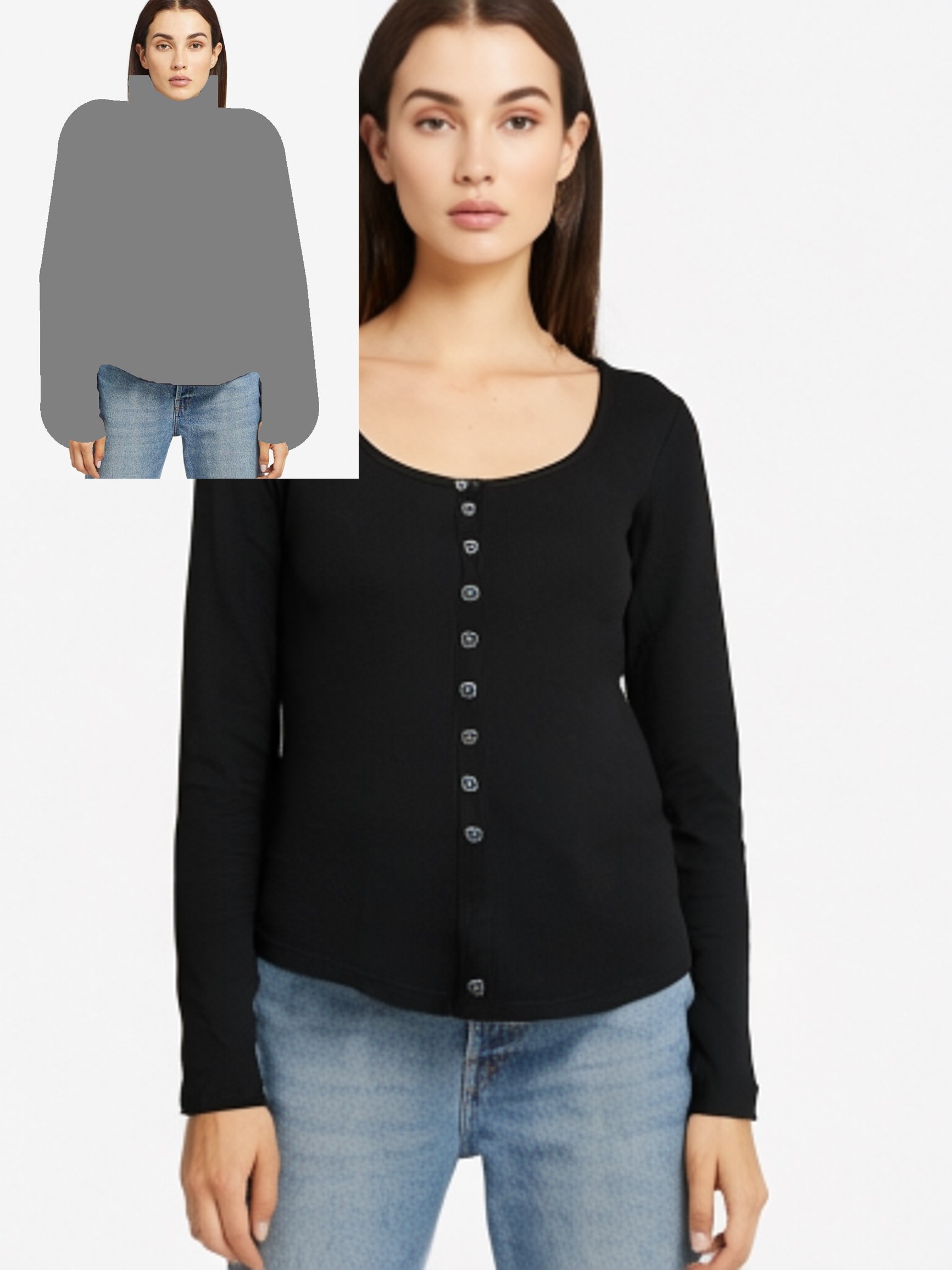} & 
\includegraphics[width=0.32\linewidth]{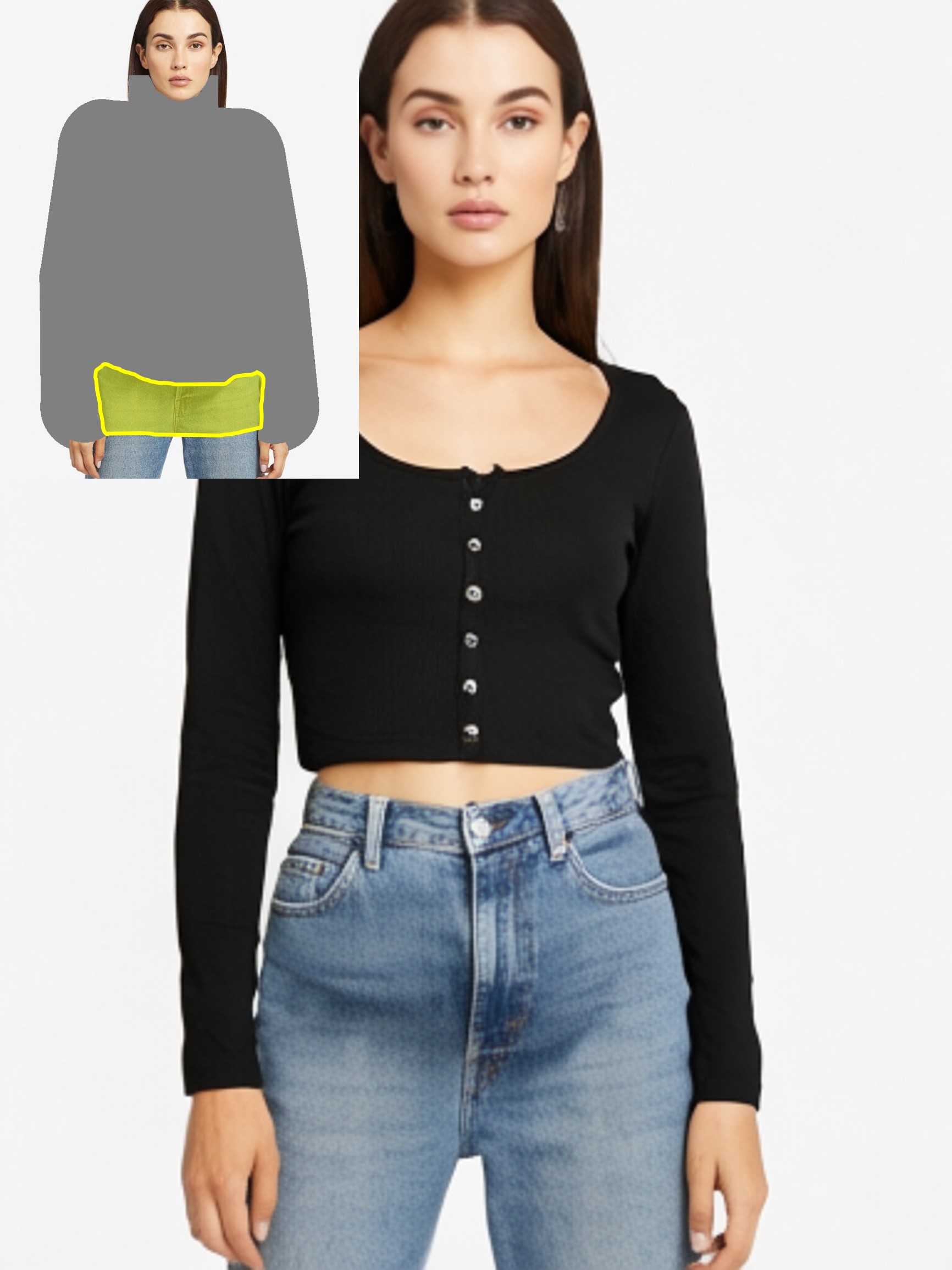}\\
\end{tabular}
}
\caption{Comparison results from training with different masks.}
\label{fig:ablation}
\end{figure*}

\tit{Ablation analysis} To verify the effectiveness of our adaptive mask training paradigm, we conducted training with three different masks in addition to our adaptive mask training paradigm. Small Mask: following the mask approach in DCI-VITON\cite{DCI-VTON}, we inflate the semantic area of the top clothing to serve as the mask; Big Mask: following the method in TryOnDiffusion\cite{TryonDiffusion}, we use a rectangular area below the face and above the waist as the mask; Normal Mask: we adopt the same mask approach as StableVITON\cite{StableVITON}. As shown in Fig. \ref{fig:ablation}, only our training paradigm effectively addresses the flaw of the lower boundary of the top clothing, preserving the type of the target clothing. Table \ref{tab:mask_ablation} presents the quantitative comparison. The adaptive mask paradigm maintains the same advantages in the SDR and S-LPIPS metrics as seen in the visualization results. We will explain in the section \ref{sec:metric_eff}  why the FID and KID scores for the adp-mask under Cross-27 testing are inferior to those of other masks.

\begin{figure*}[t]
  \centering
  \includegraphics[width=0.6\linewidth]{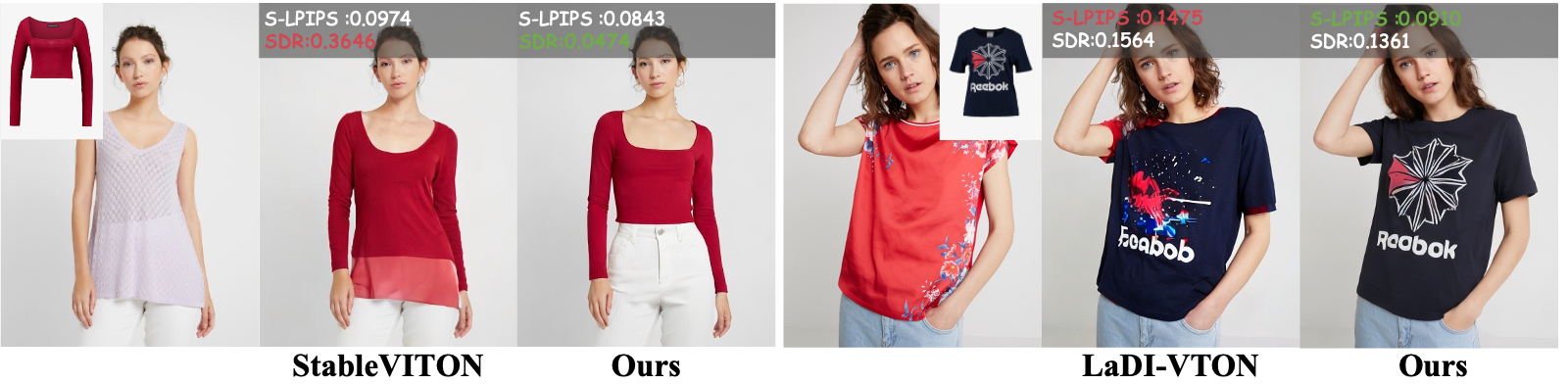}
   \caption{Visual analysis of SDR and S-LPIPS metrics.}
   \label{fig:metric_analysis_reb}
\end{figure*}

\begin{figure*}[h]
    \centering
    \begin{subfigure}[b]{0.4\textwidth}
        \includegraphics[width=\textwidth]{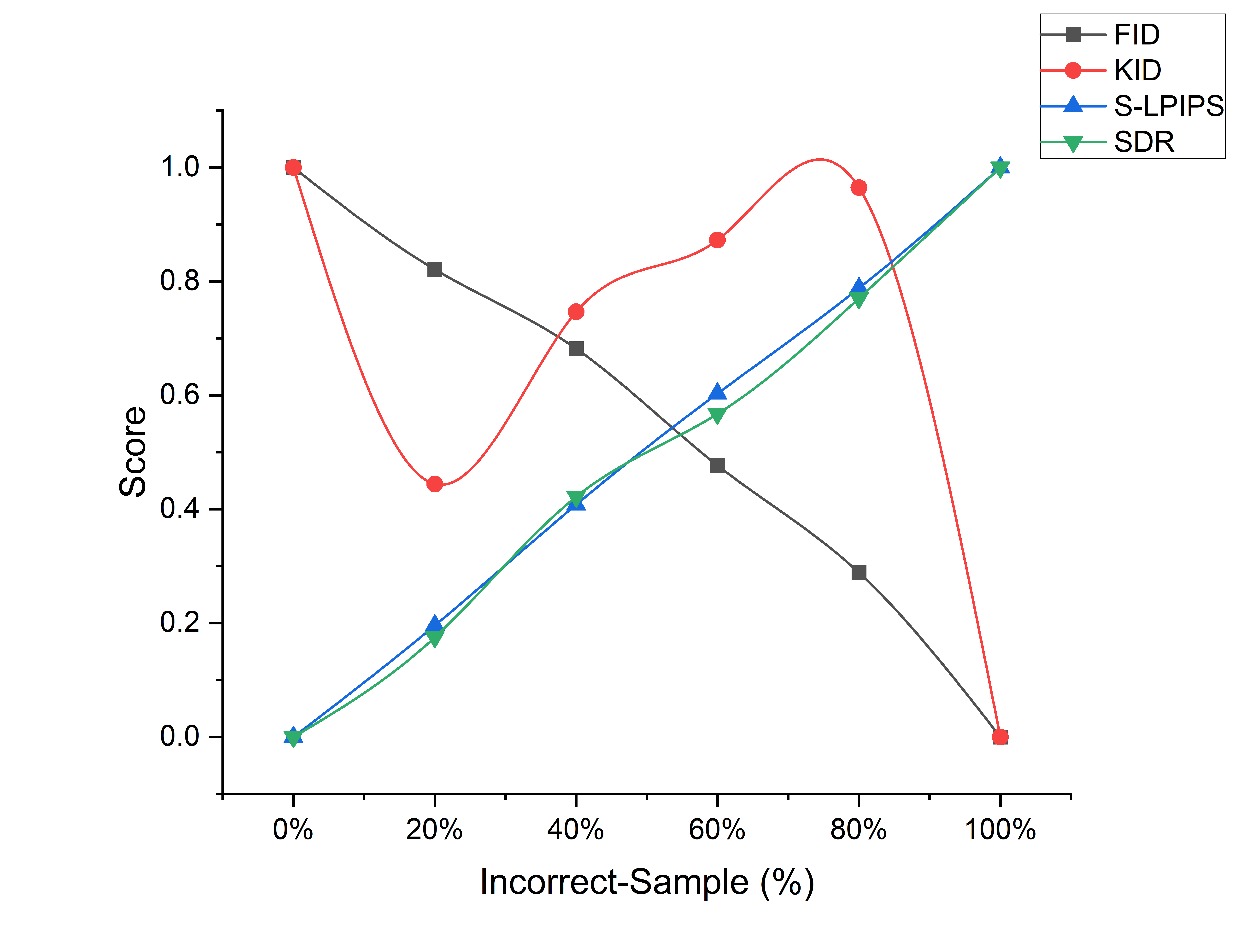}
        \caption{Unpair-2032}
        \label{fig:metric_unpair2032}
    \end{subfigure}
    \begin{subfigure}[b]{0.4\textwidth}
        \includegraphics[width=\textwidth]{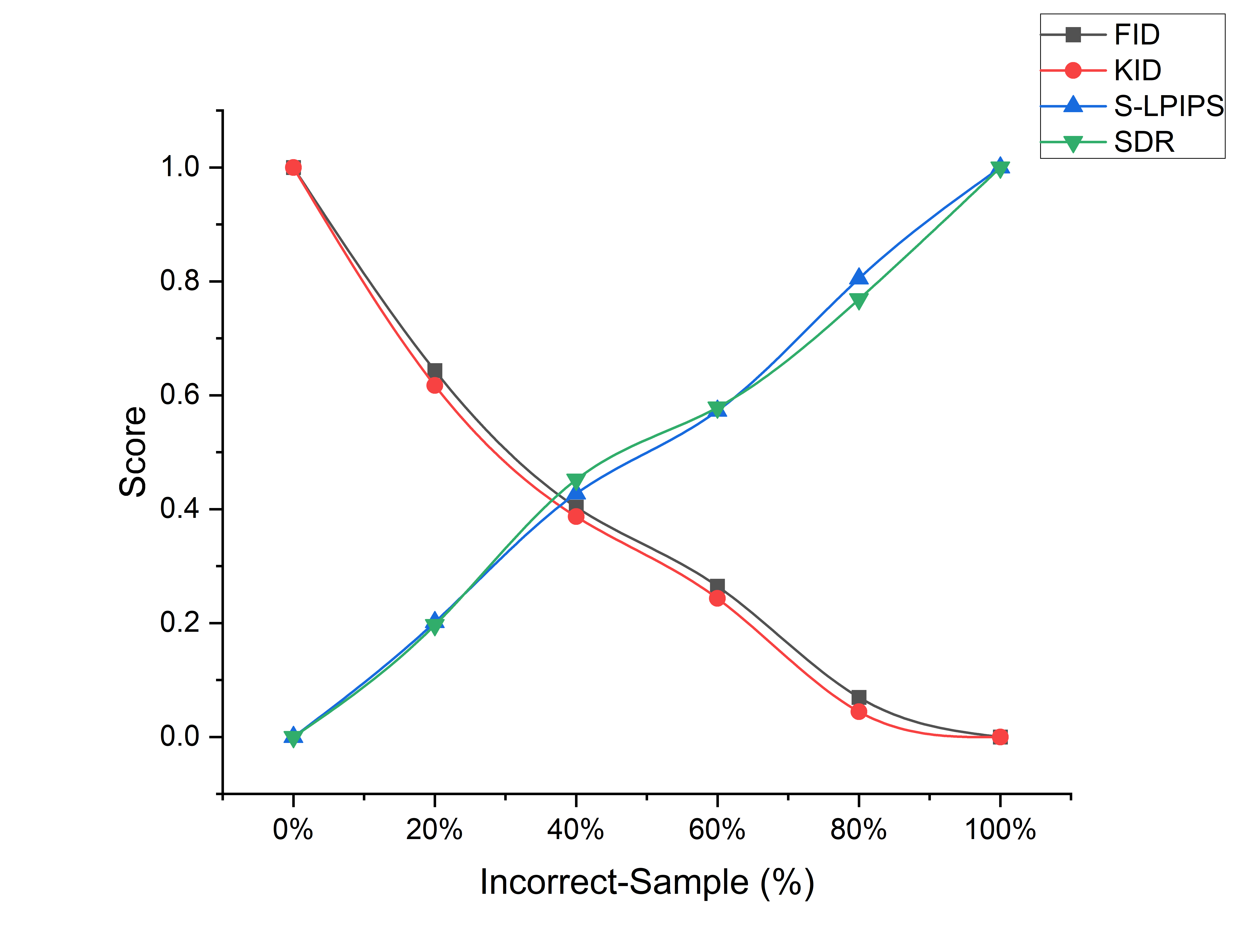}
        \caption{Cross-27}
        \label{fig:sub2}
    \end{subfigure}
    \caption{Comparison results after normalization of various metrics.}
    \label{fig:metric_benchmark}
\label{fig:metric_effective}
\end{figure*}

\begin{table*}[t]
\centering
\caption{Comparison of scores with different proportions of \textit{Incorrect samples}.}
\begin{tabular}{lc cccc c cccc}
\toprule
\textbf{Testsets}& & \multicolumn{4}{c}{\textbf{Unpair-2032}}&&\multicolumn{4}{c}{\textbf{Cross-27}}\\
\cmidrule{3-6} \cmidrule{8-11}
\textbf{Incorrect} 
& & \textbf{FID} $\downarrow$ & \textbf{KID} $\downarrow$ & \textbf{SDR} $\downarrow$ & \textbf{S-LPIPS} $\downarrow$ & & \textbf{FID} $\downarrow$ & \textbf{KID} $\downarrow$ & \textbf{SDR} $\downarrow$ & \textbf{S-LPIPS} $\downarrow$\\
\midrule
\textbf{0\%}       & & 9.517 & 1.38 & \textbf{0.1946} & \textbf{0.0991} & & 38.783 & 10.79 & \textbf{0.2240} & \textbf{0.0904} \\
\textbf{20\%}      & & 9.092 & \underline{1.33} & \underline{0.2330} & \underline{0.1079} & & 30.241 &  7.19 & \underline{0.2994} & \underline{0.1013} \\
\textbf{40\%}      & & 8.761 & 1.36 & 0.2874 & 0.1175 & & 24.504 &  4.98 & 0.3976 & 0.1136 \\
\textbf{60\%}      & & 8.275 & 1.37 & 0.3193 & 0.1263 & & 21.171 &  3.60 & 0.4463 & 0.1215 \\
\textbf{80\%}      & & \underline{7.828} & 1.38 & 0.3640 & 0.1347 & & \underline{16.496} &  \underline{1.70} & 0.5194 & 0.1341 \\
\textbf{100\%}     & & \textbf{7.143} & \textbf{1.29} & 0.4145 & 0.1442 & & \textbf{14.836} &  \textbf{1.27} & 0.6084 & 0.1447 \\
\bottomrule
\end{tabular}
\label{tab:metric_effective}
\end{table*}

\subsection{Analysis of Metric Effectiveness}
\label{sec:metric_eff}
\textcolor{black}{Since FID\cite{FID} and KID\cite{KID} measure the similarity between the distributions of try-on results and test dataset, the correctness of the try-on is not reflected in the similarity between the two sets of data distributions. This leads to the uncertainty in the evaluation results of FID and KID. SDR and S-LPIPS calculate each try-on instance, allowing for a more accurate and stable reflection of the quality of the try-on results.
We designed an experiment to demonstrate the advantages of our metrics compared to FID and KID. Initially, we define the paired try-on images (model's clothes remain unchanged) as \textit{Incorrect samples} in the unpaired testing. We then mix these \textit{Incorrect samples} into the unpaired try-on results (generated by our method) at varying proportions for metric calculation. It is evident that as the proportion of \textit{Incorrect samples} increases, the overall quality of the unpaired try-on results deteriorates, and the evaluation metrics should reflect this phenomenon.}

\textcolor{black}{We conducted experiments on both Unpair-2032 and Cross-27, mixing \textit{Incorrect samples} into the unpaired results at intervals of 20\%, with the calculation results of all metrics shown in Table \ref{tab:metric_effective}. To more intuitively reflect the sensitivity of each metric to \textit{Incorrect samples}, we normalized the score and plotted it in Fig. \ref{fig:metric_effective}. FID and KID both fail to accurately reflect the negative impact of \textit{Incorrect samples} on try-on results. The inclusion of \textit{Incorrect samples} actually makes the data distribution of try-on results closer to that of the test dataset, leading to a trend in FID that is contrary to the actual situation on both test sets. Compared to FID, KID's stronger and more precise comparison capability does not benefit the evaluation of try-on results but rather results in chaotic outcomes on the Unpair-2032. Overall, try-on results that tend to preserve the original clothing information of the model paradoxically achieve lower FID and KID scores (indicating better performance) which is clearly incorrect. Our proposed SDR and S-LPIPS metrics consistently reflected the impact of incorrect samples on try-on quality under all experimental conditions, significantly outperforming FID and KID in evaluating the accuracy of unpaired try-on.}

\begin{figure*}[t]
    \centering
    \begin{subfigure}[b]{0.24\textwidth}
        \includegraphics[width=\textwidth]{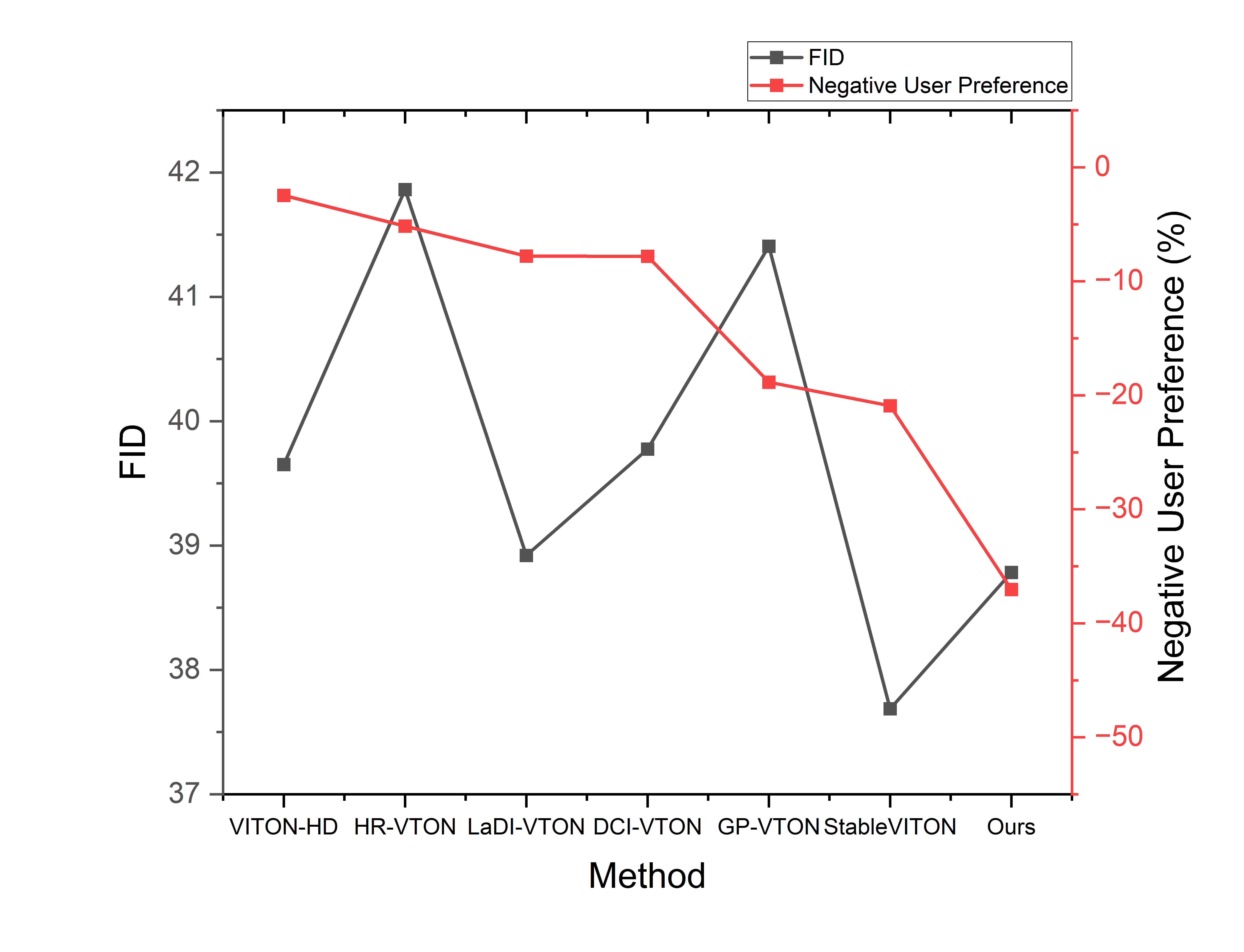}
        \caption{FID}
        \label{fig:metric_FID}
    \end{subfigure}
    \begin{subfigure}[b]{0.24\textwidth}
        \includegraphics[width=\textwidth]{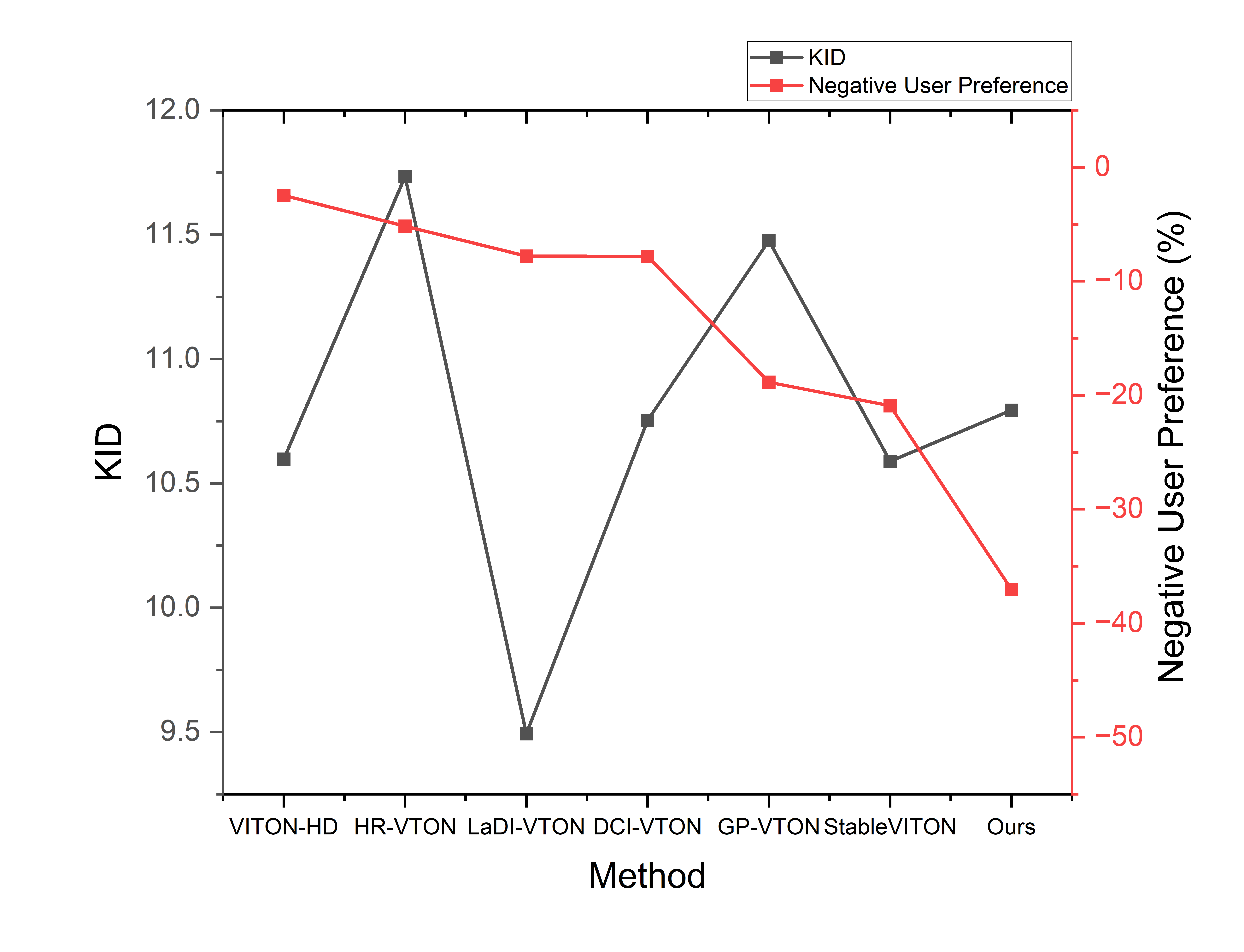}
        \caption{KID}
        \label{fig:metric_KID}
    \end{subfigure}
    \begin{subfigure}[b]{0.24\textwidth}
        \includegraphics[width=\textwidth]{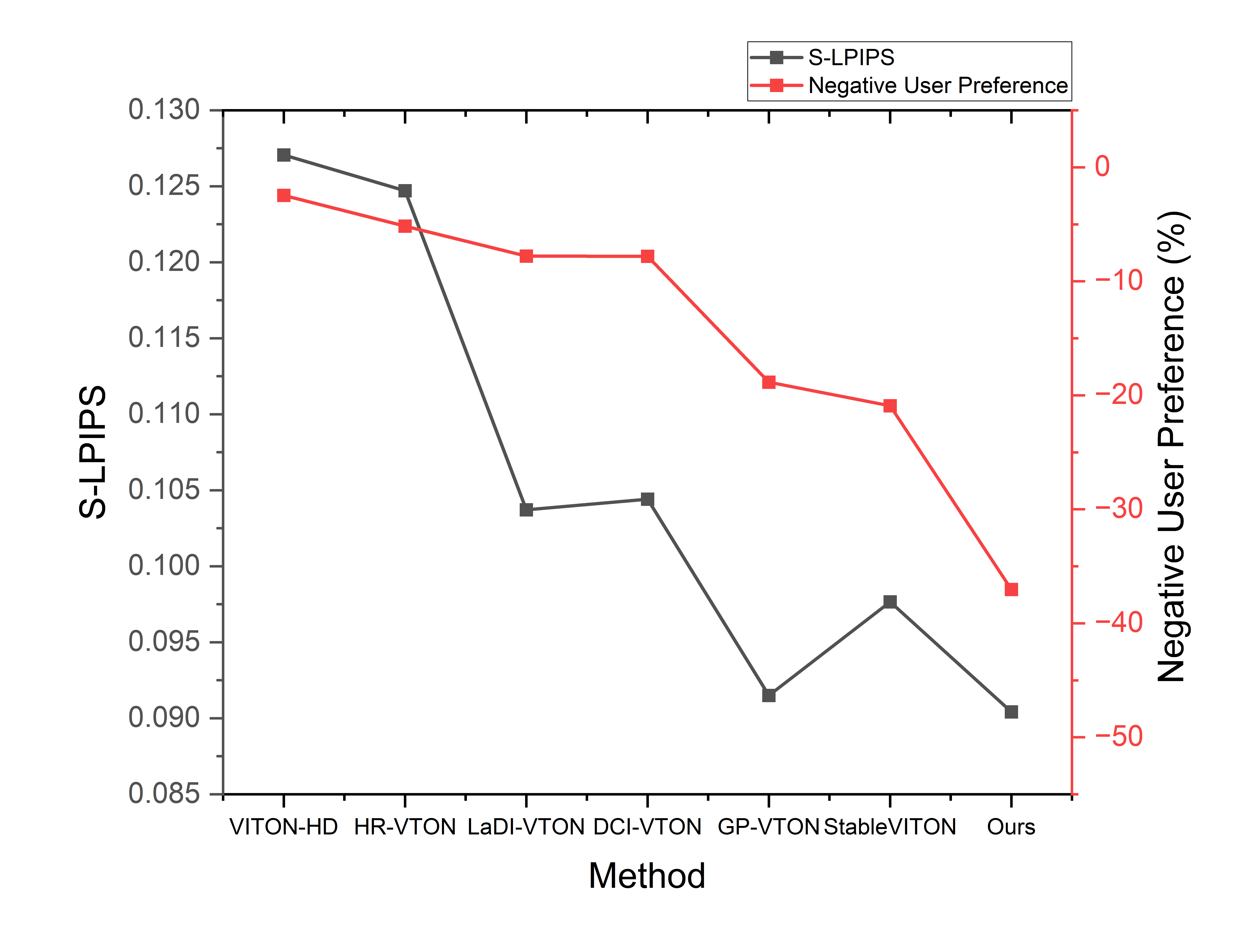}
        \caption{S-LPIPS}
        \label{fig:metric_SLPIPS}
    \end{subfigure}
    \begin{subfigure}[b]{0.24\textwidth}
        \includegraphics[width=\textwidth]{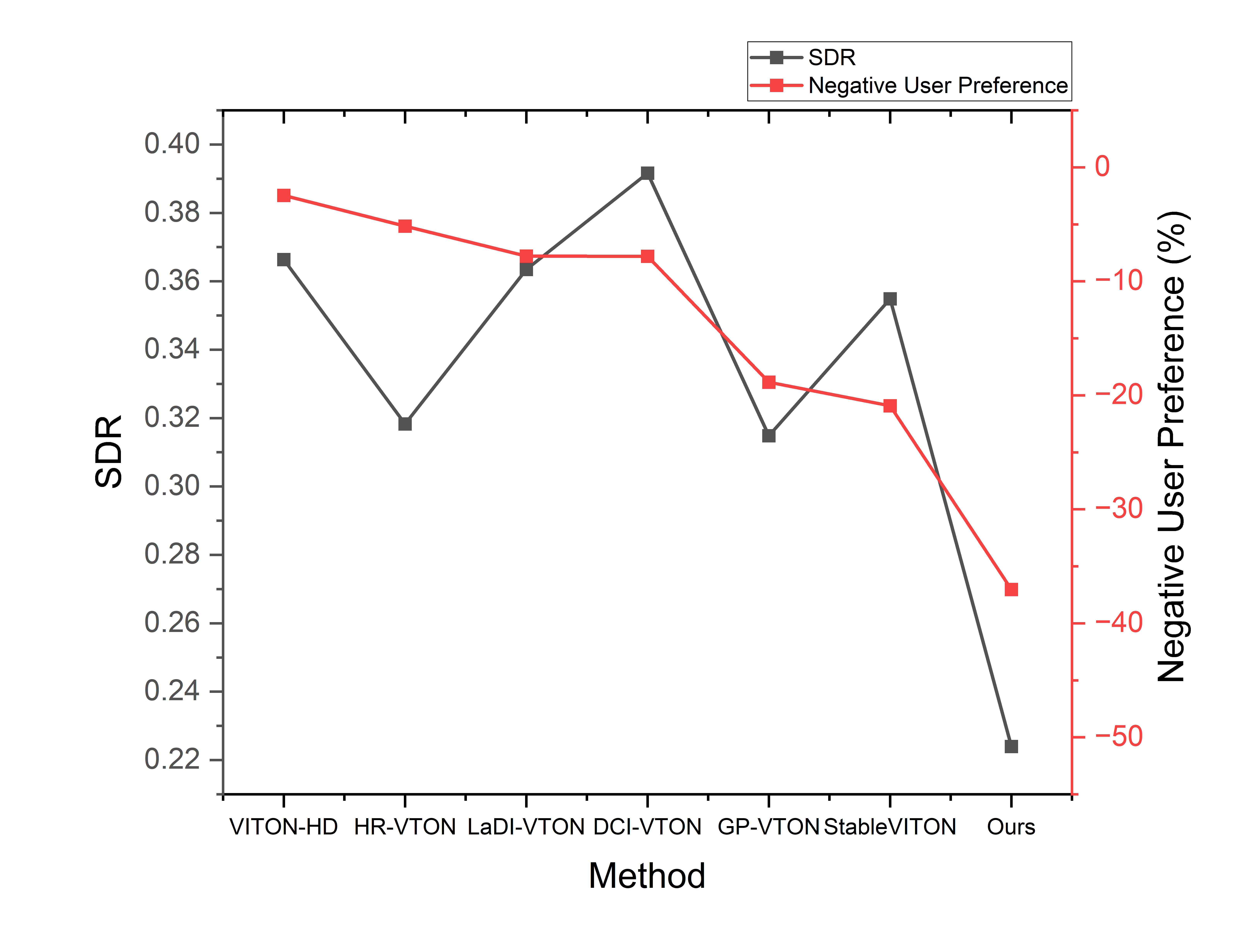}
        \caption{SDR}
        \label{fig:metric_SDR}
    \end{subfigure}
    \caption{Comparison of FID, KID, S-LPIPS, SDR scores, and negative user preference rates for each method.}
    \label{fig:metric_benchmark2}
    
\label{fig:metric_effective}
\end{figure*}

\begin{figure}[t]
\centering
\scriptsize
\setlength{\tabcolsep}{.2em}
\resizebox{\linewidth}{!}{
\begin{tabular}{cccccccc}

\includegraphics[width=0.16\linewidth]{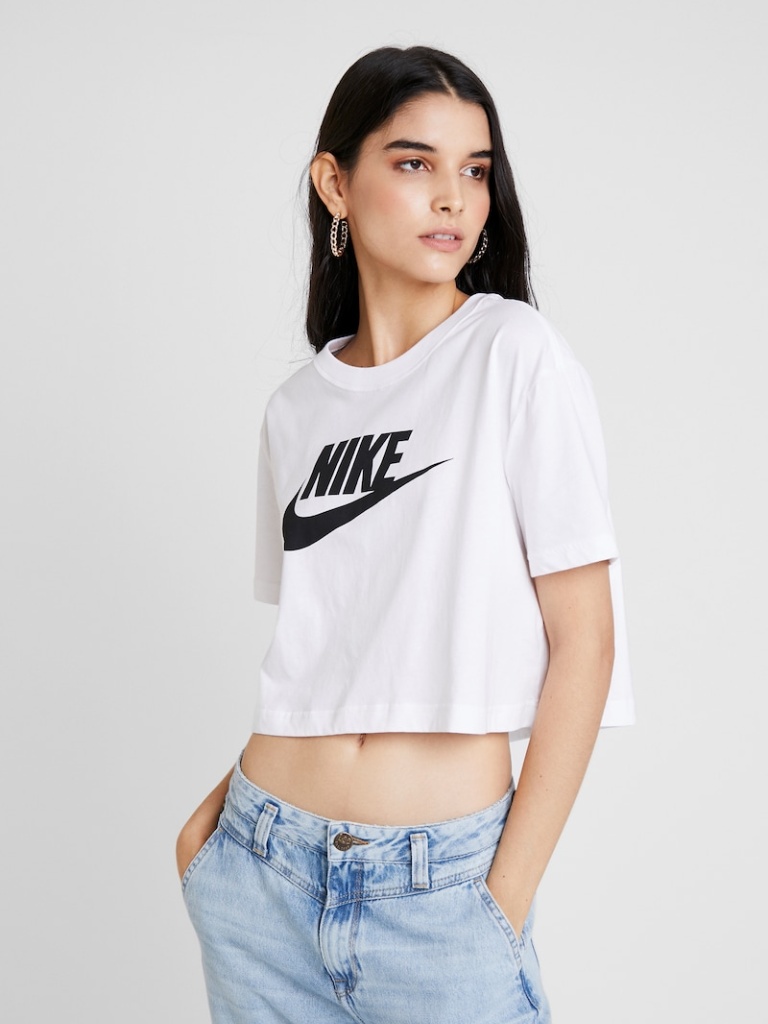} &
\includegraphics[width=0.16\linewidth]{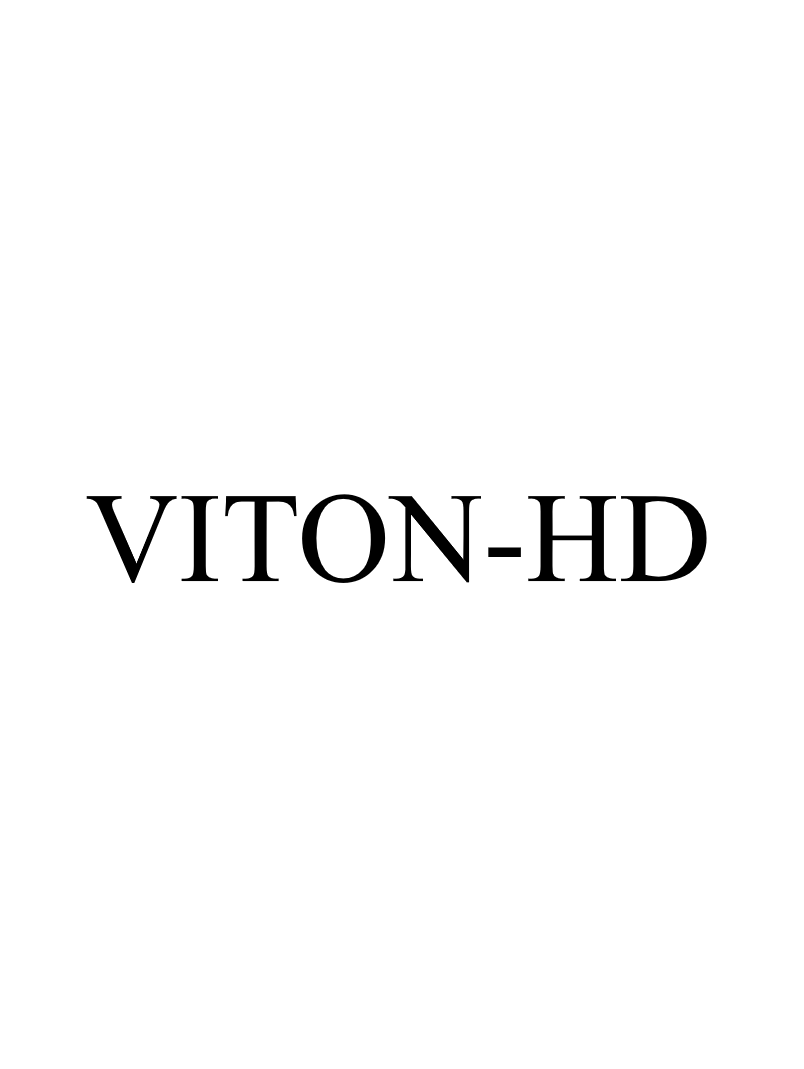} &
\includegraphics[width=0.16\linewidth]{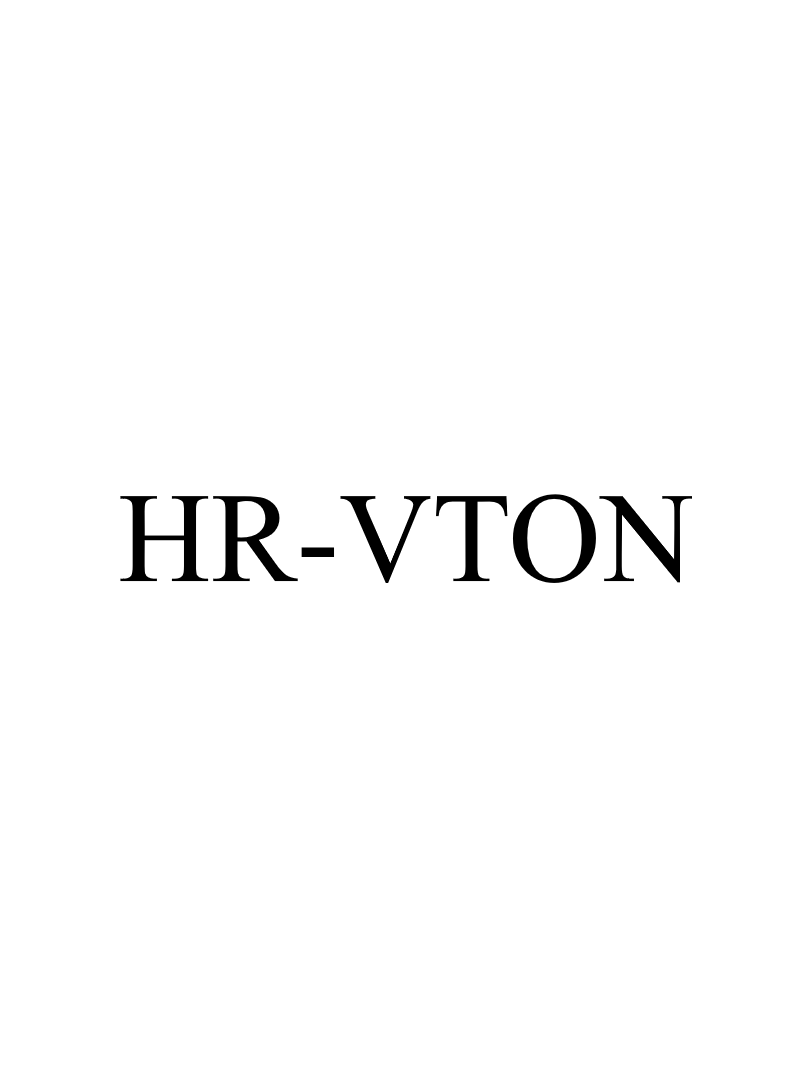} & 
\includegraphics[width=0.16\linewidth]{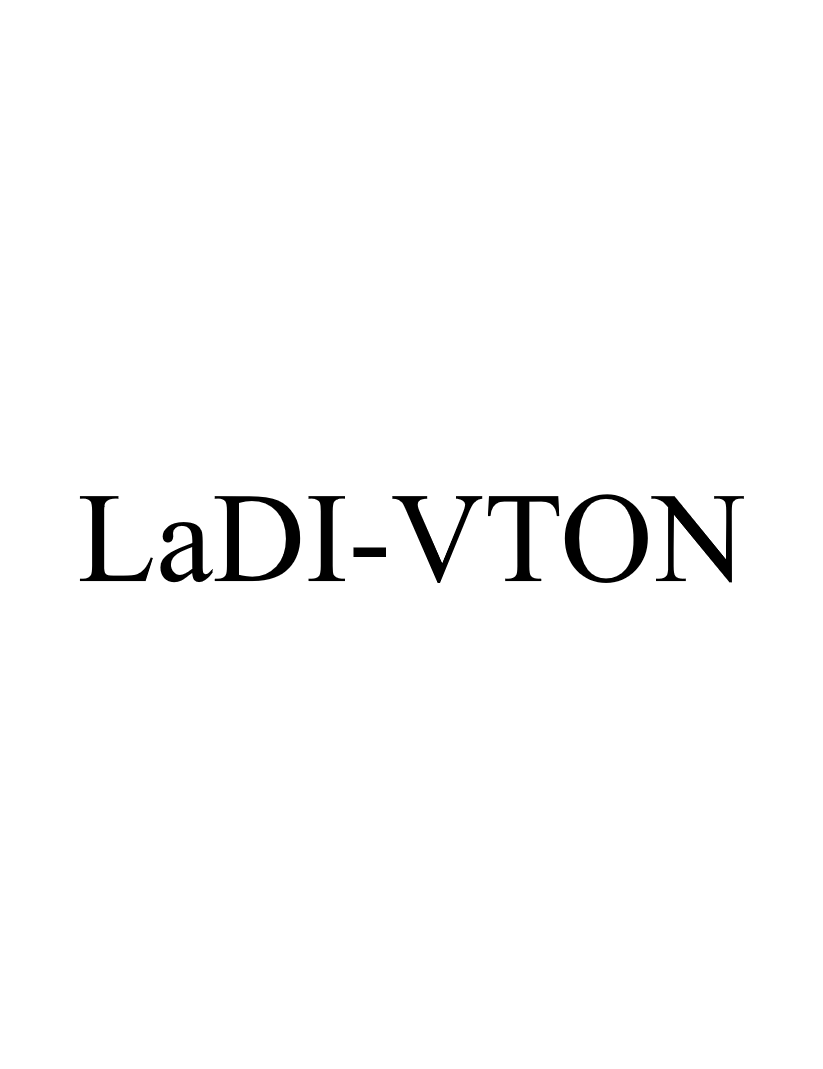} &
\includegraphics[width=0.16\linewidth]{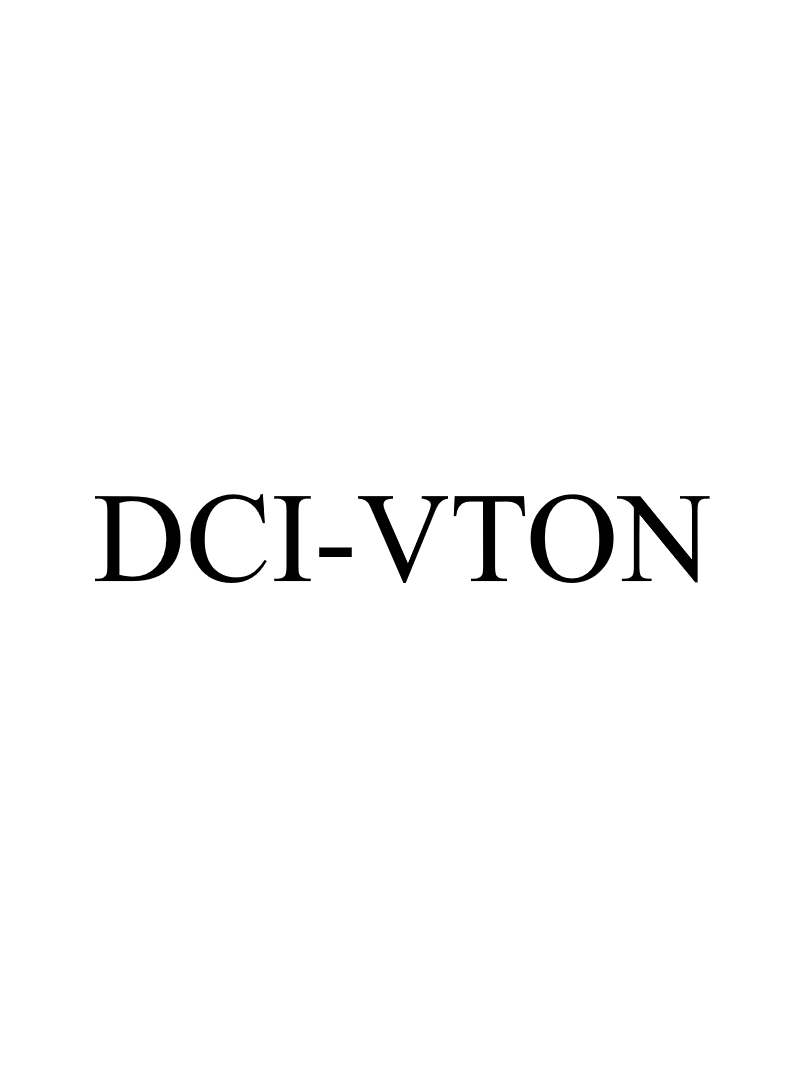} &
\includegraphics[width=0.16\linewidth]{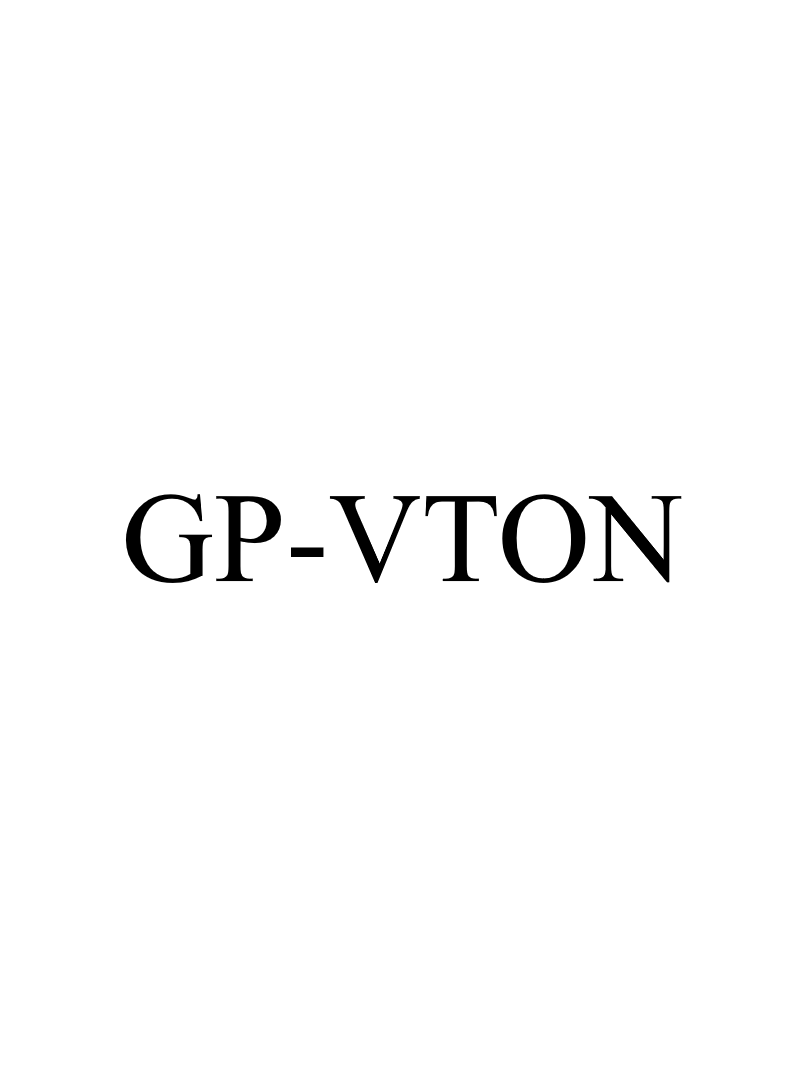} & 
\includegraphics[width=0.16\linewidth]{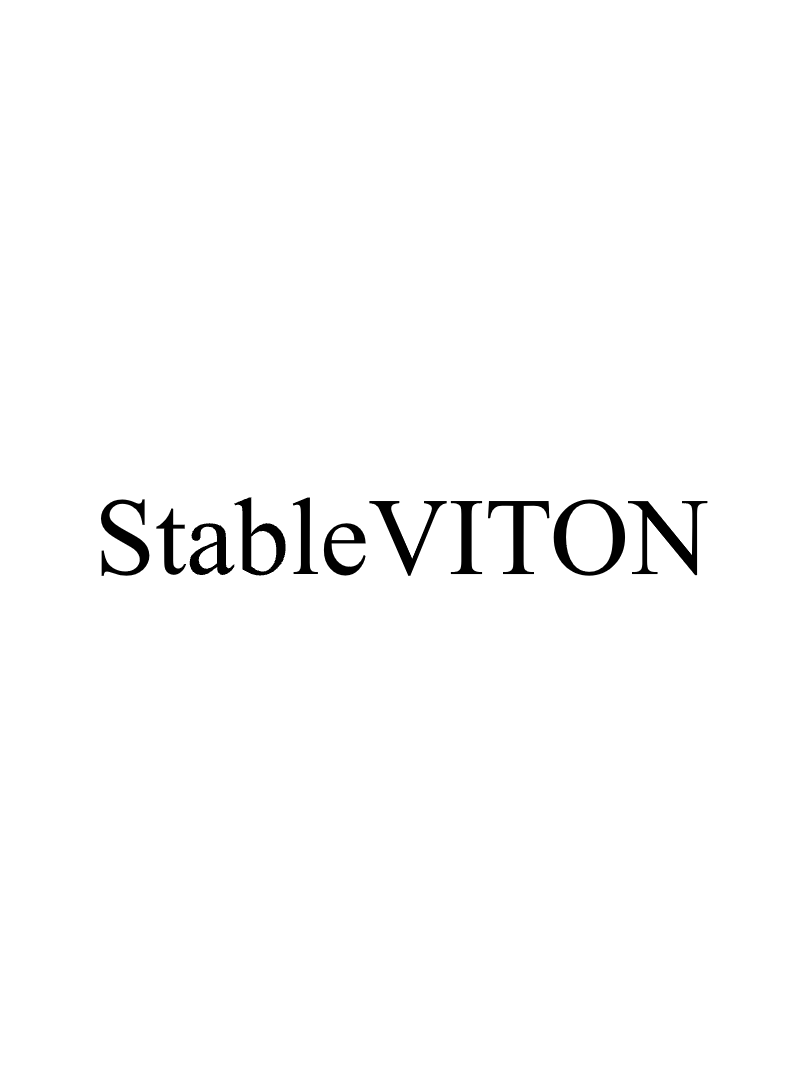} &
\includegraphics[width=0.16\linewidth]{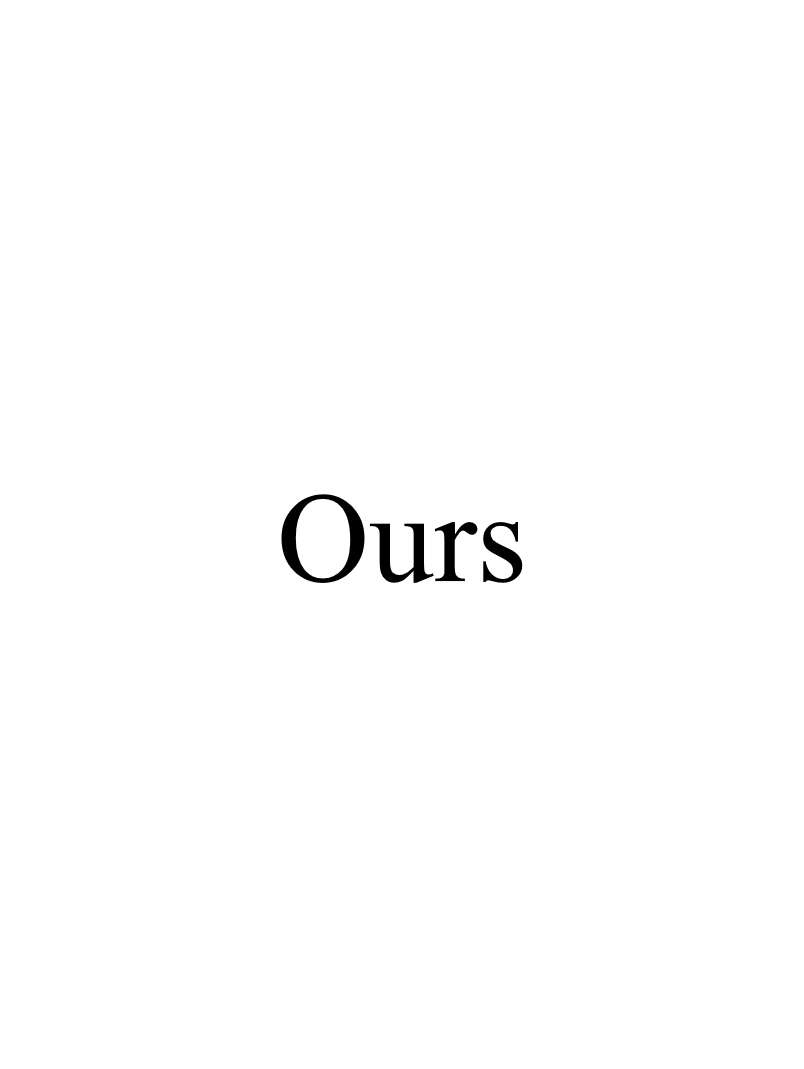} \\

\includegraphics[width=0.16\linewidth]{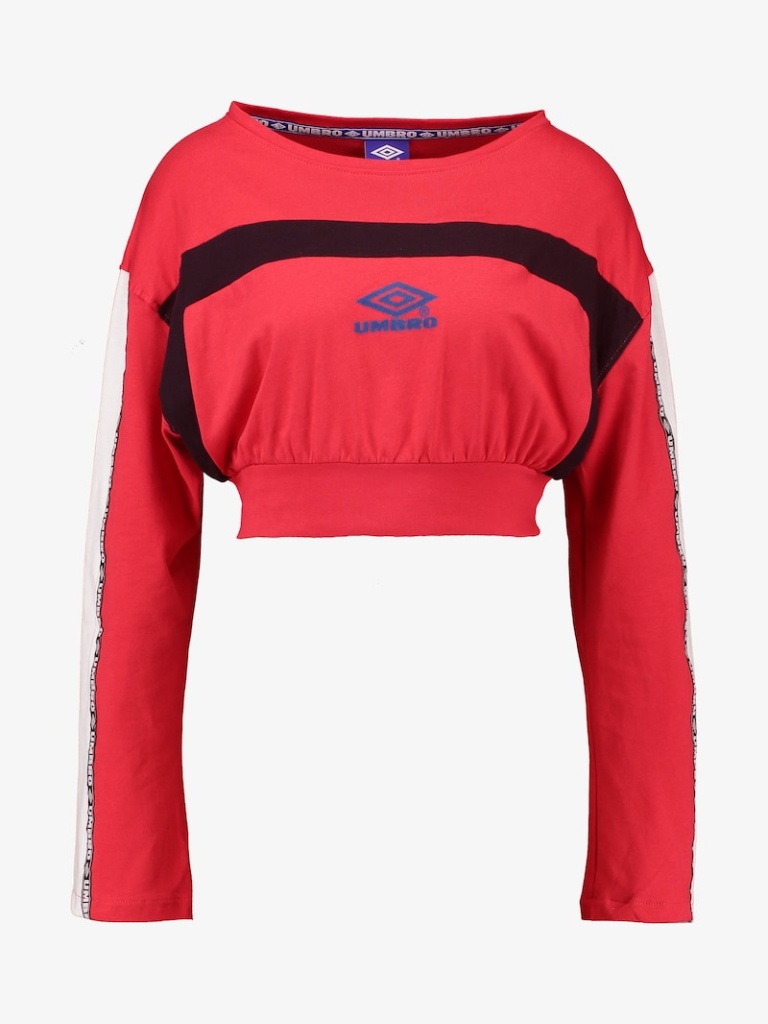} &
\includegraphics[width=0.16\linewidth]{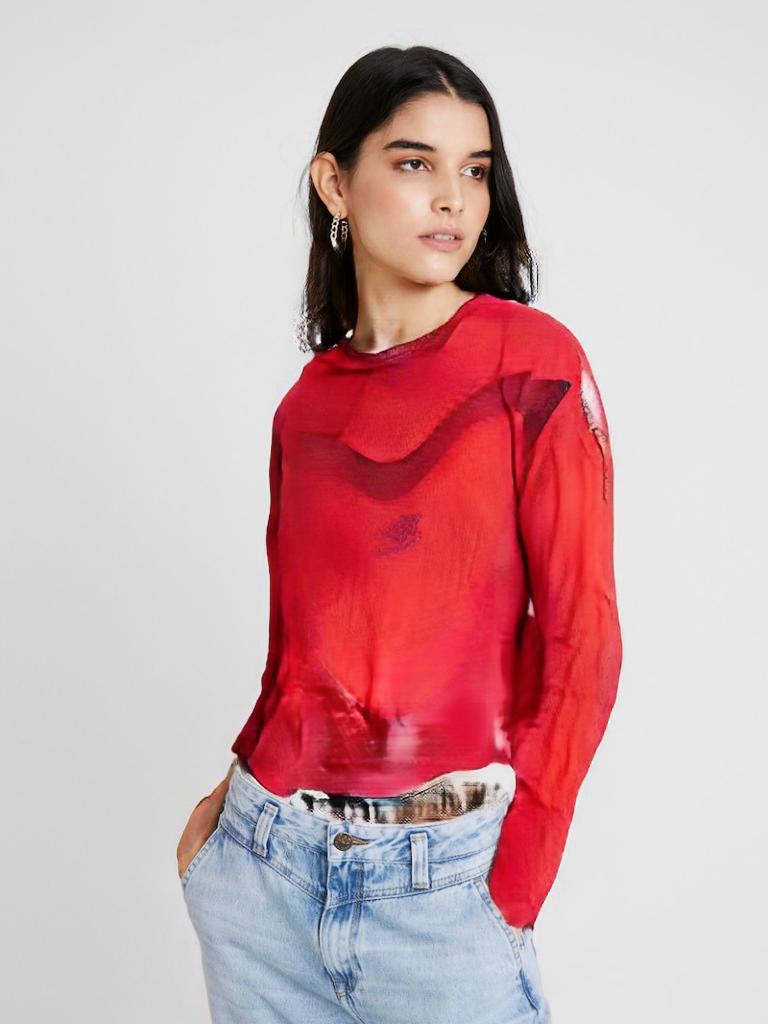} &
\includegraphics[width=0.16\linewidth]{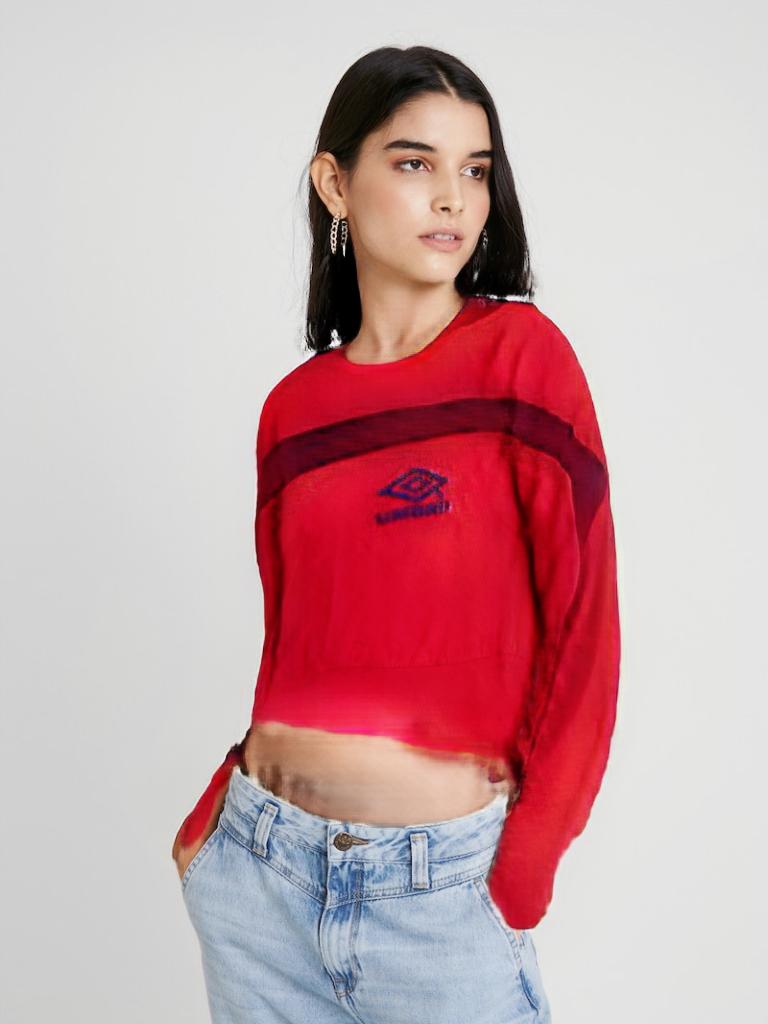} & 
\includegraphics[width=0.16\linewidth]{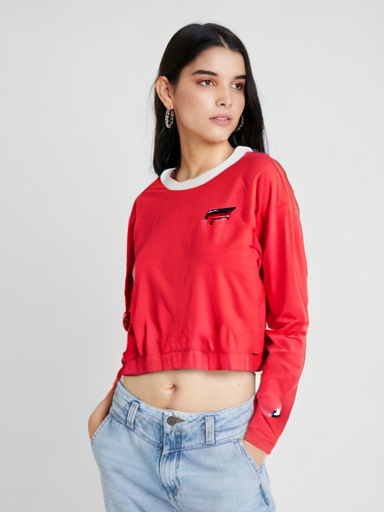} &
\includegraphics[width=0.16\linewidth]{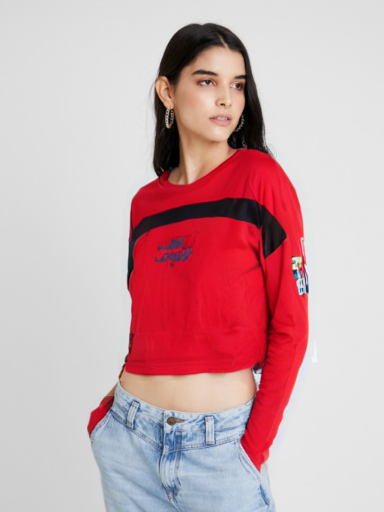} &
\includegraphics[width=0.16\linewidth]{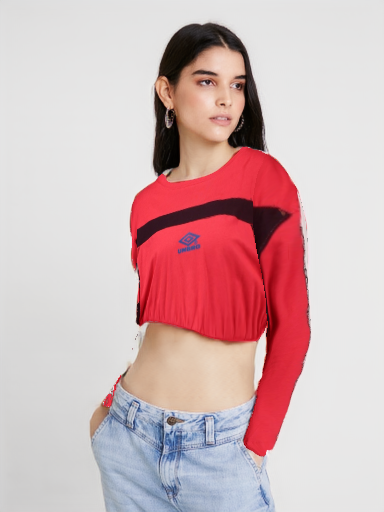} & 
\includegraphics[width=0.16\linewidth]{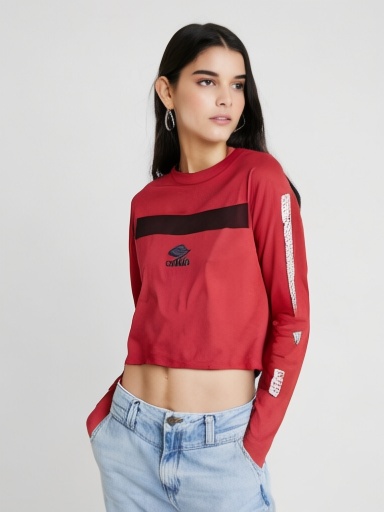} &
\includegraphics[width=0.16\linewidth]{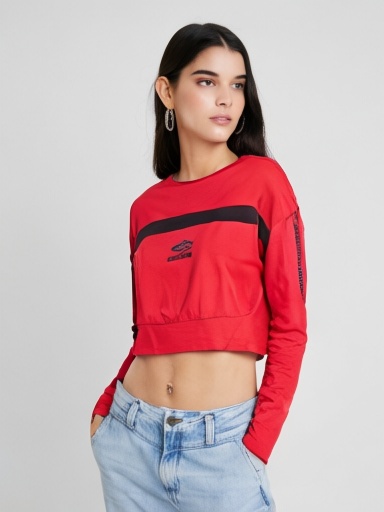} \\

\includegraphics[width=0.16\linewidth]{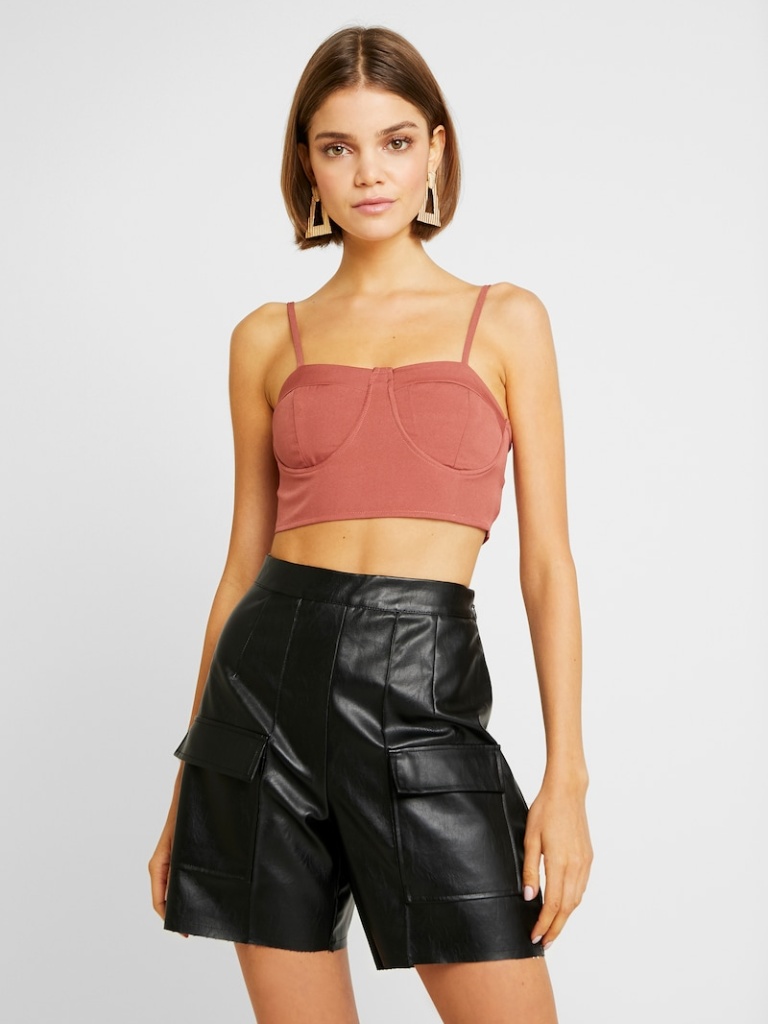} &
\includegraphics[width=0.16\linewidth]{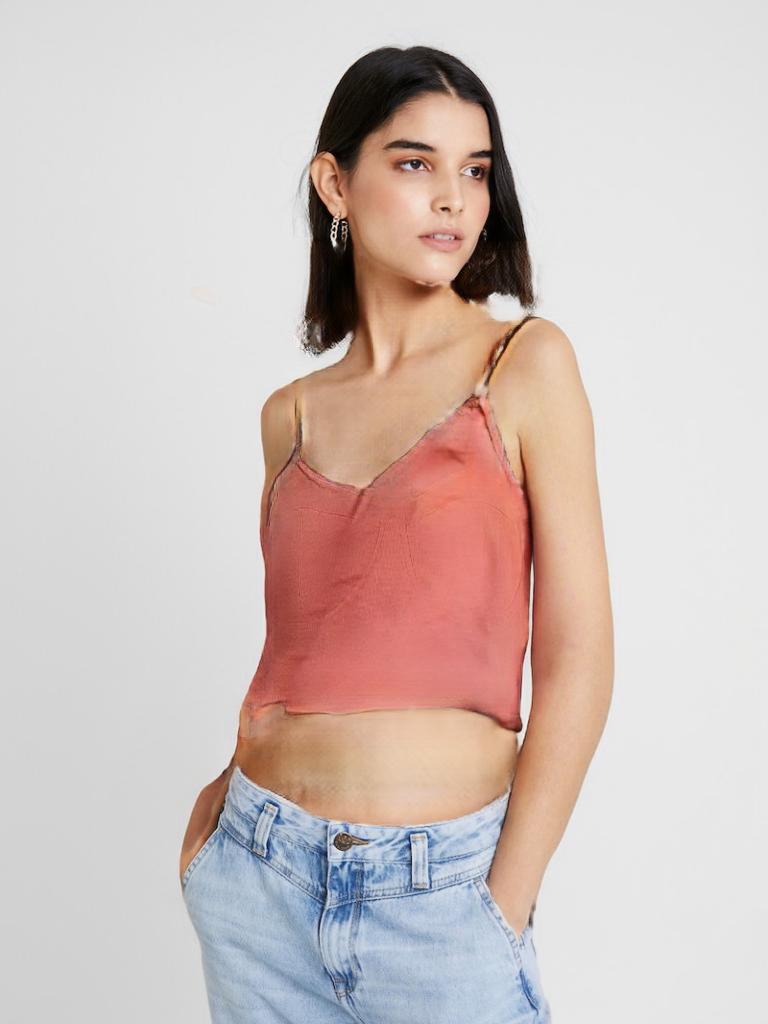} &
\includegraphics[width=0.16\linewidth]{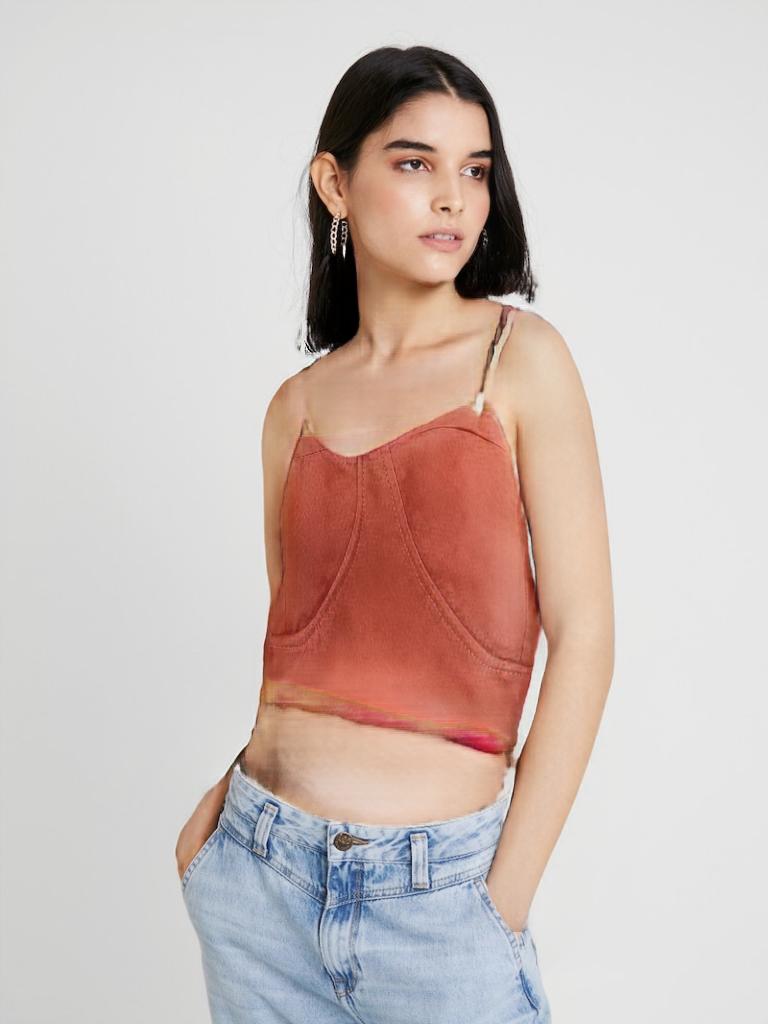} & 
\includegraphics[width=0.16\linewidth]{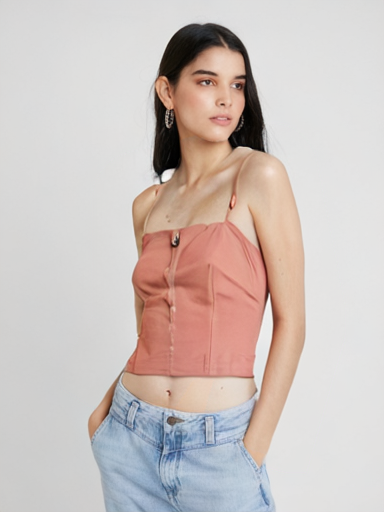} &
\includegraphics[width=0.16\linewidth]{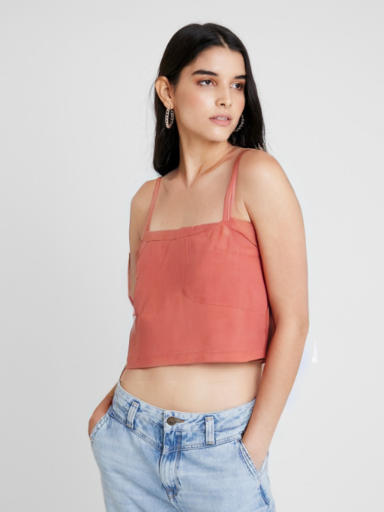} &
\includegraphics[width=0.16\linewidth]{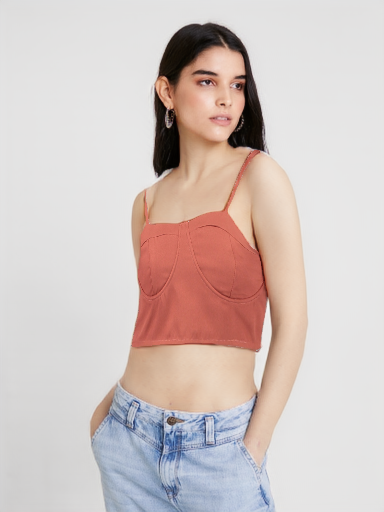} & 
\includegraphics[width=0.16\linewidth]{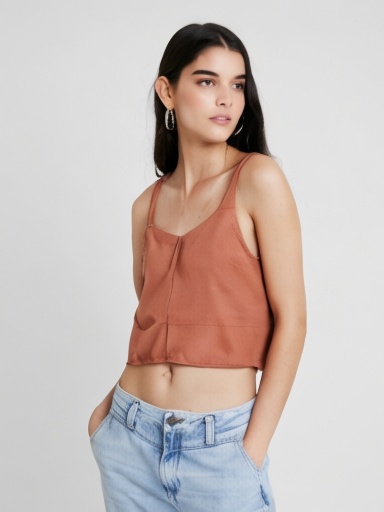} &
\includegraphics[width=0.16\linewidth]{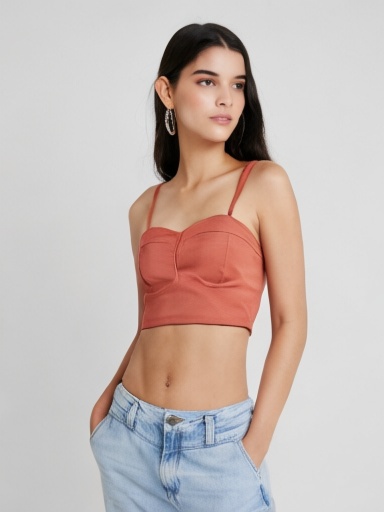} \\

\includegraphics[width=0.16\linewidth]{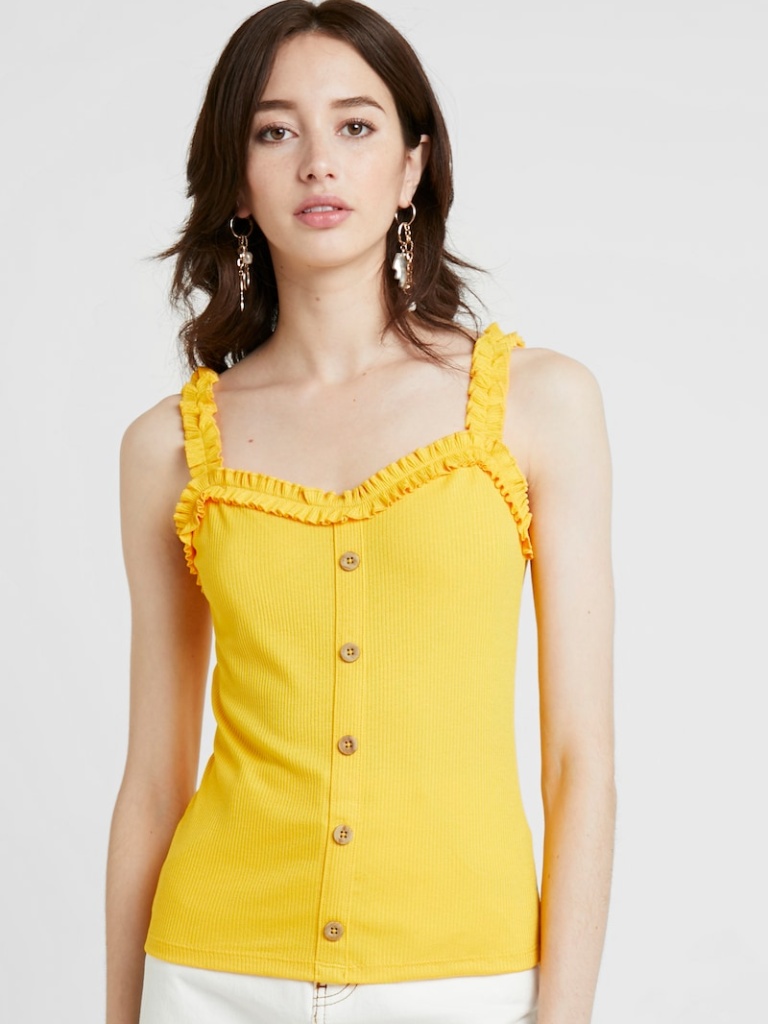} &
\includegraphics[width=0.16\linewidth]{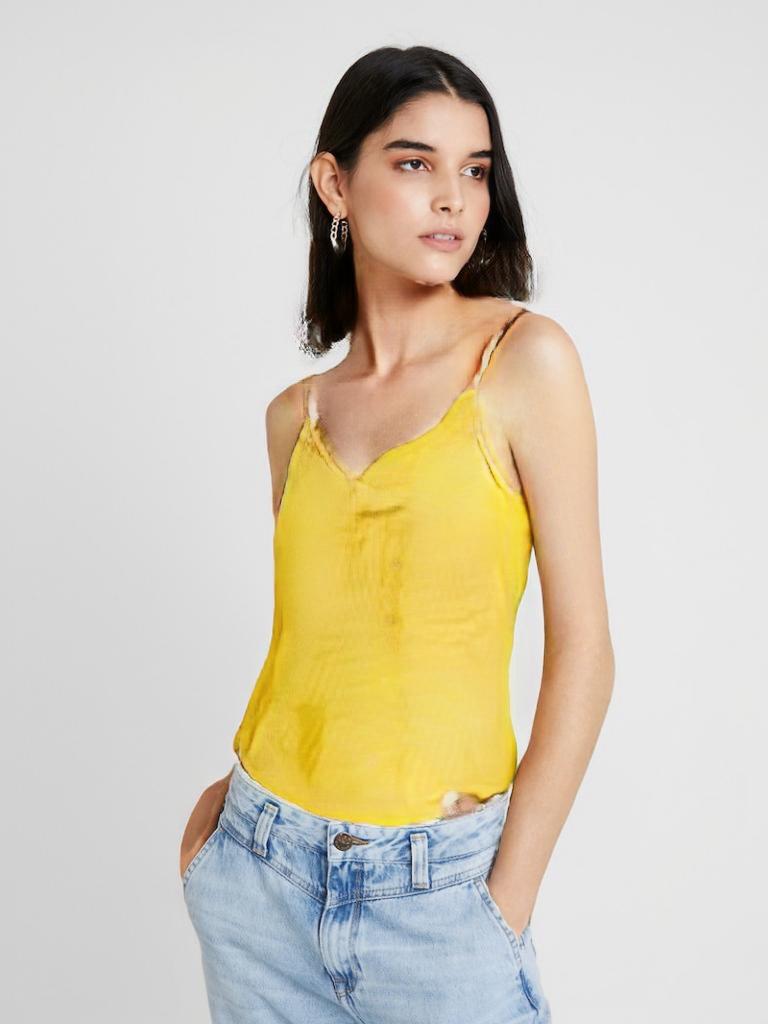} &
\includegraphics[width=0.16\linewidth]{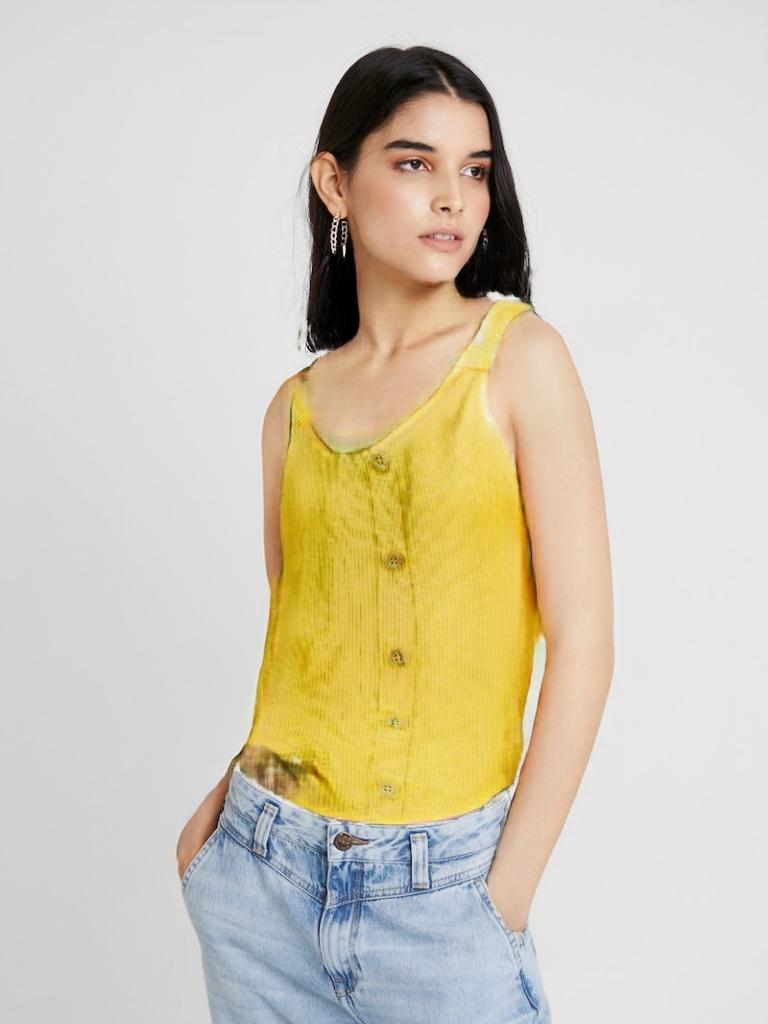} & 
\includegraphics[width=0.16\linewidth]{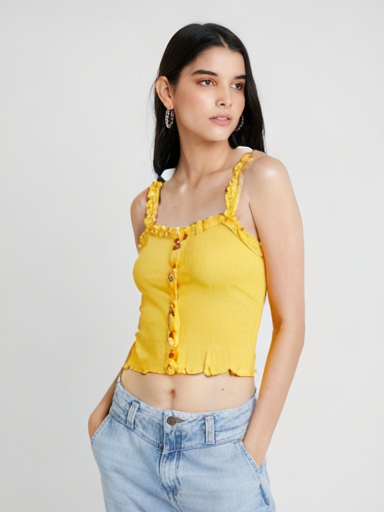} &
\includegraphics[width=0.16\linewidth]{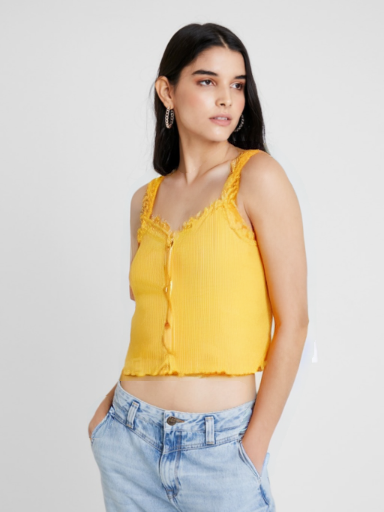} &
\includegraphics[width=0.16\linewidth]{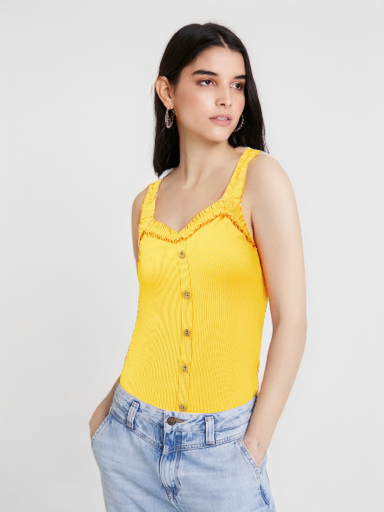} & 
\includegraphics[width=0.16\linewidth]{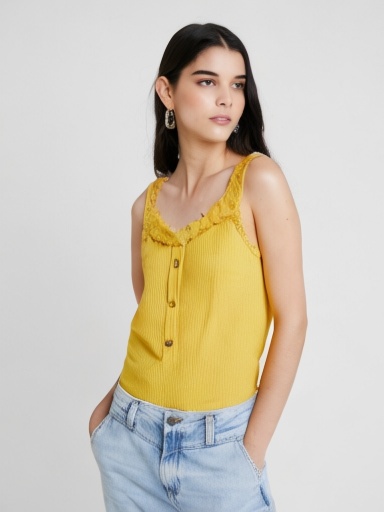} &
\includegraphics[width=0.16\linewidth]{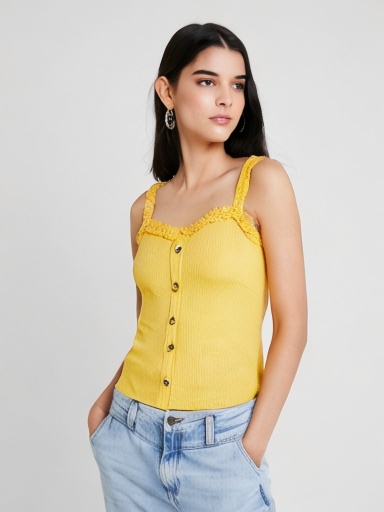} \\

\includegraphics[width=0.16\linewidth]{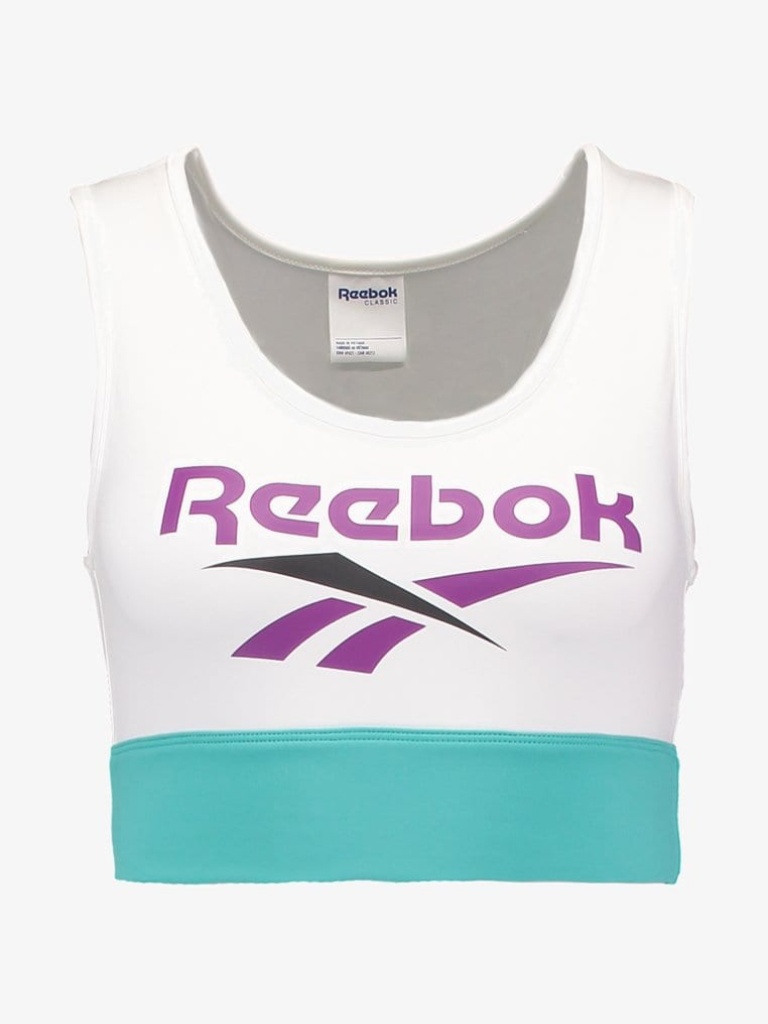} &
\includegraphics[width=0.16\linewidth]{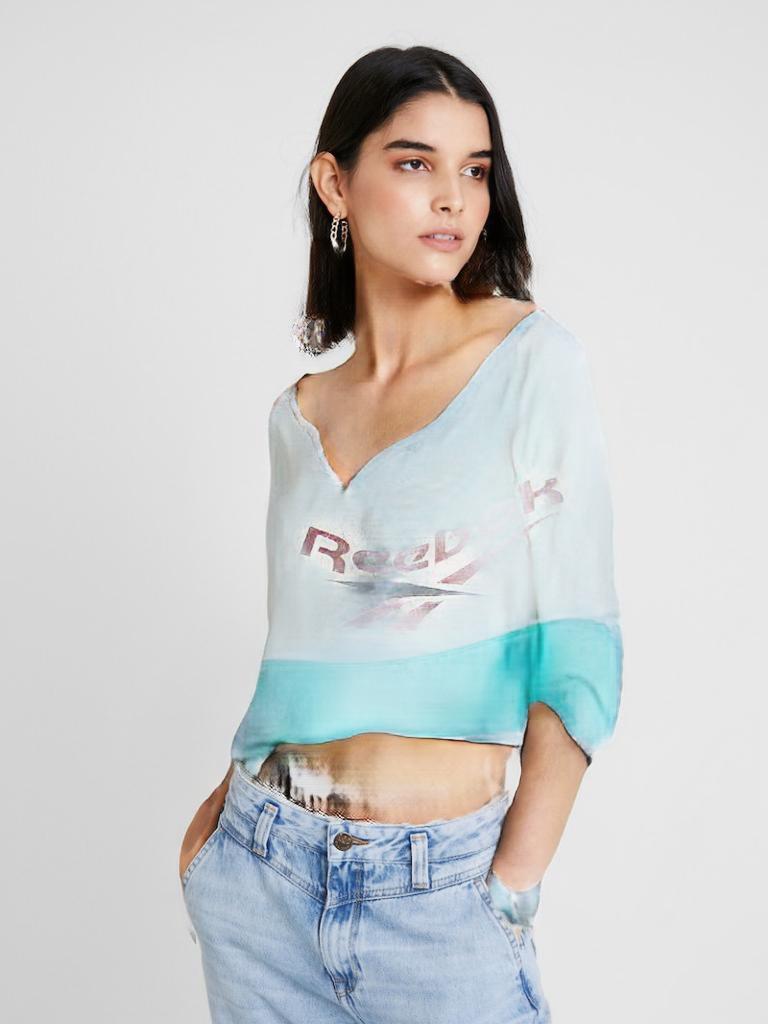} &
\includegraphics[width=0.16\linewidth]{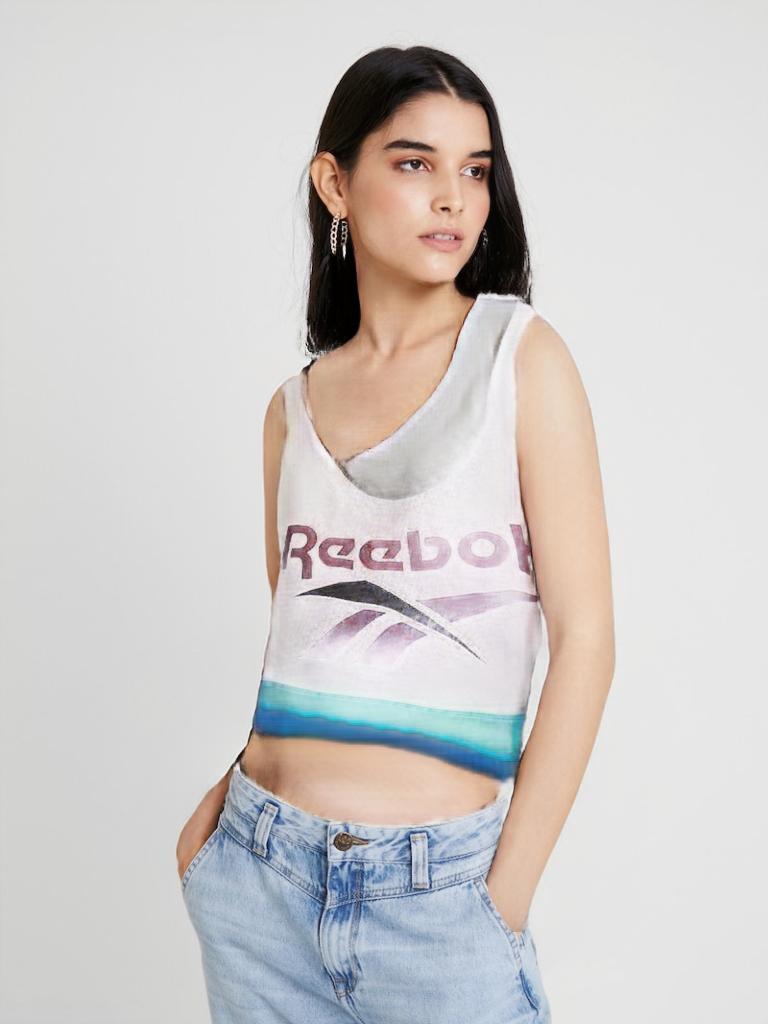} & 
\includegraphics[width=0.16\linewidth]{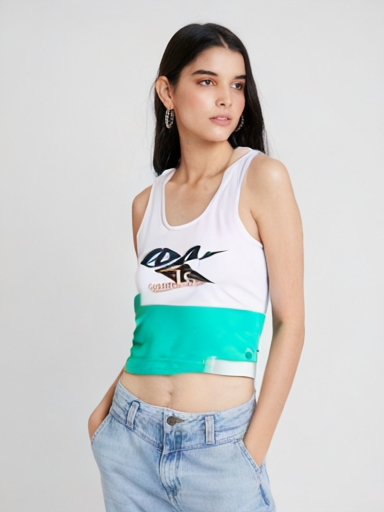} &
\includegraphics[width=0.16\linewidth]{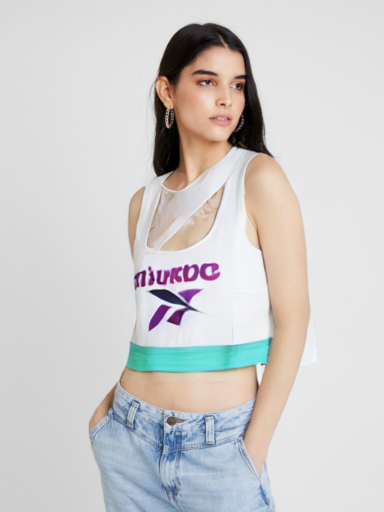} &
\includegraphics[width=0.16\linewidth]{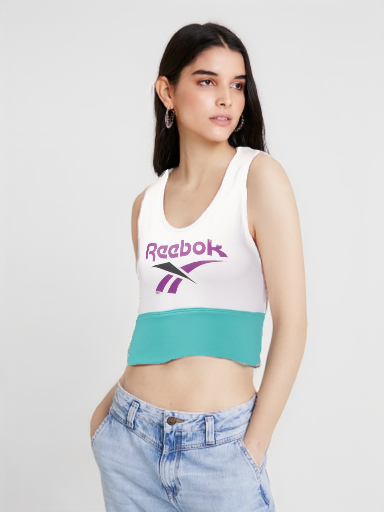} & 
\includegraphics[width=0.16\linewidth]{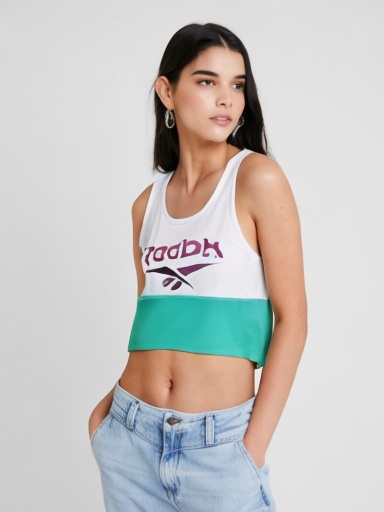} &
\includegraphics[width=0.16\linewidth]{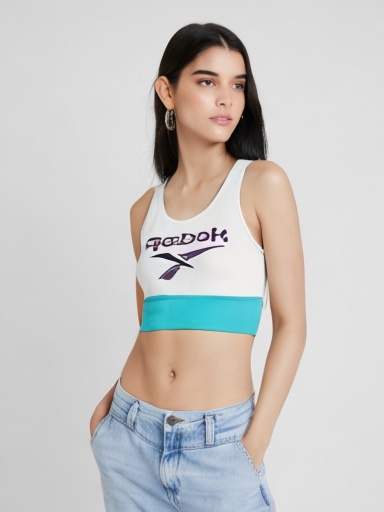} \\

\includegraphics[width=0.16\linewidth]{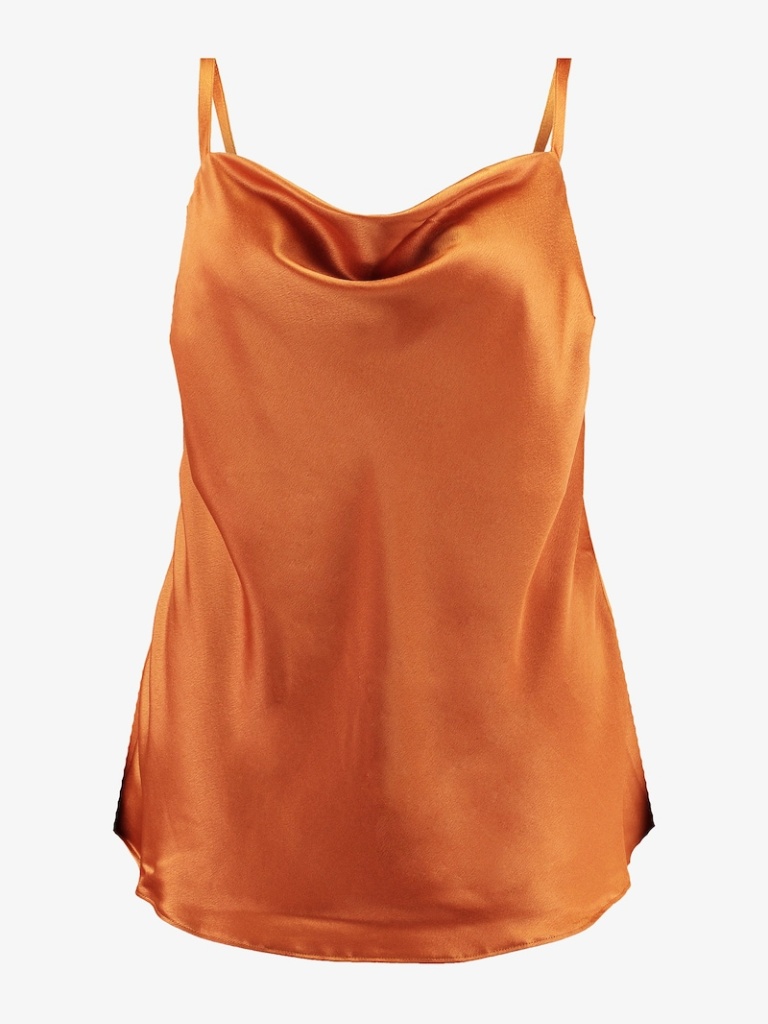} &
\includegraphics[width=0.16\linewidth]{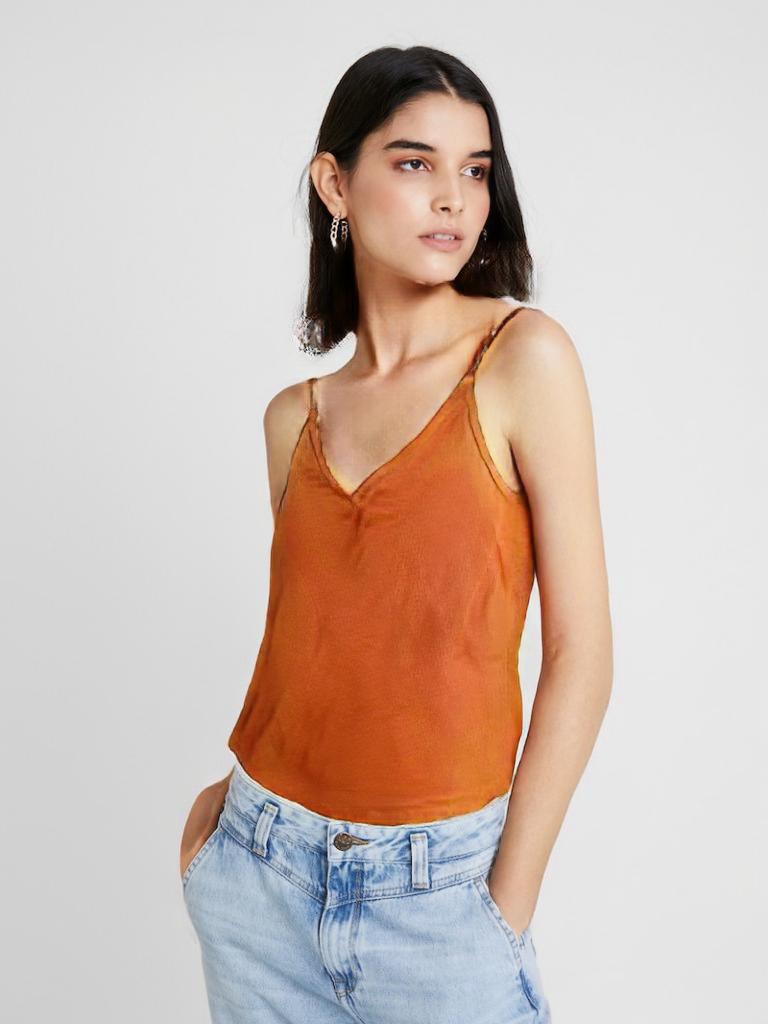} &
\includegraphics[width=0.16\linewidth]{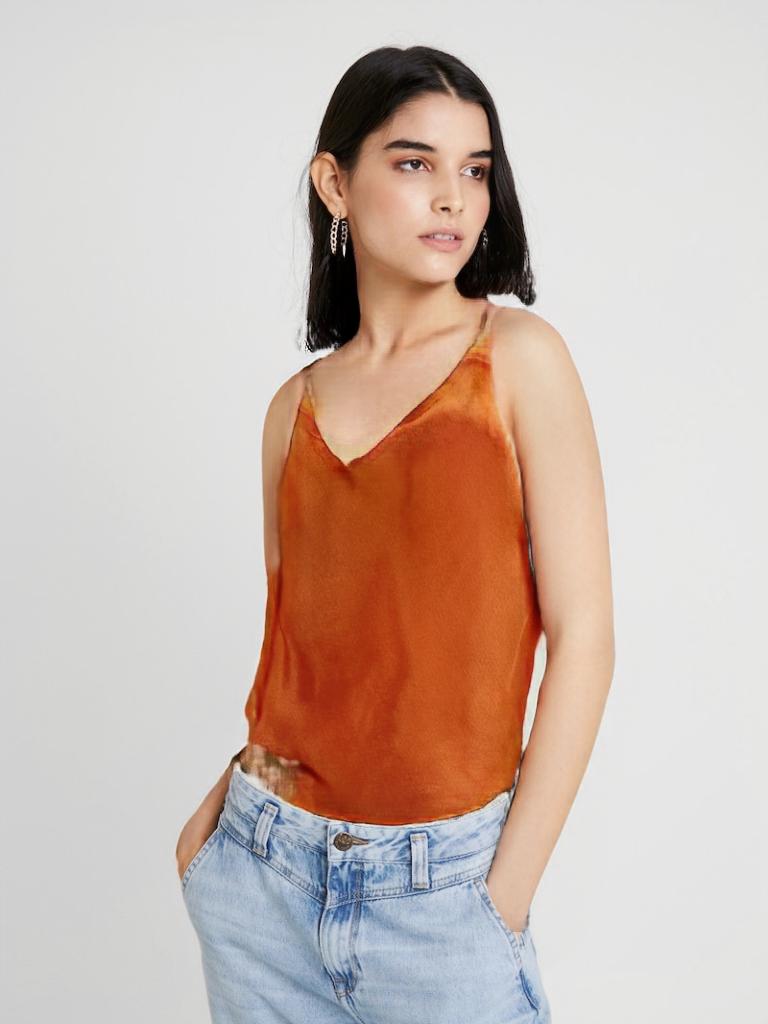} & 
\includegraphics[width=0.16\linewidth]{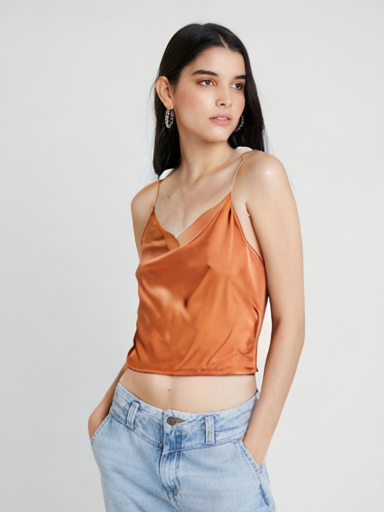} &
\includegraphics[width=0.16\linewidth]{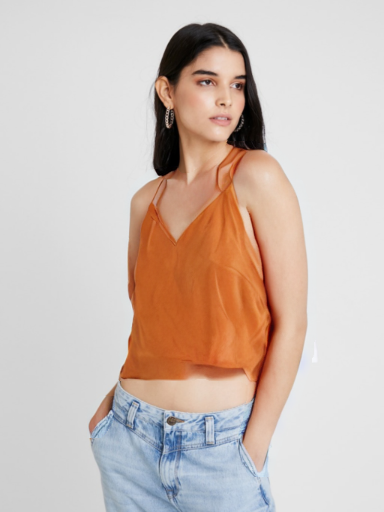} &
\includegraphics[width=0.16\linewidth]{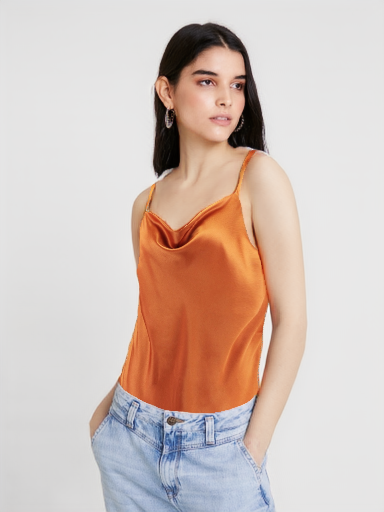} & 
\includegraphics[width=0.16\linewidth]{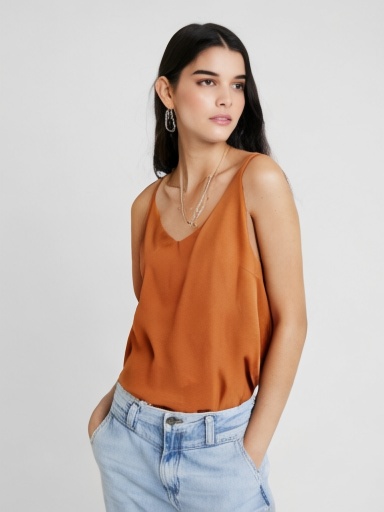} &
\includegraphics[width=0.16\linewidth]{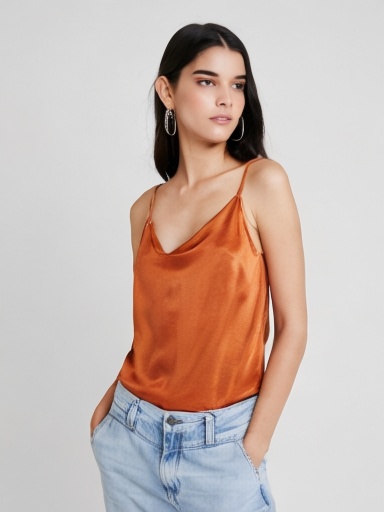} \\

\includegraphics[width=0.16\linewidth]{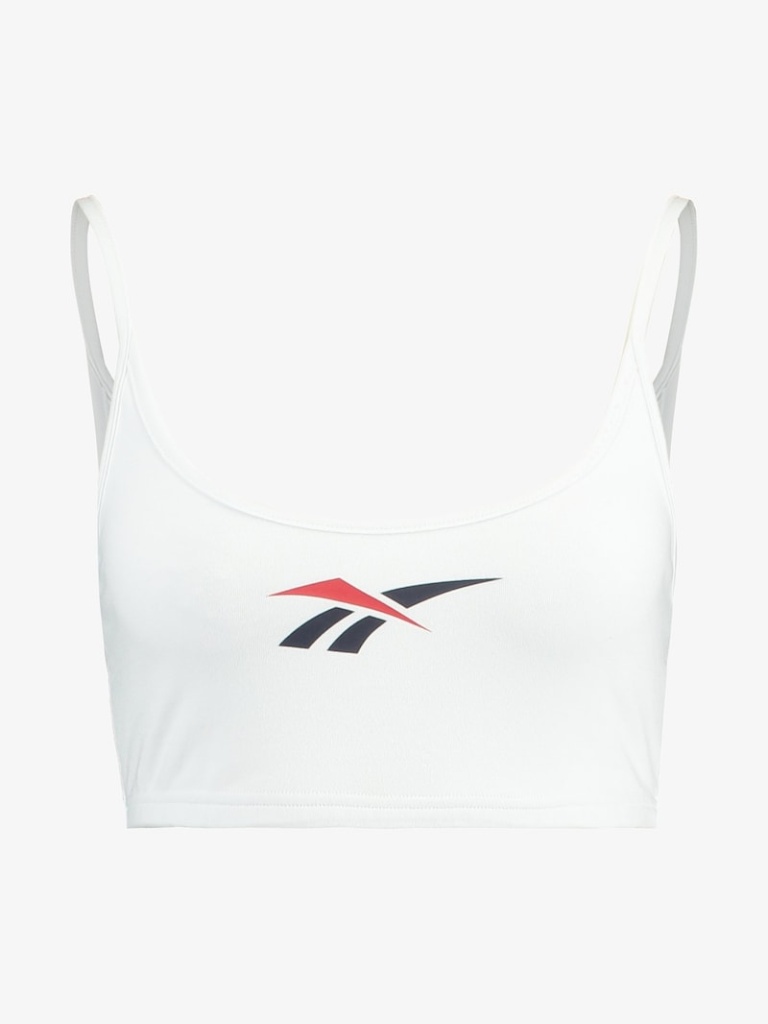} &
\includegraphics[width=0.16\linewidth]{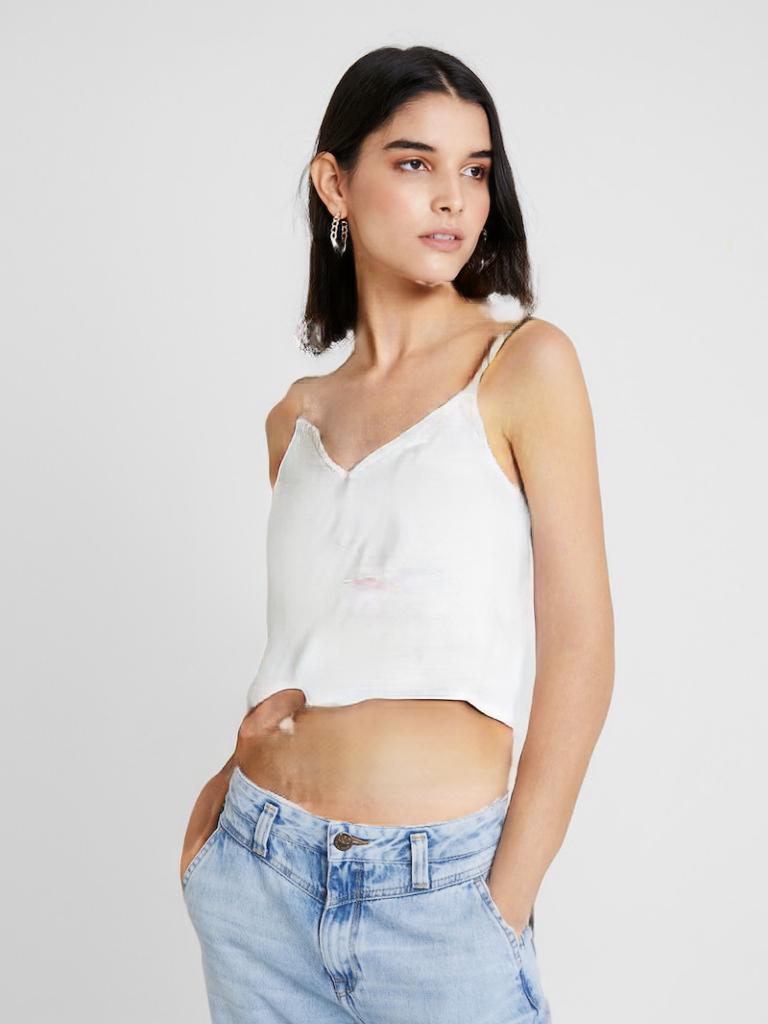} &
\includegraphics[width=0.16\linewidth]{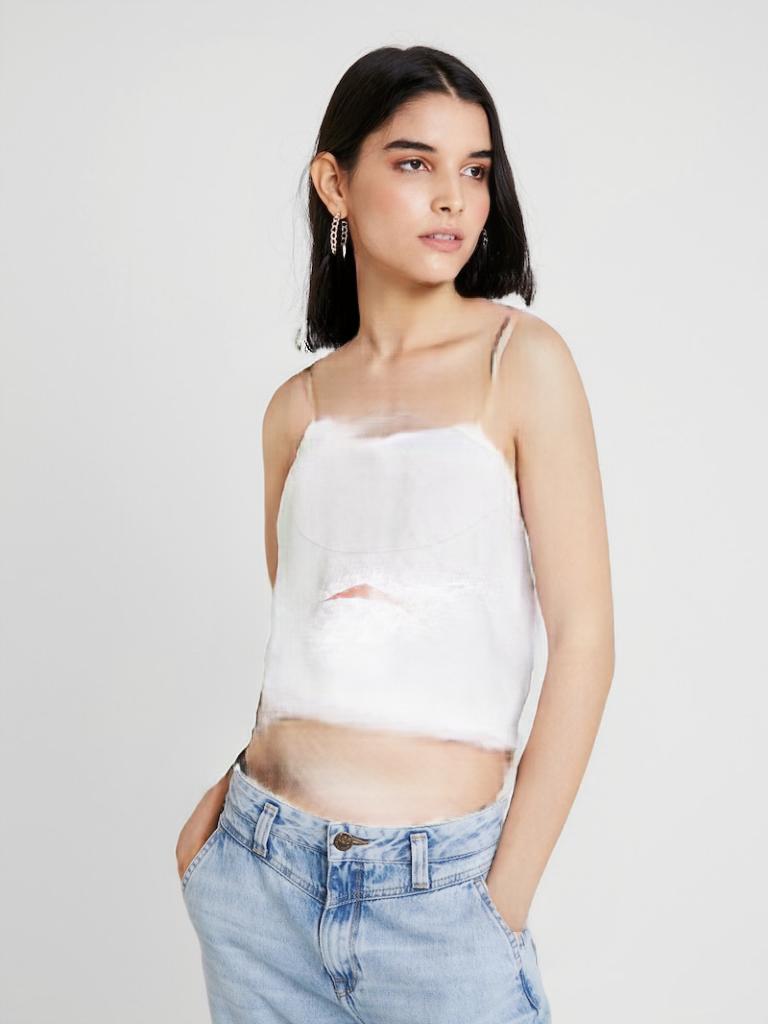} & 
\includegraphics[width=0.16\linewidth]{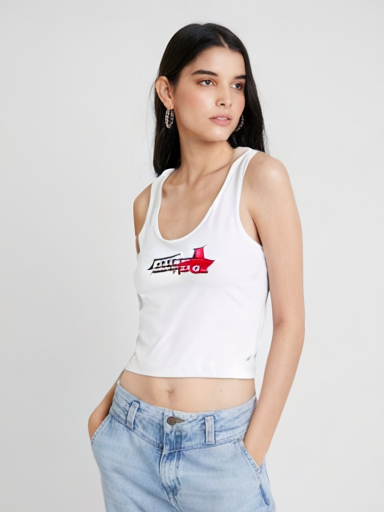} &
\includegraphics[width=0.16\linewidth]{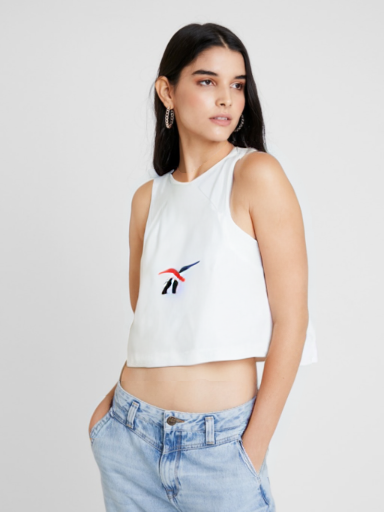} &
\includegraphics[width=0.16\linewidth]{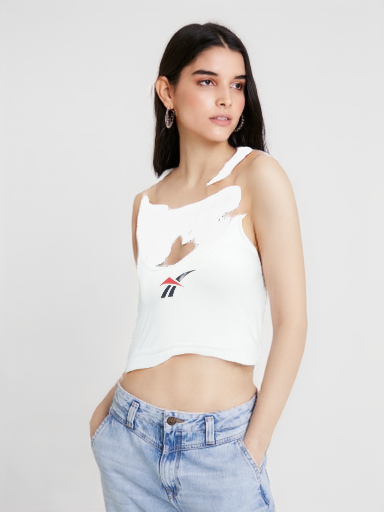} & 
\includegraphics[width=0.16\linewidth]{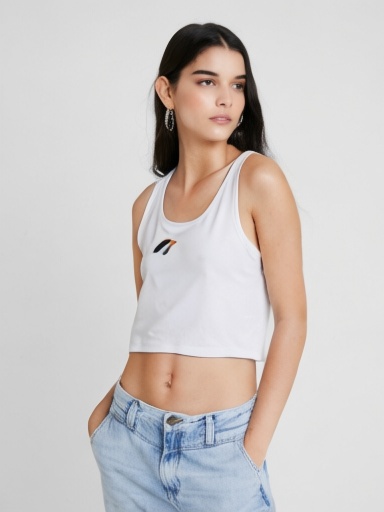} &
\includegraphics[width=0.16\linewidth]{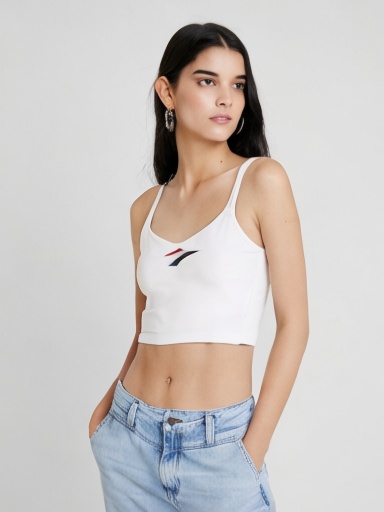} \\

\includegraphics[width=0.16\linewidth]{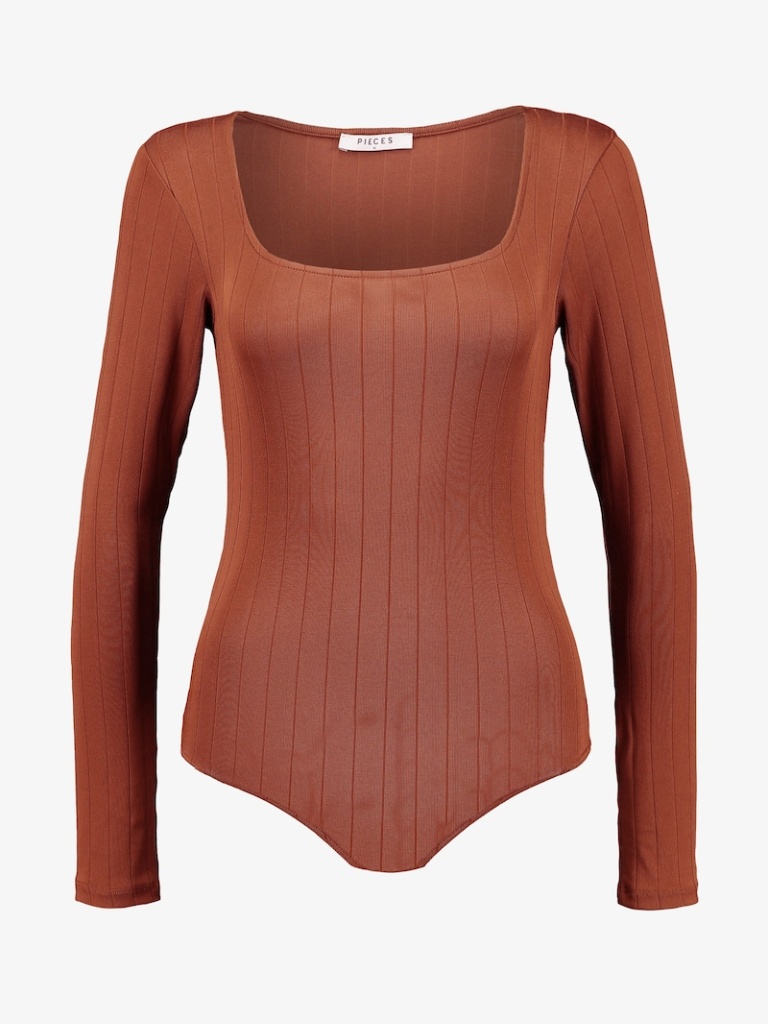} &
\includegraphics[width=0.16\linewidth]{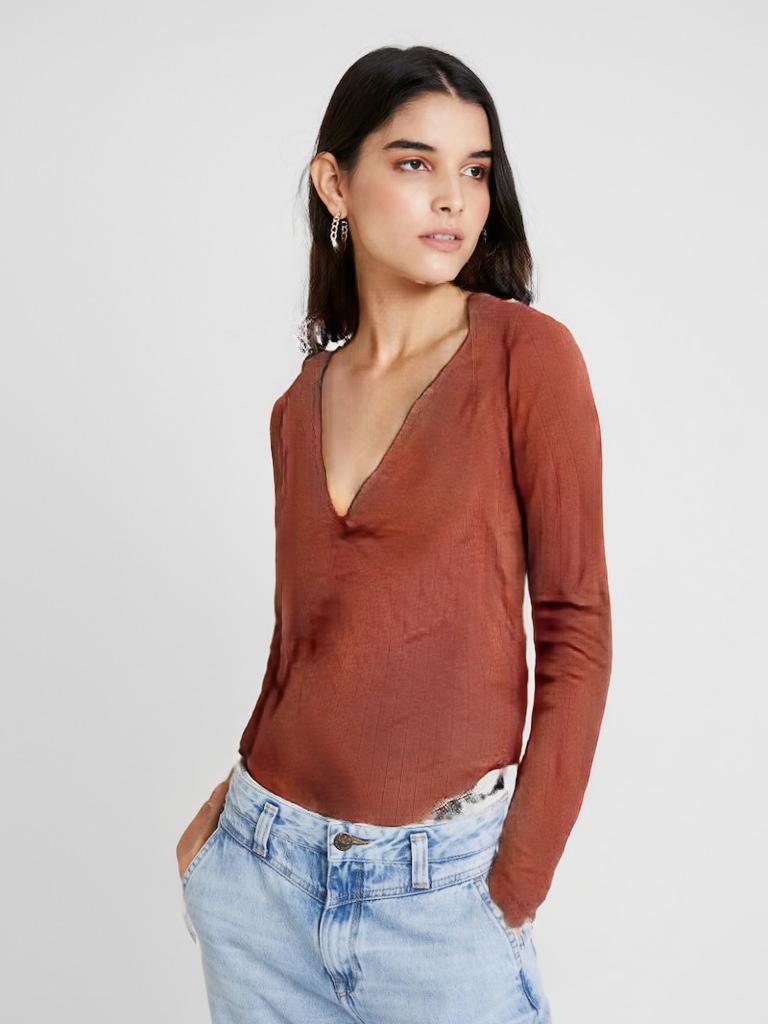} &
\includegraphics[width=0.16\linewidth]{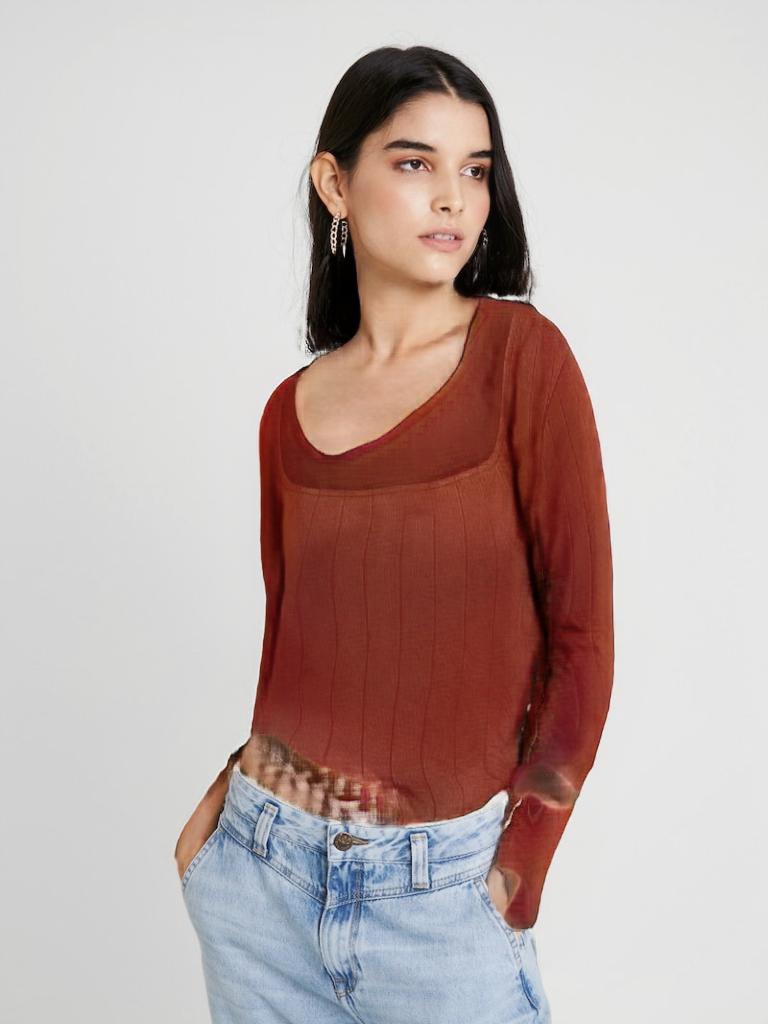} & 
\includegraphics[width=0.16\linewidth]{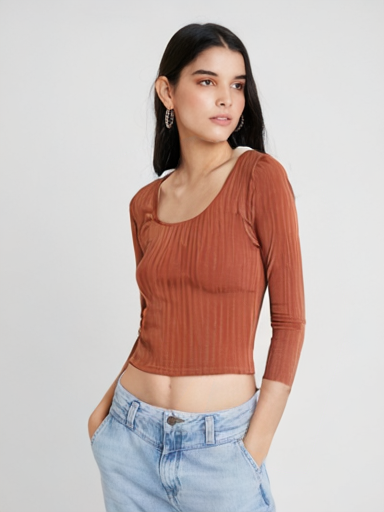} &
\includegraphics[width=0.16\linewidth]{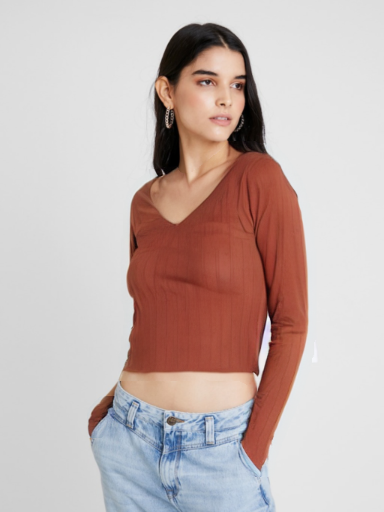} &
\includegraphics[width=0.16\linewidth]{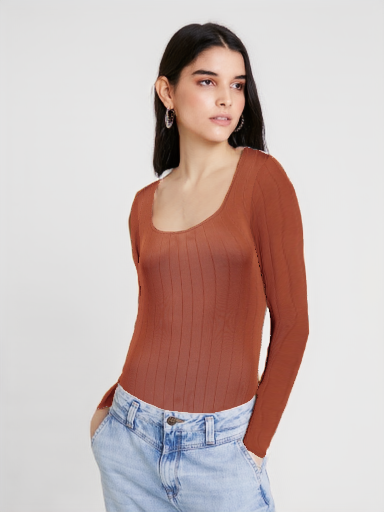} & 
\includegraphics[width=0.16\linewidth]{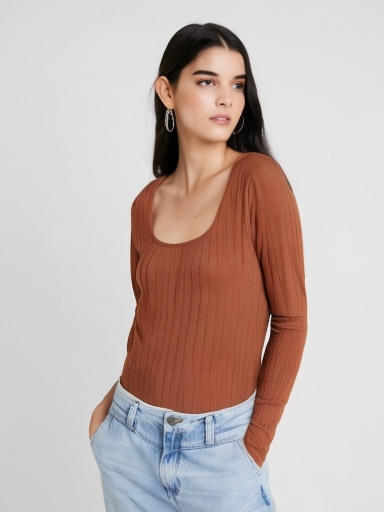} &
\includegraphics[width=0.16\linewidth]{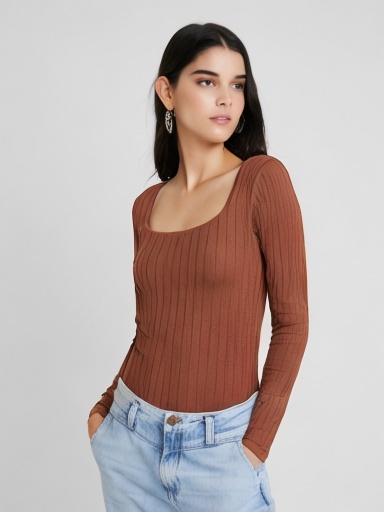} \\

\includegraphics[width=0.16\linewidth]{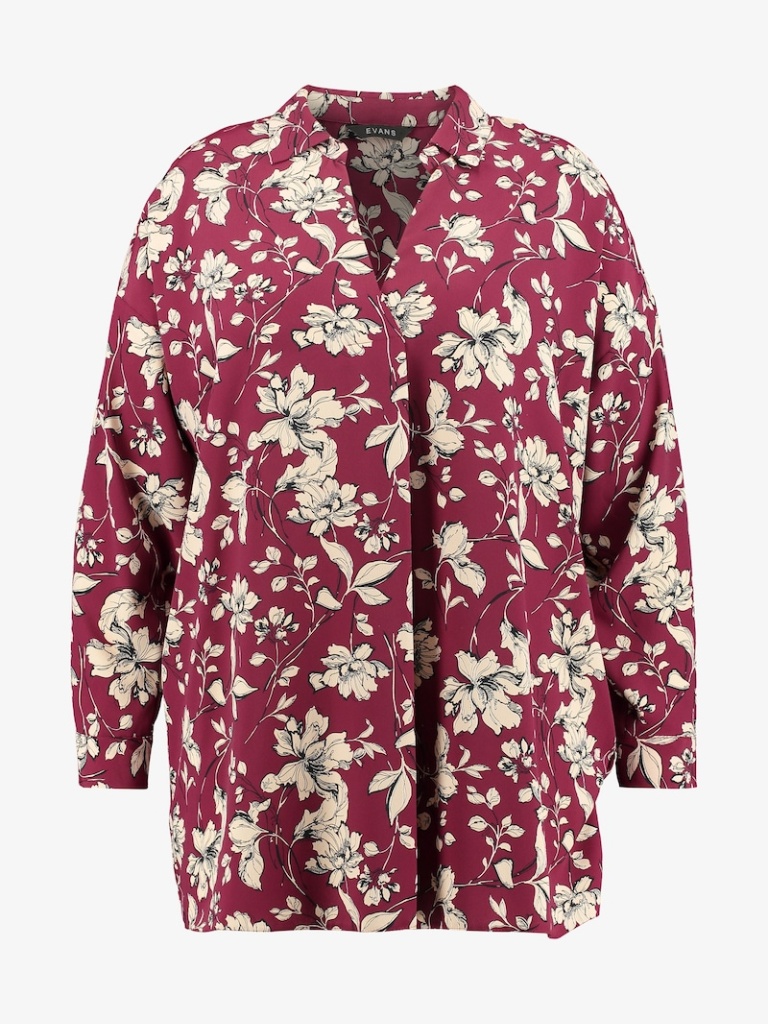} &
\includegraphics[width=0.16\linewidth]{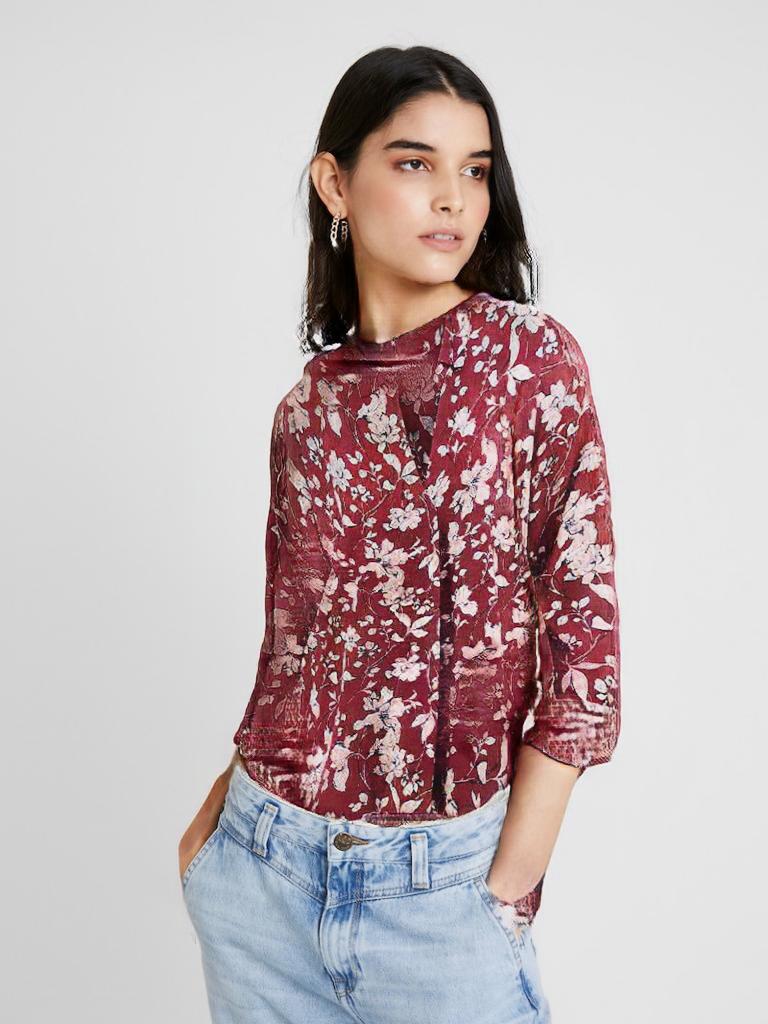} &
\includegraphics[width=0.16\linewidth]{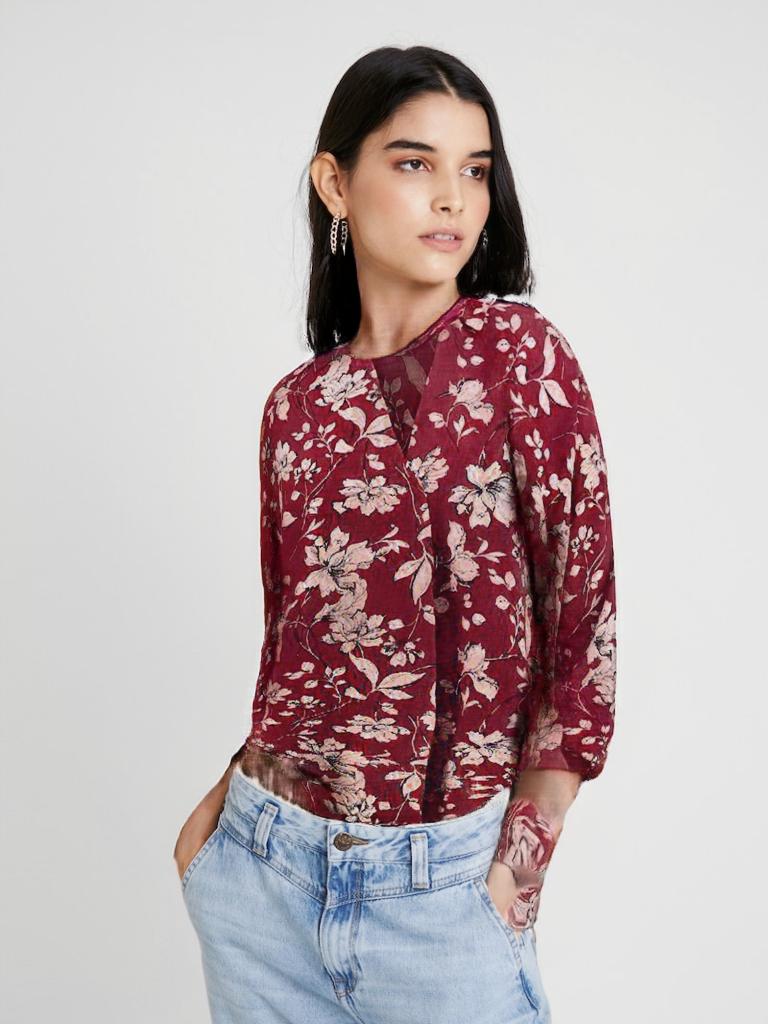} & 
\includegraphics[width=0.16\linewidth]{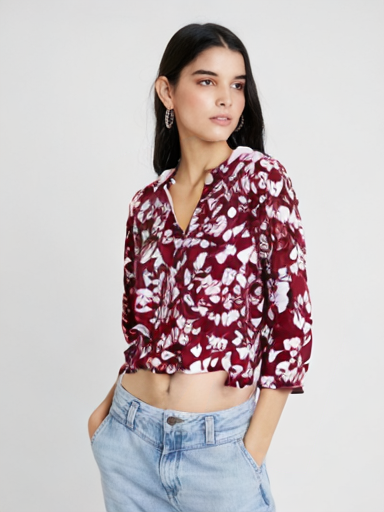} &
\includegraphics[width=0.16\linewidth]{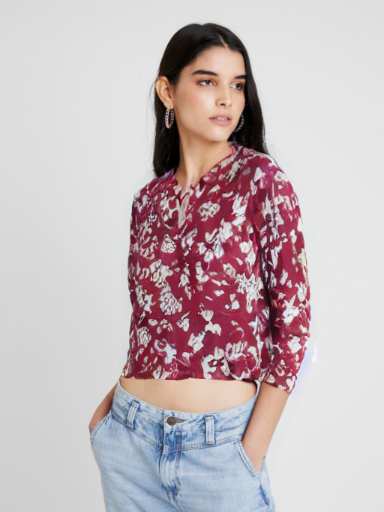} &
\includegraphics[width=0.16\linewidth]{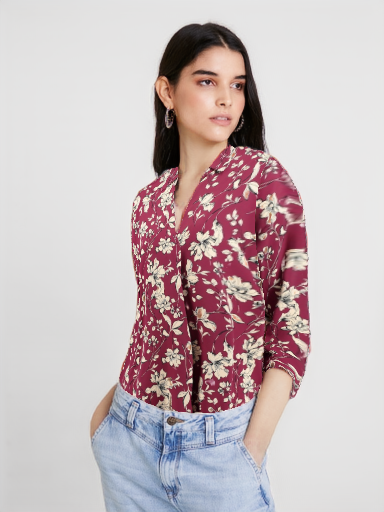} & 
\includegraphics[width=0.16\linewidth]{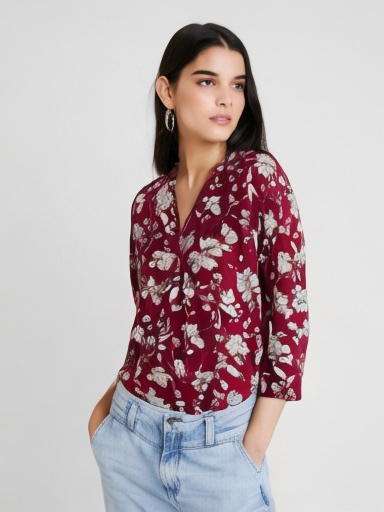} &
\includegraphics[width=0.16\linewidth]{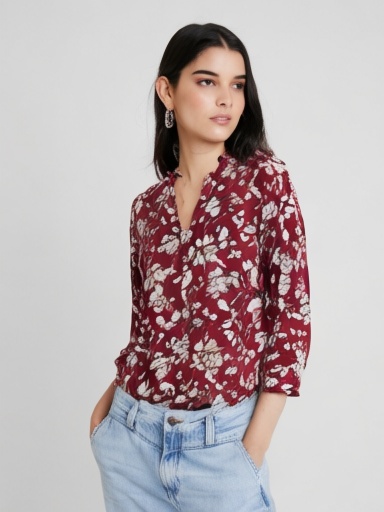} \\

\end{tabular}
}
\caption{Comparison results of the same model trying on different clothing.}
\label{fig:comparison_1}
\end{figure}

\begin{figure}[t]
\centering
\scriptsize
\setlength{\tabcolsep}{.2em}
\resizebox{\linewidth}{!}{
\begin{tabular}{cccccccc}

\includegraphics[width=0.16\linewidth]{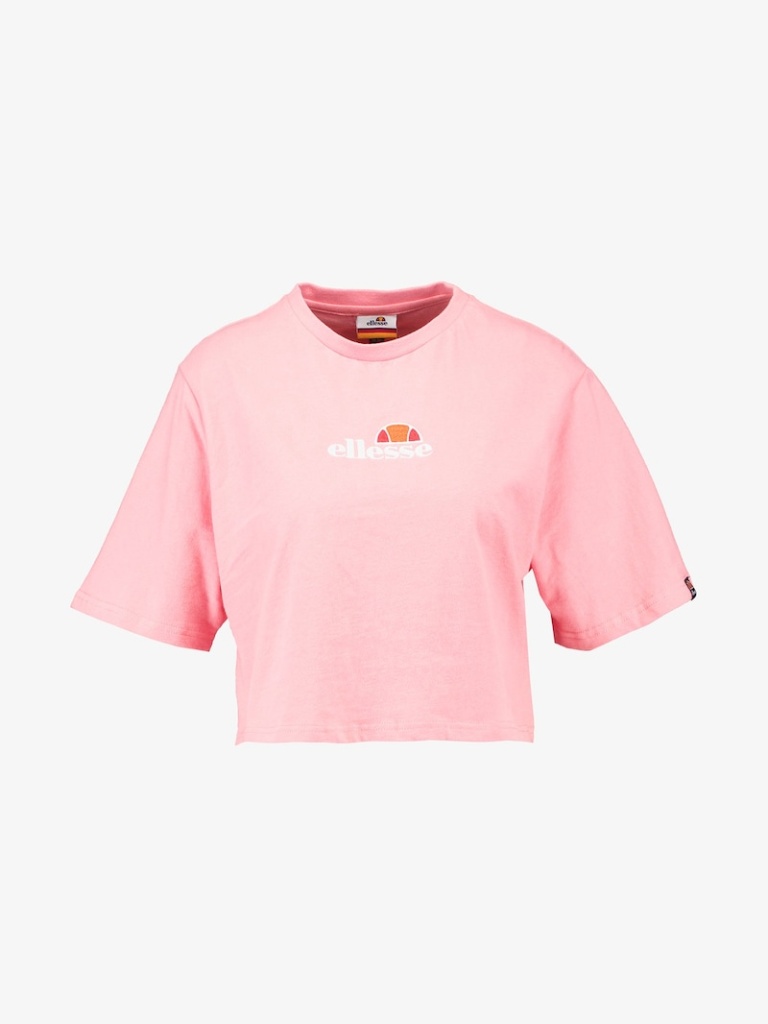} &
\includegraphics[width=0.16\linewidth]{imags/comparison/hd_label.png} &
\includegraphics[width=0.16\linewidth]{imags/comparison/hr_label.png} & 
\includegraphics[width=0.16\linewidth]{imags/comparison/ladi_label.png} &
\includegraphics[width=0.16\linewidth]{imags/comparison/dci_label.png} &
\includegraphics[width=0.16\linewidth]{imags/comparison/gp_label.png} & 
\includegraphics[width=0.16\linewidth]{imags/comparison/sd_label.png} &
\includegraphics[width=0.16\linewidth]{imags/comparison/ours_label.png} \\

\includegraphics[width=0.16\linewidth]{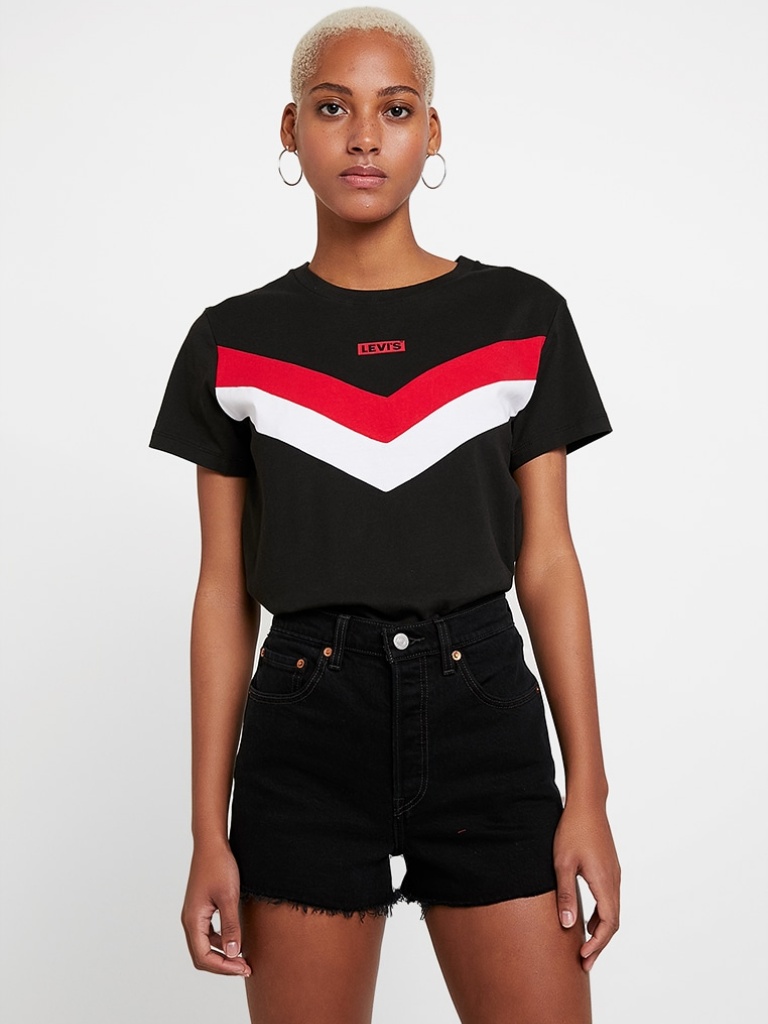} &
\includegraphics[width=0.16\linewidth]{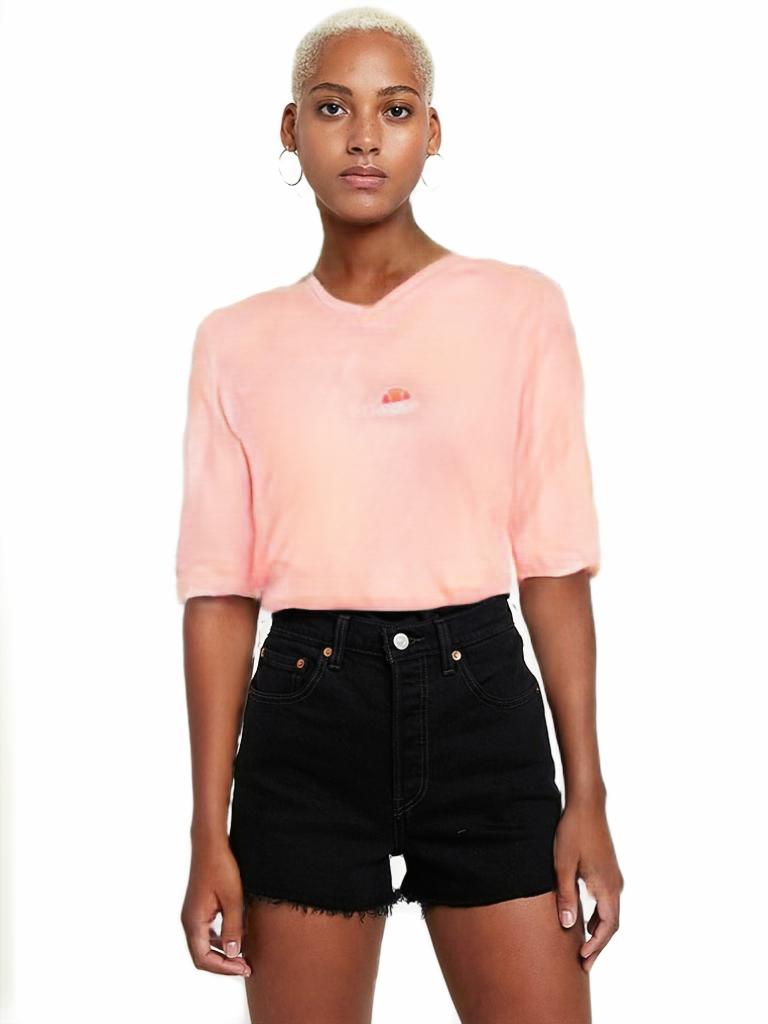} &
\includegraphics[width=0.16\linewidth]{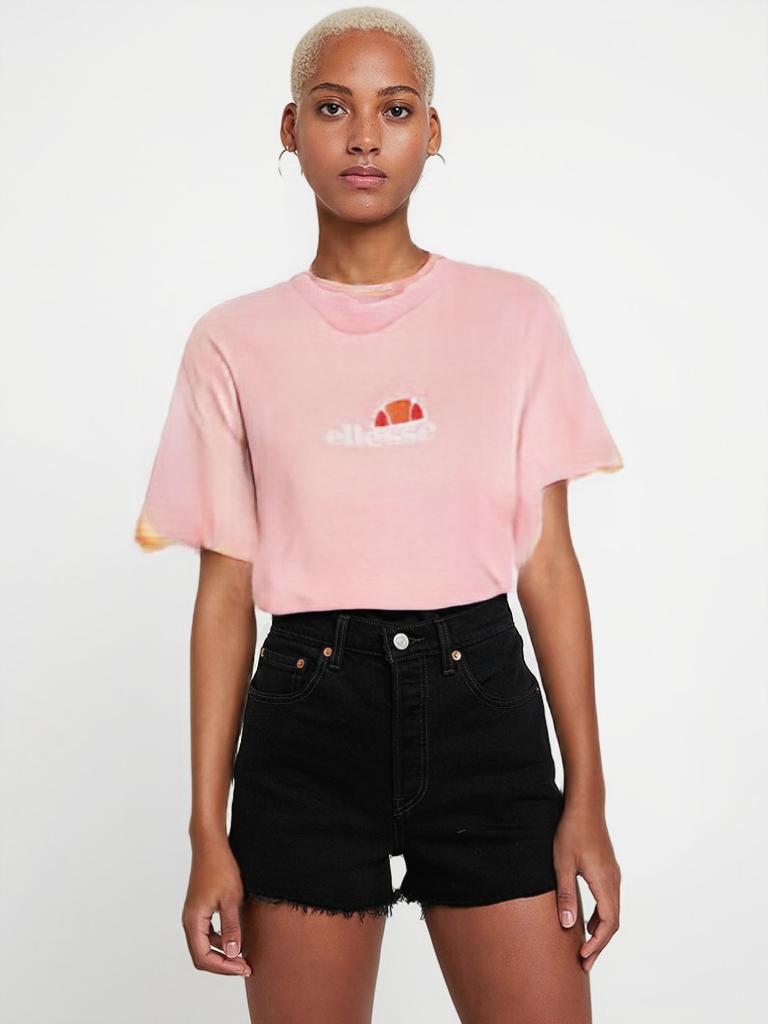} & 
\includegraphics[width=0.16\linewidth]{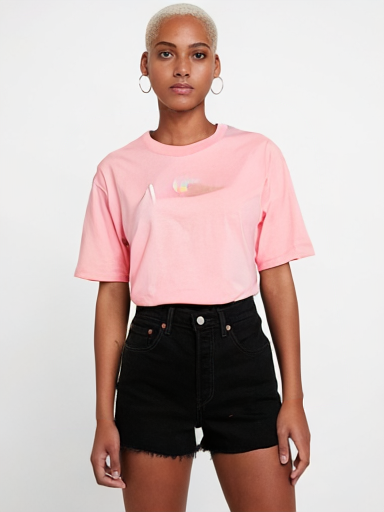} &
\includegraphics[width=0.16\linewidth]{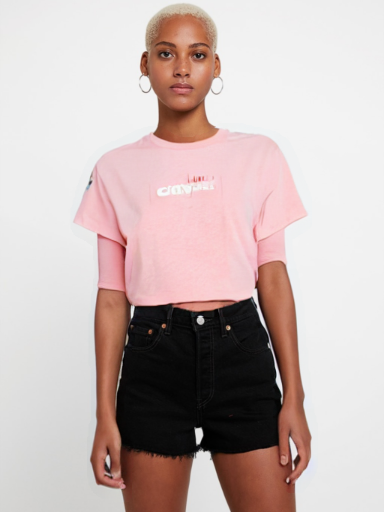} &
\includegraphics[width=0.16\linewidth]{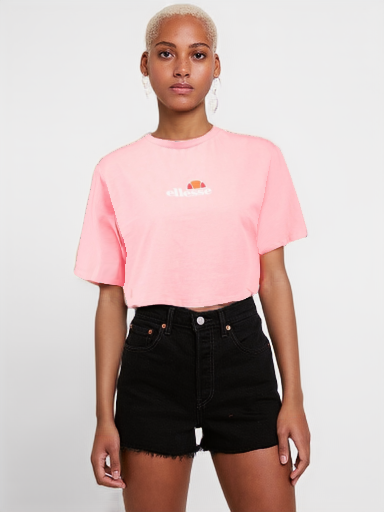} & 
\includegraphics[width=0.16\linewidth]{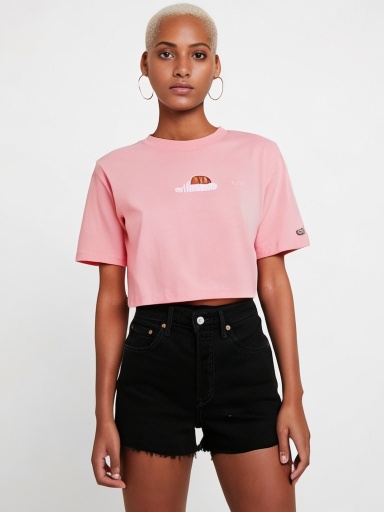} &
\includegraphics[width=0.16\linewidth]{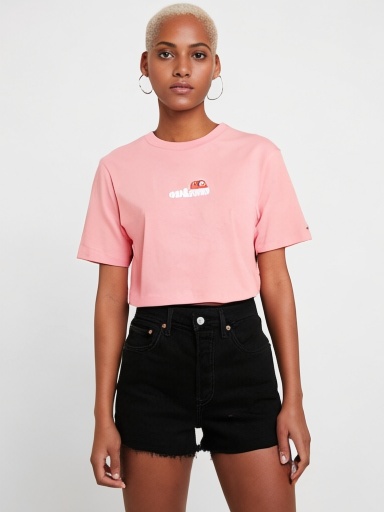} \\

\includegraphics[width=0.16\linewidth]{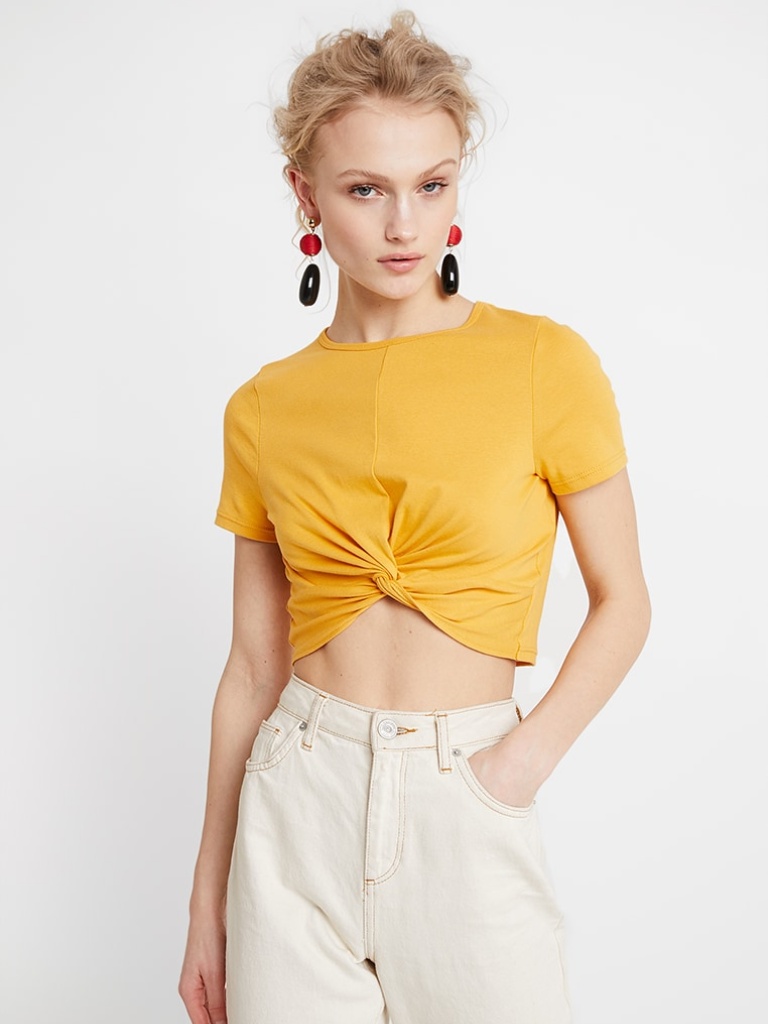} &
\includegraphics[width=0.16\linewidth]{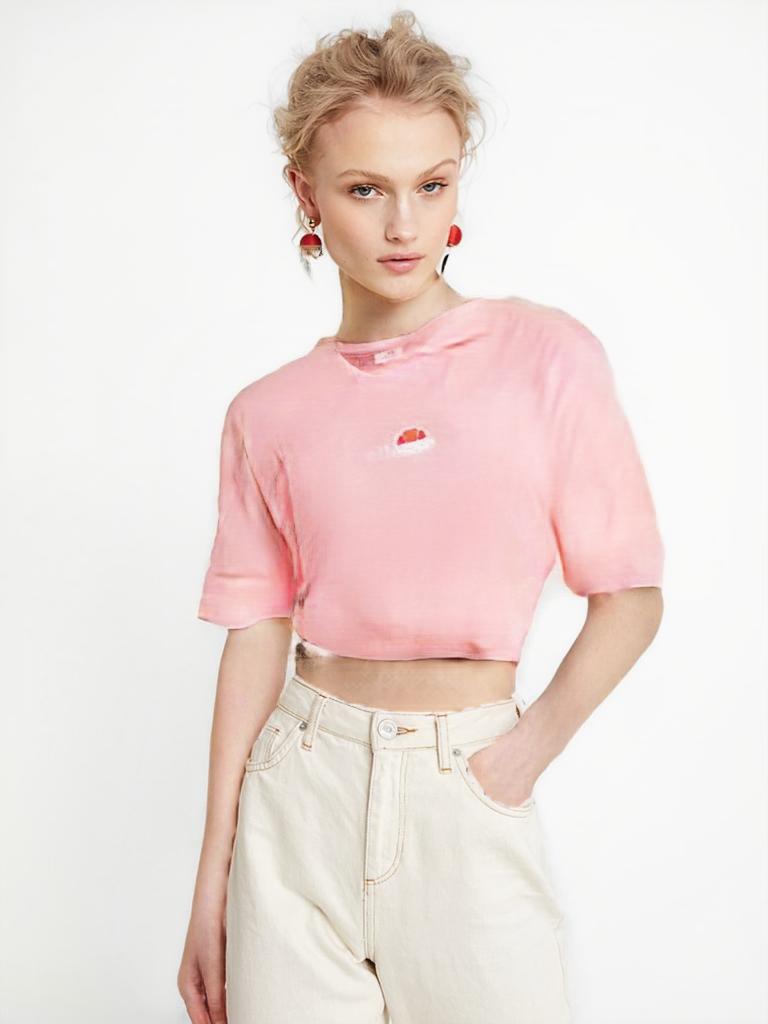} &
\includegraphics[width=0.16\linewidth]{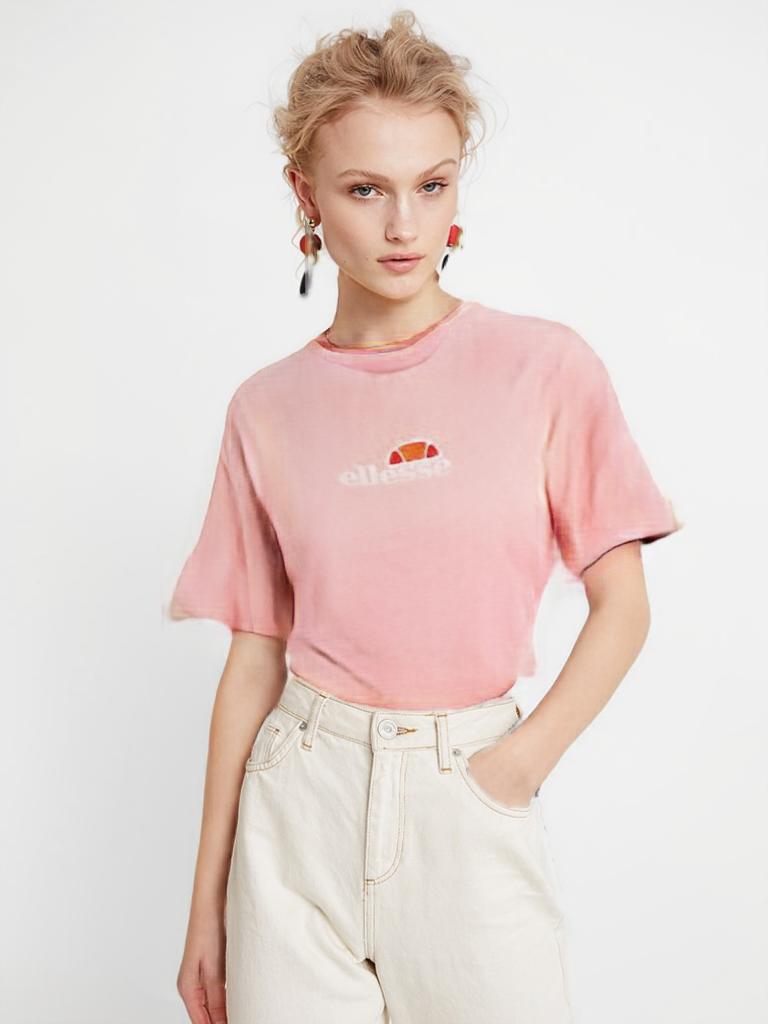} & 
\includegraphics[width=0.16\linewidth]{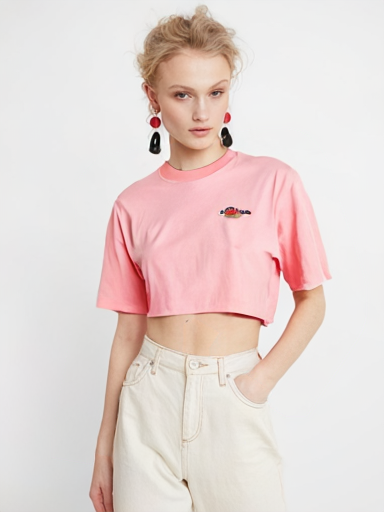} &
\includegraphics[width=0.16\linewidth]{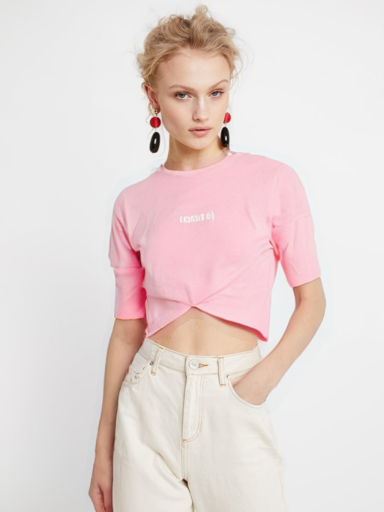} &
\includegraphics[width=0.16\linewidth]{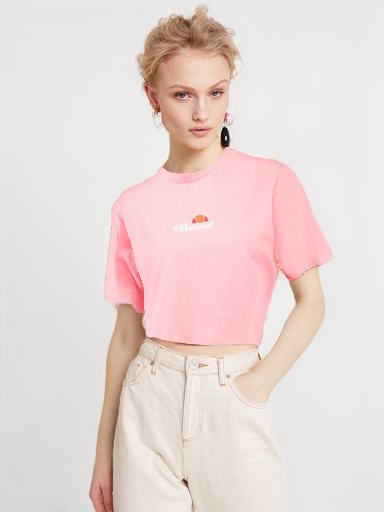} & 
\includegraphics[width=0.16\linewidth]{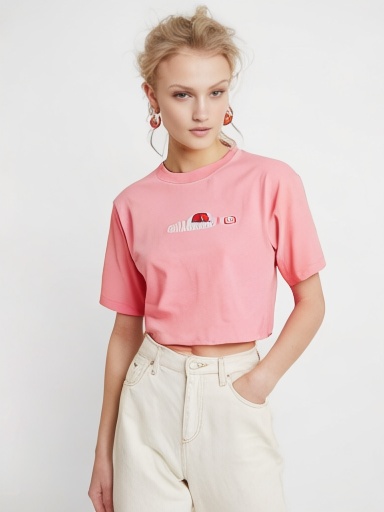} &
\includegraphics[width=0.16\linewidth]{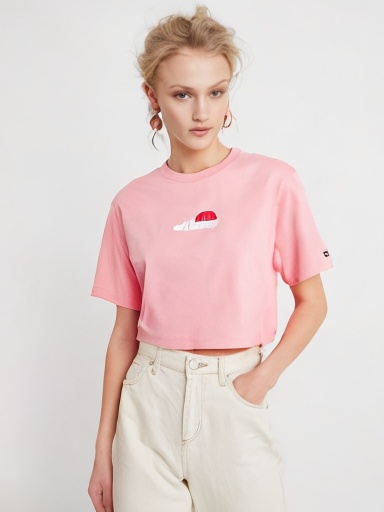} \\

\includegraphics[width=0.16\linewidth]{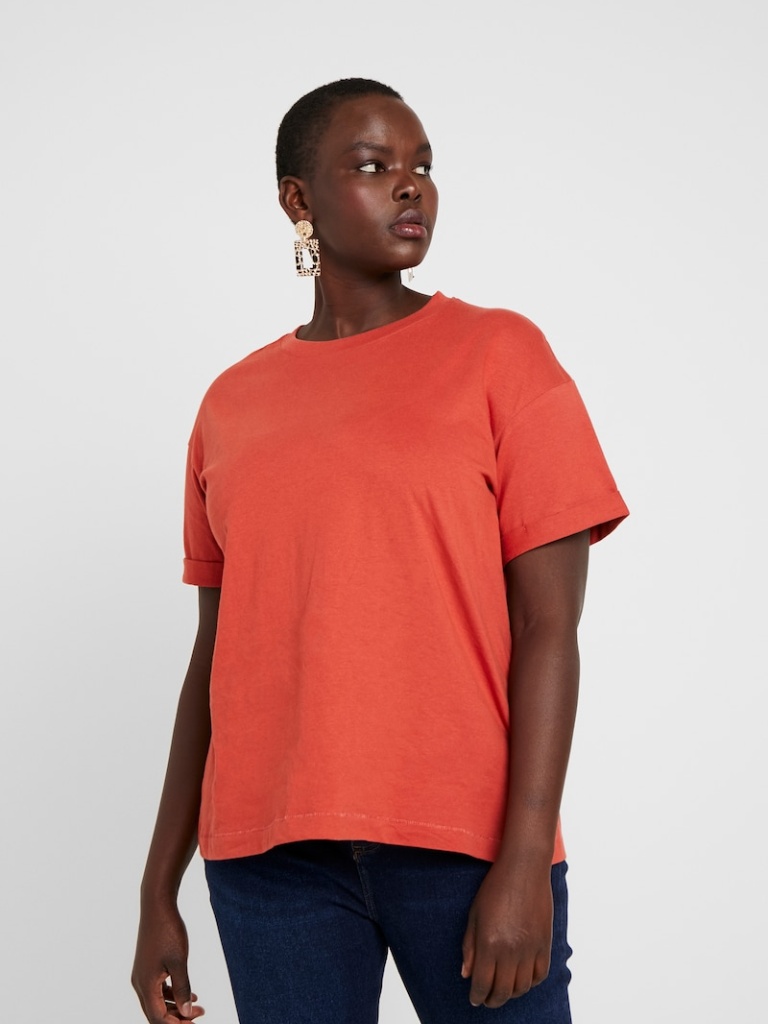} &
\includegraphics[width=0.16\linewidth]{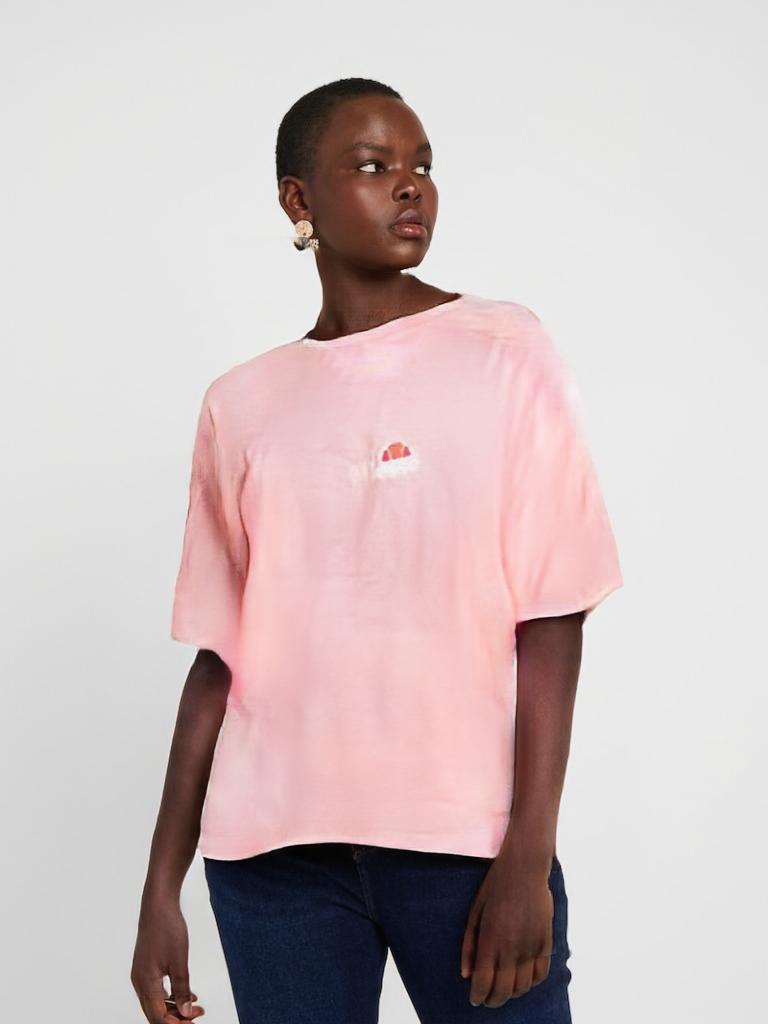} &
\includegraphics[width=0.16\linewidth]{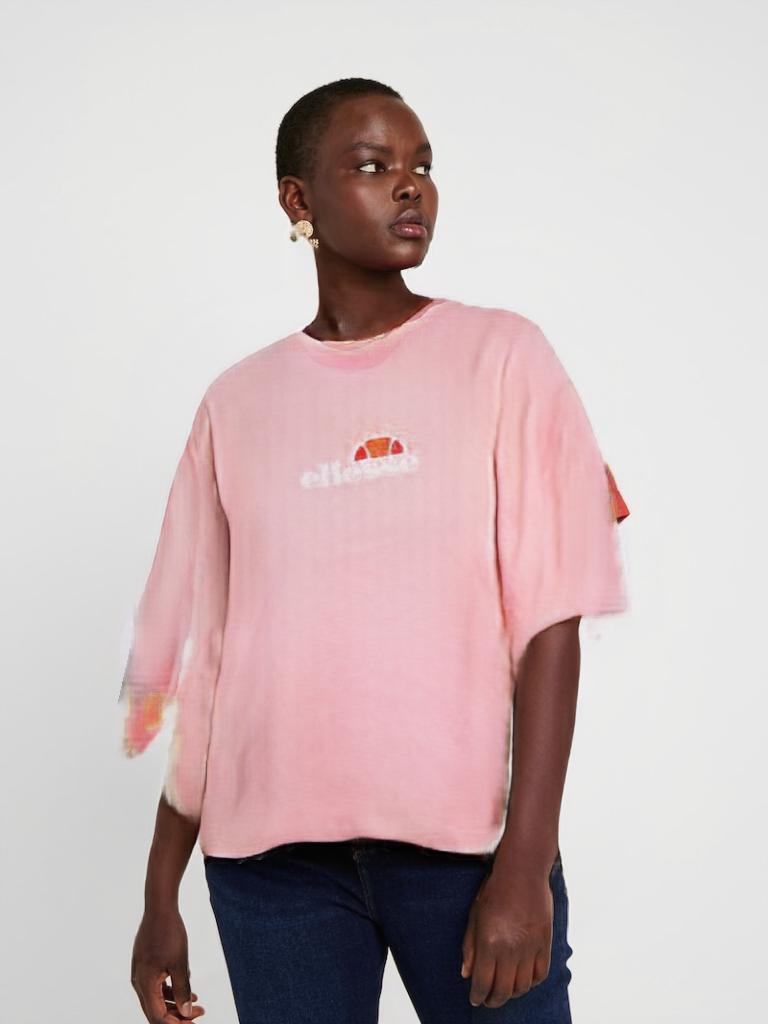} & 
\includegraphics[width=0.16\linewidth]{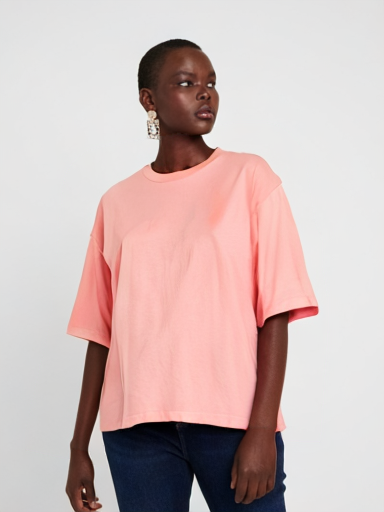} &
\includegraphics[width=0.16\linewidth]{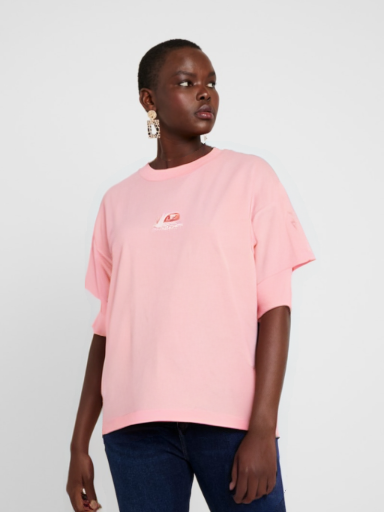} &
\includegraphics[width=0.16\linewidth]{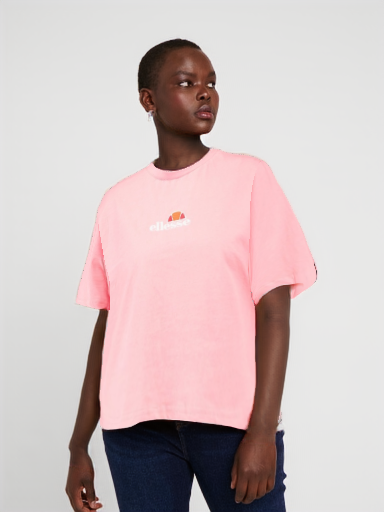} & 
\includegraphics[width=0.16\linewidth]{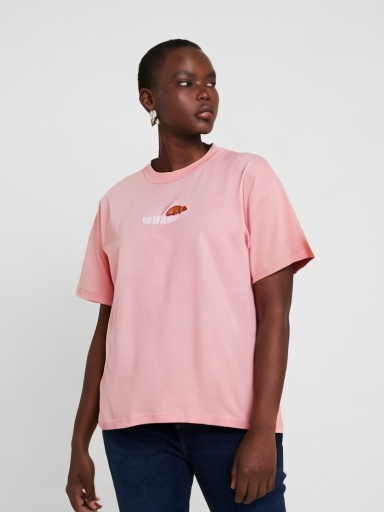} &
\includegraphics[width=0.16\linewidth]{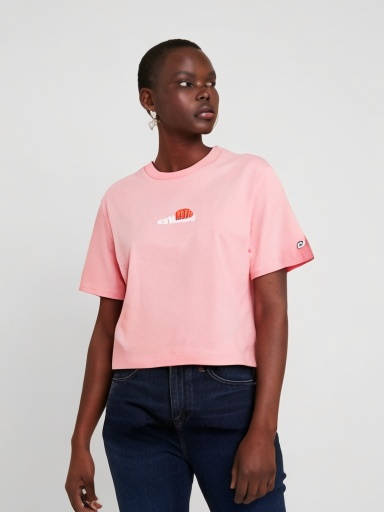} \\

\includegraphics[width=0.16\linewidth]{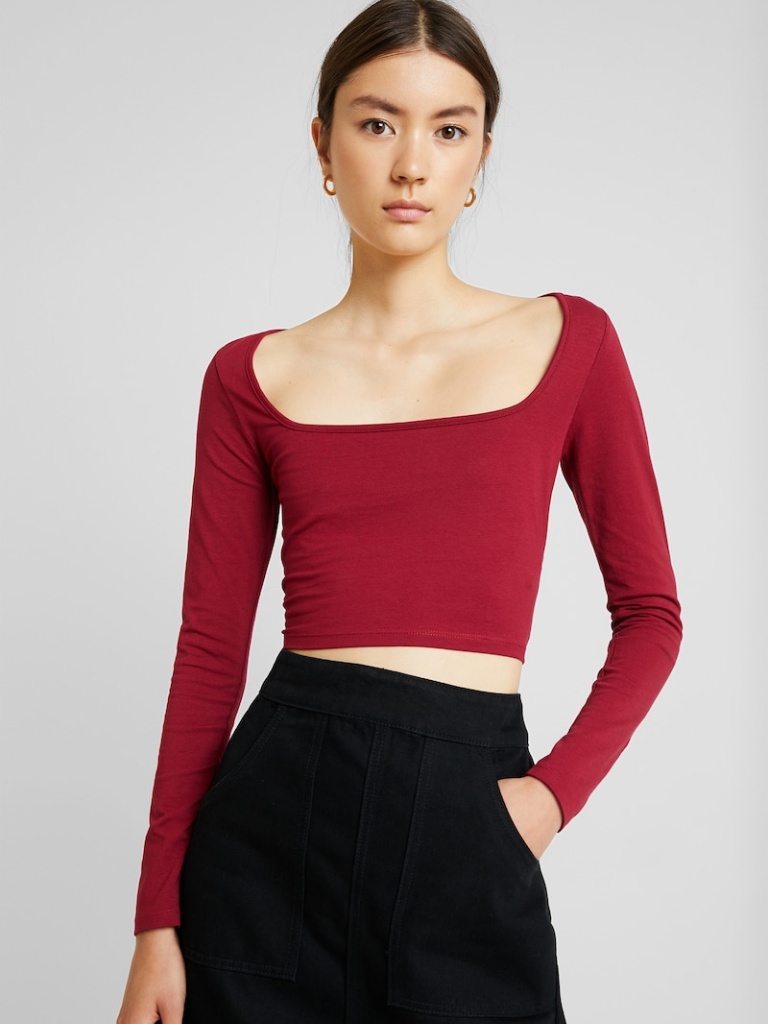} &
\includegraphics[width=0.16\linewidth]{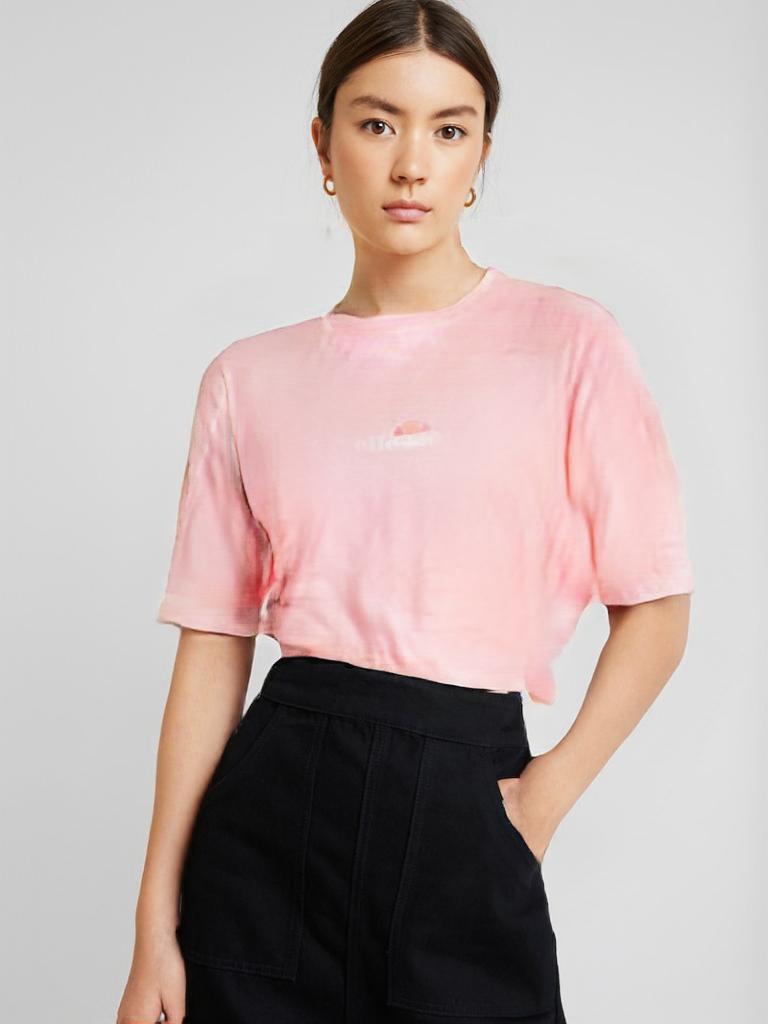} &
\includegraphics[width=0.16\linewidth]{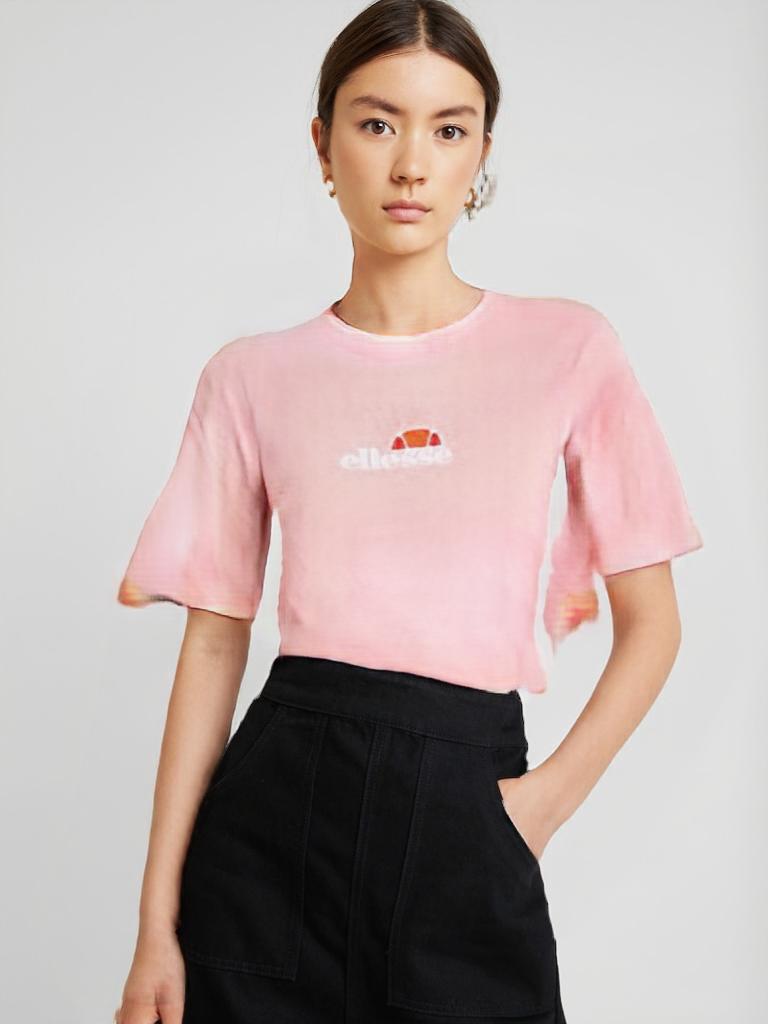} & 
\includegraphics[width=0.16\linewidth]{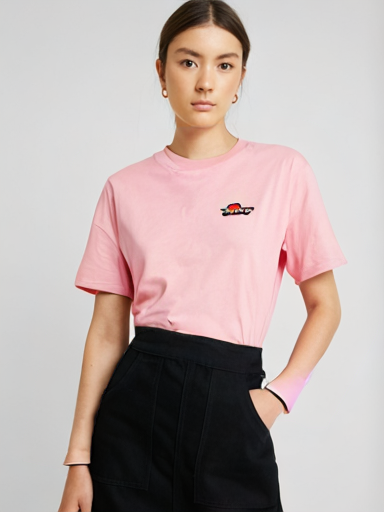} &
\includegraphics[width=0.16\linewidth]{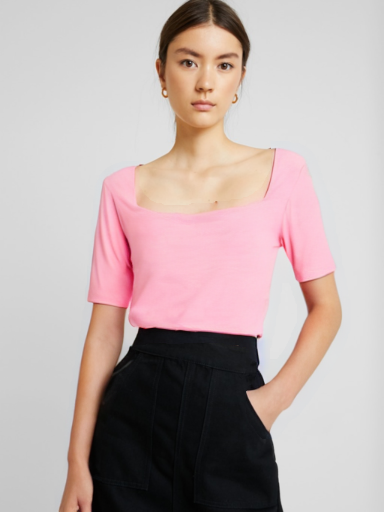} &
\includegraphics[width=0.16\linewidth]{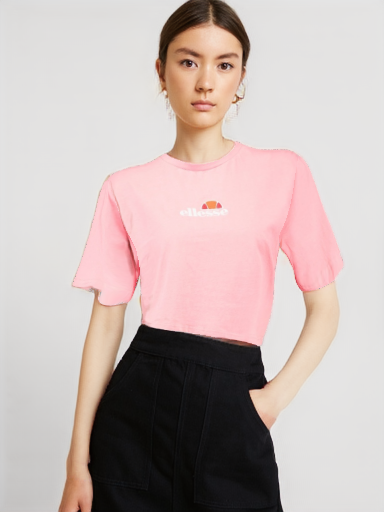} & 
\includegraphics[width=0.16\linewidth]{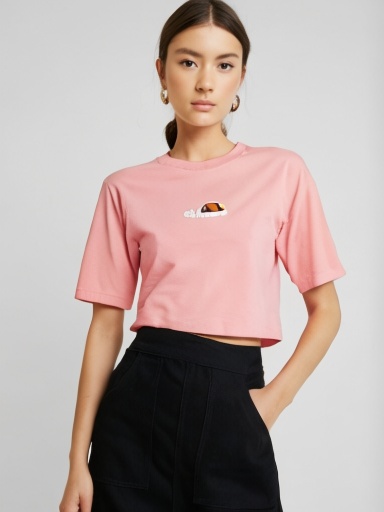} &
\includegraphics[width=0.16\linewidth]{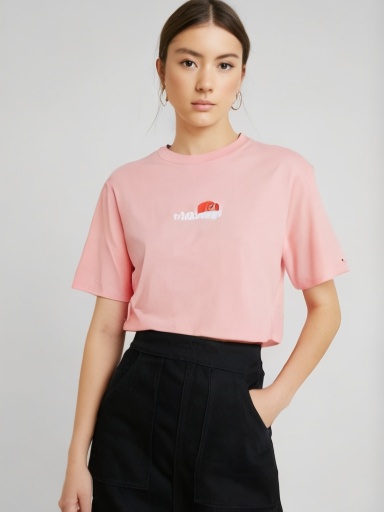} \\

\includegraphics[width=0.16\linewidth]{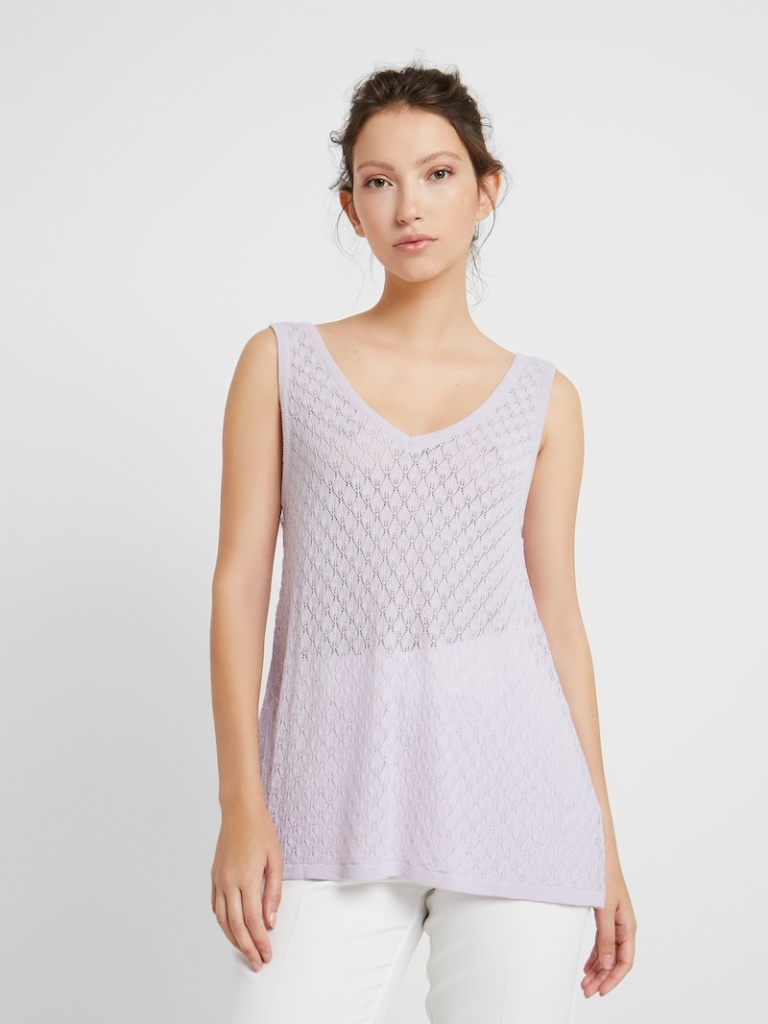} &
\includegraphics[width=0.16\linewidth]{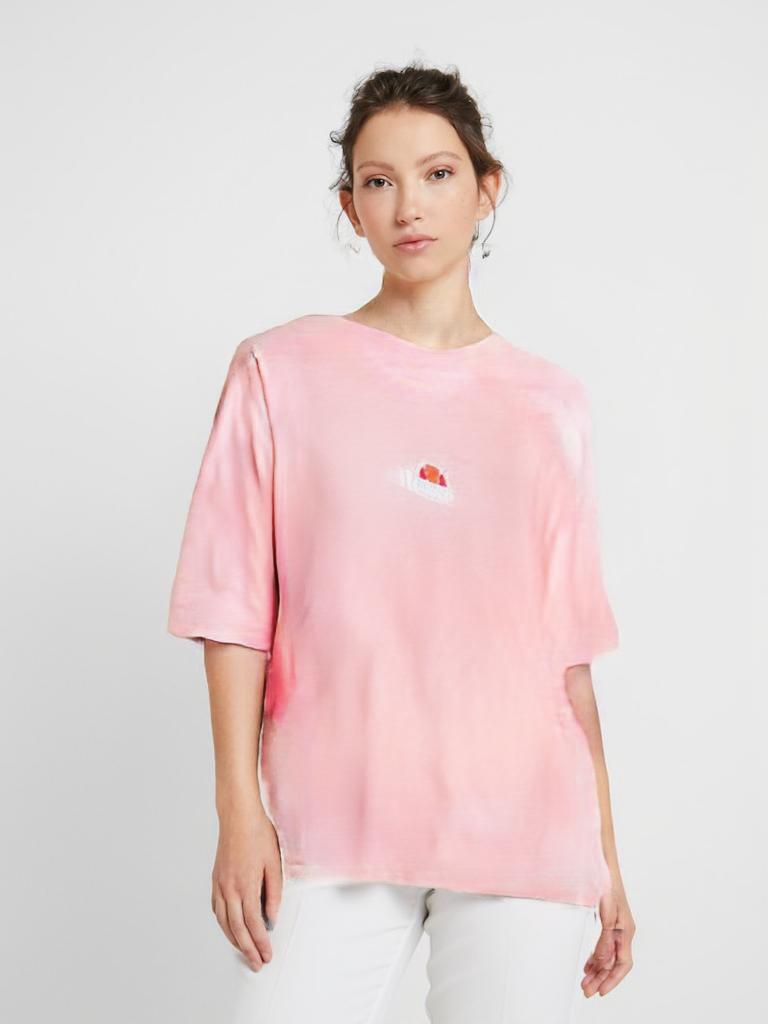} &
\includegraphics[width=0.16\linewidth]{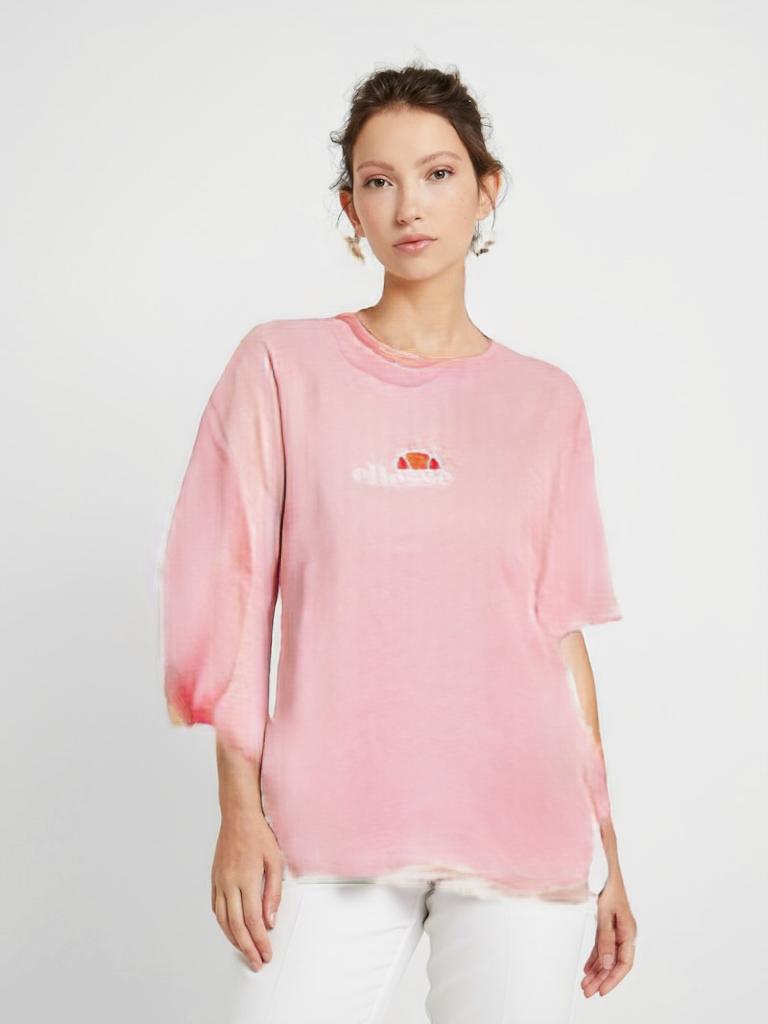} & 
\includegraphics[width=0.16\linewidth]{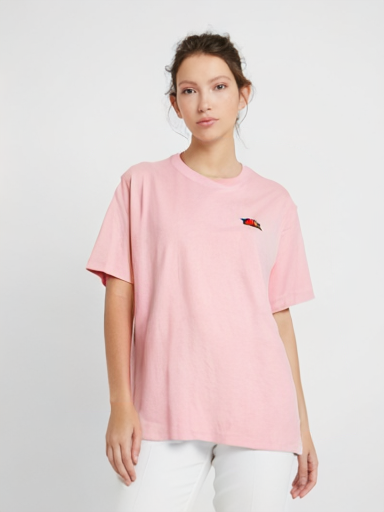} &
\includegraphics[width=0.16\linewidth]{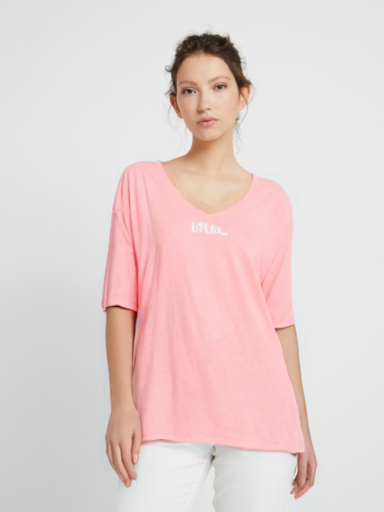} &
\includegraphics[width=0.16\linewidth]{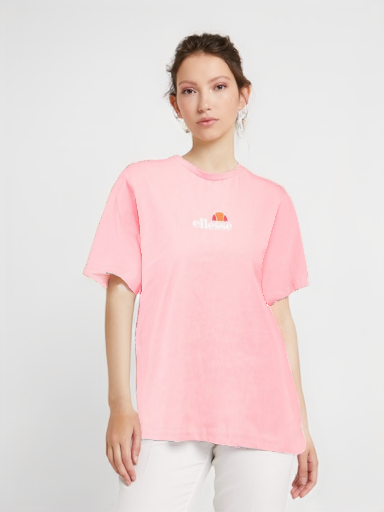} & 
\includegraphics[width=0.16\linewidth]{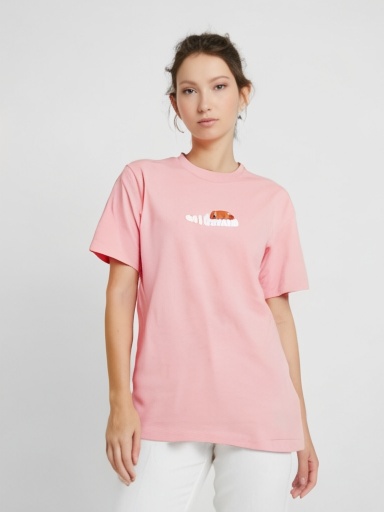} &
\includegraphics[width=0.16\linewidth]{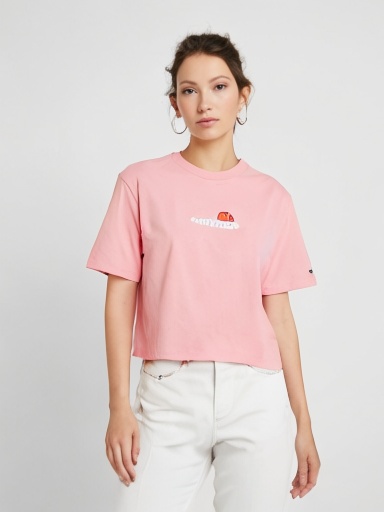} \\

\includegraphics[width=0.16\linewidth]{imags/comparison/03731_00.jpg} &
\includegraphics[width=0.16\linewidth]{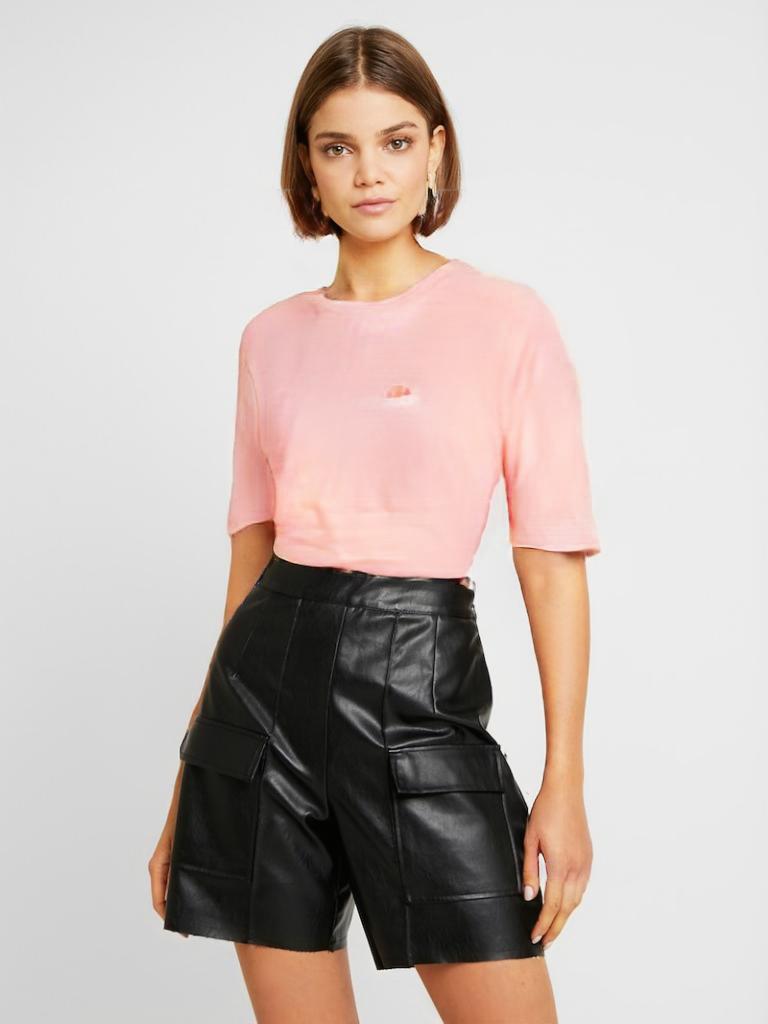} &
\includegraphics[width=0.16\linewidth]{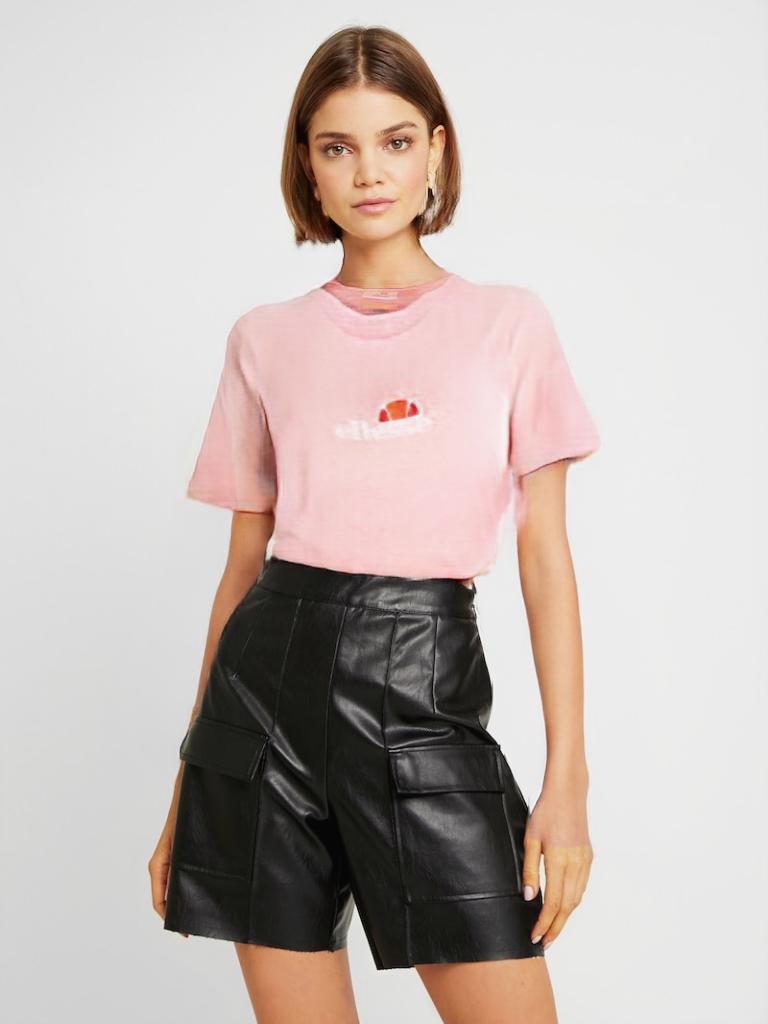} & 
\includegraphics[width=0.16\linewidth]{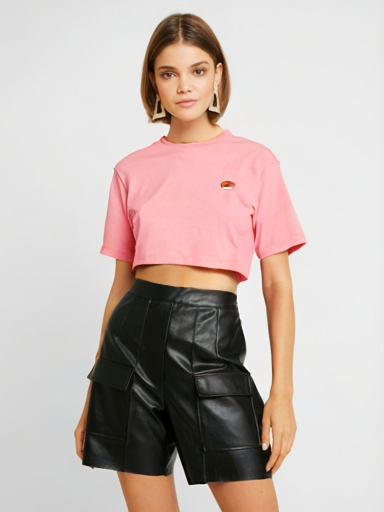} &
\includegraphics[width=0.16\linewidth]{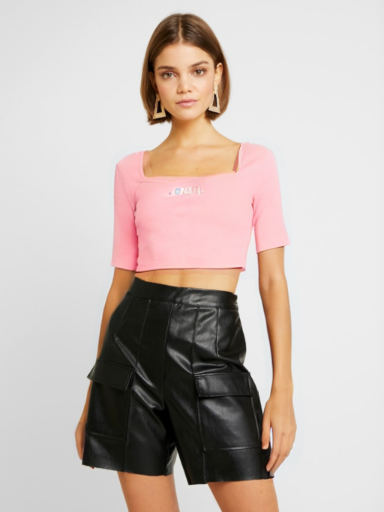} &
\includegraphics[width=0.16\linewidth]{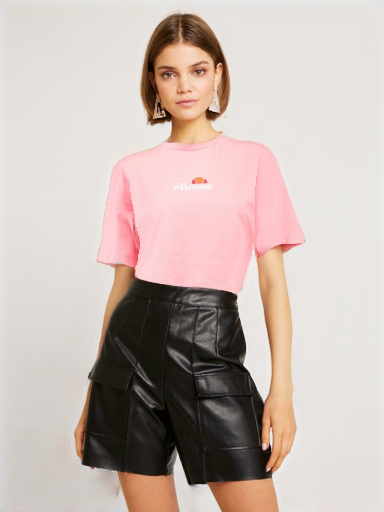} & 
\includegraphics[width=0.16\linewidth]{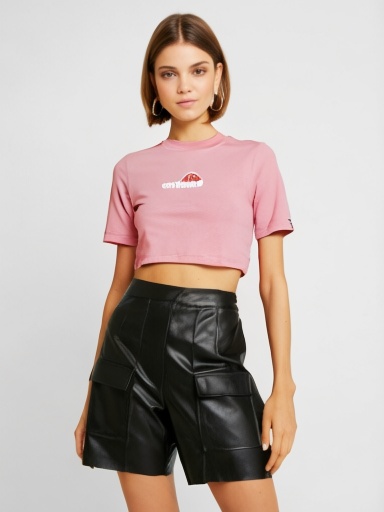} &
\includegraphics[width=0.16\linewidth]{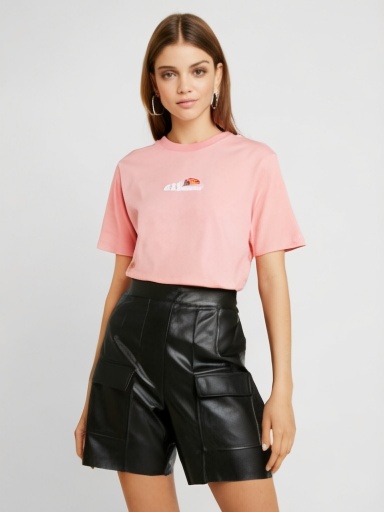} \\

\includegraphics[width=0.16\linewidth]{imags/comparison/04096_00.jpg} &
\includegraphics[width=0.16\linewidth]{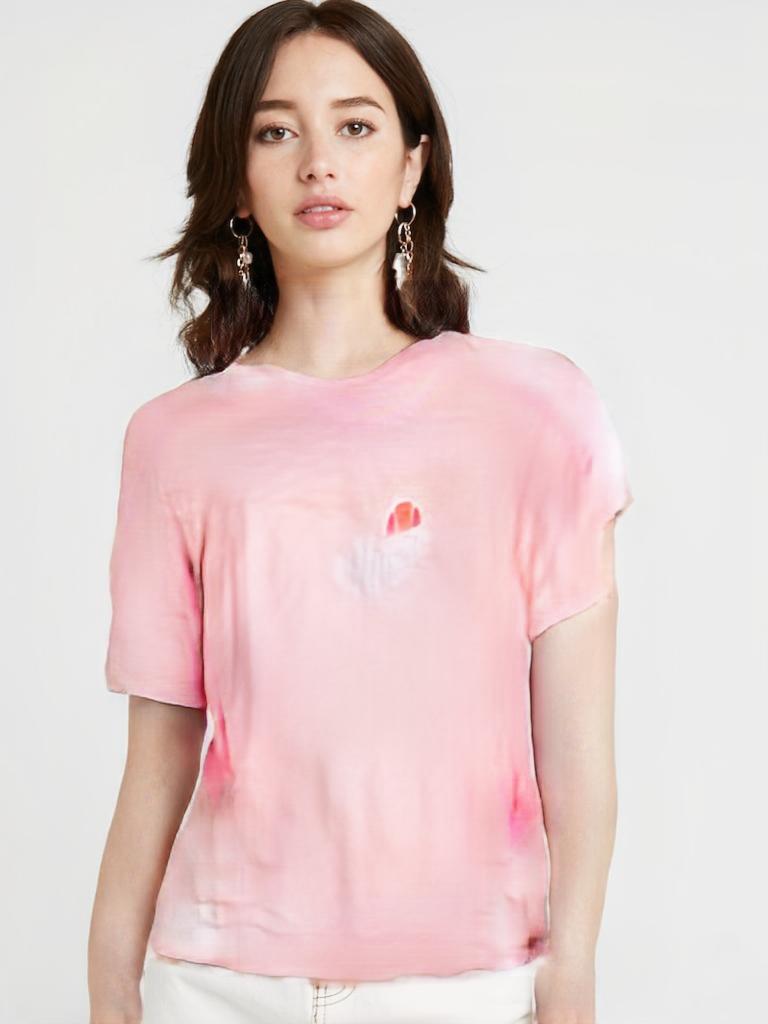} &
\includegraphics[width=0.16\linewidth]{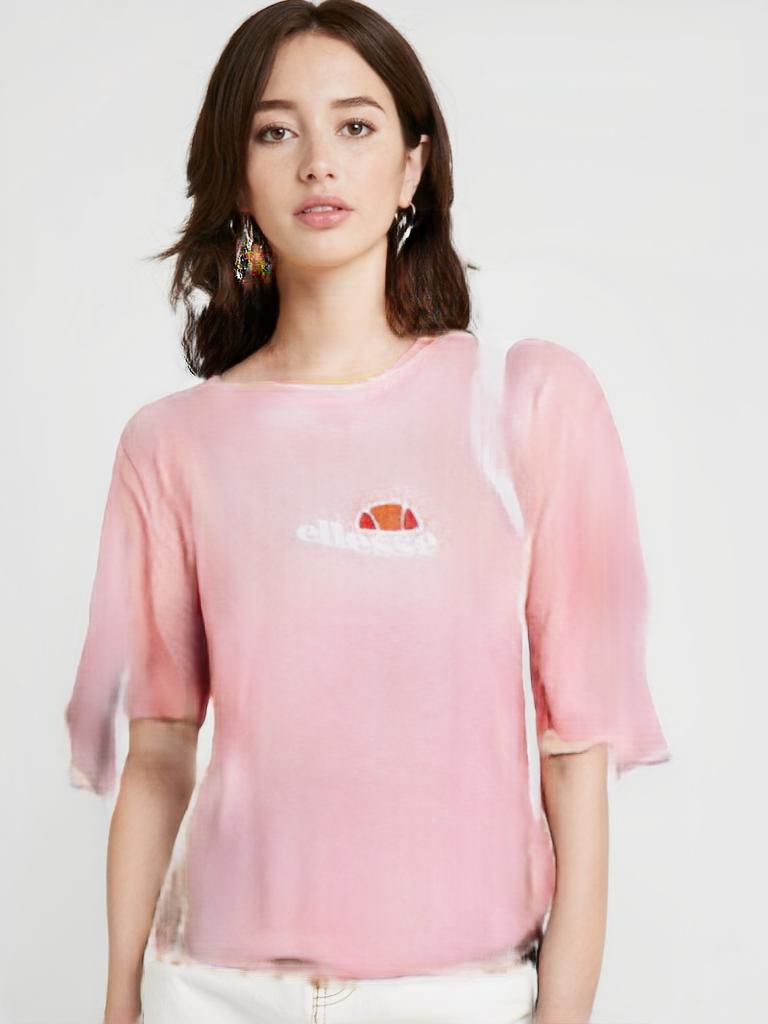} & 
\includegraphics[width=0.16\linewidth]{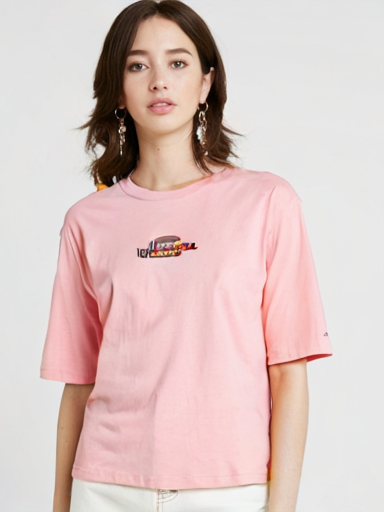} &
\includegraphics[width=0.16\linewidth]{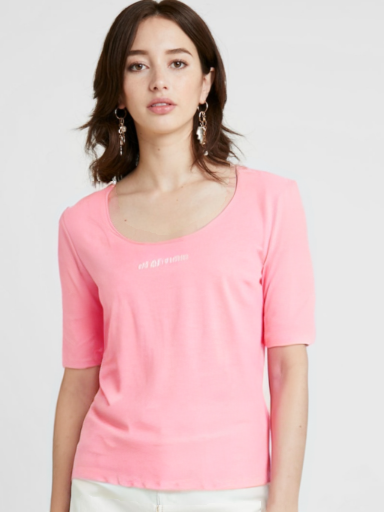} &
\includegraphics[width=0.16\linewidth]{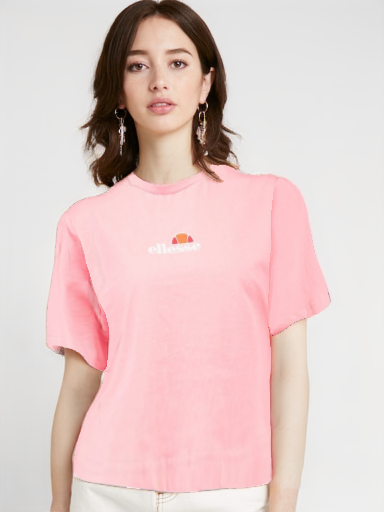} & 
\includegraphics[width=0.16\linewidth]{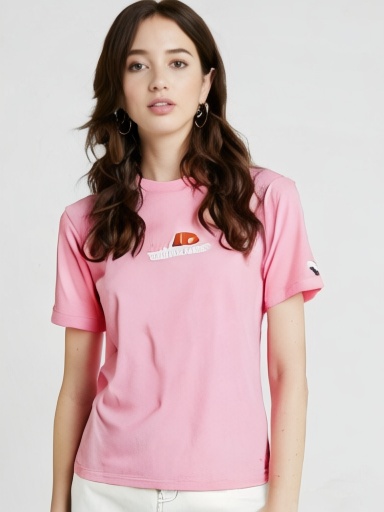} &
\includegraphics[width=0.16\linewidth]{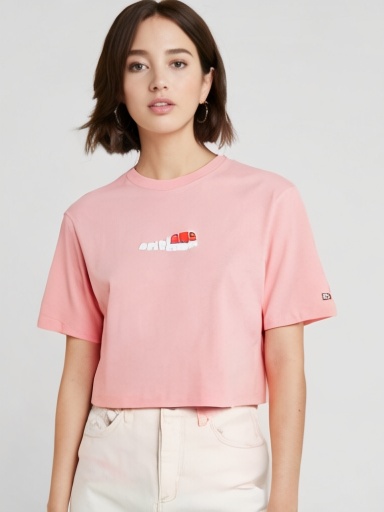} \\

\includegraphics[width=0.16\linewidth]{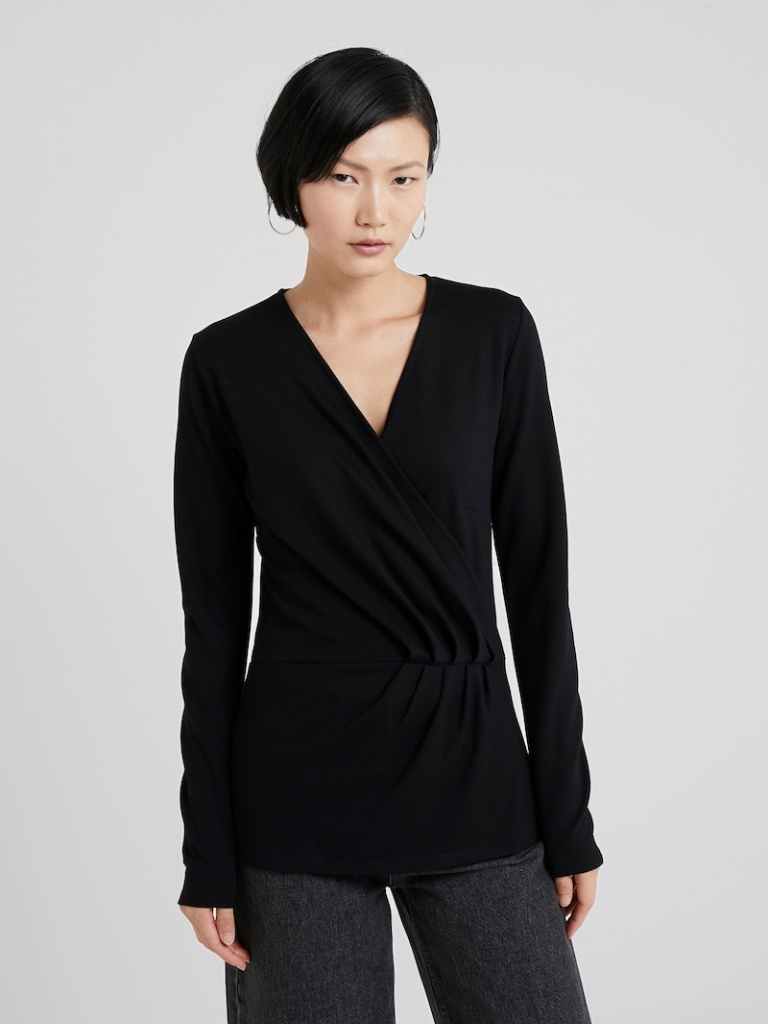} &
\includegraphics[width=0.16\linewidth]{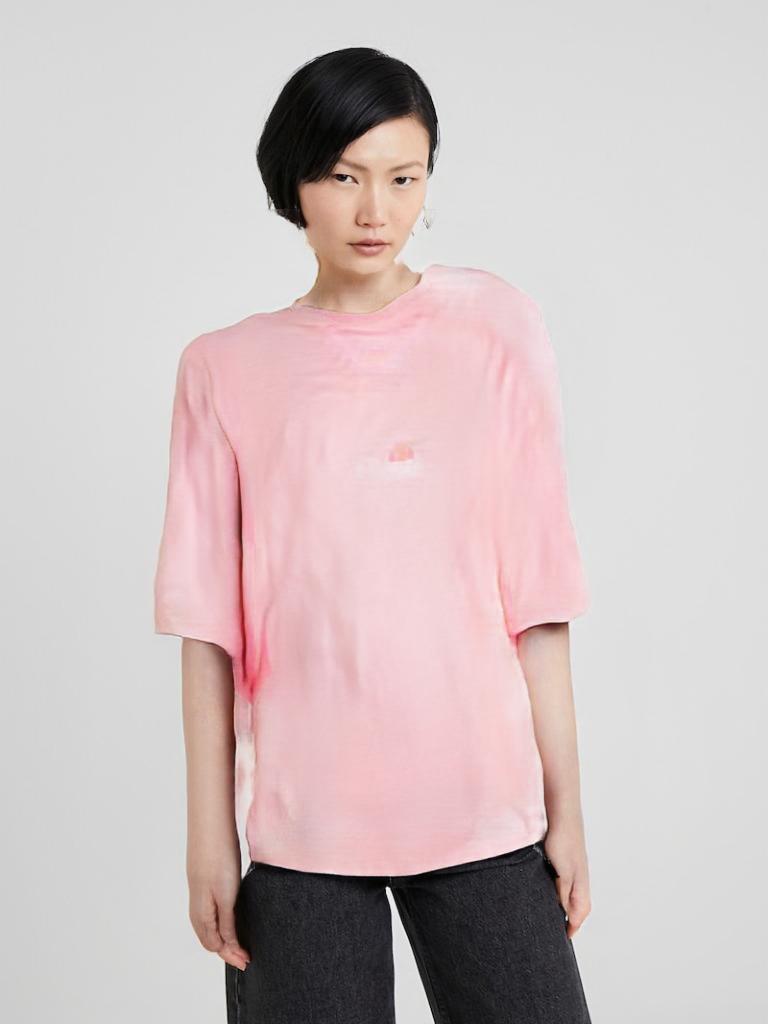} &
\includegraphics[width=0.16\linewidth]{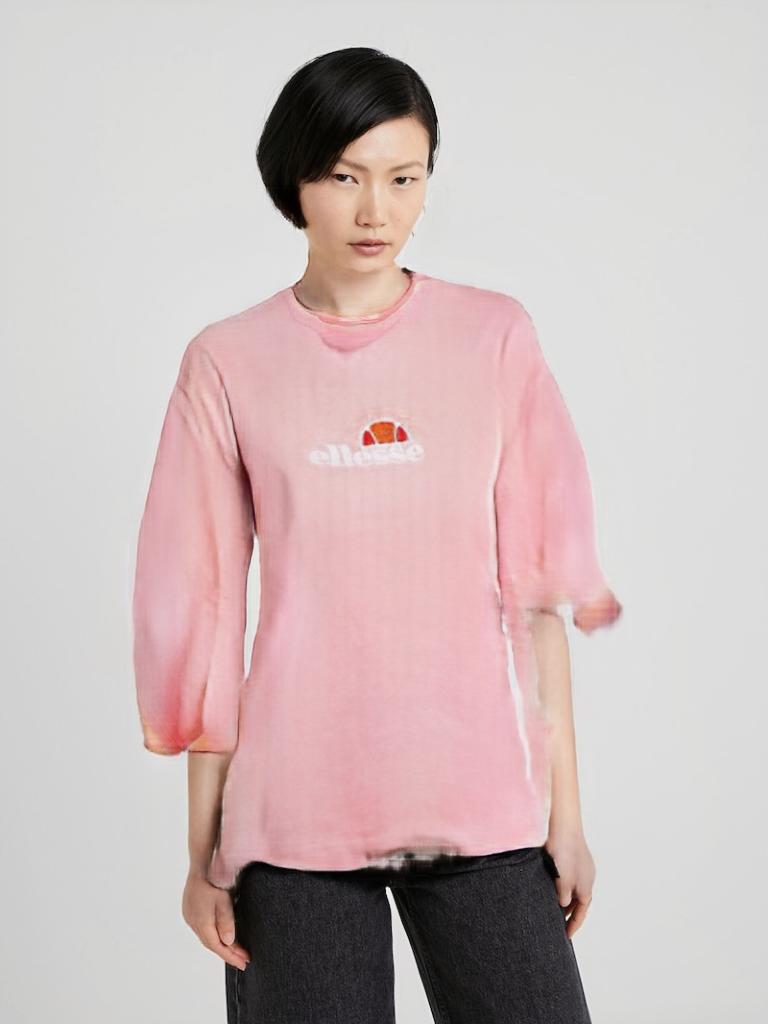} & 
\includegraphics[width=0.16\linewidth]{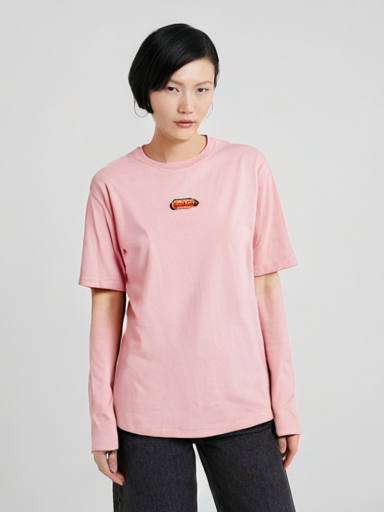} &
\includegraphics[width=0.16\linewidth]{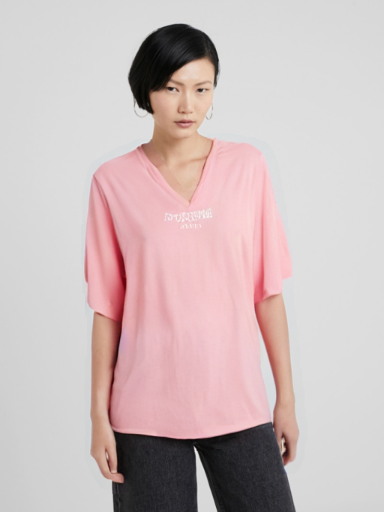} &
\includegraphics[width=0.16\linewidth]{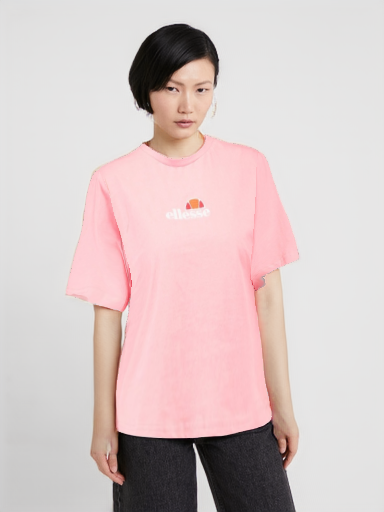} & 
\includegraphics[width=0.16\linewidth]{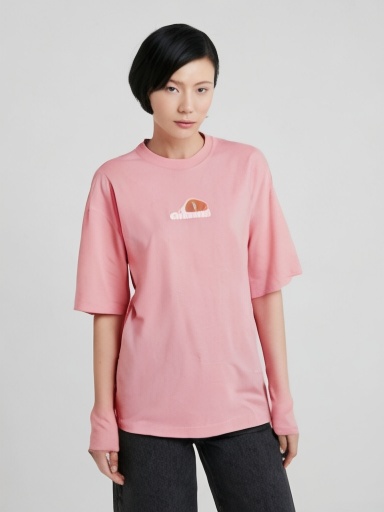} &
\includegraphics[width=0.16\linewidth]{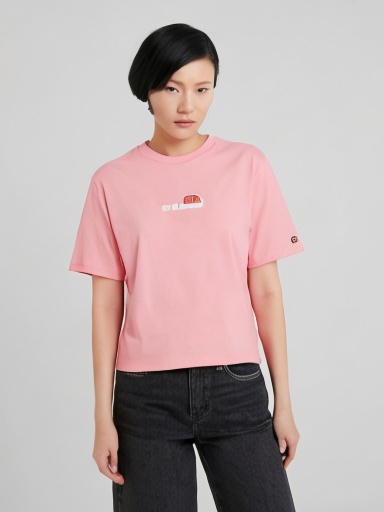} \\

\end{tabular}
}
\caption{Comparison results of different models trying on the same clothing.}
\label{fig:comparison_2}
\end{figure}

Fig. \ref{fig:metric_analysis_reb} shows two examples containing both good fit and not good fit results, where the try-on quality is also reflected by the proposed metrics. 
The main advantage in try-on results against previous methods is the preservation of clothing types, which is measured by SDR. The try-on situation that demands highly in type preservation is marked with green in Fig. \ref{fig:benchmark_reb}, which occupies a small portion in VITON-HD. Therefore, our results acquire marginal improvement on Unpair-2032, but significant improvement in SDR on Cross-27. Besides comparing clothing  types by SDR, S-LPIPS is proposed to compare clothing  texture, both of which are proposed to make up the drawbacks of existing metrics in unpaired evaluation.

\begin{figure}[t]
\centering
\scriptsize
\setlength{\tabcolsep}{.1em}
\begin{tabular}{c c c}

\includegraphics[width=0.16\linewidth]{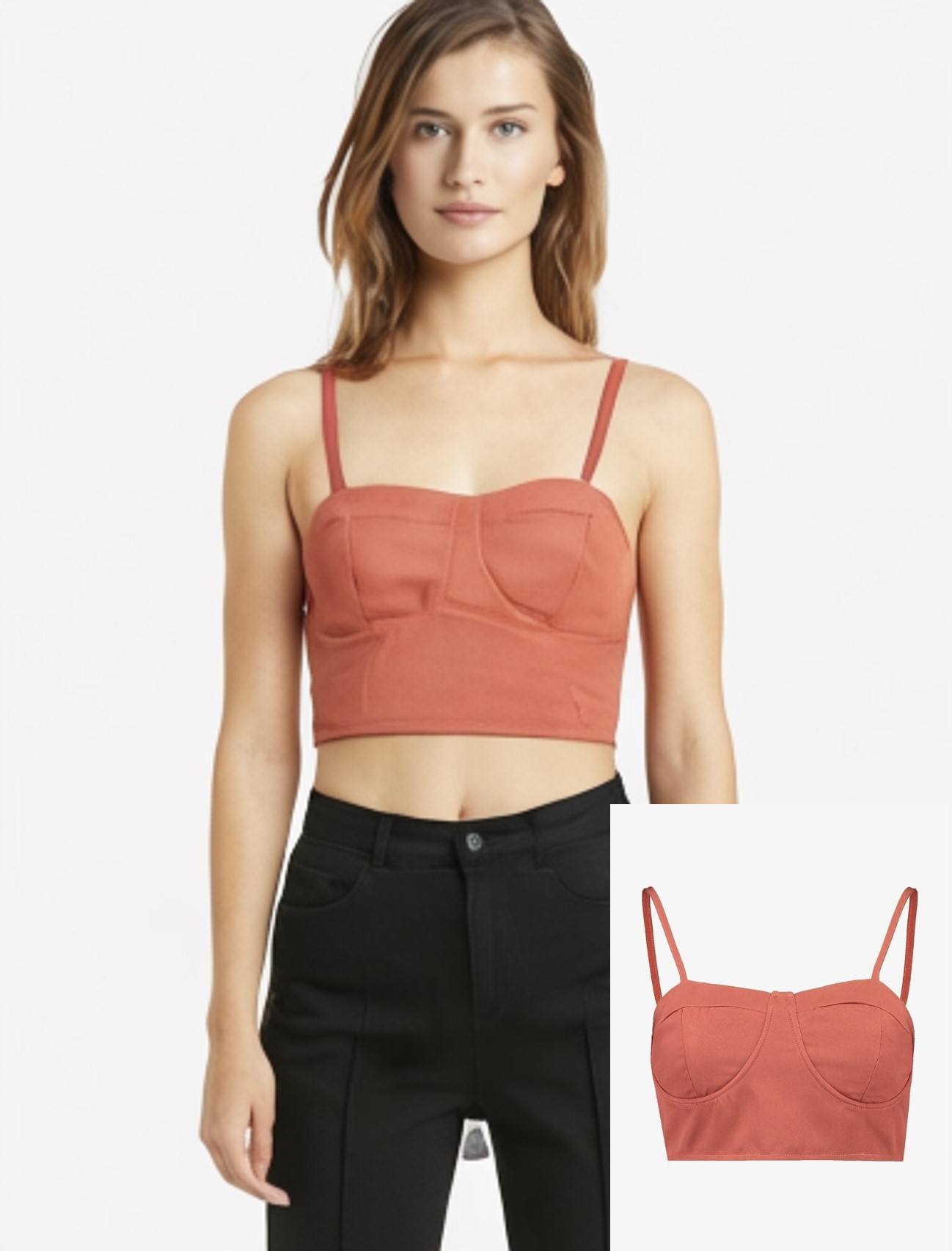} & 
\includegraphics[width=0.16\linewidth]{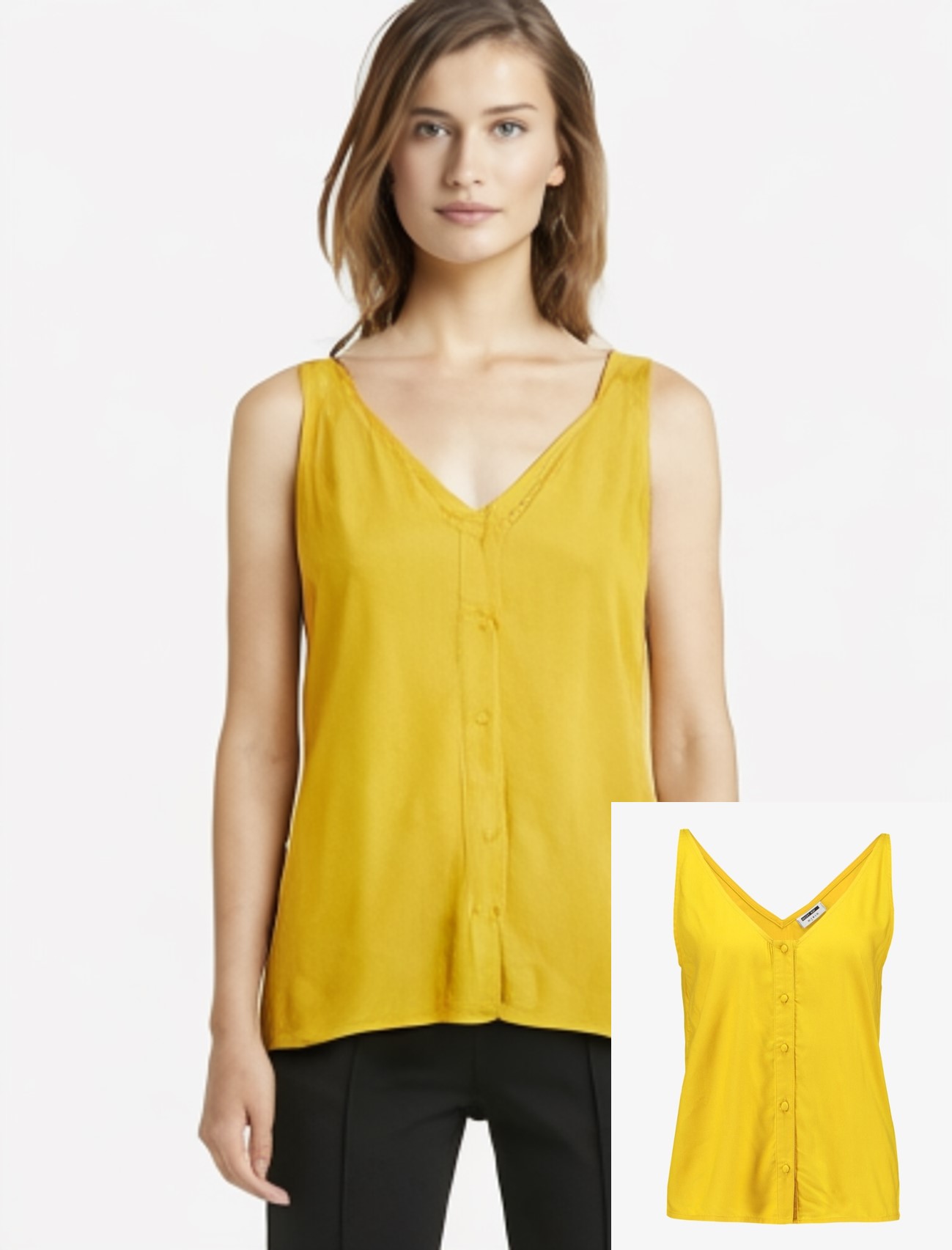} &
\includegraphics[width=0.16\linewidth]{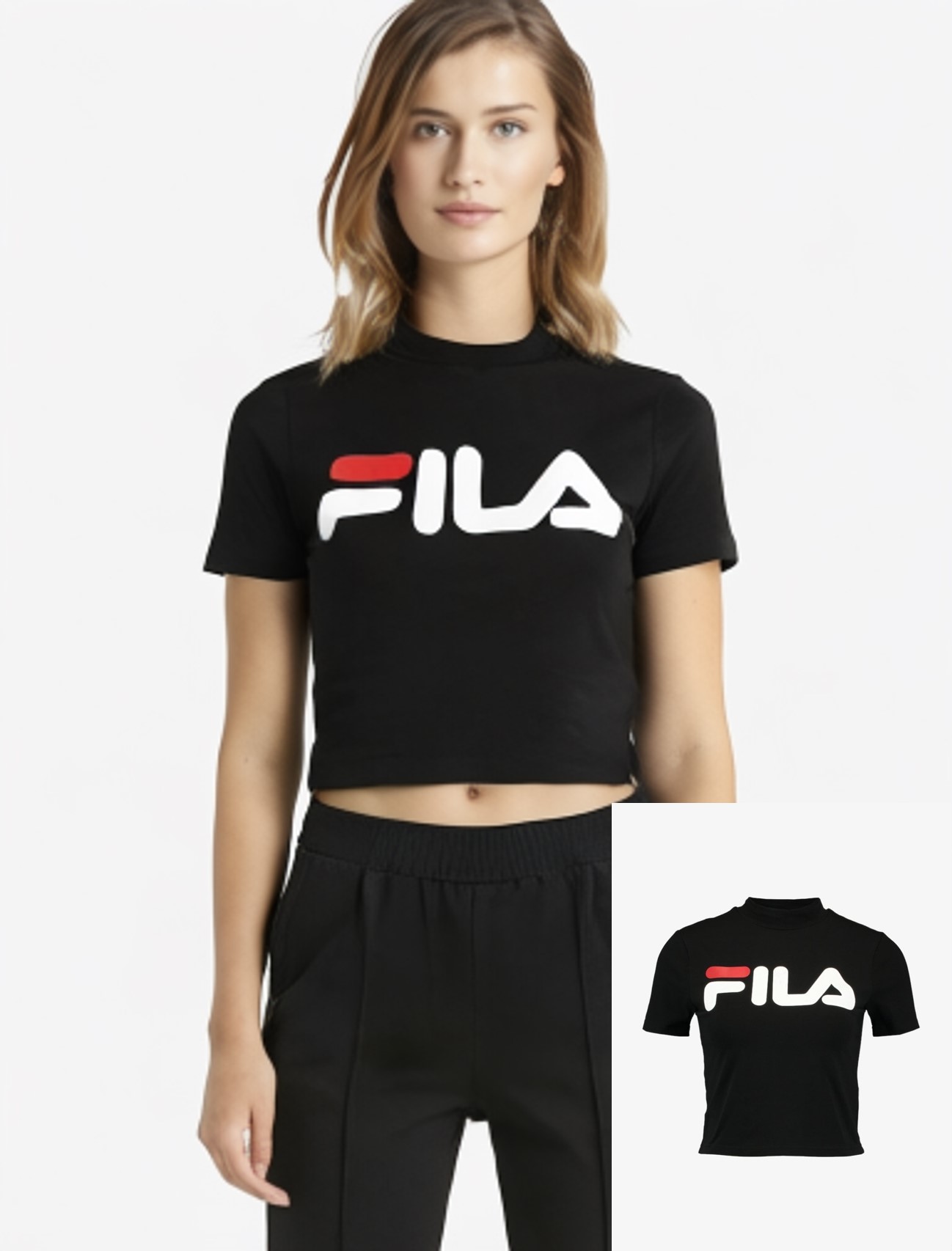} \\

\includegraphics[width=0.16\linewidth]{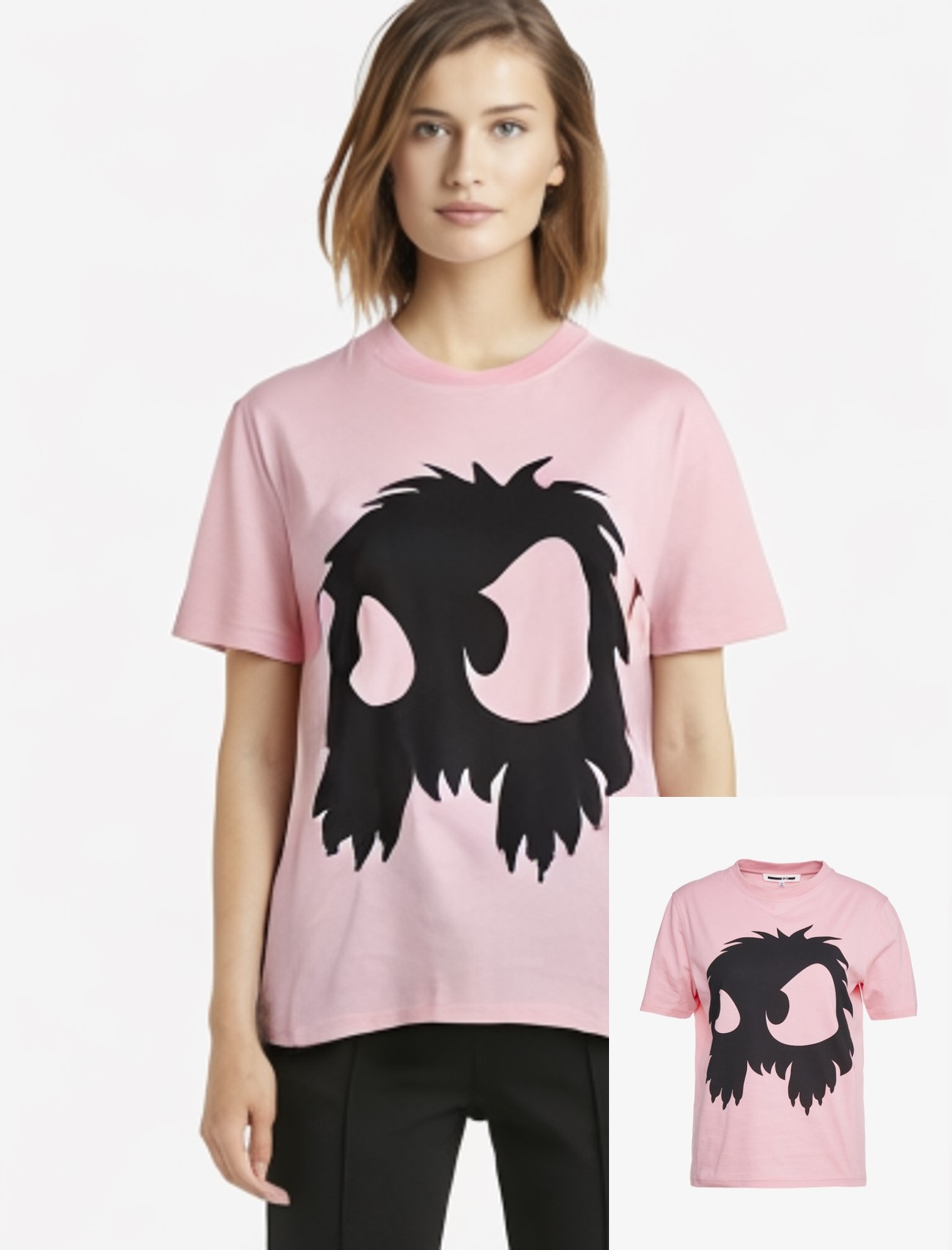} & 
\includegraphics[width=0.16\linewidth]{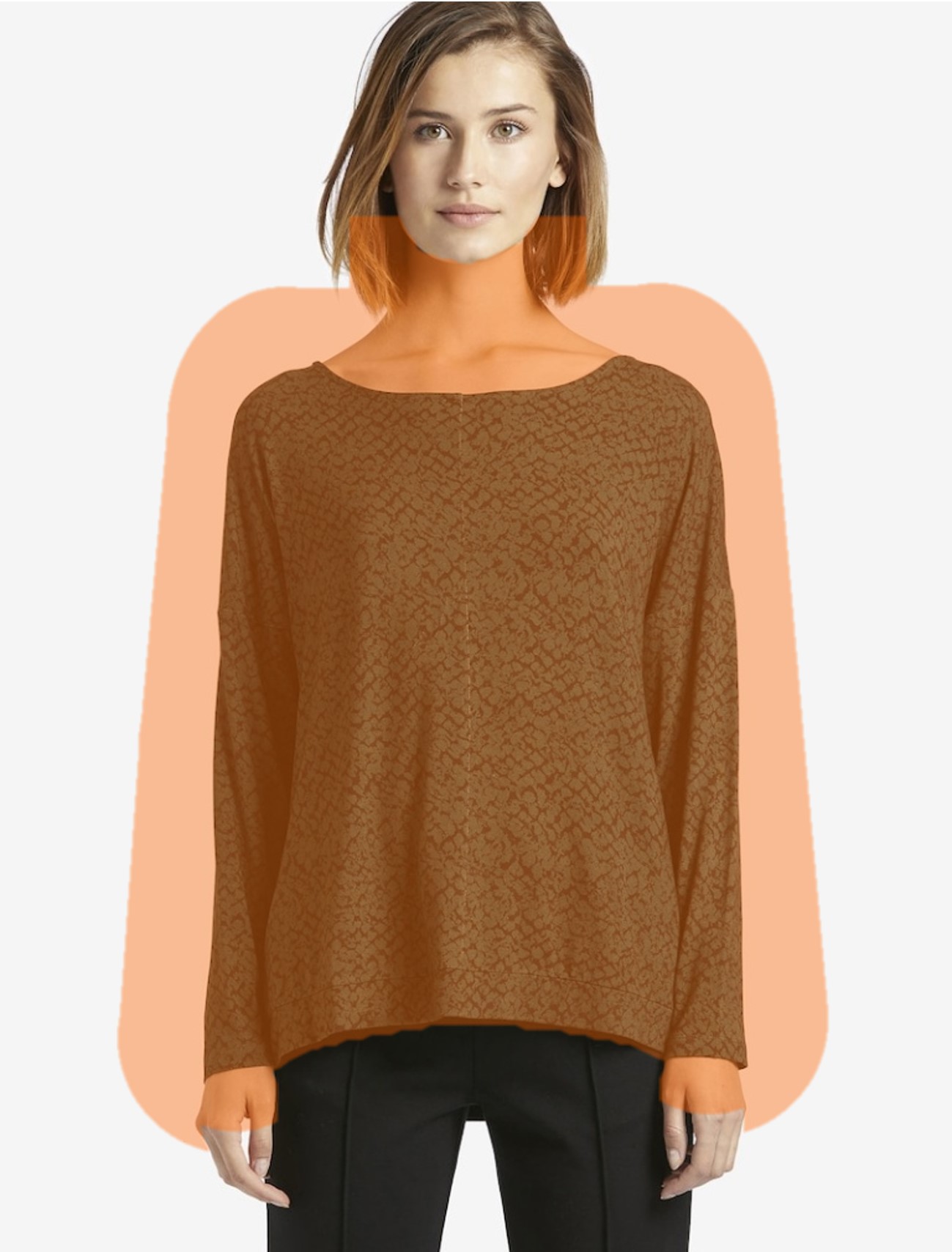} &
\includegraphics[width=0.16\linewidth]{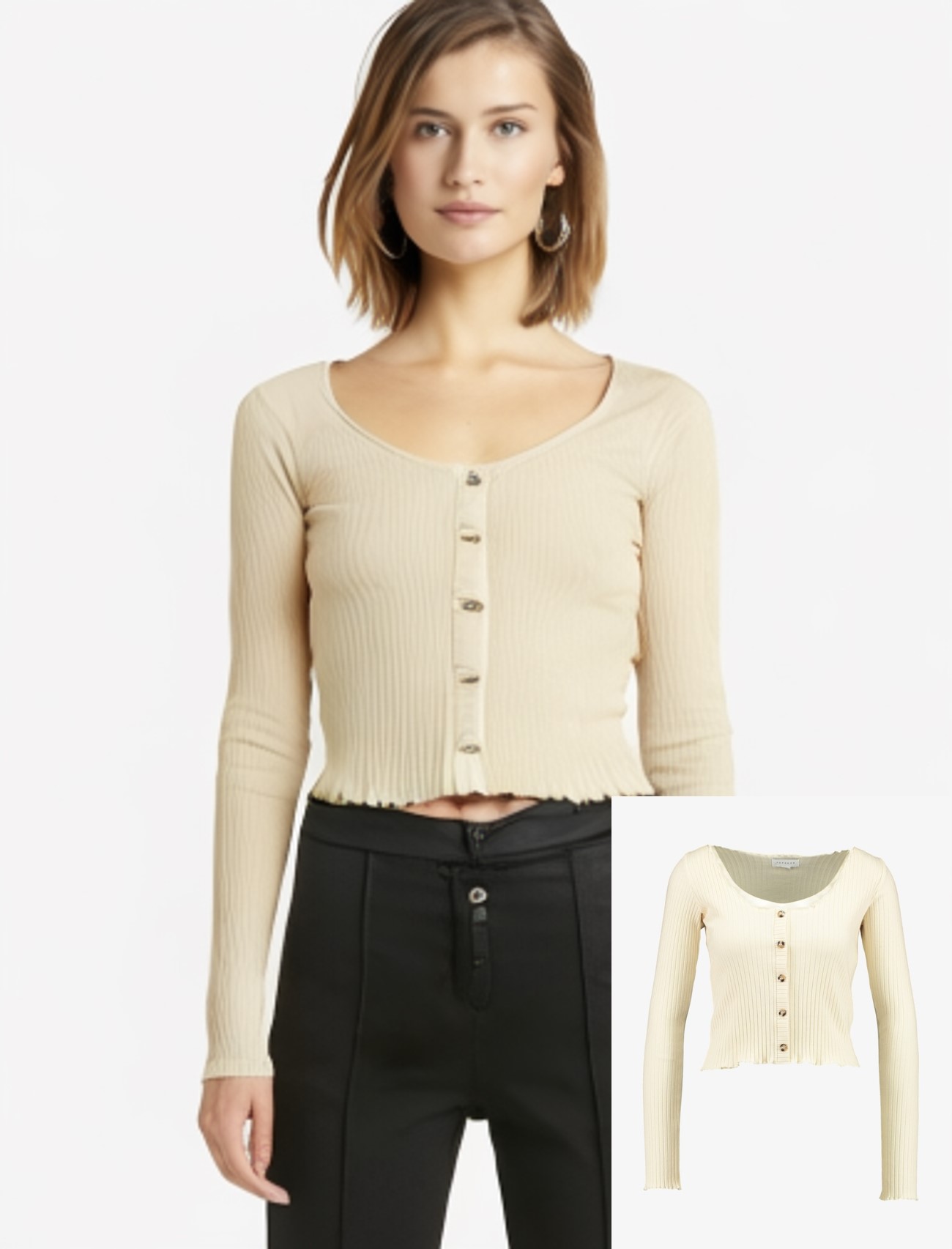} \\

\includegraphics[width=0.16\linewidth]{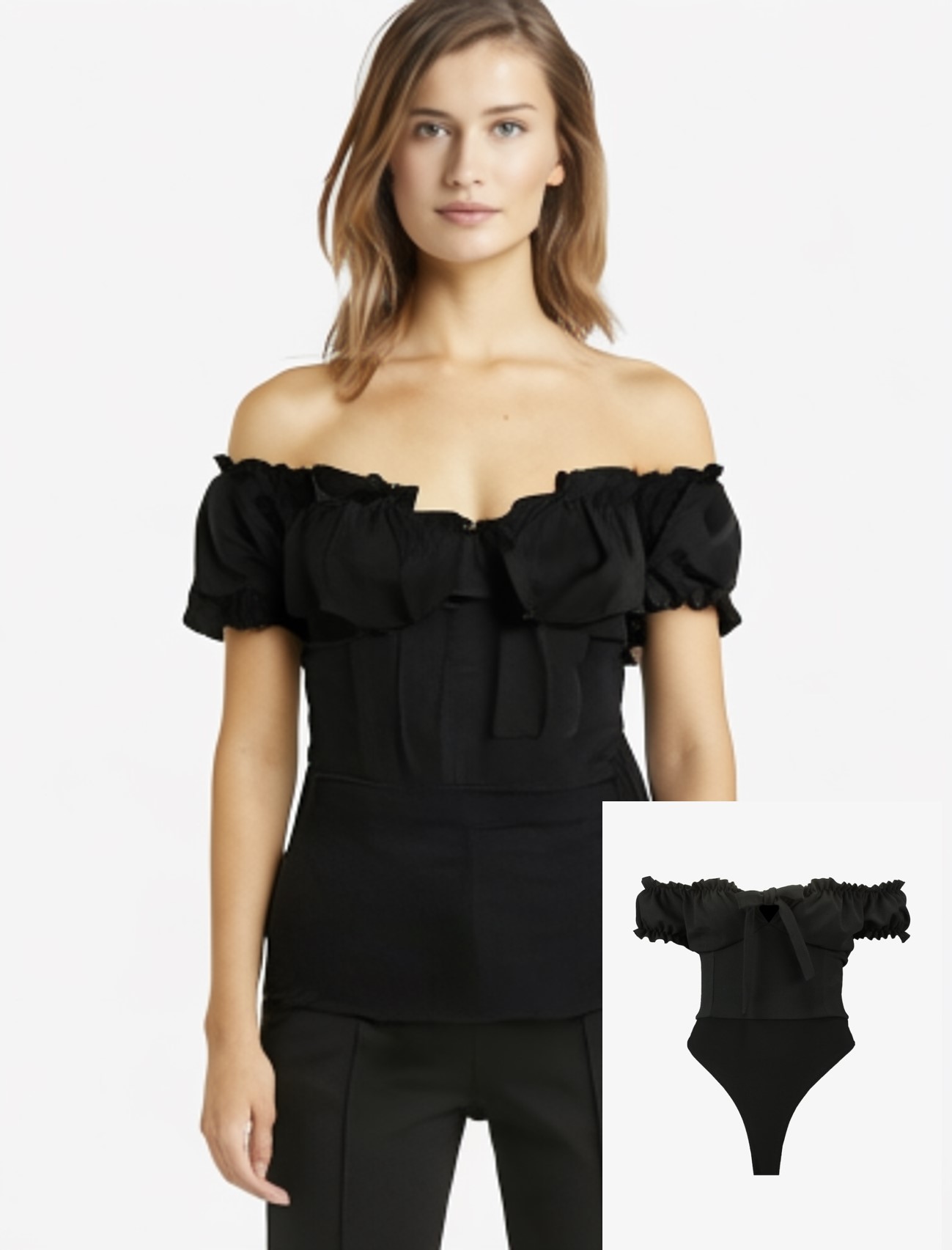} & 
\includegraphics[width=0.16\linewidth]{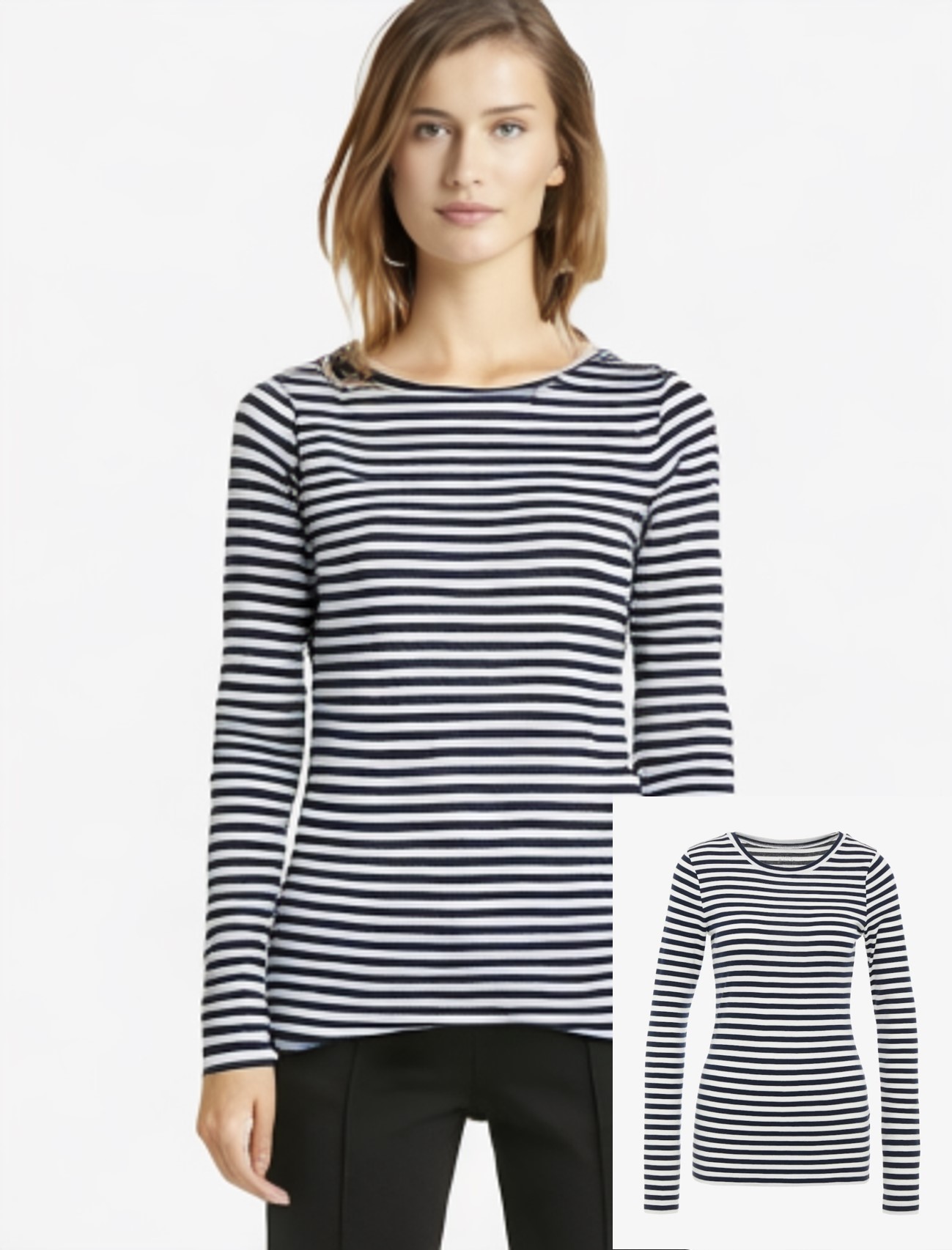} &
\includegraphics[width=0.16\linewidth]{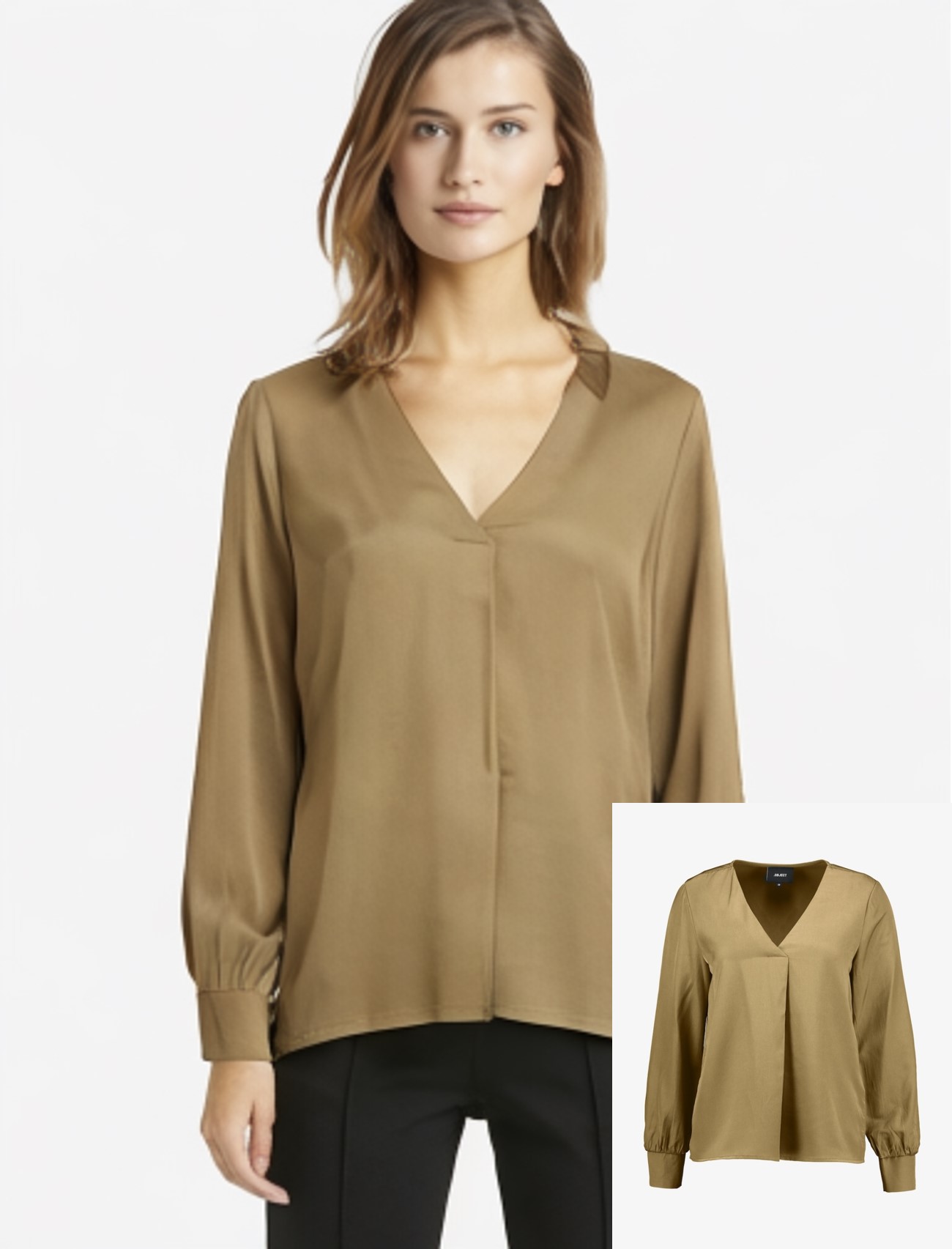} \\

\end{tabular}
\caption{Our method achieves realistic visual effects when trying on various types of clothing.}
\label{fig:more_results}
\end{figure}

\subsection{User Study}
\label{sec:metric}
We conducted a survey with 40 human users using the results generated on the Cross-27. Among the try-on results of the seven methods (in Fig. \ref{fig:metric_benchmark2}), we randomly selected the try-on results of three methods each time. To reflect the relative performance of seven methods, users were asked to choose the "most realistic" results among three randomly selected samples instead of seven. If making one choice from seven methods, we probably get one or two recent methods with higher scores while the others with poor performance. Each user completed 50 such choices. We calculate the user preference rate (\%) of each method as the proportion of times that it was chosen as "most realistic" out of the total number of choices. A higher percentage indicates that users consider the results generated by that method to be more realistic.

Since the scores of the metrics are negatively correlated with the quality of the try-on, for a more convenient comparison, we take the negative of the user preference rate to obtain the "Negative User Preference Rate." As shown by the red line in Fig. \ref{fig:metric_benchmark2}, our method achieved the lowest negative user preference rate, which means the highest user preference rate, among all methods. In addition, when comparing the scores of FID, KID, S-LPIPS, and SDR metrics with the negative user preference rate, the relationship of superiority and inferiority among methods reflected by FID and KID shows a considerable deviation from the user preference. In contrast, our proposed S-LPIPS and SDR metrics are more consistent with the user preference, as shown in Fig. \ref{fig:metric_SLPIPS} and Fig. \ref{fig:metric_SDR}. SDR is designed to reflect the extent of changing clothing type via try-on, with amplification for poor results. However, user evaluation assigns the same weight for each testing case. Therefore, compared with user study, the performance of seven methods on SDR experiences fluctuation, but shares a similar trend (e.g., the high score of user study becomes much higher for SDR at DCI-VTON and StableVITON while the low score of user study becomes much lower for SDR at GP-VTON and Ours.

\subsection{More Visual Results}
\label{sec:results}
We present more comparison results in Fig. \ref{fig:comparison_1} and Fig. \ref{fig:comparison_2}, where our method shows significant improvements in preserving the type of clothing compared to existing work. Our method is well compatible with trying on clothing of different types, as shown in Fig. \ref{fig:more_results}.

\subsection{Limitations \& Discussion}
\label{sec:limits}
\begin{figure}[t]
\centering
\scriptsize
\setlength{\tabcolsep}{.2em}
\begin{tabular}{c c cc}

\includegraphics[width=0.16\linewidth]{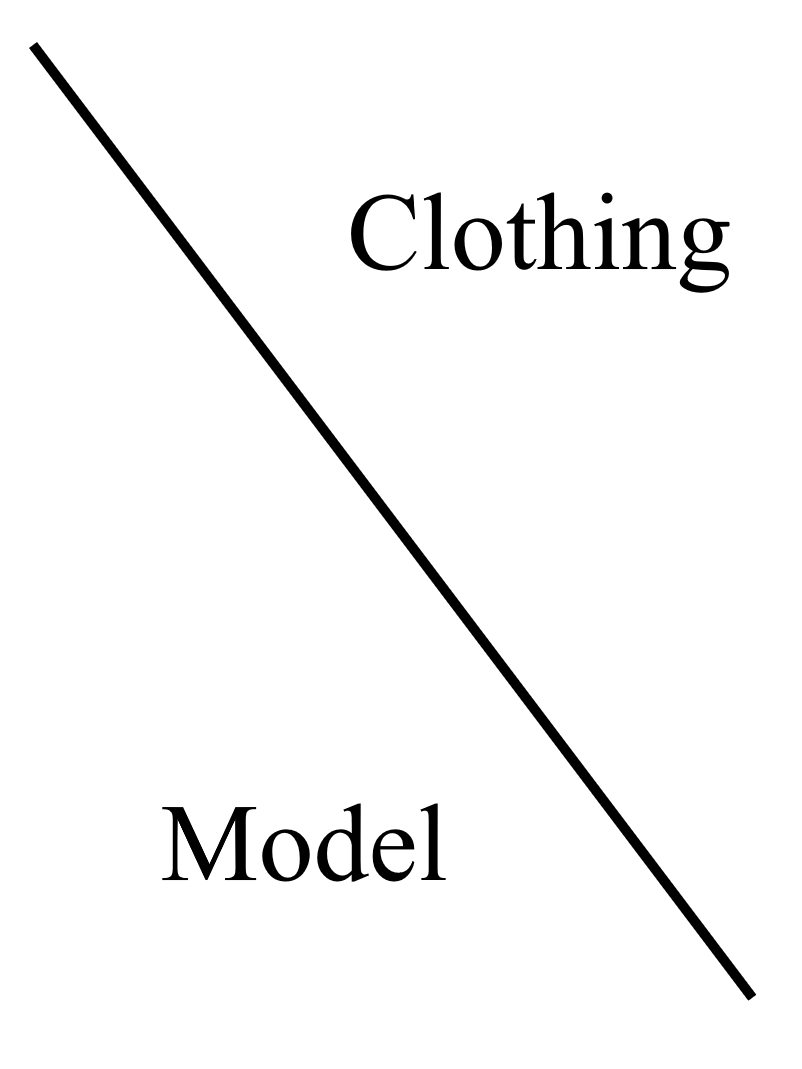} & &
\includegraphics[width=0.16\linewidth]{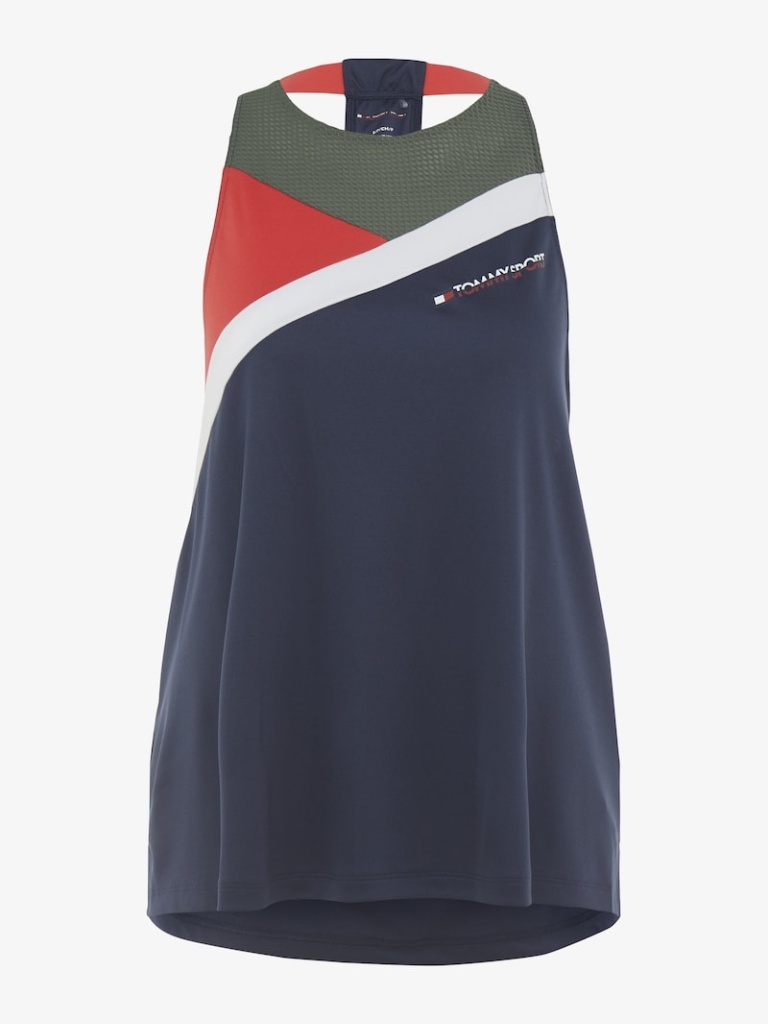} &
\includegraphics[width=0.16\linewidth]{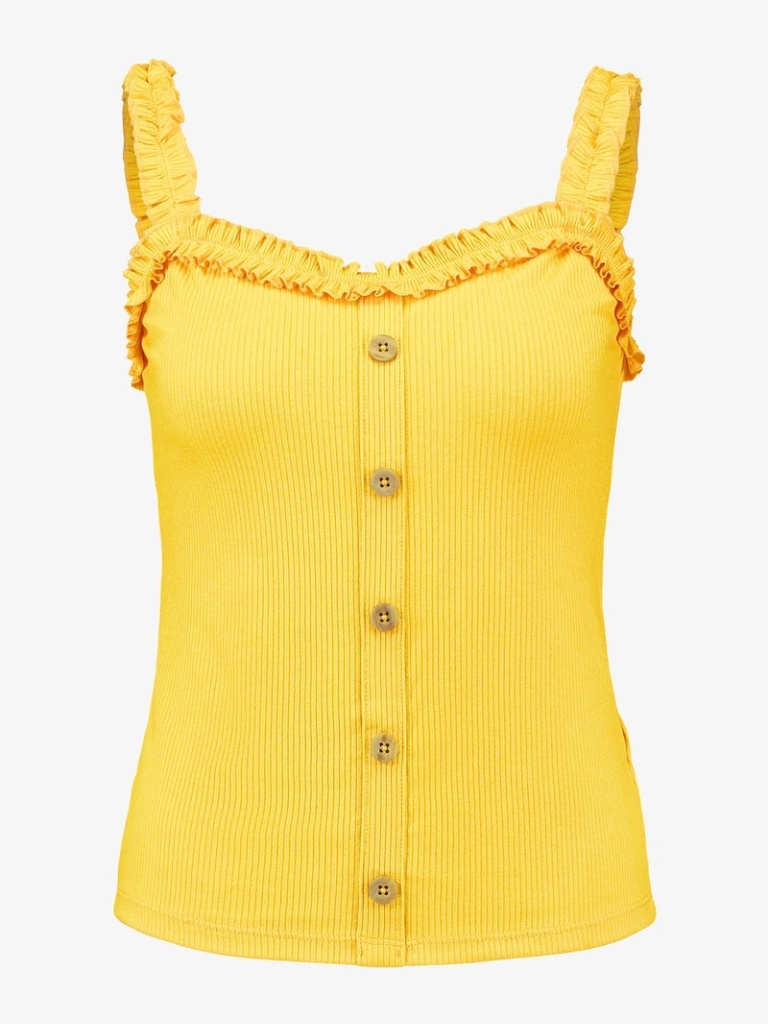} \\

\includegraphics[width=0.16\linewidth]{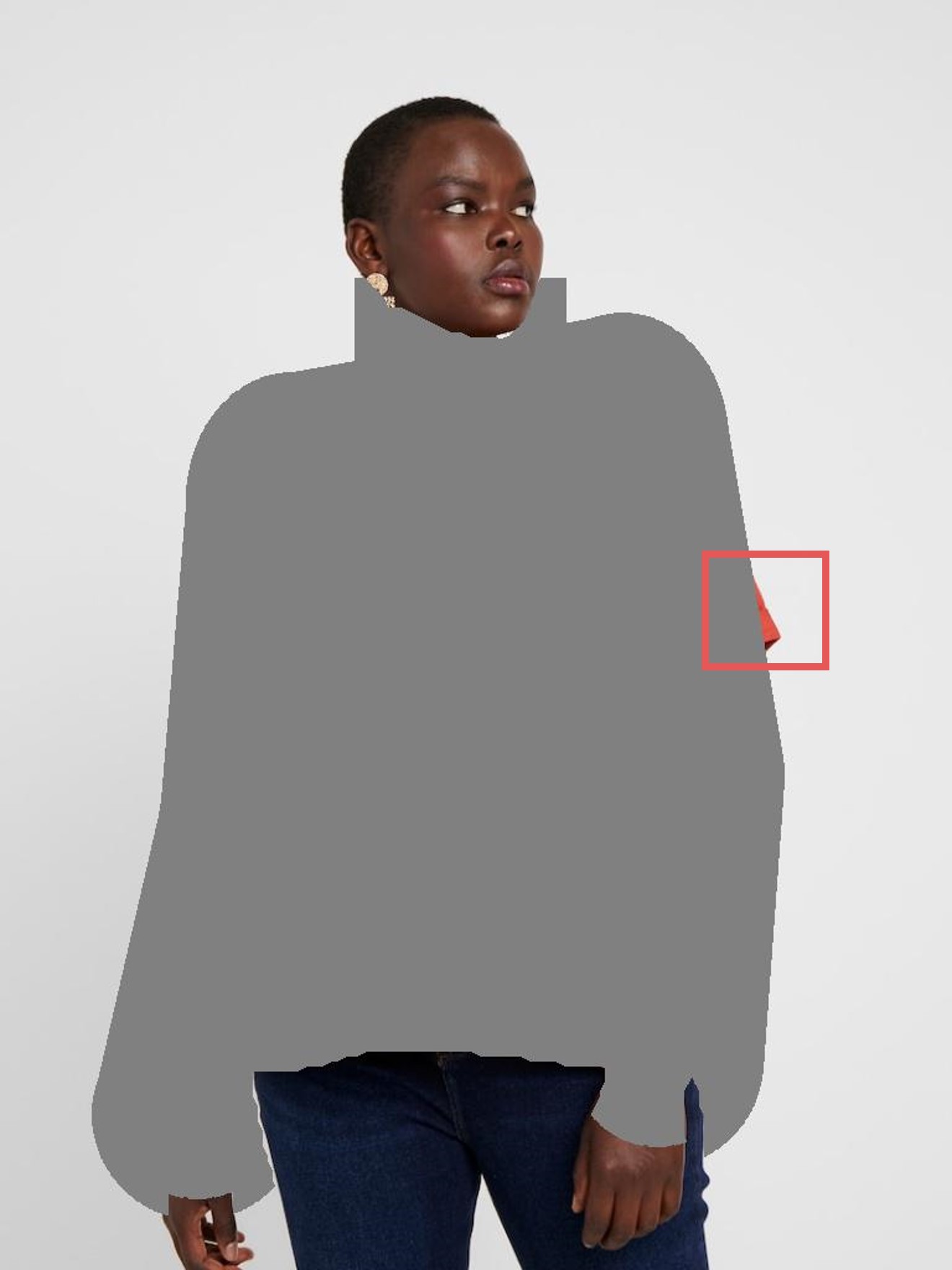} & &
\includegraphics[width=0.16\linewidth]{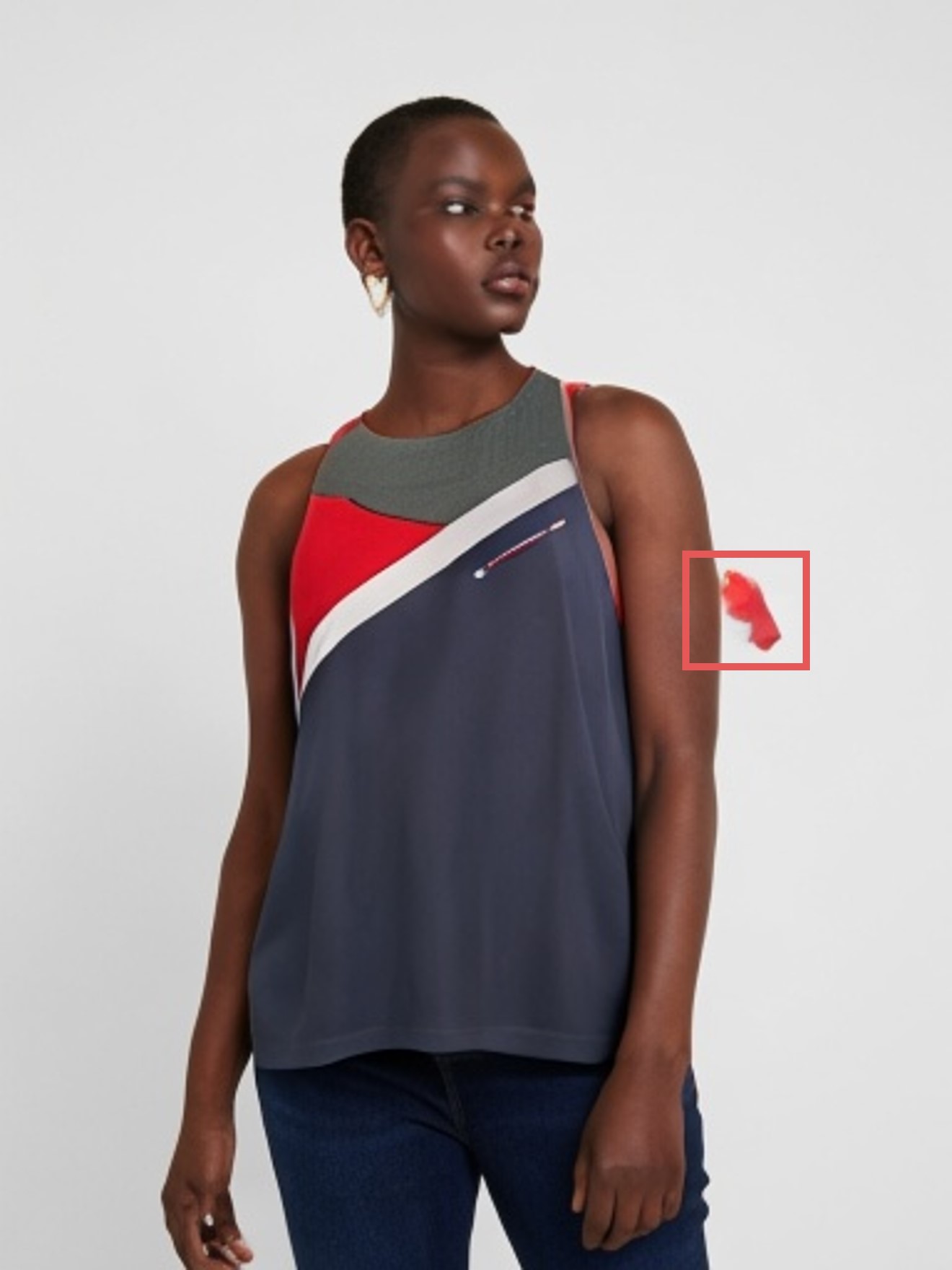} &
\includegraphics[width=0.16\linewidth]{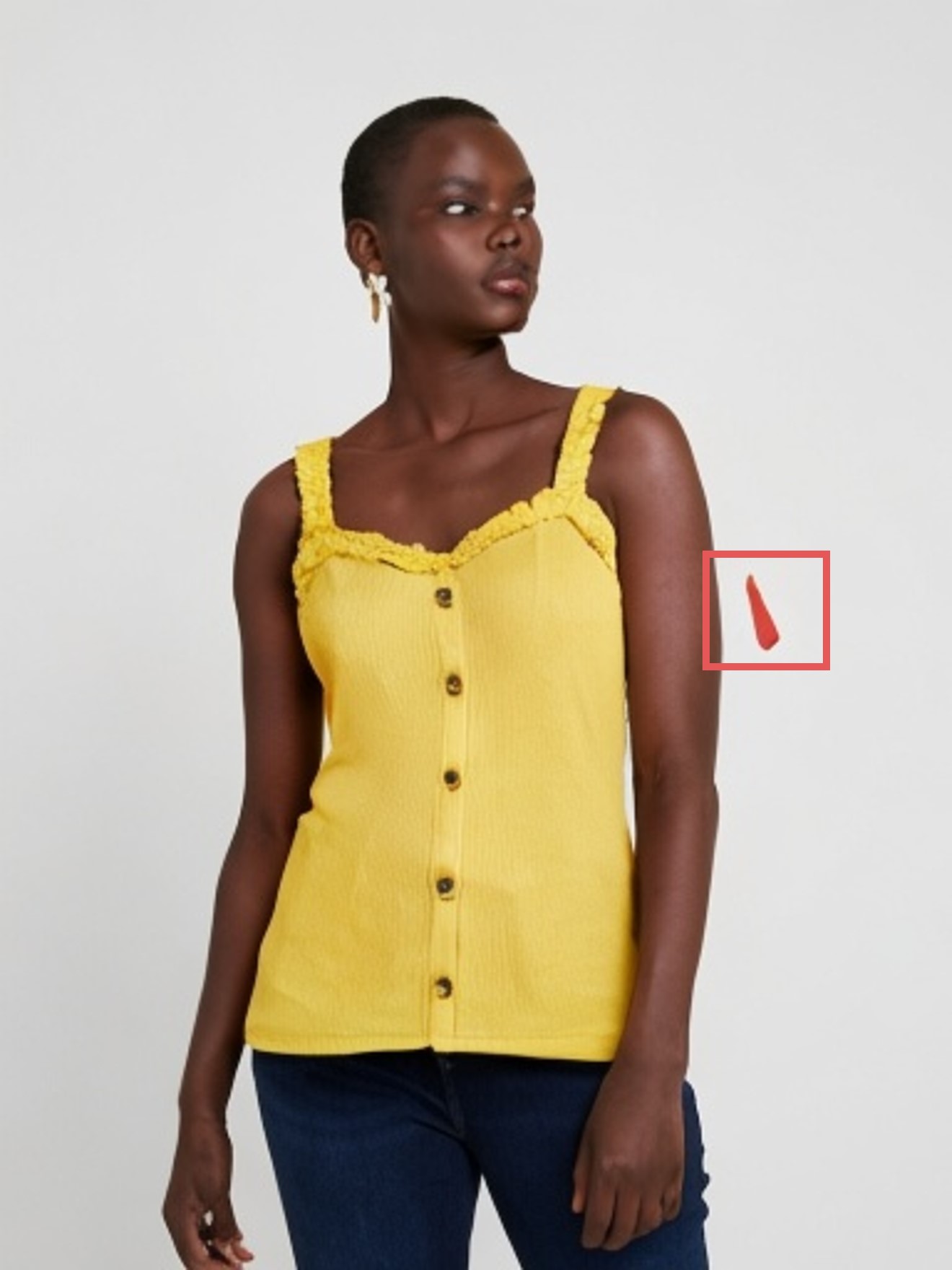} \\

\includegraphics[width=0.16\linewidth]{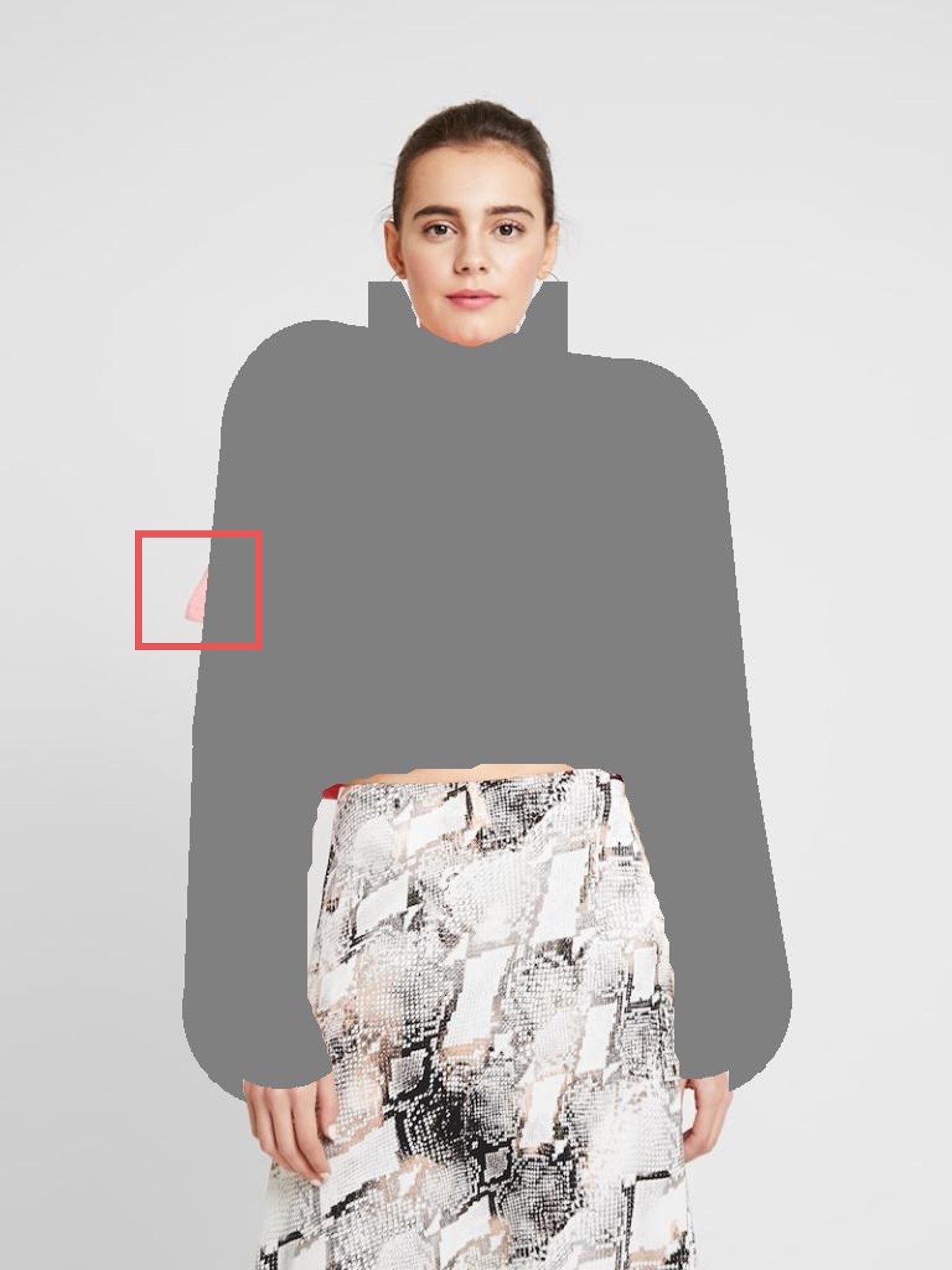} & &
\includegraphics[width=0.16\linewidth]{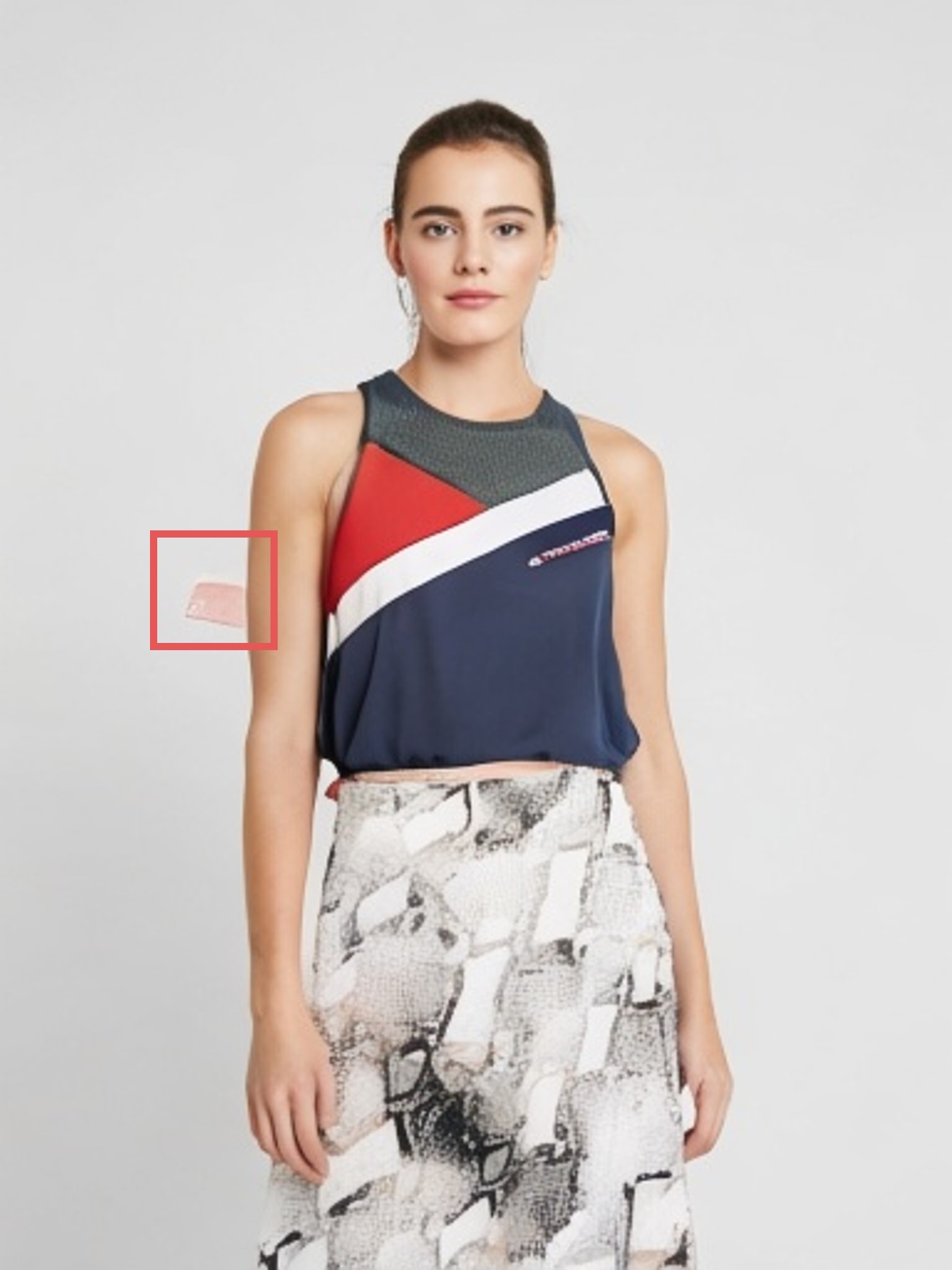} &
\includegraphics[width=0.16\linewidth]{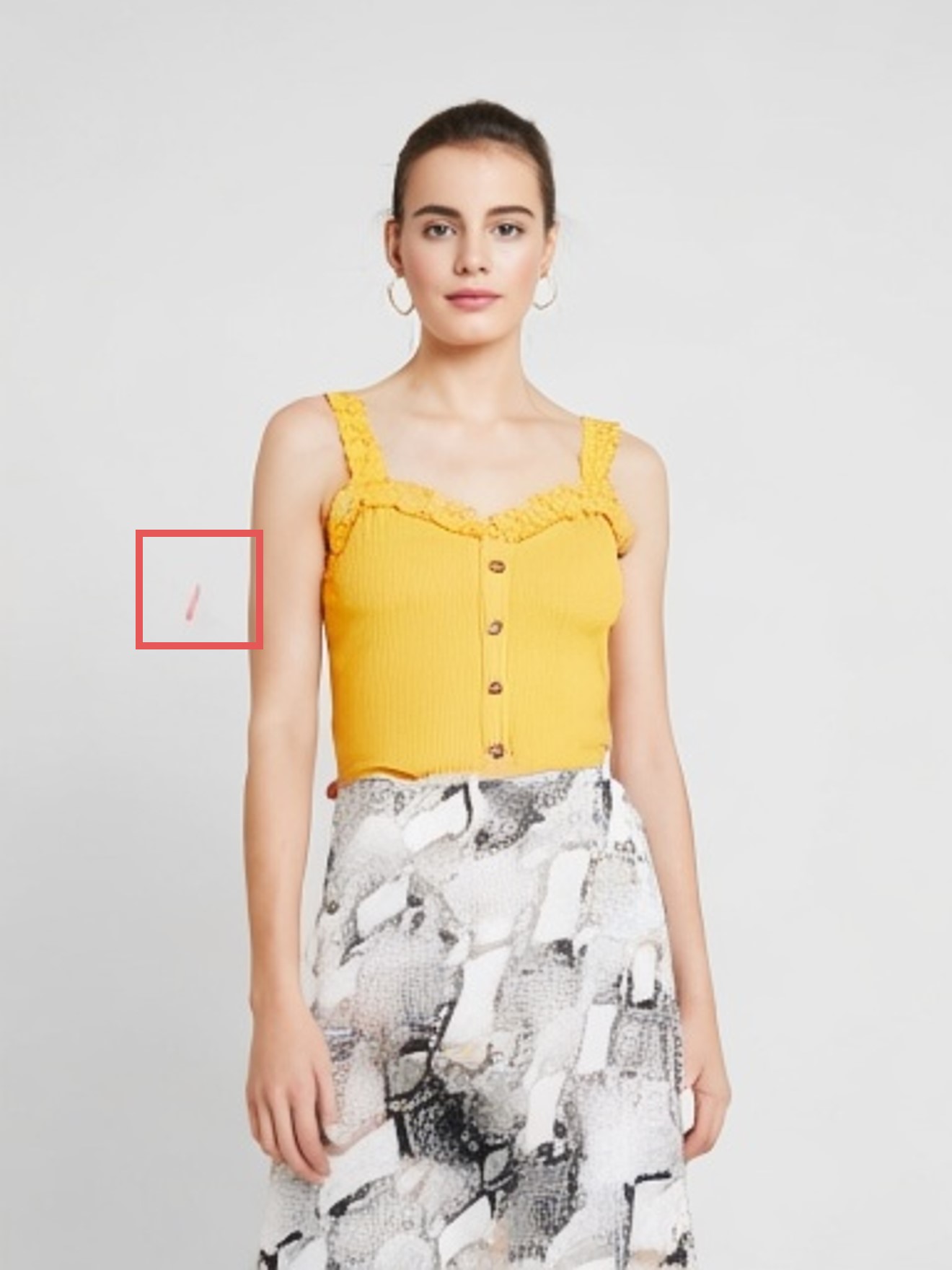} \\

\end{tabular}
\caption{The model's original clothing is too cumbersome to be completely masked, resulting in undesired remaining in the generated images.}
\label{fig:limitation}
\end{figure}

Although our method can well maintain the type of the clothing, when the model's original clothing is relatively cumbersome (covering a larger area), our mask cannot completely eliminate the original clothing. These residual clothing pixels will appear as flaws in the final generated results, as shown within the red box in Fig. \ref{fig:limitation}. In our subsequent work, we will further explore more refined methods of mask creation to eliminate all original clothing information while preserving the non-try-on parts.

\begin{figure}[t]
  \centering
  \includegraphics[width=1\linewidth]{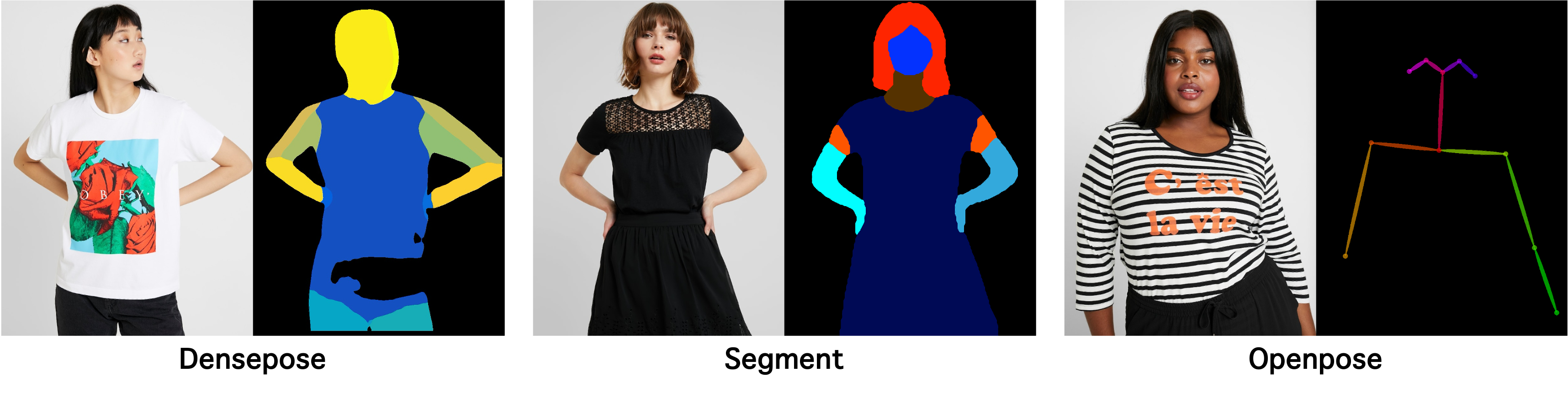}
   \caption{Limitation: parsing flaws}
   \label{fig:limitation_reb}
\end{figure}

The proposed evaluation metrics also rely on parsing. We give examples of parsing flaws in Fig. \ref{fig:limitation_reb}, where Densepose contains holes due to the color of background/foreground, cloth segment is too long due to the same color of top/bottom garment, and pose detector misses some keypoints. As failed parsing cases occupy a very small portion, the training is not affected with abundant training  samples. With the help of corrections in our implementation, e.g., 1) Densepose and semantic segmentation cooperate with each other to fill holes and 2) keypoints fitering as illustrated in Sec. \ref{sec:metric_LPIPS}, the limitations on failed parsing can be alleviated, but not totally overcome.

\section{Conclusions}
\textcolor{black}{In this paper, we focus on the unpaired try-on task in virtual try-on, where we introduce an adaptive mask training paradigm that effectively addresses the flaw of training in existing methods. Furthermore, to fill the lacks in benchmarks and evaluation metrics in the unpaired try-on, we propose the Cross-27 and the SDR, S-LPIPS metrics, respectively. Experiments demonstrate that Cross-27 can test the performance of try-on methods more comprehensively than existing test set. Additionally, through \textit{Incorrect sample} mixing experiments, we have demonstrated the superiority of the SDR and S-LPIPS metrics compared to the existing FID and KID metrics. The limitations of our work will be further analyzed in the supplement.}

\bibliographystyle{ieeetr}
\bibliography{main}

\begin{thebibliography}{10}

\bibitem{fashion_customization}
D.~Song, J.~Zeng, M.~Liu, X.~Li, and A.~Liu, ``Fashion customization: Image generation based on editing clue,'' {\em {IEEE} Trans. Circuits Syst. Video Technol.}, vol.~34, no.~6, pp.~4434--4444, 2024.

\bibitem{fashion_retrieval}
H.~Su, P.~Wang, L.~Liu, H.~Li, Z.~Li, and Y.~Zhang, ``Where to look and how to describe: Fashion image retrieval with an attentional heterogeneous bilinear network,'' {\em {IEEE} Trans. Circuits Syst. Video Technol.}, vol.~31, no.~8, pp.~3254--3265, 2021.

\bibitem{Viton}
X.~Han, Z.~Wu, Z.~Wu, R.~Yu, and L.~S. Davis, ``Viton: An image-based virtual try-on network,'' in {\em Proceedings of the IEEE conference on computer vision and pattern recognition}, pp.~7543--7552, 2018.

\bibitem{CP-VTON}
B.~Wang, H.~Zheng, X.~Liang, Y.~Chen, L.~Lin, and M.~Yang, ``Toward characteristic-preserving image-based virtual try-on network,'' in {\em Proceedings of the European conference on computer vision (ECCV)}, pp.~589--604, 2018.

\bibitem{CP-VTON+}
M.~R. Minar, T.~T. Tuan, H.~Ahn, P.~Rosin, and Y.-K. Lai, ``Cp-vton+: Clothing shape and texture preserving image-based virtual try-on,'' in {\em CVPR Workshops}, vol.~3, pp.~10--14, 2020.

\bibitem{ACGPN}
H.~Yang, R.~Zhang, X.~Guo, W.~Liu, W.~Zuo, and P.~Luo, ``Towards photo-realistic virtual try-on by adaptively generating-preserving image content,'' in {\em Proceedings of the IEEE/CVF conference on computer vision and pattern recognition}, pp.~7850--7859, 2020.

\bibitem{PFAFN}
Y.~Ge, Y.~Song, R.~Zhang, C.~Ge, W.~Liu, and P.~Luo, ``Parser-free virtual try-on via distilling appearance flows,'' in {\em Proceedings of the IEEE/CVF conference on computer vision and pattern recognition}, pp.~8485--8493, 2021.

\bibitem{VITON-HD}
S.~Choi, S.~Park, M.~Lee, and J.~Choo, ``Viton-hd: High-resolution virtual try-on via misalignment-aware normalization,'' in {\em Proceedings of the IEEE/CVF conference on computer vision and pattern recognition}, pp.~14131--14140, 2021.

\bibitem{HR-VITON}
S.~Lee, G.~Gu, S.~Park, S.~Choi, and J.~Choo, ``High-resolution virtual try-on with misalignment and occlusion-handled conditions,'' {\em arXiv preprint arXiv:2206.14180}, 2022.

\bibitem{GP-VTON}
X.~Zhenyu, H.~Zaiyu, D.~Xin, Z.~Fuwei, D.~Haoye, Z.~Xijin, Z.~Feida, and L.~Xiaodan, ``Gp-vton: Towards general purpose virtual try-on via collaborative local-flow global-parsing learning,'' in {\em Proceedings of the IEEE/CVF Conference on Computer Vision and Pattern Recognition (CVPR)}, June 2023.

\bibitem{StableVITON}
J.~Kim, G.~Gu, M.~Park, S.~Park, and J.~Choo, ``Stableviton: Learning semantic correspondence with latent diffusion model for virtual try-on,'' {\em CoRR}, vol.~abs/2312.01725, 2023.

\bibitem{DDIM}
J.~Song, C.~Meng, and S.~Ermon, ``Denoising diffusion implicit models,'' in {\em International Conference on Learning Representations}, 2021.

\bibitem{DDPM}
J.~Ho, A.~Jain, and P.~Abbeel, ``Denoising diffusion probabilistic models,'' {\em Advances in neural information processing systems}, vol.~33, pp.~6840--6851, 2020.

\bibitem{IDDPM}
A.~Q. Nichol and P.~Dhariwal, ``Improved denoising diffusion probabilistic models,'' in {\em Proceedings of the 38th International Conference on Machine Learning, {ICML} 2021, 18-24 July 2021, Virtual Event} (M.~Meila and T.~Zhang, eds.), vol.~139 of {\em Proceedings of Machine Learning Research}, pp.~8162--8171, {PMLR}, 2021.

\bibitem{LDM}
R.~Rombach, A.~Blattmann, D.~Lorenz, P.~Esser, and B.~Ommer, ``High-resolution image synthesis with latent diffusion models,'' in {\em Proceedings of the IEEE/CVF conference on computer vision and pattern recognition}, pp.~10684--10695, 2022.

\bibitem{LaDI-VTON}
D.~Morelli, A.~Baldrati, G.~Cartella, M.~Cornia, M.~Bertini, and R.~Cucchiara, ``Ladi-vton: Latent diffusion textual-inversion enhanced virtual try-on,'' {\em arXiv preprint arXiv:2305.13501}, 2023.

\bibitem{DCI-VTON}
J.~Gou, S.~Sun, J.~Zhang, J.~Si, C.~Qian, and L.~Zhang, ``Taming the power of diffusion models for high-quality virtual try-on with appearance flow,'' {\em arXiv preprint arXiv:2308.06101}, 2023.

\bibitem{MGD}
A.~Baldrati, D.~Morelli, G.~Cartella, M.~Cornia, M.~Bertini, and R.~Cucchiara, ``Multimodal garment designer: Human-centric latent diffusion models for fashion image editing,'' {\em arXiv preprint arXiv:2304.02051}, 2023.

\bibitem{DressCode}
D.~Morelli, M.~Fincato, M.~Cornia, F.~Landi, F.~Cesari, and R.~Cucchiara, ``Dress code: High-resolution multi-category virtual try-on,'' in {\em Proceedings of the IEEE/CVF Conference on Computer Vision and Pattern Recognition}, pp.~2231--2235, 2022.

\bibitem{FID}
M.~Heusel, H.~Ramsauer, T.~Unterthiner, B.~Nessler, and S.~Hochreiter, ``Gans trained by a two time-scale update rule converge to a local nash equilibrium,'' {\em Advances in neural information processing systems}, vol.~30, 2017.

\bibitem{CleanFID}
G.~Parmar, R.~Zhang, and J.-Y. Zhu, ``On aliased resizing and surprising subtleties in gan evaluation,'' in {\em CVPR}, 2022.

\bibitem{KID}
M.~Bińkowski, D.~J. Sutherland, M.~Arbel, and A.~Gretton, ``Demystifying {MMD} {GAN}s,'' in {\em International Conference on Learning Representations}, 2018.

\bibitem{SSIM}
Z.~Wang, A.~C. Bovik, H.~R. Sheikh, and E.~P. Simoncelli, ``Image quality assessment: from error visibility to structural similarity,'' {\em IEEE transactions on image processing}, vol.~13, no.~4, pp.~600--612, 2004.

\bibitem{LPIPS}
R.~Zhang, P.~Isola, A.~A. Efros, E.~Shechtman, and O.~Wang, ``The unreasonable effectiveness of deep features as a perceptual metric,'' in {\em CVPR}, 2018.

\bibitem{text_to_image_syntheis}
J.~Liang, W.~Pei, and F.~Lu, ``Layout-bridging text-to-image synthesis,'' {\em {IEEE} Trans. Circuits Syst. Video Technol.}, vol.~33, no.~12, pp.~7438--7451, 2023.

\bibitem{pose_guided_image_generation}
P.~Zhang, L.~Yang, X.~Xie, and J.~Lai, ``Lightweight texture correlation network for pose guided person image generation,'' {\em {IEEE} Trans. Circuits Syst. Video Technol.}, vol.~32, no.~7, pp.~4584--4598, 2022.

\bibitem{style-flow}
S.~He, Y.~Song, and T.~Xiang, ``Style-based global appearance flow for virtual try-on,'' in {\em {IEEE/CVF} Conference on Computer Vision and Pattern Recognition, {CVPR} 2022, New Orleans, LA, USA, June 18-24, 2022}, pp.~3460--3469, {IEEE}, 2022.

\bibitem{GANs}
M.~M. B. X.-D. Warde and F.~S. O. A.~C. Ian, ``J. goodfellow, jean pouget-abadie and yoshua bengio. generative adversarial nets,'' 2014.

\bibitem{FlowNet}
Y.~Li, C.~Huang, and C.~C. Loy, ``Dense intrinsic appearance flow for human pose transfer,'' in {\em Proceedings of the IEEE/CVF Conference on Computer Vision and Pattern Recognition}, pp.~3693--3702, 2019.

\bibitem{TPS}
J.~Duchon, ``Splines minimizing rotation-invariant semi-norms in sobolev spaces,'' in {\em Constructive Theory of Functions of Several Variables: Proceedings of a Conference Held at Oberwolfach April 25--May 1, 1976}, pp.~85--100, Springer, 1977.

\bibitem{STN}
M.~Jaderberg, K.~Simonyan, A.~Zisserman, {\em et~al.}, ``Spatial transformer networks,'' {\em Advances in neural information processing systems}, vol.~28, 2015.

\bibitem{beats_gan}
P.~Dhariwal and A.~Nichol, ``Diffusion models beat gans on image synthesis,'' {\em Advances in neural information processing systems}, vol.~34, pp.~8780--8794, 2021.

\bibitem{PBE}
B.~Yang, S.~Gu, B.~Zhang, T.~Zhang, X.~Chen, X.~Sun, D.~Chen, and F.~Wen, ``Paint by example: Exemplar-based image editing with diffusion models,'' in {\em Proceedings of the IEEE/CVF Conference on Computer Vision and Pattern Recognition}, pp.~18381--18391, 2023.

\bibitem{TryonDiffusion}
L.~Zhu, D.~Yang, T.~Zhu, F.~Reda, W.~Chan, C.~Saharia, M.~Norouzi, and I.~Kemelmacher{-}Shlizerman, ``Tryondiffusion: {A} tale of two unets,'' in {\em {IEEE/CVF} Conference on Computer Vision and Pattern Recognition, {CVPR} 2023, Vancouver, BC, Canada, June 17-24, 2023}, pp.~4606--4615, {IEEE}, 2023.

\bibitem{Dior}
A.~Cui, D.~McKee, and S.~Lazebnik, ``Dressing in order: Recurrent person image generation for pose transfer, virtual try-on and outfit editing,'' in {\em 2021 {IEEE/CVF} International Conference on Computer Vision, {ICCV} 2021, Montreal, QC, Canada, October 10-17, 2021}, pp.~14618--14627, {IEEE}, 2021.

\bibitem{IS}
T.~Hinz, M.~Fisher, O.~Wang, and S.~Wermter, ``Improved techniques for training single-image gans,'' in {\em {IEEE} Winter Conference on Applications of Computer Vision, {WACV} 2021, Waikoloa, HI, USA, January 3-8, 2021}, pp.~1299--1308, {IEEE}, 2021.

\bibitem{OpenPose}
Z.~Cao, T.~Simon, S.-E. Wei, and Y.~Sheikh, ``Realtime multi-person 2d pose estimation using part affinity fields,'' in {\em Proceedings of the IEEE conference on computer vision and pattern recognition}, pp.~7291--7299, 2017.

\bibitem{CIHP}
K.~Gong, X.~Liang, Y.~Li, Y.~Chen, M.~Yang, and L.~Lin, ``Instance-level human parsing via part grouping network,'' in {\em Computer Vision - {ECCV} 2018 - 15th European Conference, Munich, Germany, September 8-14, 2018, Proceedings, Part {IV}} (V.~Ferrari, M.~Hebert, C.~Sminchisescu, and Y.~Weiss, eds.), vol.~11208 of {\em Lecture Notes in Computer Science}, pp.~805--822, Springer, 2018.

\bibitem{DensePose}
R.~A. G{\"u}ler, N.~Neverova, and I.~Kokkinos, ``Densepose: Dense human pose estimation in the wild,'' in {\em Proceedings of the IEEE conference on computer vision and pattern recognition}, pp.~7297--7306, 2018.

\bibitem{VGG}
K.~Simonyan and A.~Zisserman, ``Very deep convolutional networks for large-scale image recognition,'' in {\em 3rd International Conference on Learning Representations, {ICLR} 2015, San Diego, CA, USA, May 7-9, 2015, Conference Track Proceedings} (Y.~Bengio and Y.~LeCun, eds.), 2015.

\end{thebibliography}

\begin{IEEEbiography}[{\includegraphics[width=1in,height=1.25in,clip,keepaspectratio]{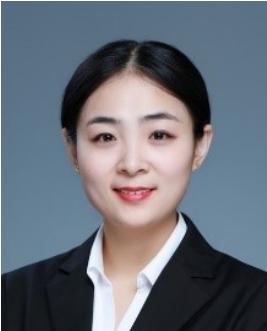}}]{Dan Song} received the Ph.D. degree in computer science and technology from Zhejiang University, China, in 2018. She was an academic visitor at the National Centre for Computer Animation (NCCA), United Kingdom. She is currently an Associate Professor with the School of Electrical and Information Engineering, Tianjin University. Her research interests include virtual try-on and multimedia information processing.
\end{IEEEbiography}

\begin{IEEEbiography}[{\includegraphics[width=1in,height=1.25in,clip,keepaspectratio]{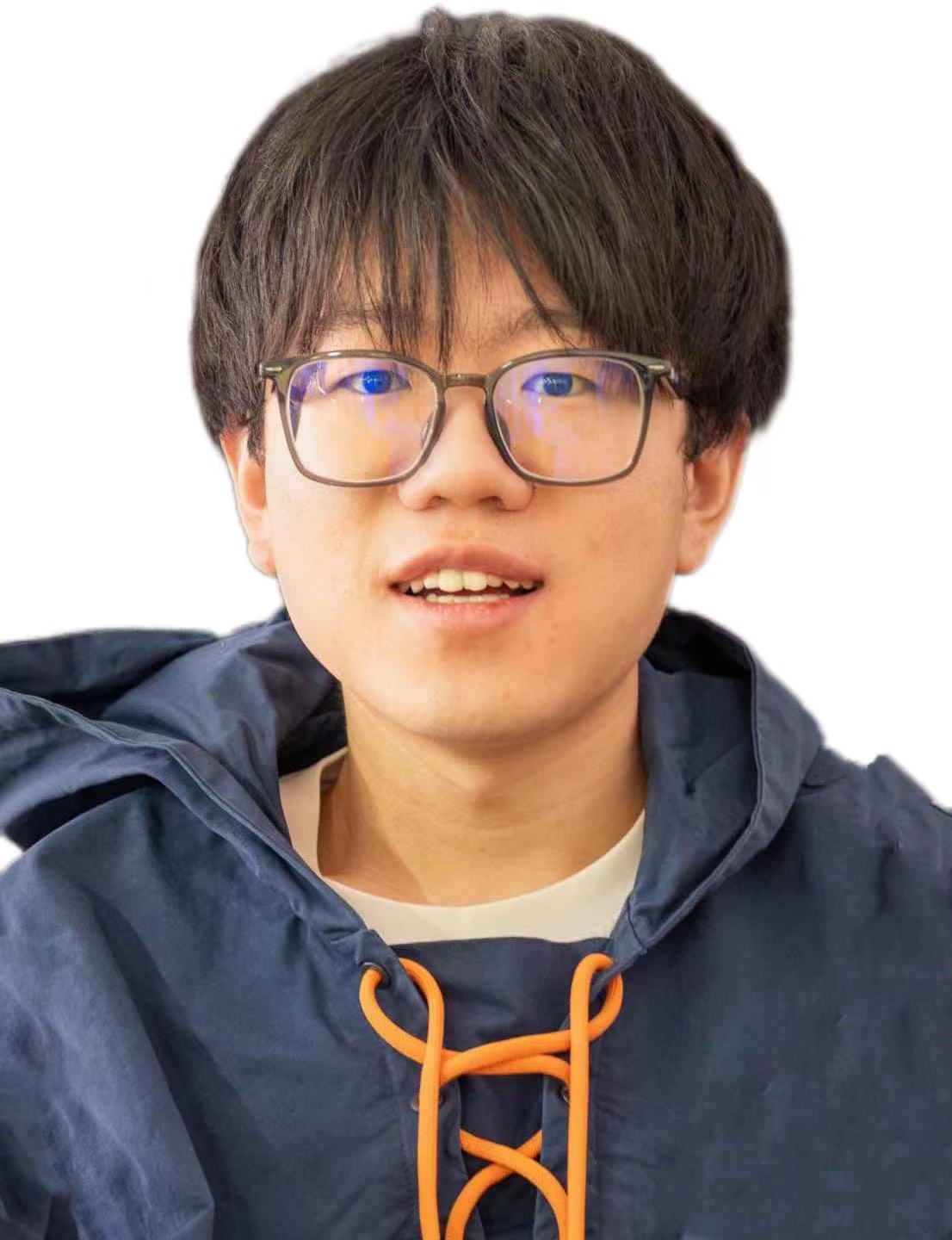}}]{Xuanpu Zhang}
is currently pursuing the master's degree with Tianjin University, Tianjin, China. His research interests include computer vision and Artificial Intelligence Generative Component (AIGC).
\end{IEEEbiography}

\begin{IEEEbiography}[{\includegraphics[width=1in,height=1.25in,clip,keepaspectratio]{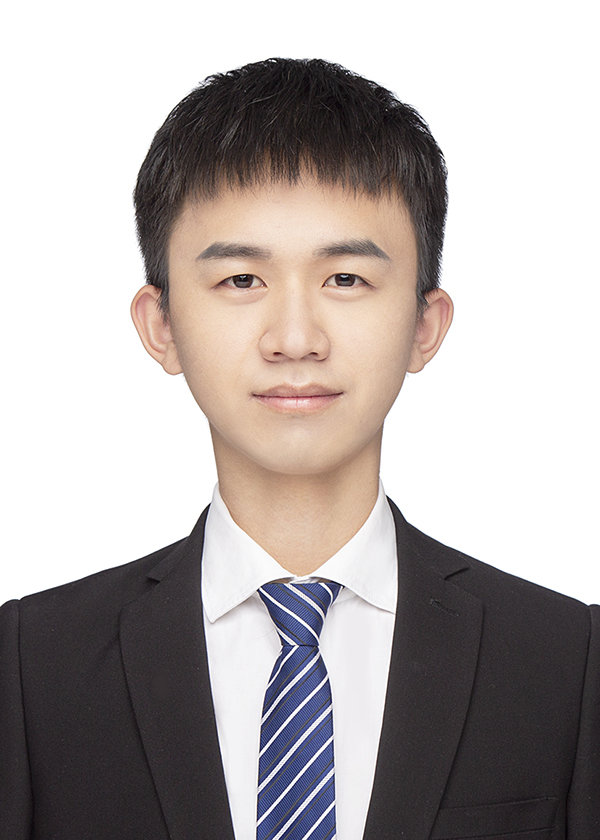}}]{Jianhao Zeng} received the master's degree in electronic engineering from Tianjin University, Tianjin, China. His research interests include computer vision and Artificial Intelligence Generative Component (AIGC).
\end{IEEEbiography}

\begin{IEEEbiography}[{\includegraphics[width=1in,height=1.25in,clip,keepaspectratio]{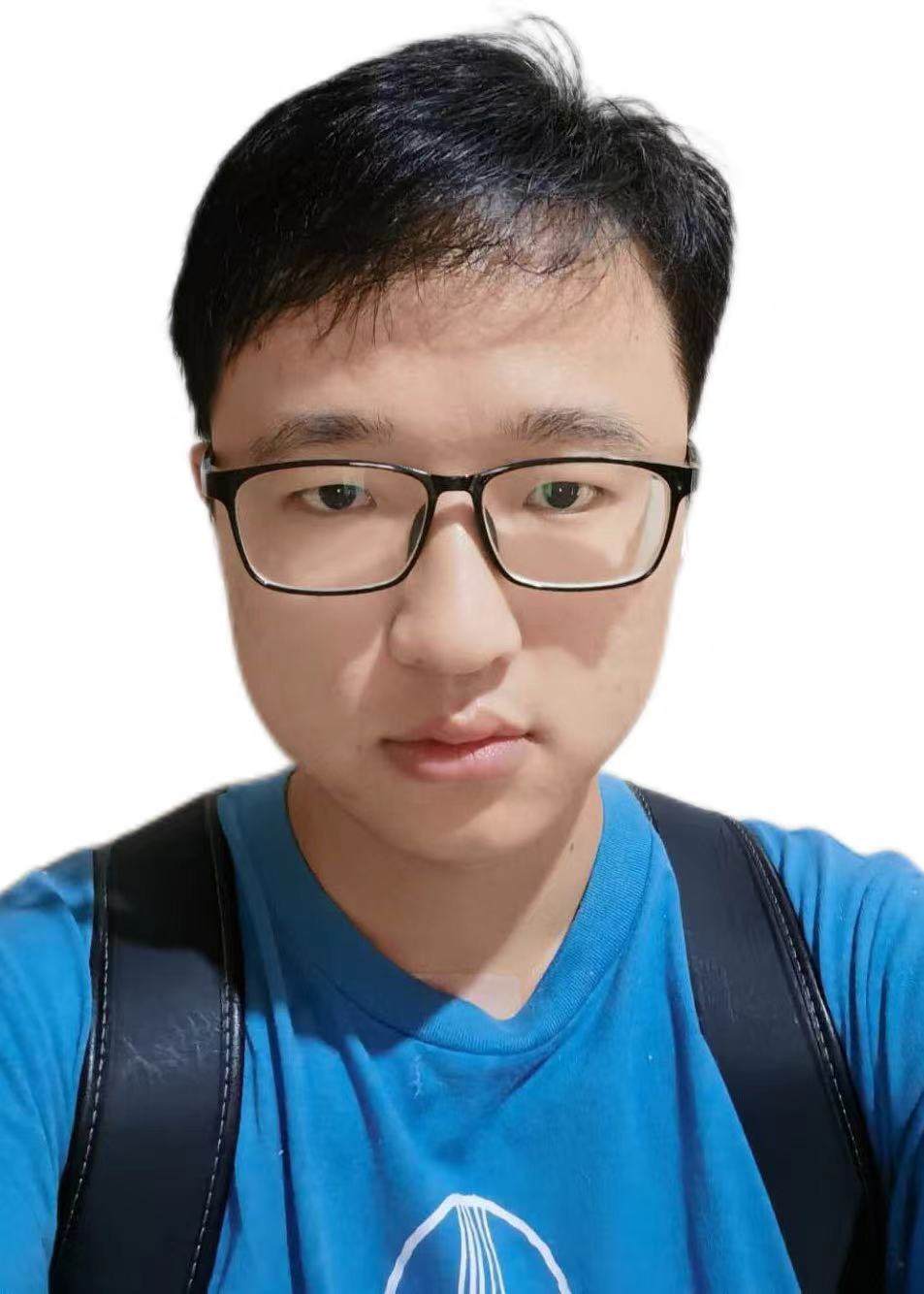}}]{Pengxin Zhan} obtained the master’s degree in Computer Science from Zhejiang University, Hangzhou, China. He is currently a research staff engineer in Alibaba International Digital Commerce Group. His research interests include computer vision and image generation.
\end{IEEEbiography}

\begin{IEEEbiography}[{\includegraphics[width=1in,height=1.25in,clip,keepaspectratio]{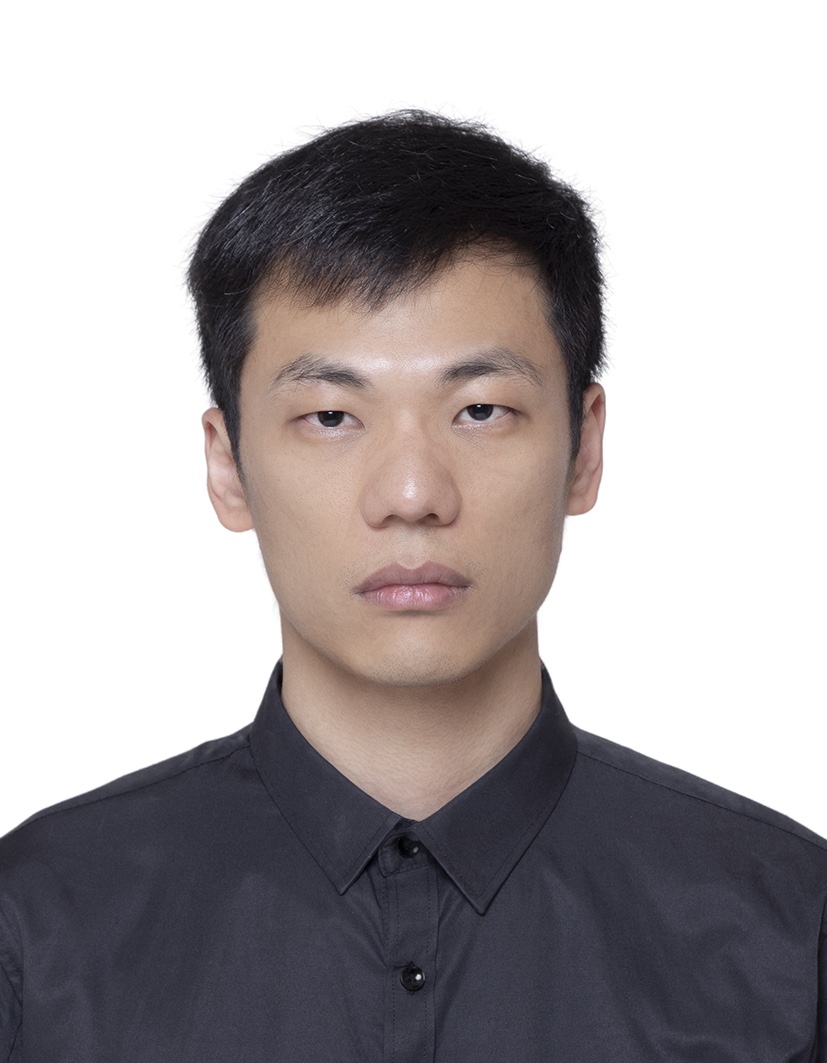}}]{Qingguo Chen} received the master degree in computer science from Nanjing University, China, in 2015. He is currently a research staff engineer in Alibaba International Digital Commerce Group. His research interests include recommendation system, computer vision, LLM and multimodal LLM.
\end{IEEEbiography}

\begin{IEEEbiography}[{\includegraphics[width=1in,height=1.25in,clip,keepaspectratio]{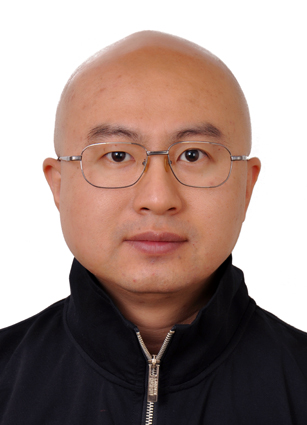}}]{Weihua Luo} joined Alibaba International Digital Commerce Group in 2024 as a researcher. His current research interests are mainly on LLM, machine translation and multimodal LLM. He has published over 50 papers in international journals and top conferences such as ACL, AAAI, EMNLP, ICLR, etc. Weihua Luo obtained his Ph.D. degree in Institute of Computing Technology, Chinese Academy of Sciences. And he received the Bachelor and Master degree in Tsinghua University.
\end{IEEEbiography}

\begin{IEEEbiography}[{\includegraphics[width=1in,height=1.25in,clip,keepaspectratio]{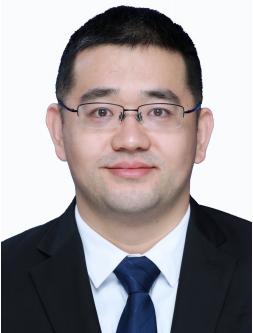}}]{An-An Liu} (Senior Member, IEEE) received the Ph.D. degree in electronic engineering from Tianjin University, Tianjin, China, in 2010. He was a Visiting Professor with the SeSaMe Research Centre, National University of Singapore, Singapore, and a Visiting Scholar with the Robotics Institute, Carnegie Mellon University, Pittsburgh, PA, USA. He is currently a Professor with the School of Electrical and Information Engineering, Tianjin University. His current research interests include computer vision and multimedia information processing.
\end{IEEEbiography}

\end{document}